\pdfoutput=1
\documentclass[11pt]{article}

\usepackage[preprint]{acl}

\usepackage{times}
\usepackage{latexsym}
\usepackage{paralist}

\usepackage[T1]{fontenc}
\usepackage[utf8]{inputenc}

\usepackage{microtype}

\usepackage{inconsolata}

\usepackage{graphicx}

\usepackage[utf8]{inputenc} % allow utf-8 input
\usepackage[T1]{fontenc}    % use 8-bit T1 fonts
\usepackage{hyperref}       % hyperlinks
\usepackage{url}            % simple URL typesetting
\usepackage{booktabs}       % professional-quality tables
\usepackage{amsfonts}       % blackboard math symbols
\usepackage{nicefrac}       % compact symbols for 1/2, etc.
\usepackage{microtype}      % microtypography
\usepackage{xcolor}         % colors
\usepackage{footnotebackref}

\usepackage{amsmath,amsfonts,amssymb}
\usepackage{amsthm}
\usepackage{graphicx}
\usepackage{color}
\usepackage[normalem]{ulem}
\usepackage{diagbox}
\usepackage{makecell}
\usepackage{multicol}
\usepackage{multirow}
\usepackage{tabularx}
\usepackage{paralist}
\usepackage{array}
\usepackage{setspace}
\usepackage{lipsum}
\usepackage[misc]{ifsym}

\usepackage{listings}
\usepackage{color}

\usepackage{xargs}                      % Use more than one optional parameter in a new commands
\usepackage[colorinlistoftodos,prependcaption,textsize=tiny]{todonotes}

\newcommandx{\unsure}[2][1=]{\todo[linecolor=red,backgroundcolor=red!25,bordercolor=red,#1]{#2}}
\newcommandx{\change}[2][1=]{\todo[linecolor=blue,backgroundcolor=blue!25,bordercolor=blue,#1]{#2}}
\newcommandx{\info}[2][1=]{\todo[linecolor=OliveGreen,backgroundcolor=OliveGreen!25,bordercolor=OliveGreen,#1]{#2}}
\newcommandx{\improvement}[2][1=]{\todo[linecolor=Plum,backgroundcolor=Plum!25,bordercolor=Plum,#1]{#2}}

\newcommand\blfootnote[1]{%
  \begingroup
  \renewcommand\thefootnote{}\footnote{#1}%
  \addtocounter{footnote}{-1}%
  \endgroup
}

\title{Scaling Laws for Linear Complexity Language Models}

\author{
{\normalsize
$^{\star}$Xuyang Shen$^{1}$, $^{\star}$Dong Li$^{1}$, $^{\star}$Ruitao Leng$^{1,2}$, Zhen Qin$^{3}$, Weigao Sun$^{1}$, $^\textrm{\Letter}$Yiran Zhong$^{1}$
}\\
$^{1}$OpenNLPLab, $^{2}$Australian National University, $^{3}$TapTap\\
\ \ \texttt{https://github.com/OpenNLPLab/ScalingLaws} 
}

\begin{document}
\maketitle
\begin{abstract}
The interest in linear complexity models for large language models is on the rise, although their scaling capacity remains uncertain. In this study, we present the scaling laws for linear complexity language models to establish a foundation for their scalability. Specifically, we examine the scaling behaviors of three efficient linear architectures. These include TNL~\cite{qin2024lightning}, a linear attention model with data-independent decay; HGRN2~\cite{qin2024hgrn2}, a linear RNN with data-dependent decay; and cosFormer2~\cite{zhen2022cosformer,qin2024you}, a linear attention model without decay. We also include LLaMA as a baseline architecture for softmax attention for comparison. These models were trained with six variants, ranging from 70M to 7B parameters on a 300B-token corpus, and evaluated with a total of 1,376 intermediate checkpoints on various downstream tasks. These tasks include validation loss, commonsense reasoning, and information retrieval and generation. The study reveals that existing linear complexity language models exhibit similar scaling capabilities as conventional transformer-based models while also demonstrating superior linguistic proficiency and knowledge retention.
\blfootnote{\noindent $^{\star}$Equal contribution. $^\textrm{\Letter}$ The corresponding author (Email: \textit{zhongyiran@gmail.com}).}
\end{abstract}

\section{Introduction}\label{introduction}

\renewcommand{\dblfloatpagefraction}{.9}
\begin{figure*}[htbp]
\vspace{-3mm}
    \centering
    % loss, NC, DC
        \includegraphics[width=0.30\linewidth]{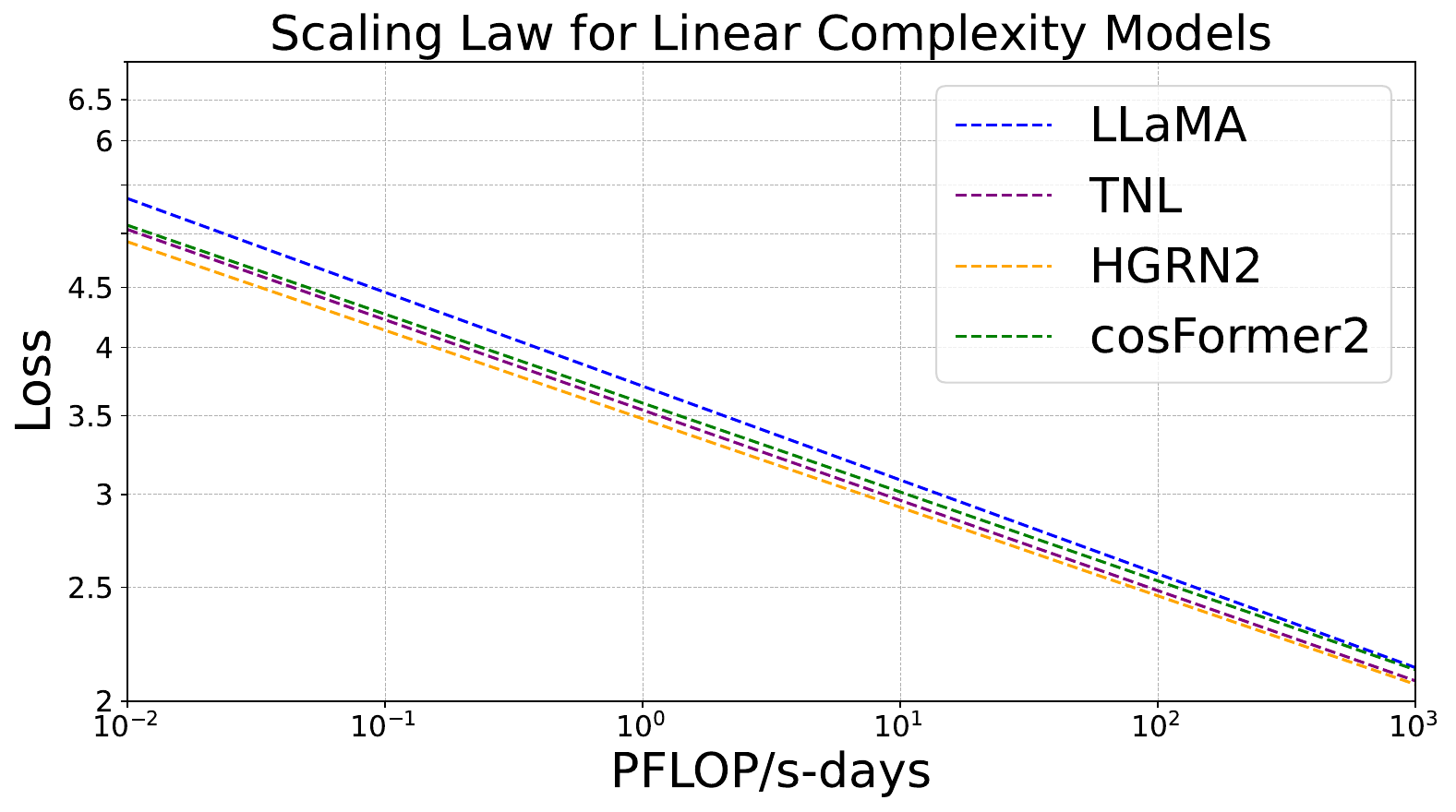} 
        \includegraphics[width=0.30\linewidth]{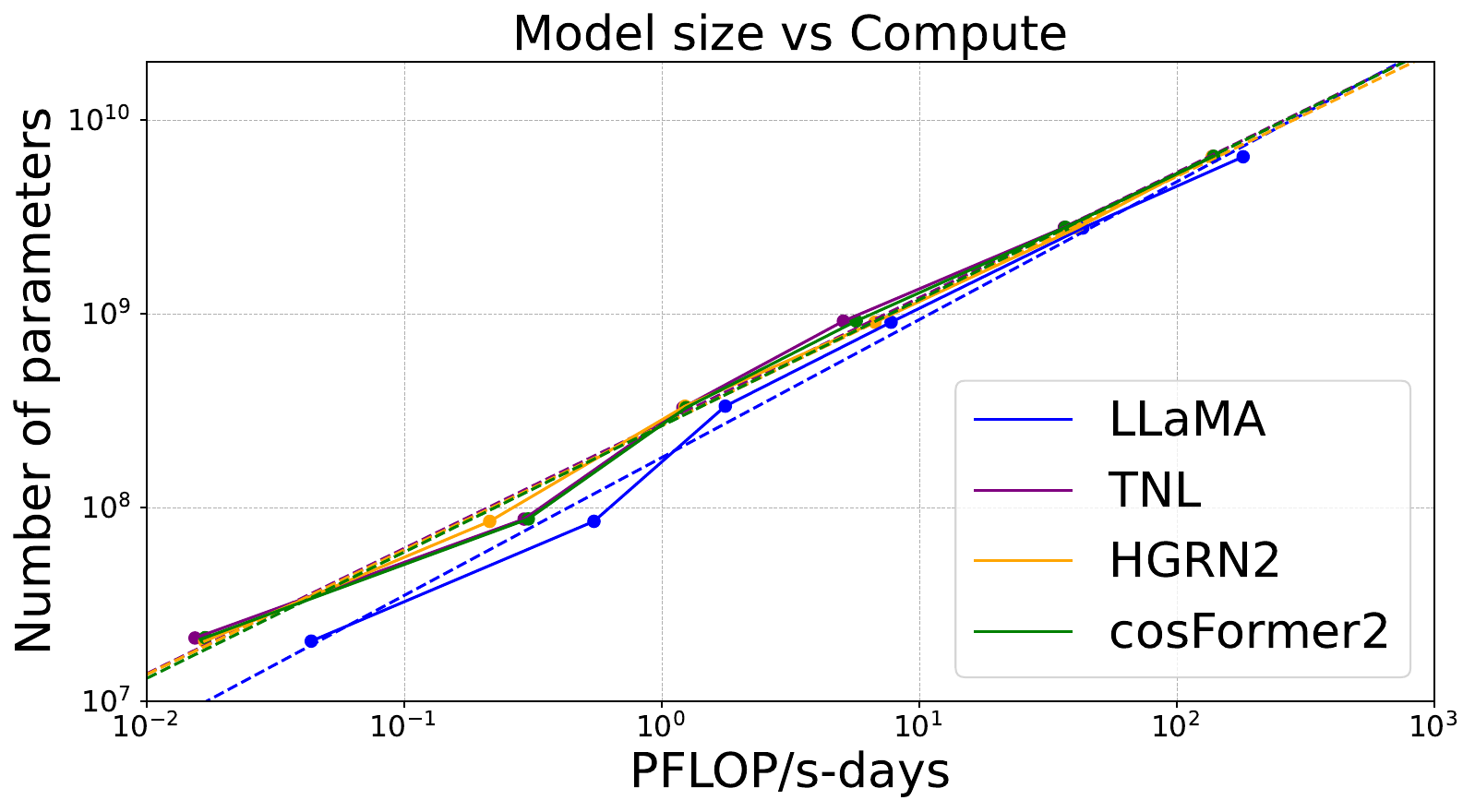} 
        \includegraphics[width=0.30\linewidth]{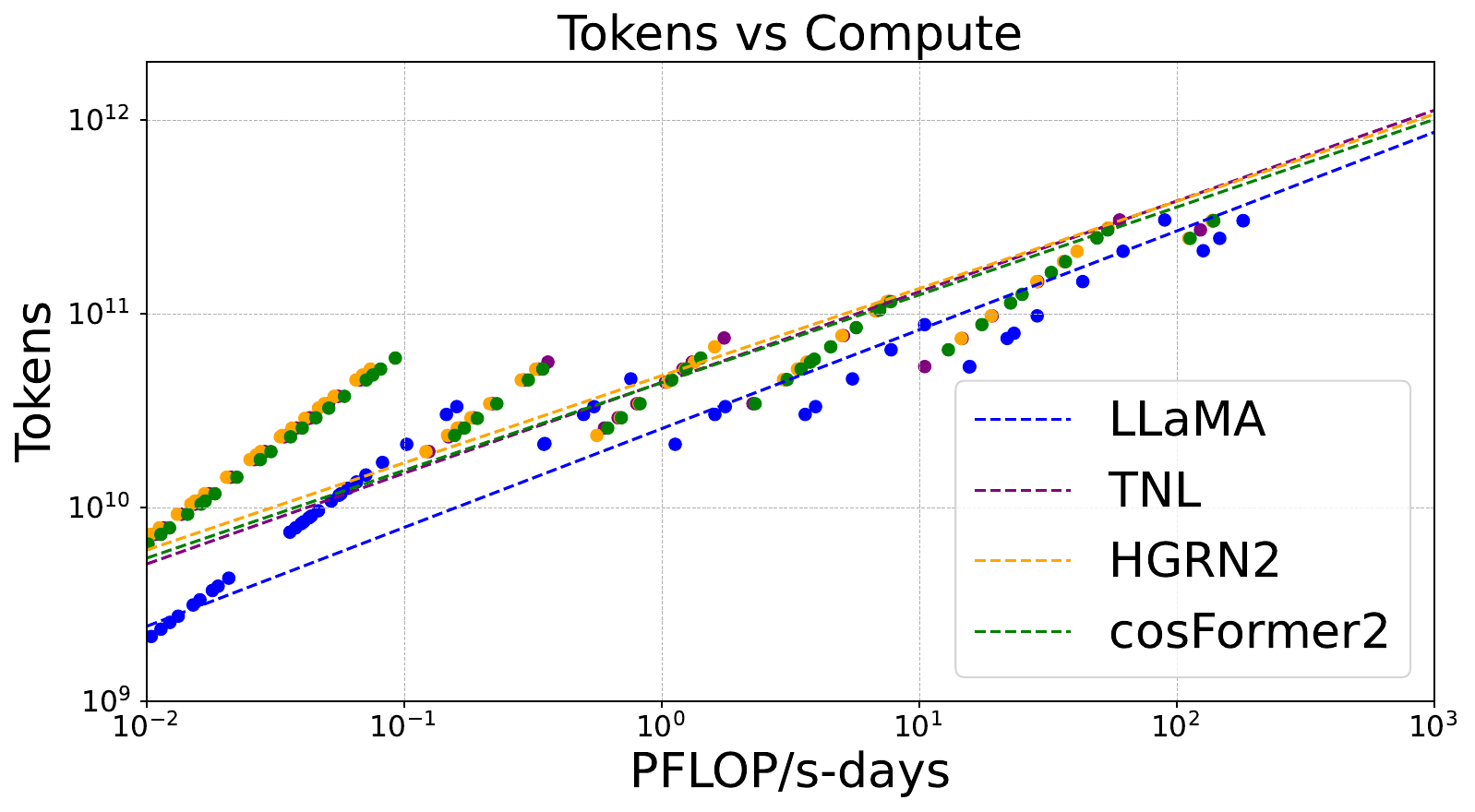} 
        \includegraphics[width=0.30\linewidth]{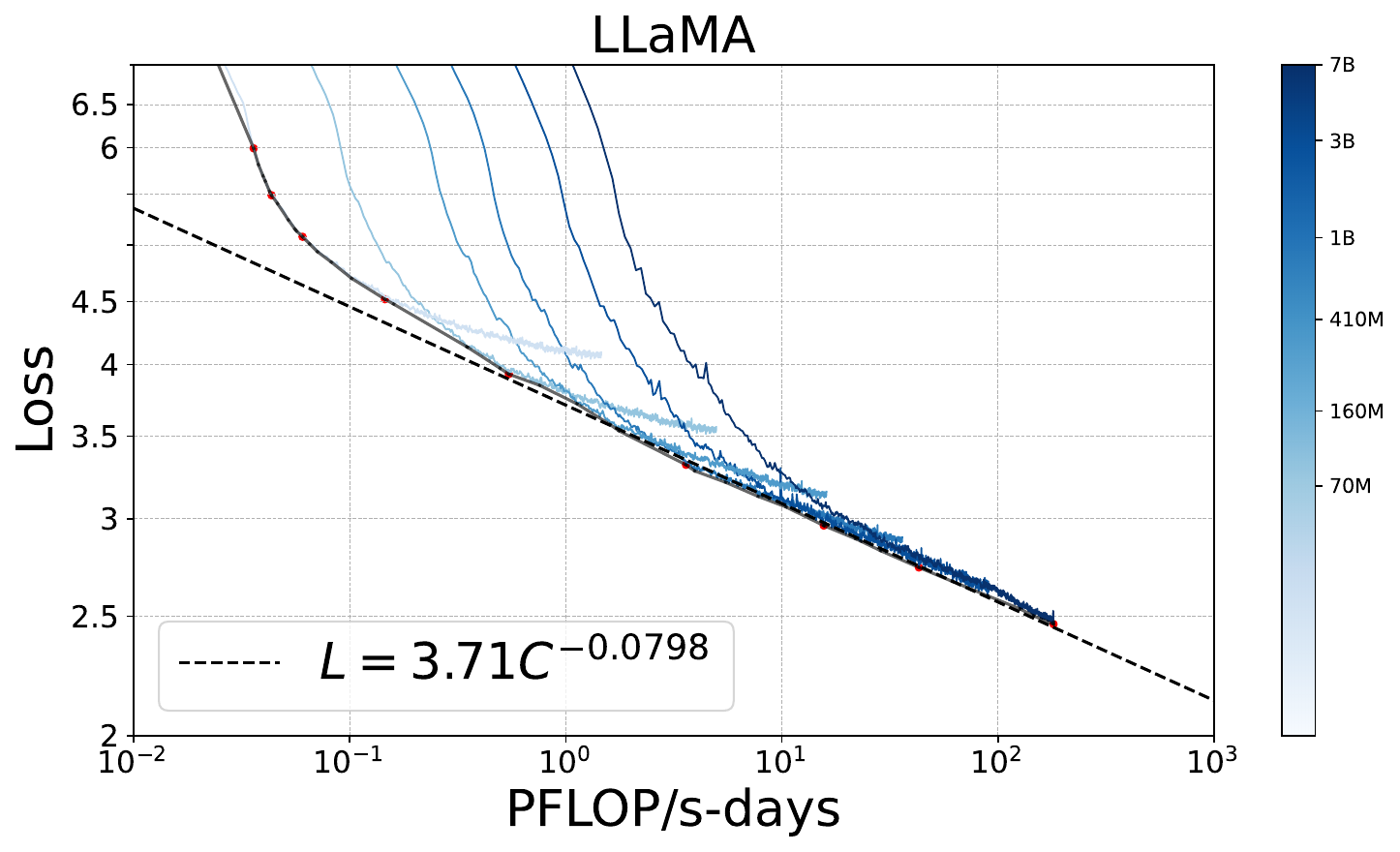} 
        \includegraphics[width=0.30\linewidth]{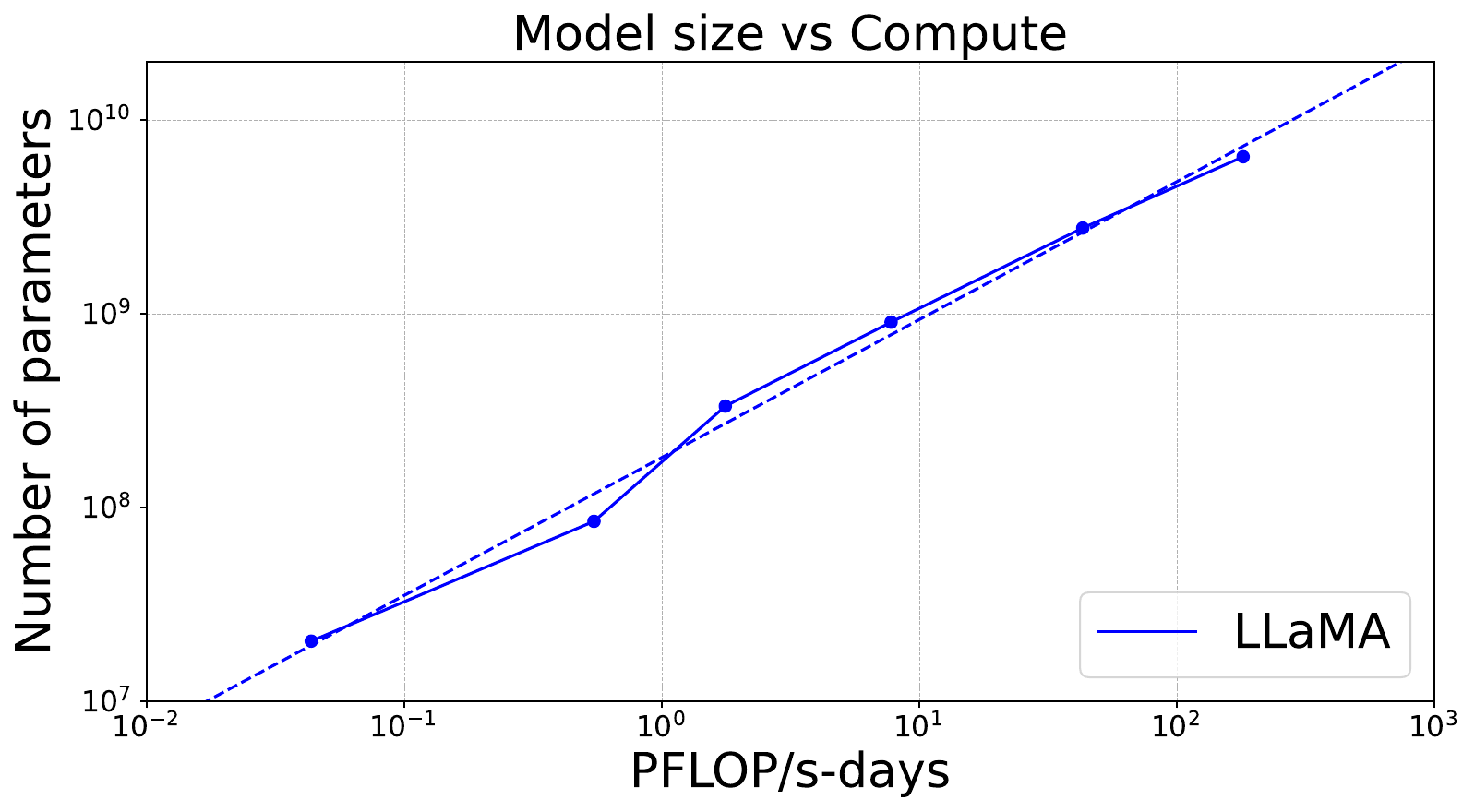} 
        \includegraphics[width=0.30\linewidth]{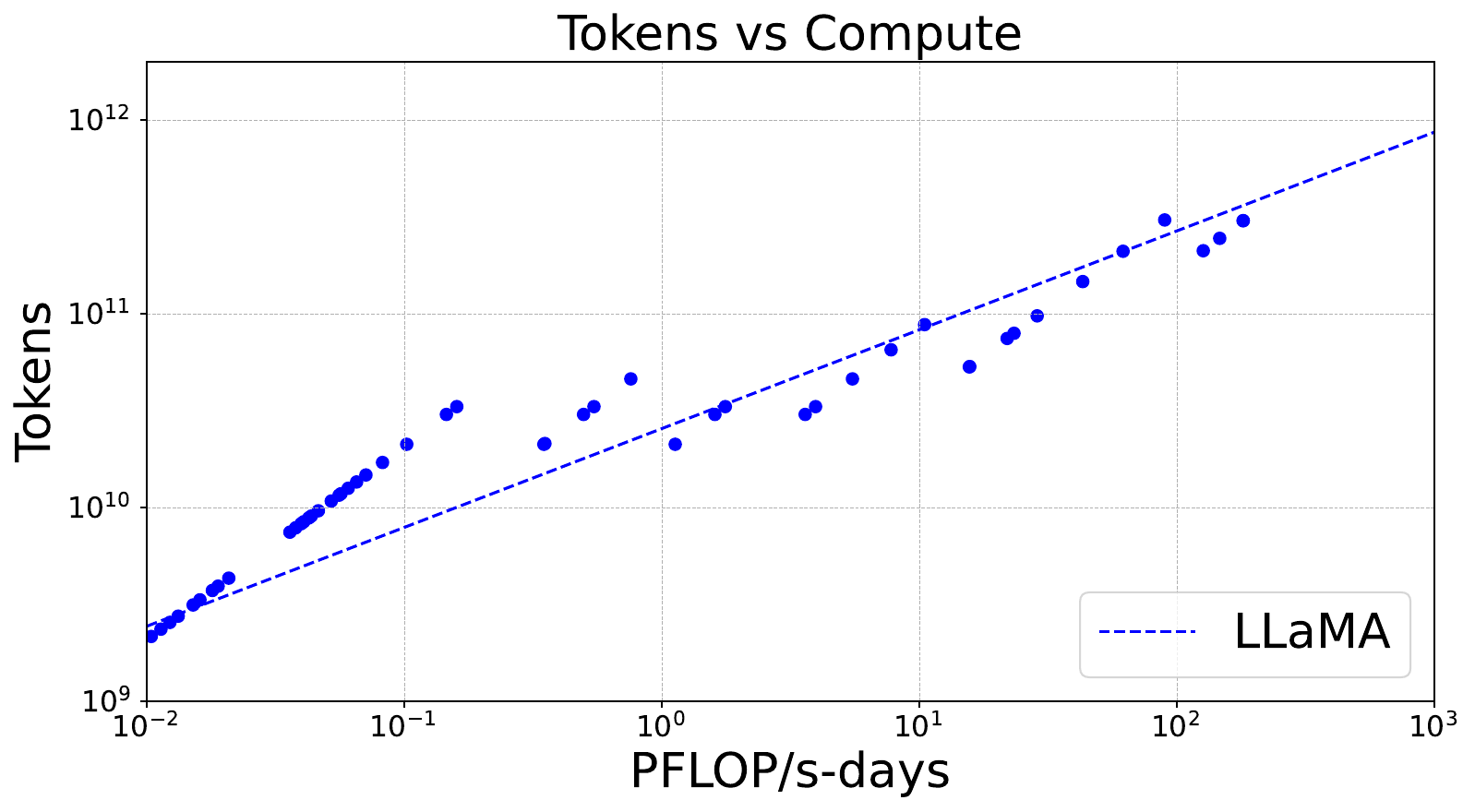} 
        \includegraphics[width=0.30\linewidth]{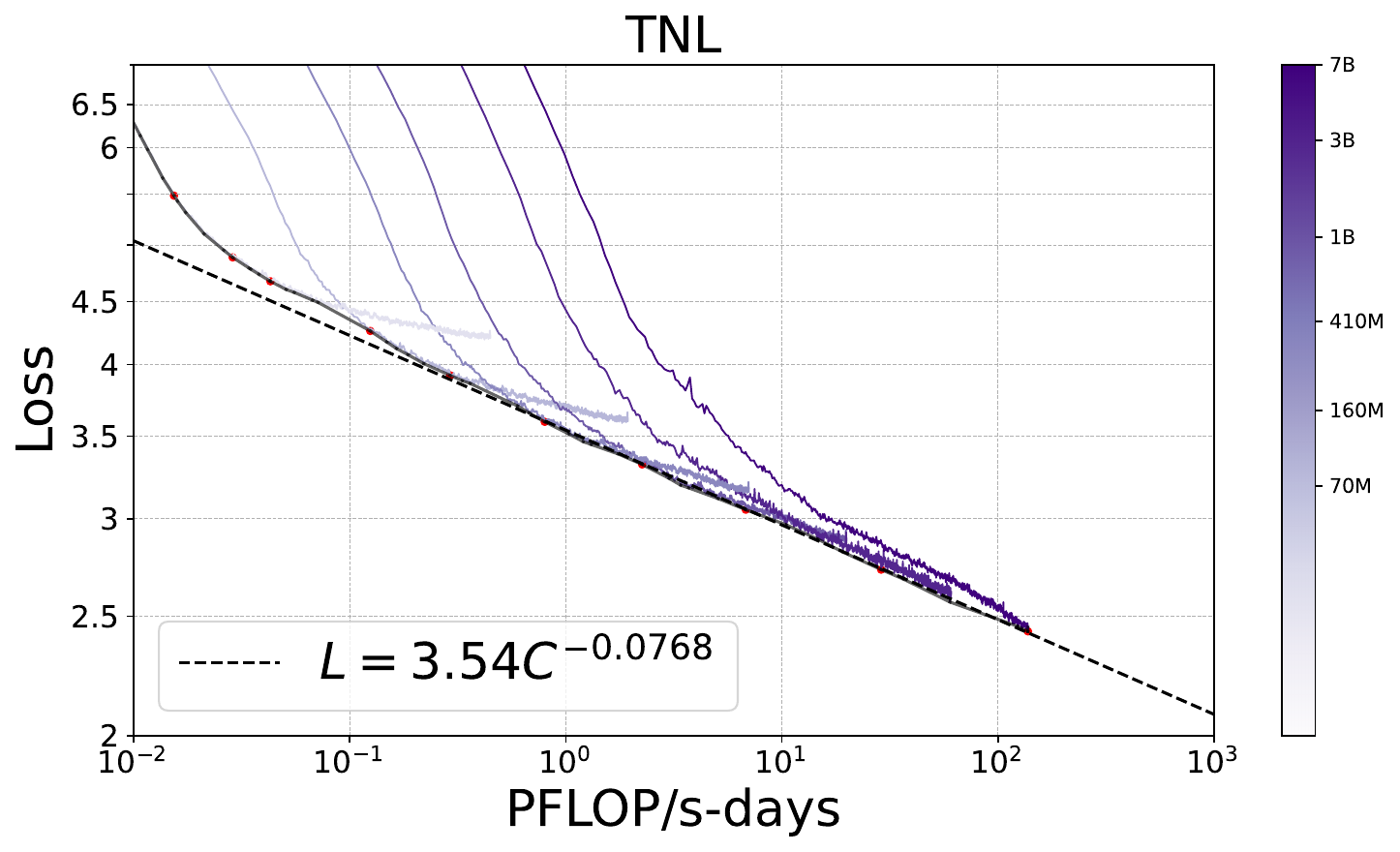} 
        \includegraphics[width=0.30\linewidth]{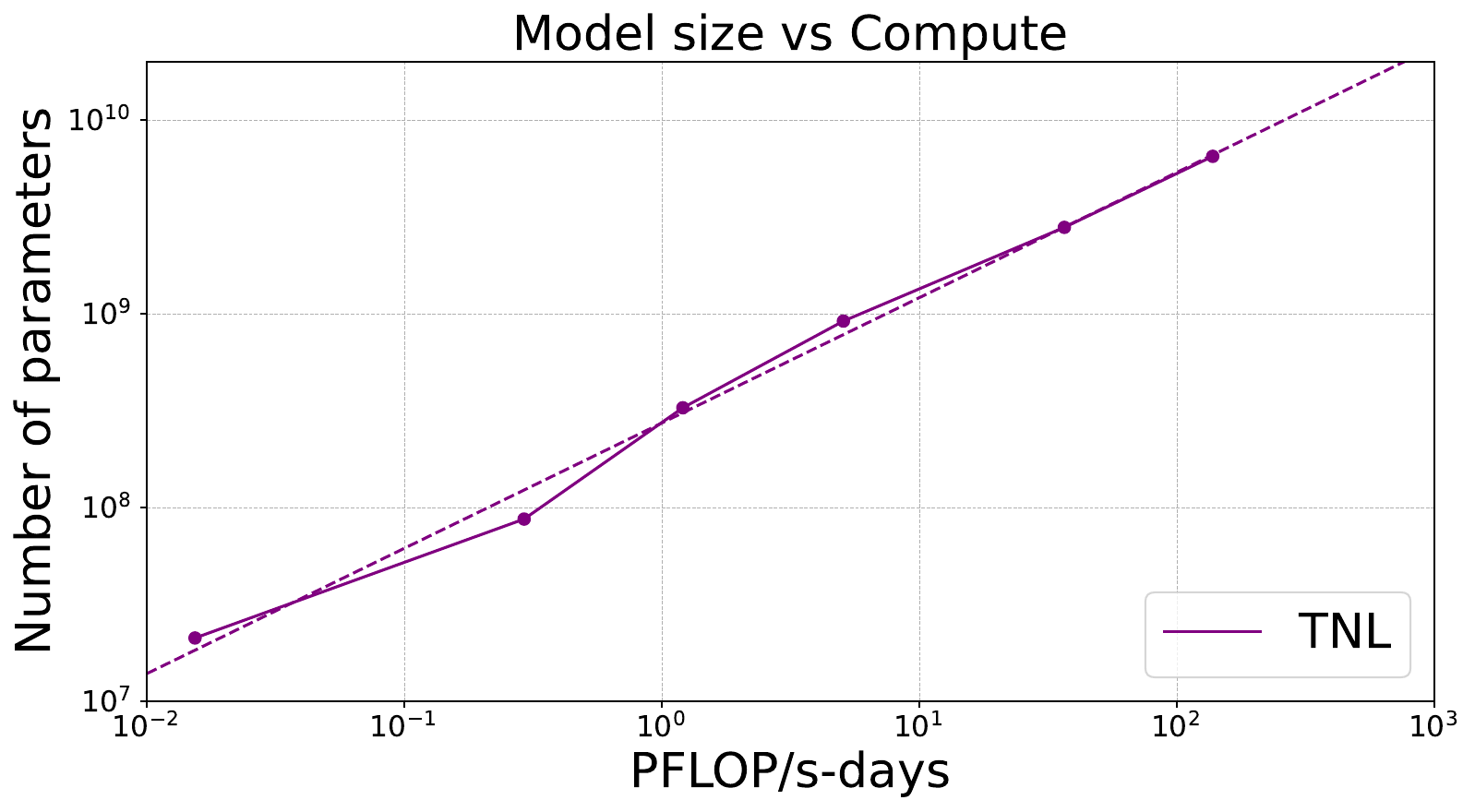} 
        \includegraphics[width=0.30\linewidth]{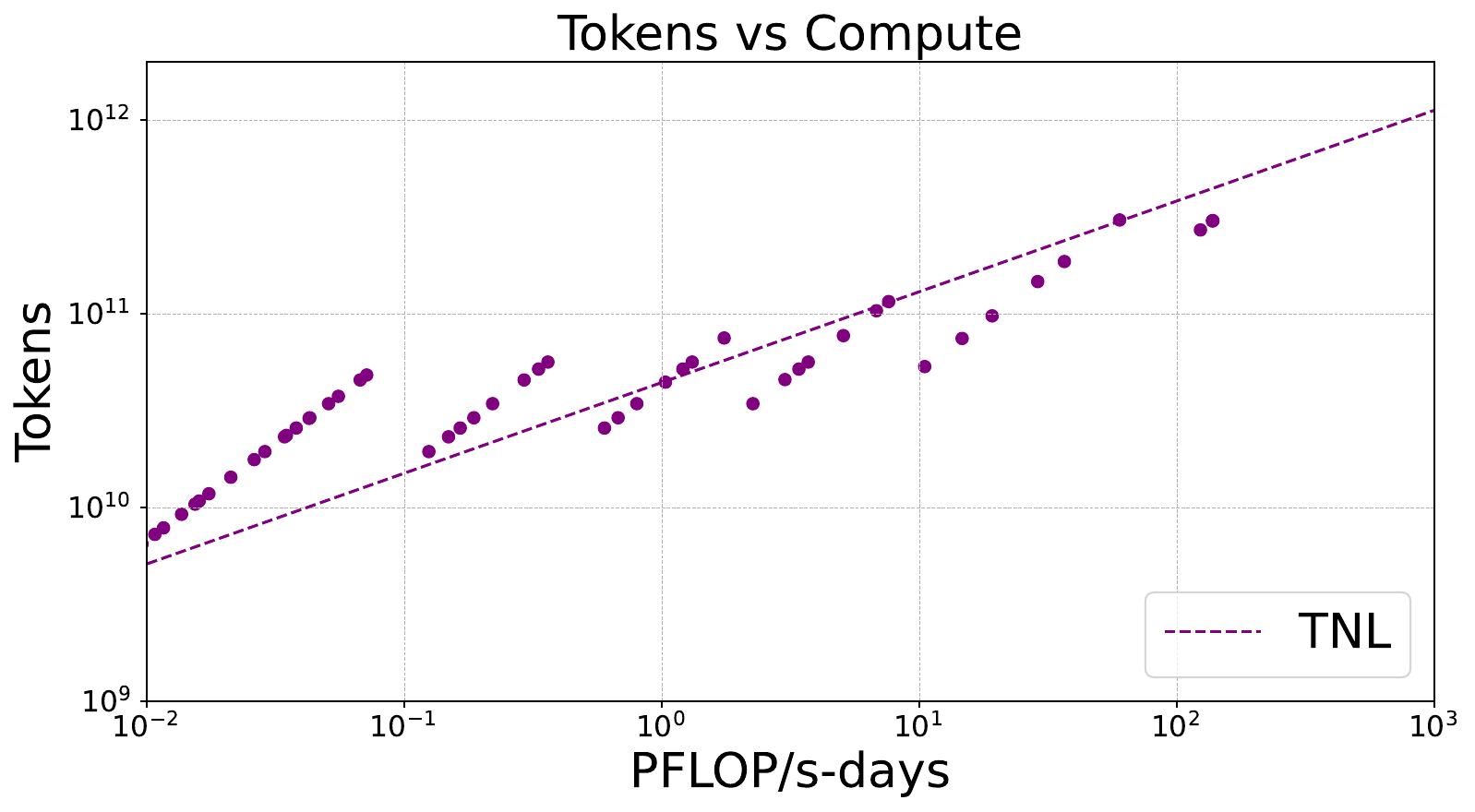} 
        \includegraphics[width=0.30\linewidth]{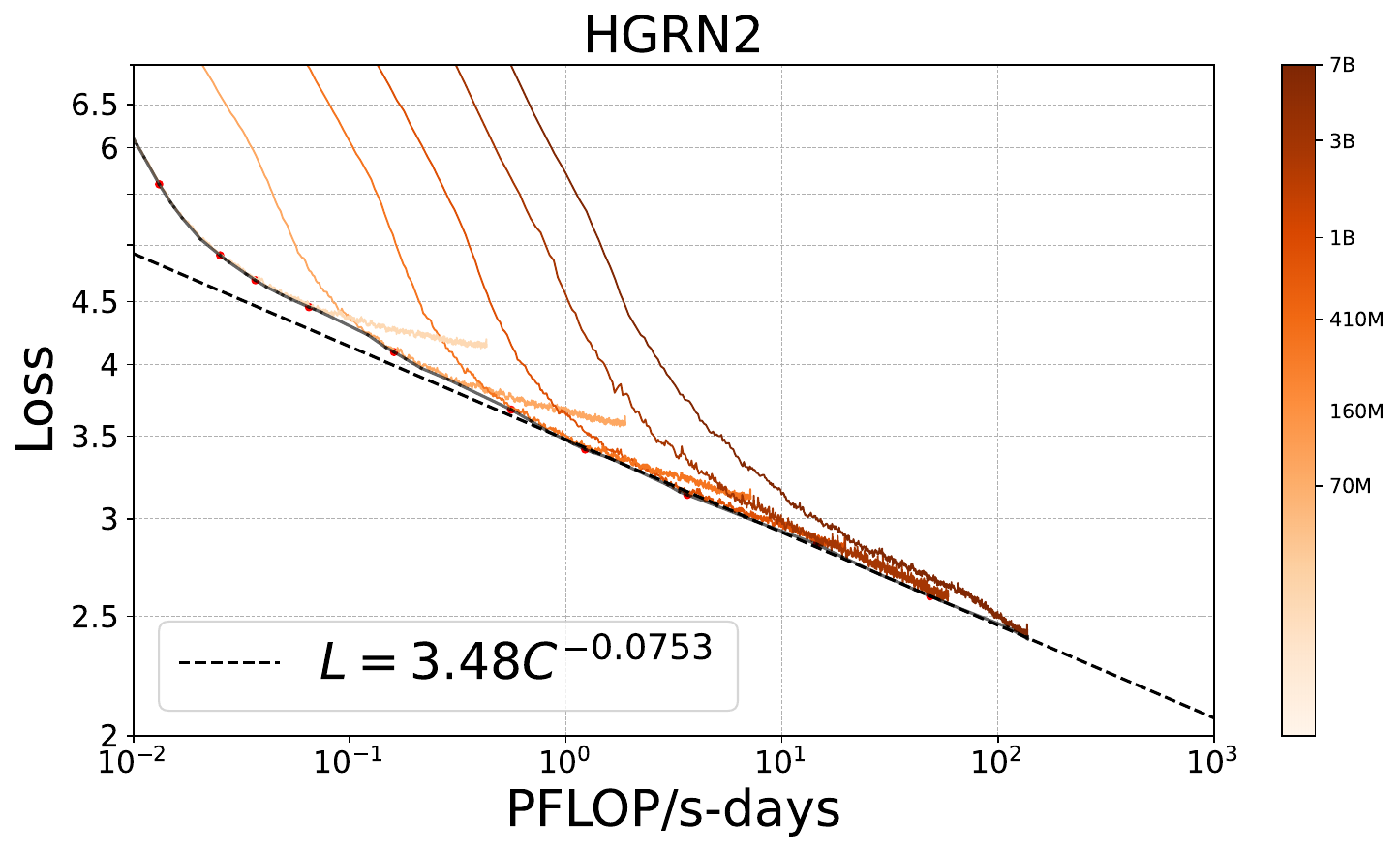} 
        \includegraphics[width=0.30\linewidth]{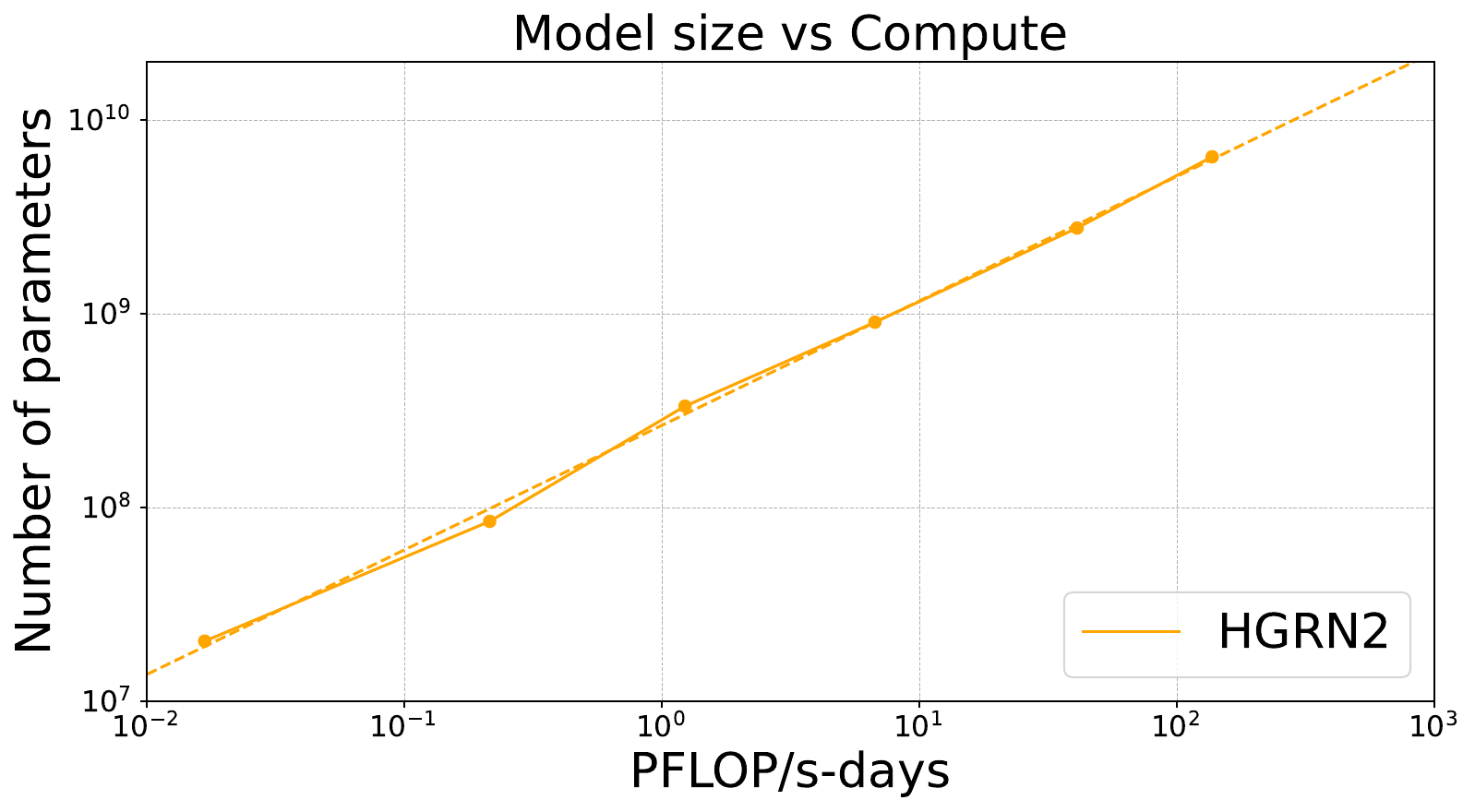} 
        \includegraphics[width=0.30\linewidth]{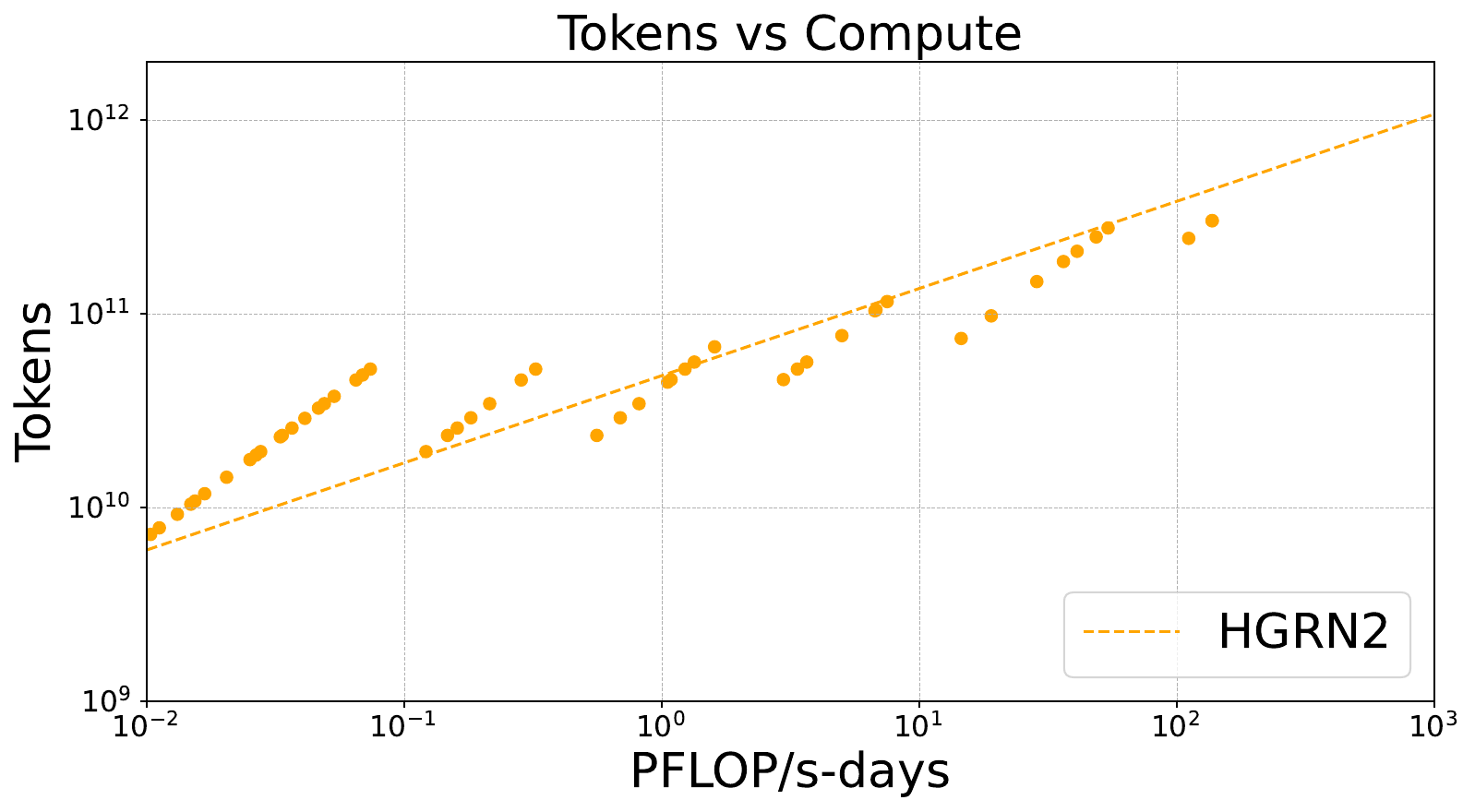} 
        \includegraphics[width=0.30\linewidth]{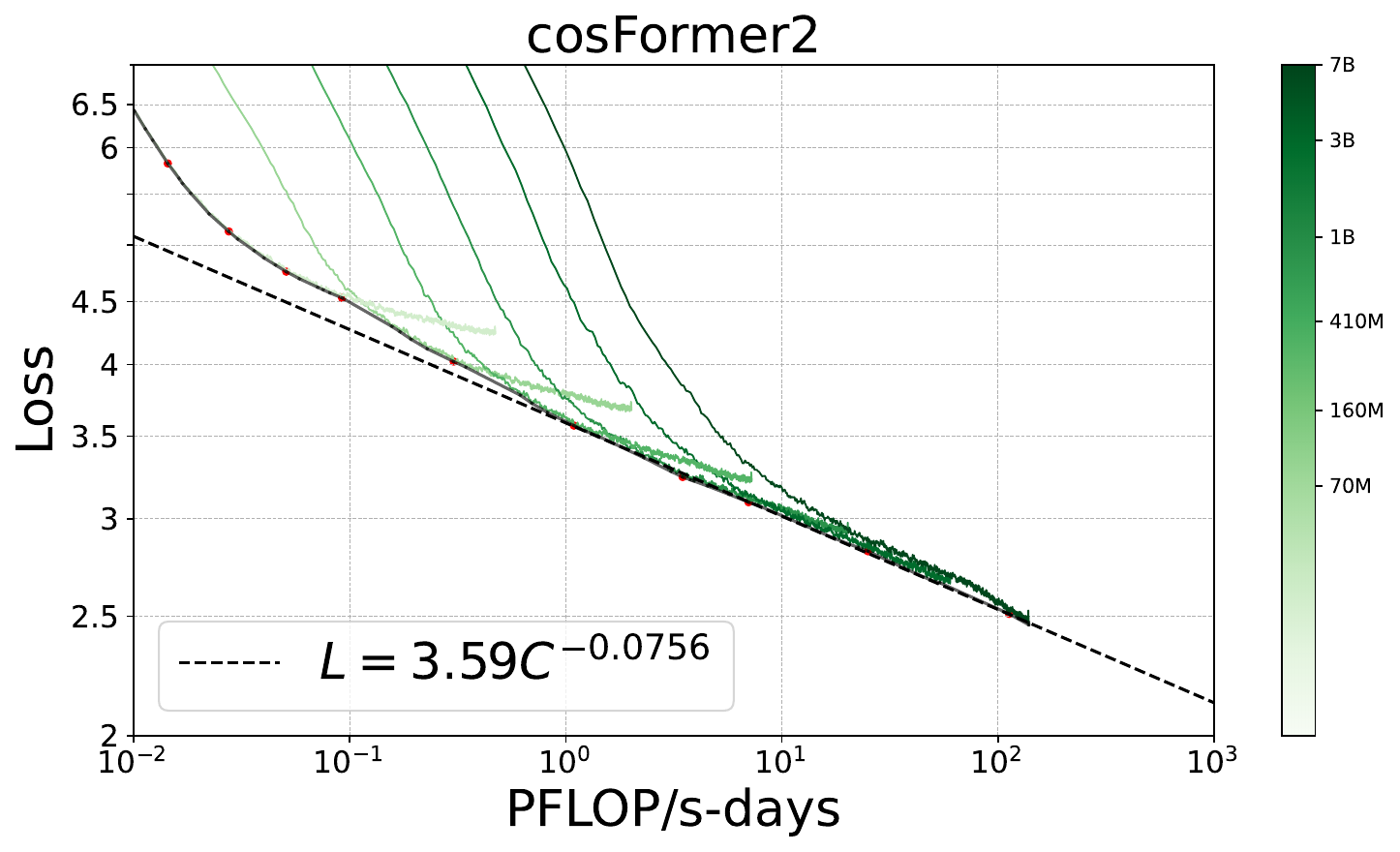} 
        \includegraphics[width=0.30\linewidth]{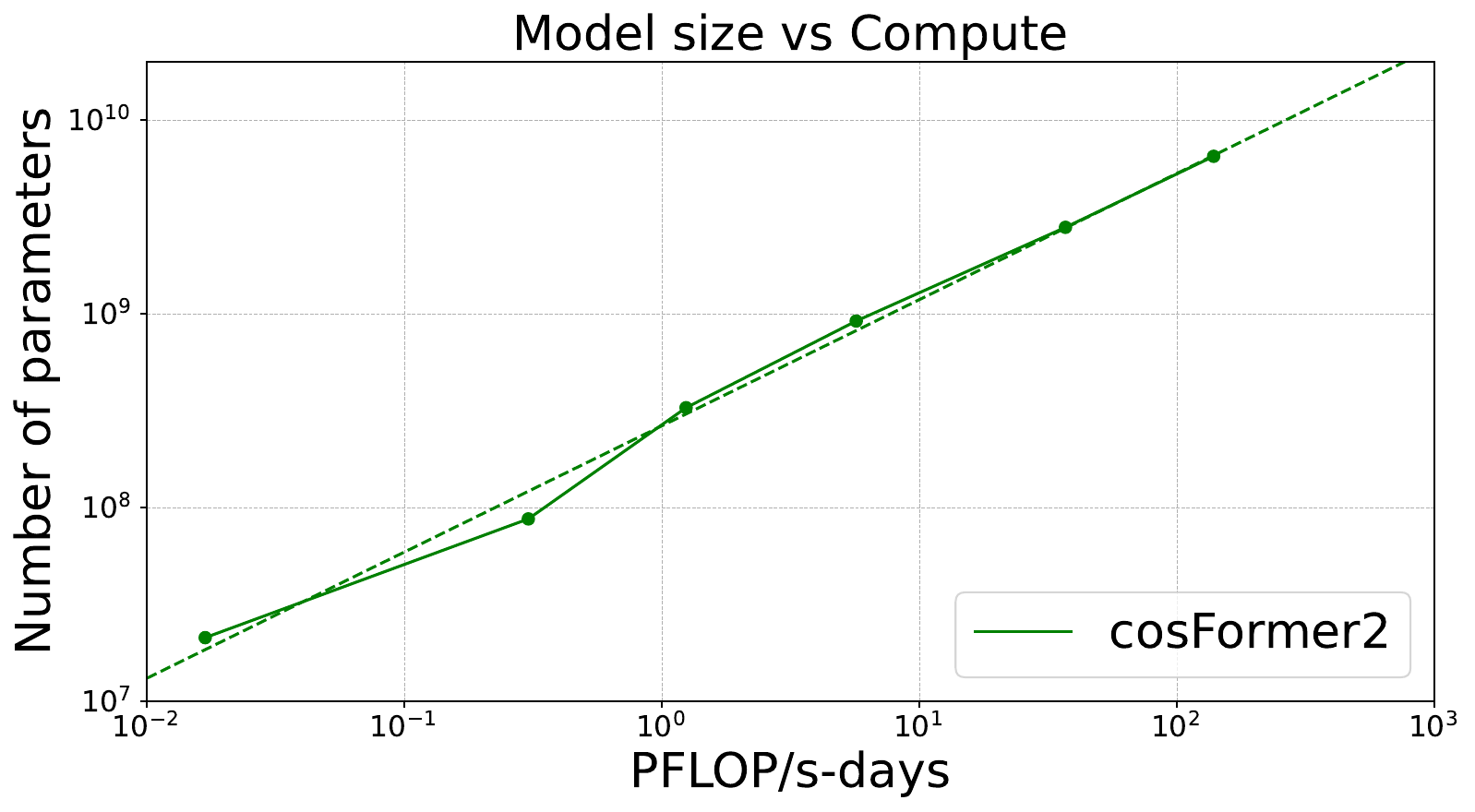} 
        \includegraphics[width=0.30\linewidth]{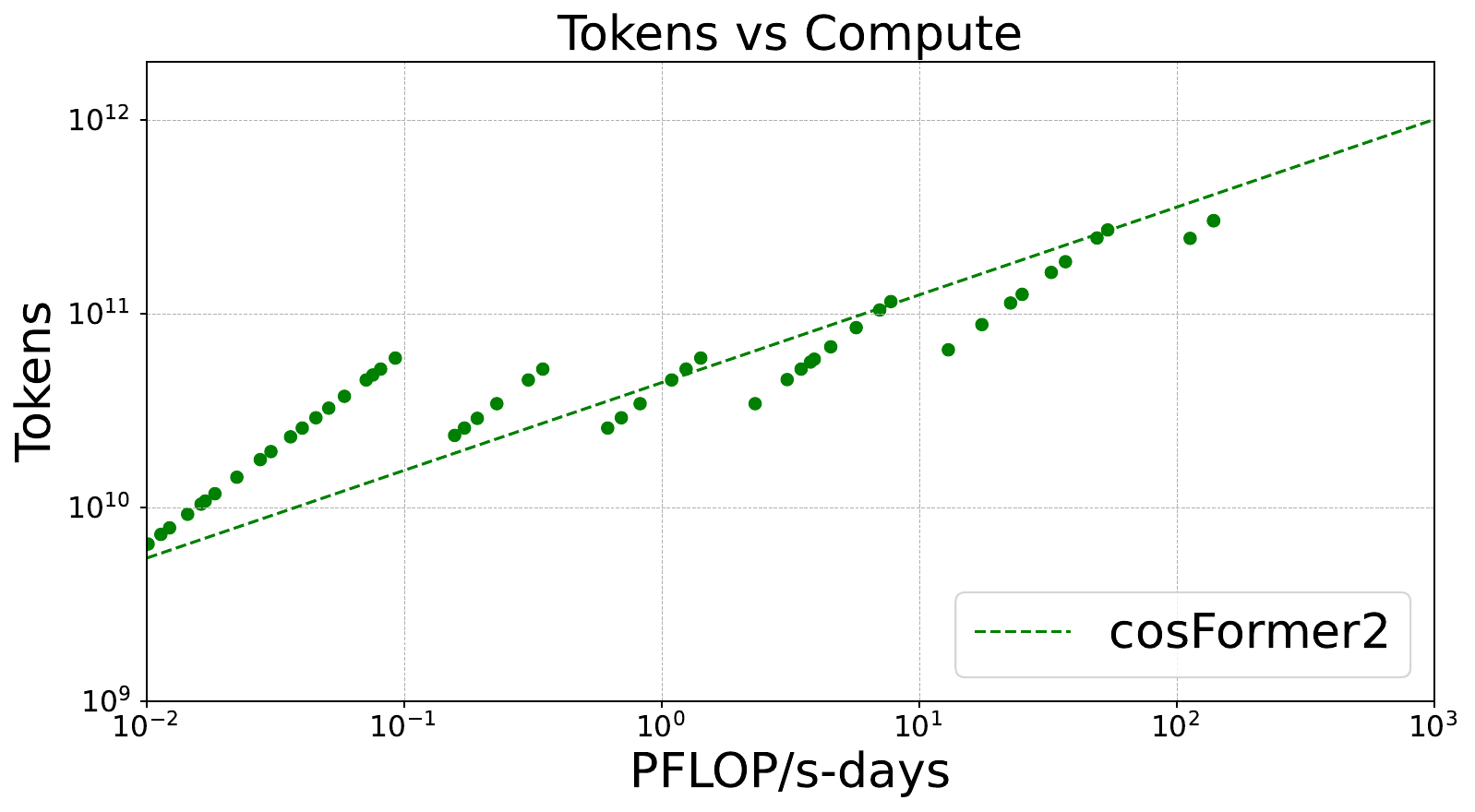}  
\vspace{-4mm}
    \caption{\textbf{Training Curve Fitting for Four Architectures.} In the master row, we present predicted training curves for various architectures, with each subsequent row representing a different architecture. On the left, the training curves for models ranging from 70M to 7B parameters are displayed. From these curves, we extract the envelope of minimum loss per FLOP, using these data points to estimate the optimal model size (center) for a specified compute budget, and the optimal number of training tokens (right).}
    \vspace{-3mm}
    \label{fig:scaling_curve}
\end{figure*}

The prosperity of large language models (LLMs) has necessitated the development of scaling laws~\cite{jared_scaling_law_openai_2020} to optimize the trade-off between increasing model size and expanding training data within finite computational resources. Scaling laws empirically study the correlation between model performance and factors including the number of parameters, training tokens, and FLOPs. Previous works~\cite{jared_scaling_law_openai_2020, Henighan_scaling_2020, jordan_chinchilla_2022, clark_scaling_law_moe_icml_2022} have established power laws to describe these scaling trends. Experiments are typically conducted on smaller models with relatively low training costs. From these observations, regression models are derived to guide the scaling of parameters, data, and computational resources. Establishing these scaling laws is crucial before expanding language models to the scale of LLMs, ensuring predictable results under controllable training costs. Scaling laws have guided the success of many recent LLMs, such as Chinchilla~\cite{jordan_chinchilla_2022} and GPT-4~\cite{gpt4}. It is noteworthy that existing scaling laws are predominantly established for traditional softmax attention transformers~\cite{vaswani_transformer_2017}.

Linear complexity language models~\cite{katharopoulos2020transformers, zhen2022cosformer, choromanski2021rethinking, zheng2022linear, zheng2023efficient, hua2022transformer, liu2022neural, qin2023transnormerllm, qin2024lightning, qin2024various, gu2021efficiently, gu2020hippo, gu2022parameterization, fu2022hungry, qin2023toeplitz, fu2023simple, orvieto2023resurrecting, qin2023hierarchically, qin2024hgrn2,yang2023gated,mamba,mamba2,2307.08621}, have emerged as a promising alternative to traditional transformers in causal language modeling. However, the scalability of these models remains uncertain, which limits their applicability to large language models. To address this concern, in this paper we have developed pre-training scaling laws for efficient large language models. Following the approach outlined in~\cite{jordan_chinchilla_2022}, we have used the training loss as a regression target to establish power law equations against FLOPs and infer the optimal model size and dataset size under constant computation budgets for linear complexity models. Our study focuses on investigating three efficient architectures, as detailed in Section \ref{preliminary}: TNL~\cite{qin2024lightning}, HGRN2~\cite{qin2024hgrn2}, and cosFormer2~\cite{zhen2022cosformer, qin2023linearized, qin2024you}. Additionally, LLaMA~\cite{llama2_2023} is used as a baseline to represent softmax attention transformers. For a comprehensive analysis, we compare the scaling behavior of downstream task performance across different architectures. As outlined in Section \ref{setup}, each model is evaluated in terms of linguistic proficiency, knowledge retention, and information retrieval and generation. Our findings reveal that linear complexity models exhibit similar scaling trends to conventional transformer-based models and consistently outperform LLaMA in cross-domain perplexity and average accuracy in commonsense reasoning under the same FLOPs budget but demonstrate weakness in retrieval tasks.

Our contributions are summarized as follows:
\begin{compactitem}
    \item We disclose scaling laws for linear complexity language models, focusing on three different architectures. 
    Reveling the training loss $L$,  model size $N$ and dataset size $D$ have power-law relationships with computation budget $C$. 
    \item Our experiments showcase the advantage of linear complexity language models over traditional transformers on linguistic proficiency while 
     inferior performance in retrieval tasks.
    \item We analyze the scaling trends for downstream task performance and observe the correlation of performance with computation budget.
    \item For linear models, aspect ratio (model dimension / number of layers) and context length affect model capacity. This is contradictory to previous scaling laws, where model shape makes a negligible impact. 
   \item For linear models, data-dependent decay is beneficial in retrieval tasks and is not significantly different from data-independent decay in other tasks.
\end{compactitem}

\section{Preliminary}\label{preliminary}

\subsection{Causal language modeling}
Causal language modeling forecasts the next word in a sequence by analyzing prior words, commonly used in GPT models~\cite{radford2018improving}. It employs the cross-entropy loss function to assess model accuracy by comparing the predicted and actual word distributions—a lower score suggests better performance. 
Transformers that use softmax-based attention~\cite{vaswani_transformer_2017}, are referred to as \textit{vanilla} transformers. In our experiments, LLaMA~\cite{llama2_2023,touvron2023llama} serves as the exemplar for this category of transformers.

\subsection{Linear complexity sequence models}
In order to tackle the high time complexity of traditional transformers, researchers are currently investigating new linear complexity sequence model architectures. These alternatives include linear attention~\cite{qin_cosformer_iclr_2021}, state space models (SSMs)~\cite{gu_S4_ICLR_2021}, long convolution~\cite{qin2023toeplitz}, and linear RNN~\cite{qin_hgrn_nips_2024}. According to \citet{qin2024unlocking}, SSMs can be considered as linear attention variants, and long convolution can be accurately transformed into SSMs~\cite{qin2023accelerating}. In this study, TNL~\cite{qin2023transnormerllm,qin2024lightning,qin2024various} and cosFormer2~\cite{qin2024you} serve as representatives of linear attention, while HGRN2~\cite{qin2024hgrn2} is the chosen representative of linear RNN.

\noindent
\textbf{TNL} uses \textit{data-independent decay} to enhance Linear Attention in order to address the "dilution" problem. By employing these techniques, along with Lightning Attention~\cite{qin2024lightning}, TNL outperforms traditional softmax attention models in both efficiency and accuracy.

\noindent
\textbf{HGRN2} overcomes the limited expressiveness of traditional HGRNs~\cite{qin_hgrn_nips_2024} by employing a state expansion mechanism inspired by linear attention, which enlarges the recurrent state size without extra parameters. It also incorporates \textit{data-dependent decay} in its positional encoding components. This innovation enables HGRN2 to achieve superior performance in language modeling, image classification, and Long Range Arena benchmarks, demonstrating enhanced efficiency and accuracy compared to both its predecessor and other contemporary models.

\noindent
\textbf{cosFormer2 (cos2)} represents an advanced iteration of the original cosFormer~\cite{zhen2022cosformer} model without decay, incorporating several significant enhancements that optimize its performance and functionality: $\textit{1}$. cos2 adopts complex-based LRPE positional encoding~\cite{qin2023linearized} to facilitate a per-channel $\mathrm{cos}$ reweighting mechanism, an improvement over the uniform $\mathrm{cos}$ weighting applied across all features in the original cosFormer~\cite{zhen2022cosformer}.
$\textit{2}$. cos2 enhances its handling of relative positional information through the integration of TPE~\cite{qin2024you}.
$\textit{3}$. cos2 utilizes a low-rank output gate from TNL~\cite{qin2024various}, contributing to more efficient data processing.
$\textit{4}$. cos2 employs normalization~\citep{qin_transnormer_emnlp_2022} for improved stability and performance instead of using a denominator. These enhancements enable cos2, with its \textit{no-decay positional encoding}, to match the performance of top Transformer models like LLaMA.

\subsection{Model size and FLOPs calculation}
\begin{table*}[thp]
\small
    \centering
     \caption{\textbf{Checklist of Model Parameters and FLOPs.} Detailed calculations are deduced in Appendix \ref{model_params_flops}. Here $h$ is the number of attention heads. Compared to the full equations, we exclude embedding parameters and other subleading terms in our calculation for better fitting results of the scaling law equations. 
     } 
     \vspace{-3mm}
    \setlength{\tabcolsep}{13mm}
    \renewcommand{\arraystretch}{1.5}
    \begin{tabular}{c|c|c}
    \toprule
        Architecture & Parameter count & FLOPs count \\ \midrule
        LLaMA  & $12ld^2$ & $72bnld^2(1+\frac{n}{6d}+\frac{5}{18d})$  \\ \hline
        TNL & $12ld^2 +2ld^2/h$ & $72bnld^2(1+\frac{1}{2h}+\frac{5}{18d})$  \\ \hline
        HGRN2  & $12ld^2+ld$ & $72bnld^2(1+\frac{1}{3h}+\frac{29}{72d})$  \\ \hline
        cosFormer2 & $12ld^2 +2ld^2/h+d^2/h$ & $72bnld^2(1+\frac{3}{4h}+\frac{23}{72d})$  \\ 
    \bottomrule
    \end{tabular}
    \label{model_flops}
    \vspace{-4mm}
\end{table*}
In calculating the model parameters $N$ and compute budget $C$, previous studies have employed varying levels of simplification. For clarity, we denote model specifications as follows: $l$ (number of layers), $d$ (model dimension), and $d_{f}$ (feed forward layer dimension). \citet{jared_scaling_law_openai_2020} only accounts for the weights of linear layers (excluding input and output embedding) as model parameters. When $d_{f}=4d$, the total number of model parameters is computed as: $N=12ld^2$. The total forward and backward FLOPs can be approximated as: $C\approx 6N$. On the other hand, \citet{jordan_chinchilla_2022} takes a more detailed approach by incorporating embedding parameters into the model parameter count and factoring in the computational load of softmax operation, input and output embedding in the FLOPS count. To underscore the distinction between \textit{vanilla} transformers and linear complexity sequence models, as well as the variations among linear complexity sequence models, we employ a detailed method for computing $N$ and $C$ as outlined in Table~\ref{model_flops}.

\section{Experimental setup}\label{setup}

\subsection{Corpus}
The training dataset for this work comprises 300 billion tokens sampled from a self-collected and curated corpus of approximately 2 trillion tokens. The data is bilingual, consisting of English and Chinese texts in a 2:1 ratio. Sources of data span various categories, including academic publications, books, and selected web pages. The corpus was refined using several cleaning strategies~\cite{qin2023transnormerllm,qin2024various}, such as rule-based filtering, deduplication, and a proprietary self-cleaning scheme.

\subsection{Training procedures}
Our experiments were implemented using the Metaseq training framework~\cite{zhang2022opt} built atop PyTorch~\cite{paszke2019pytorch}. The LLaMA model was equipped with FlashAttention-2~\cite{flashattention2_2023}, whereas the TNL, cosFormer2 model incorporated Lightning Attention~\cite{qin2024lightning}. HGRN2 employs Flash Linear Attention (FLA)~\cite{yang2024fla}. We conducted all experiments on H100/H800 80G GPUs. 

For all model architectures and training sequence lengths, we maintained a consistent global batch size of 4 million tokens. We utilized the Adam optimizer, with a learning rate of 3e-4 and a weight decay of 0.1. A fixed learning rate scheduler was used for all experiments within constrained computation resources. We use tiktoken~\cite{tiktoken} as the tokenizer, featuring a vocabulary size of 100,280.

\subsection{Model configurations}
We investigate four distinct model architectures: LLaMA, TNL, cosFormer2, and HGRN2, across a spectrum of scales—70M, 160M, 410M, 1B, 3B, and 7B. Each model is trained on a corpus of up to 300 billion tokens with a context length of 8192, aligning with the methodology proposed by Hoffmann et al. \cite{jordan_chinchilla_2022}, where training loss serves as a direct proxy for test loss.

In our continued exploration of linear complexity models, we have extended the pre-training context lengths for the 1B models to encompass 2048, 4096, and 16384. Additionally, we have introduced variations in the hidden dimensions of the 1B models, testing sizes 1536, 1792, 2048, and 3072, to assess the impact of these adjustments on pre-training loss and subsequent performance.

\begin{table}[ht]
\small
    \caption{\textbf{Specifications of Model Variants.} We outlines the specifications for various model variants, detailing their hidden dimensions (Hidden) and the dimensions of attention heads (H. Dim). }
\vspace{-3mm}
    \centering
    \setlength{\tabcolsep}{3.4mm}
    \begin{tabular}{cccccc}
    \toprule
        Size & Layers  & Hidden & Head & H. Dim \\ \midrule
        70M  & 6 & 512 & 8 &  128  \\ 
        160M & 12 & 768 & 8 &  128 \\ 
        410M  & 26 & 1024 & 8 & 128  \\ 
        1B & 32 & 1536 & 16 & 128  \\ 
        3B & 35 & 2560 & 20 & 128  \\ 
        7B & 32 & 4096 & 32 &  128 \\ 
    \bottomrule
    \end{tabular}
    \vspace{-4mm}
    \label{model_sizes}
\end{table}

\subsection{Evaluation metrics}
\textbf{Perplexity} is a key metric used to evaluate the word prediction capabilities of causal language models. We use training loss and validation perplexity as evaluation metrics, with WIKITEXT-2 ~\cite{merity2016wikitext} and LAMBADA~\cite{paperno2016lambada} serving as benchmarks for comprehending complex, informative text and assessing their ability in narrative comprehension and contextual prediction, respectively. Lower perplexity indicates better predictive performance, suggesting that the model can more accurately capture language structure.

\noindent
\textbf{Knowledge retention}
Common sense reasoning (CSR) measures a model's ability to reason and understand everyday scenarios, indicating its practical real-world applicability. We report BoolQ \cite{clark2019boolq}, PIQA \citep{bisk2019piqa}, SIQA \citep{sap2019socialiqa}, HellaSwag \citep{zellers2019hellaswag}, WinoGrande \citep{sakaguchi2019winogrande}, ARC easy and challenge \citep{clark2018think} and OpenBookQA \citep{mihaylov2018suit}. We report 0-shot results for all benchmarks using LM-Eval-Harness \citep{leo2021evalharness}. 
We do not use the MMLU~\cite{hendryckstest2021} benchmark as it is more suitable for instruction-tuned models.

\noindent
\textbf{Information retrieval and generation}
The Needle in A Haystack (NIAH) benchmark is designed to evaluate the in-context retrieval capabilities of LLMs. We extend NIAH to present two difficulty levels. In Easy mode, both the question and its corresponding answer (QA pair) are embedded within a lengthy text, challenging the model to identify and respond to the query by locating the QA pair. This mode is particularly accessible for base models that have not undergone instruction tuning. In contrast, the standard mode places only the answer within the long context. Here, the model must comprehend the question, locate the relevant answer in the text, and provide a response. 

We quantify NIAH using three metrics: accuracy at a specific context length, weighted average accuracy, and NIAH score, as detailed in Appendix~\ref{sec:niah_mertics}. We use weighted average accuracy as our main evaluation metric for NIAH. Previous work \cite{ruler2024} assigns weights to context lengths in a linear scale. We assign weights for both depths and context lengths using a geometric progression for clearer distinction. We adopt \( w_{d_i} = w_{d_0} \alpha_d^{i-1} \) for depth and \( w_{c_i} = w_{c_0} \alpha_c^{i-1} \) for context length, where $w_{d_i}$ and $w_{c_i}$ are the weights for the $i$-th step, $\alpha_d$ and $\alpha_c$ are constants greater than 1. These weights form a map that modifies the average accuracy calculation.

In addition to NIAH, our evaluation also includes the SCROLLS benchmark~\cite{shaham2022scrolls}. SCROLLS assesses the model's abilities in information retrieval and generation across three distinct tasks: summarization, question answering, and natural language inference. We utilize the LM-Eval-Harness~\cite{leo2021evalharness} by configuring 0-shot, greedy-search evaluations and truncation of pre-training context length. 

\begin{table*}[ht]
\small
    \centering
     \caption{\textbf{Summary of Scaling Laws:} it illustrates the relationship between loss ($L$, left), the number of parameters ($N_{opt}$, middle), and corpus size ($D_{opt}$, right) with computation budget ($C$). It can be seen intuitively that under the same computation budget, linear complexity models consume more parameters and tokens while obtain lower loss. } 
     \vspace{-3mm}
     \setlength{\tabcolsep}{8mm}
    \begin{tabular}{c|c|c|c}
    \toprule
    Arch & $L(C)$ & $N_{opt}(C)$ & $D_{opt}(C)$ \\ \midrule
    LLaMA&  $3.7087C^{-0.0798}$  &  $(1.82\times 10^8)C^{0.7118}$  &  $(2.56\times 10^{10})C^{0.5102}$ \\ \midrule
    TNL&  $3.5391C^{-0.0768}$  &  $(2.74\times 10^8)C^{0.6470}$  &  $(4.43\times 10^{10})C^{0.4684}$ \\ \midrule
    HGRN2&  $3.4788C^{-0.0753}$  &  $(2.66\times 10^8)C^{0.6427}$  &  $(4.80\times 10^{10})C^{0.4500}$ \\ \midrule
    cosFormer2&  $3.5877C^{-0.0756}$  &  $(2.65\times 10^8)C^{0.6516}$  &  $(4.23\times 10^{10})C^{0.4529}$ \\ 
    \bottomrule
    \end{tabular}
    \label{table:scaling_law}
\end{table*}

\section{Scaling laws}\label{scaling_laws}
The concept of scaling laws involves four key factors: loss $L$, specifically the cross-entropy loss in a causal language modeling setting; model size $N$, which is determined by the number of model parameters; dataset size $D$, calculated as the number of training tokens; and computation budget $C$, represented by the total FLOPs used for training. $N_{opt}$ and $D_{opt}$ are the optimal model size and dataset size given a fixed computation budget.

Initially, we establish power law equations between $L$ and $C$. In this analysis, we adopt the approach introduced by \cite{jordan_chinchilla_2022}, treating the training loss as an unbiased estimate of the test loss. Subsequently, based on the fitted curve, we ascertain the optimal loss for specific FLOPs, enabling us to obtain coefficients for $N_{opt} \propto C^{a}$ and $D_{opt} \propto C^{b}$. When modeling the scaling trend of loss against factors such as $N$, $D$, and $C$, the original scaling laws \cite{jared_scaling_law_openai_2020} utilize the power function $L(X) = (X_0/X)^{\alpha_X}$, where $X$ represents the factor of interest. Subsequent studies \cite{Henighan_scaling_2020, clark_scaling_law_moe_icml_2022, jordan_chinchilla_2022, gao2024sparse} employ a more general power-law plus constant form, $L(X) = \epsilon + (X_0/X)^{\alpha_X}$, to achieve improved fitting. Here, the constant $\epsilon$ is interpreted as irreducible loss or the entropy of natural text \cite{jordan_chinchilla_2022}. In our case, we have simplified all forms of the power law and unified them into $L(X) = \beta_X X^{\alpha_X} $, which allows for a more intuitive comparison of the scaling capabilities of different models based on coefficients $\alpha_X$ and $\beta_X$.

\subsection{Training loss}
We aim to obtain the scaling laws of different models under the condition that only the model architectures are different. Therefore, we record the training losses of all models at the same interval, classify and fit them according to the power law mentioned above, and finally obtain the relation between $L$ and $C$, shown in the left column in Fig.\ref{fig:scaling_curve} and Table \ref{table:scaling_law}.
 
\subsection{Optimal model size and dataset size}
Given a fixed computation budget, we study how to allocate it to model parameter size and dataset size. Following \cite{jordan_chinchilla_2022}, we extract the minimal loss for each FLOP and consider this as the optimal loss for a given computation budget. To find model scaling exponent $a$ and data scaling exponent $b$ that satisfy $N_{opt} \propto C^{a}$ and $D_{opt} \propto C^{b}$, we use non-embedding parameters ~\citep{jared_scaling_law_openai_2020} as our vocabulary size is large and accounts for a large proportion of the parameters of the small model. Based on the above content, we fit all models to obtain the relationship between $C$ and $N_{opt}$ and $D_{opt}$, which can be seen in the last two columns of Fig.\ref{fig:scaling_curve} and Table \ref{table:scaling_law}.

\begin{figure*}[htbp]
\vspace{-2mm}
    \centering
        \includegraphics[width=0.30\linewidth]{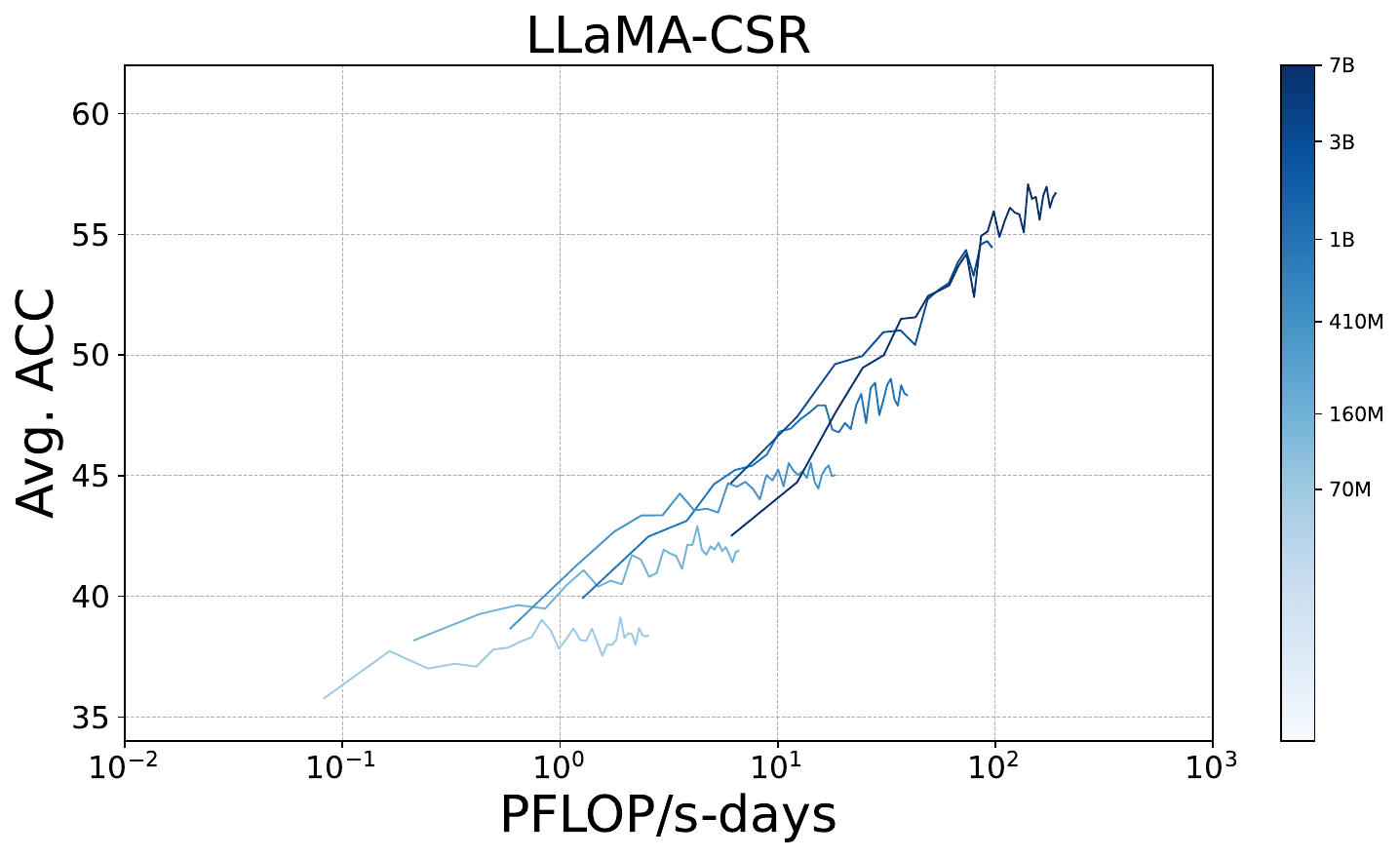} 
        \includegraphics[width=0.30\linewidth]{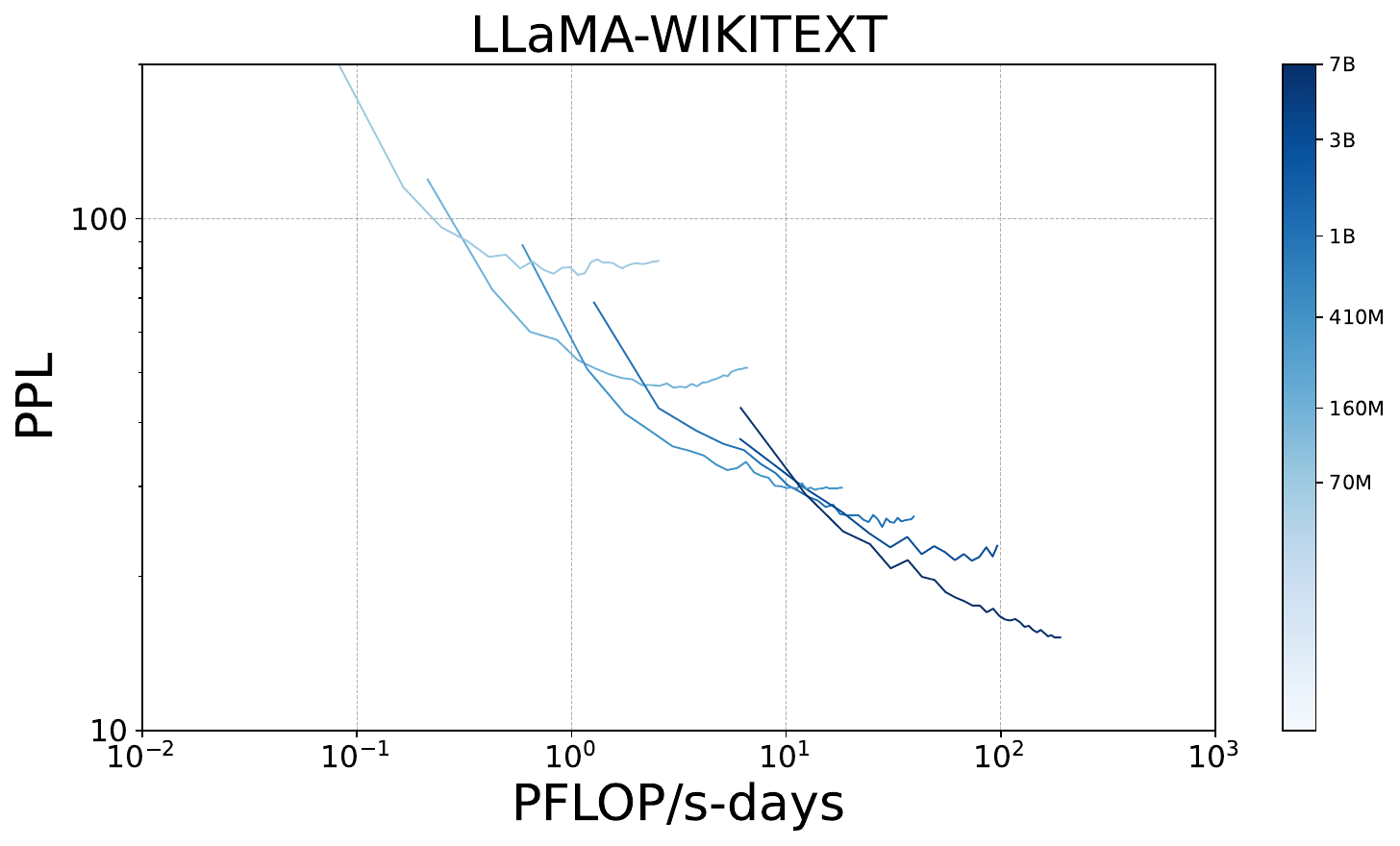} 
        \includegraphics[width=0.30\linewidth]{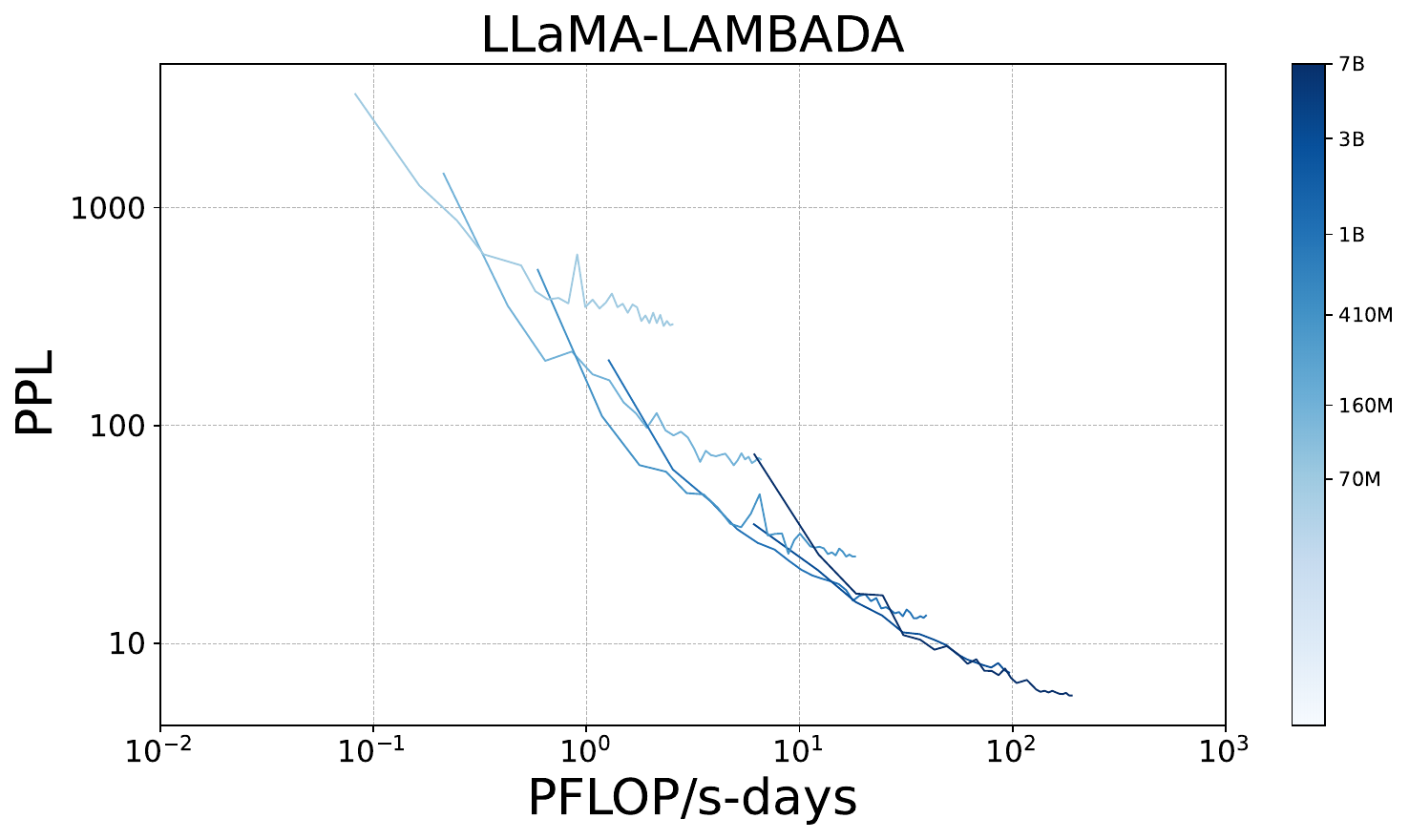} 
        \includegraphics[width=0.30\linewidth]{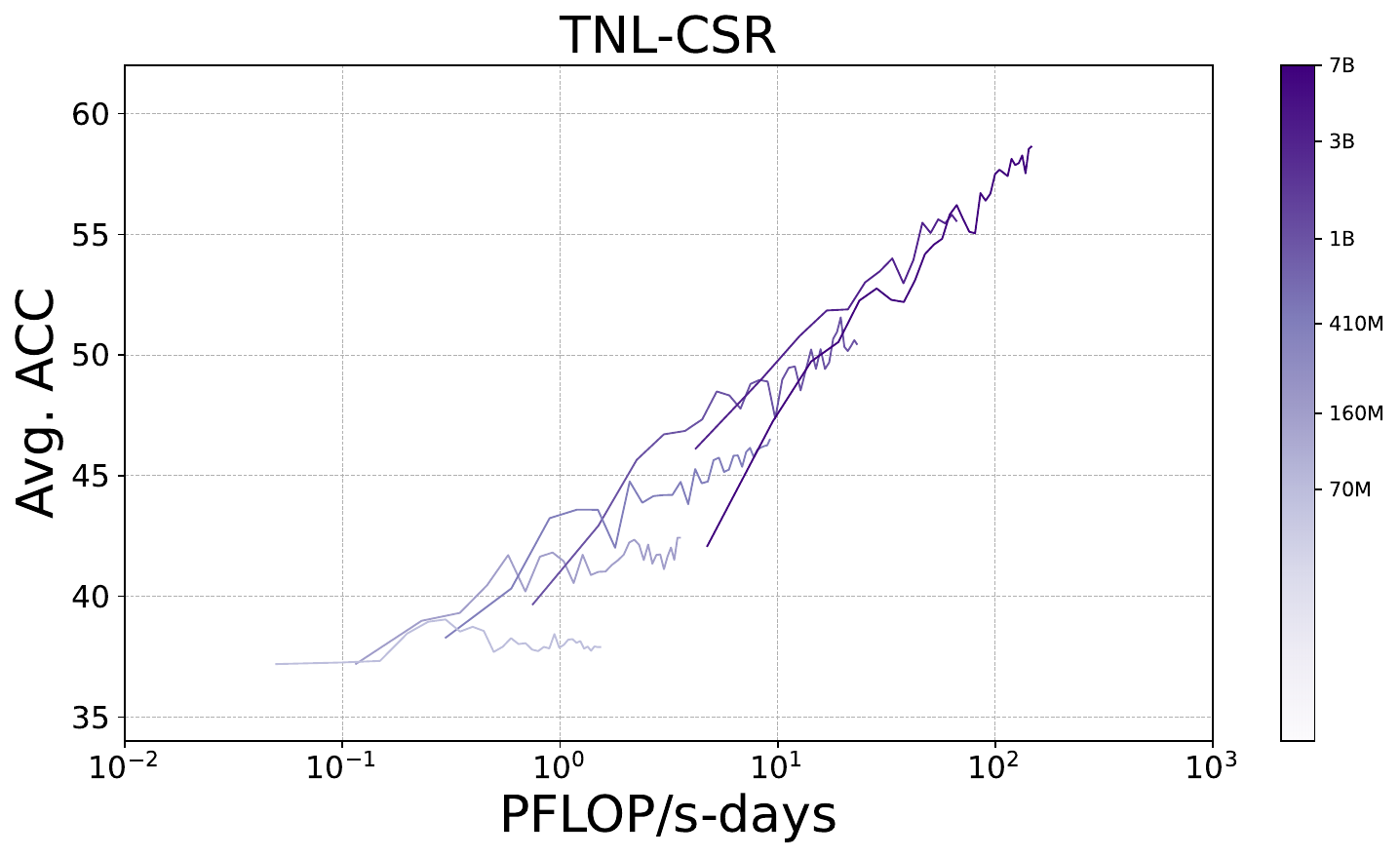} 
        \includegraphics[width=0.30\linewidth]{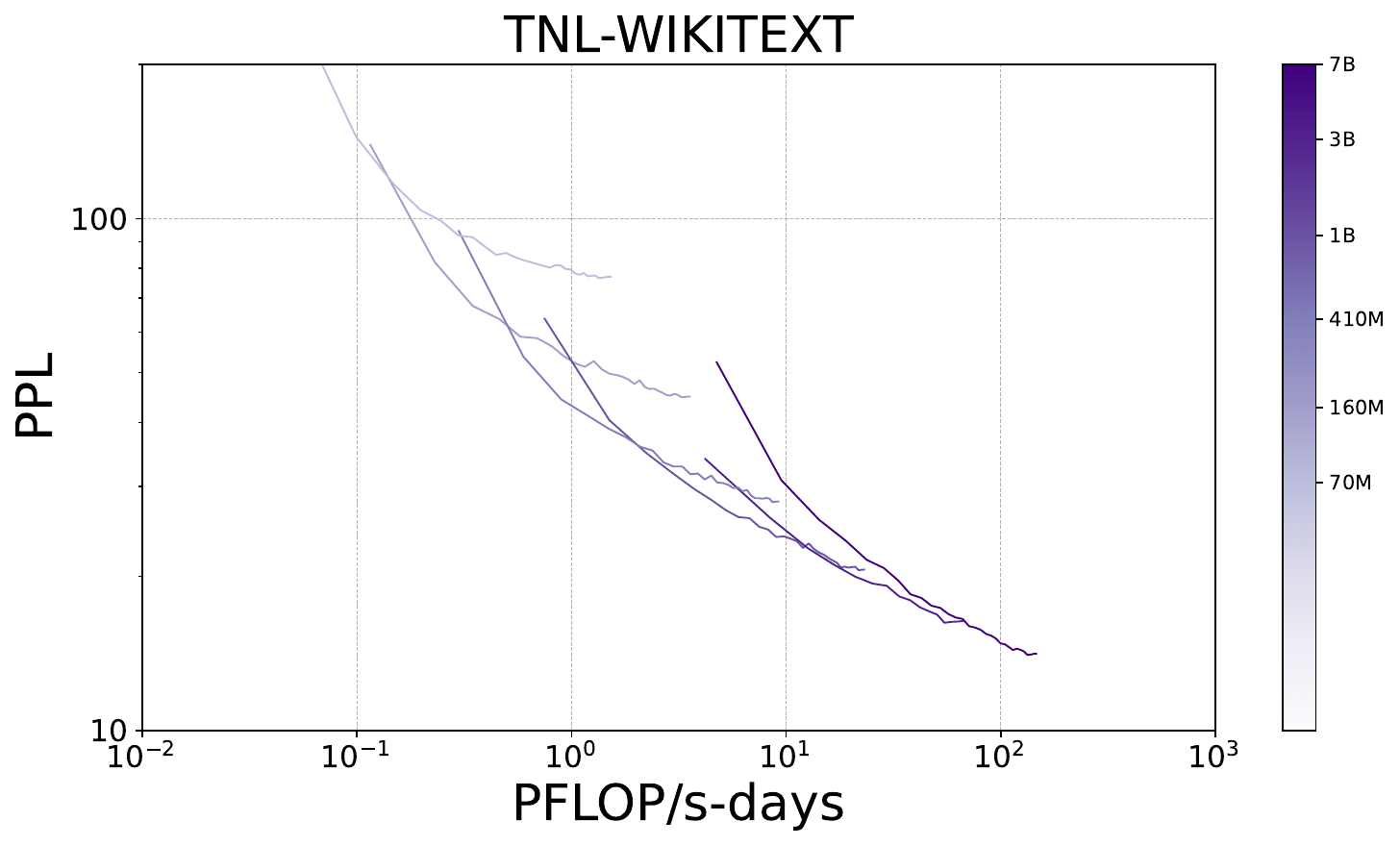} 
        \includegraphics[width=0.30\linewidth]{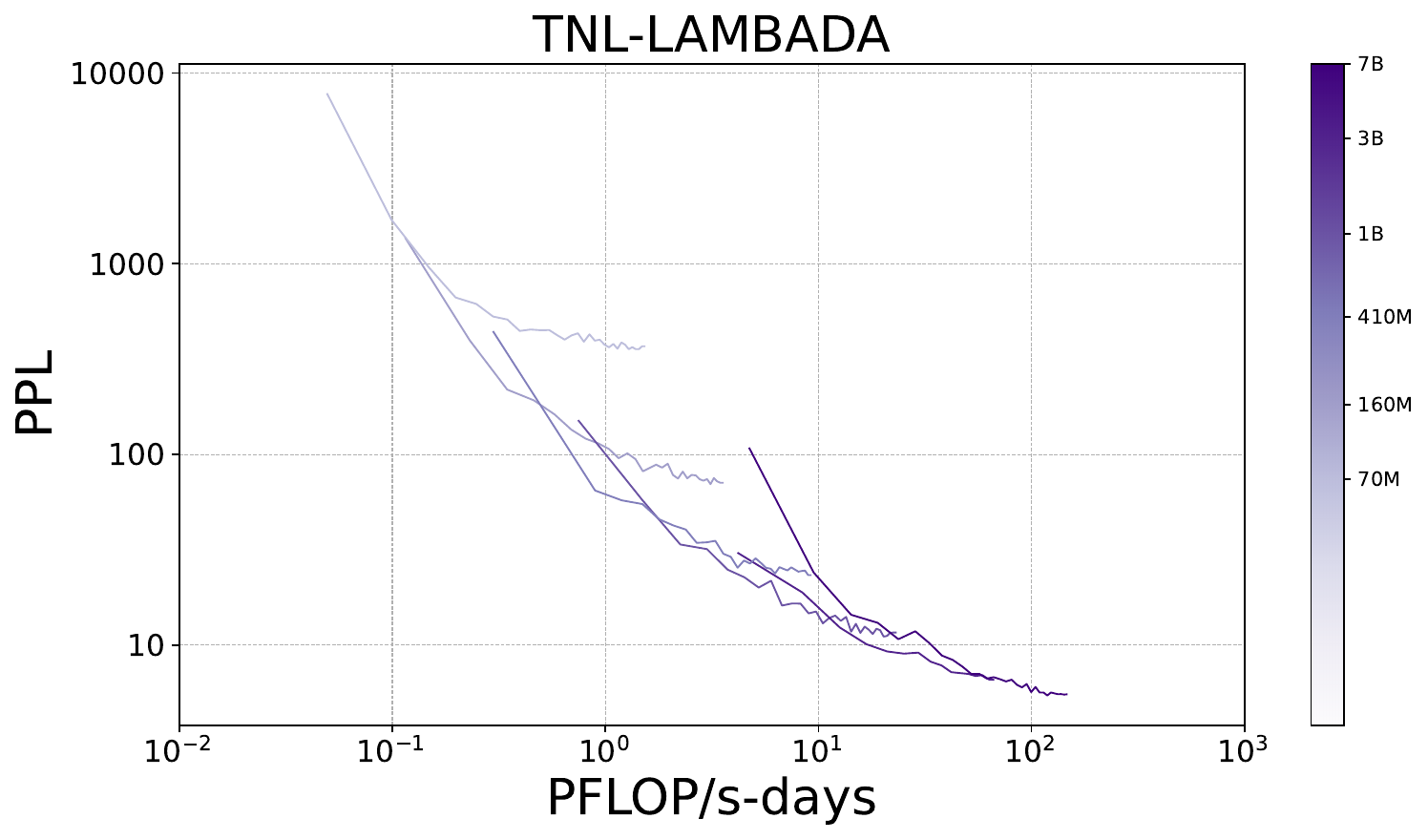} 
        \includegraphics[width=0.30\linewidth]{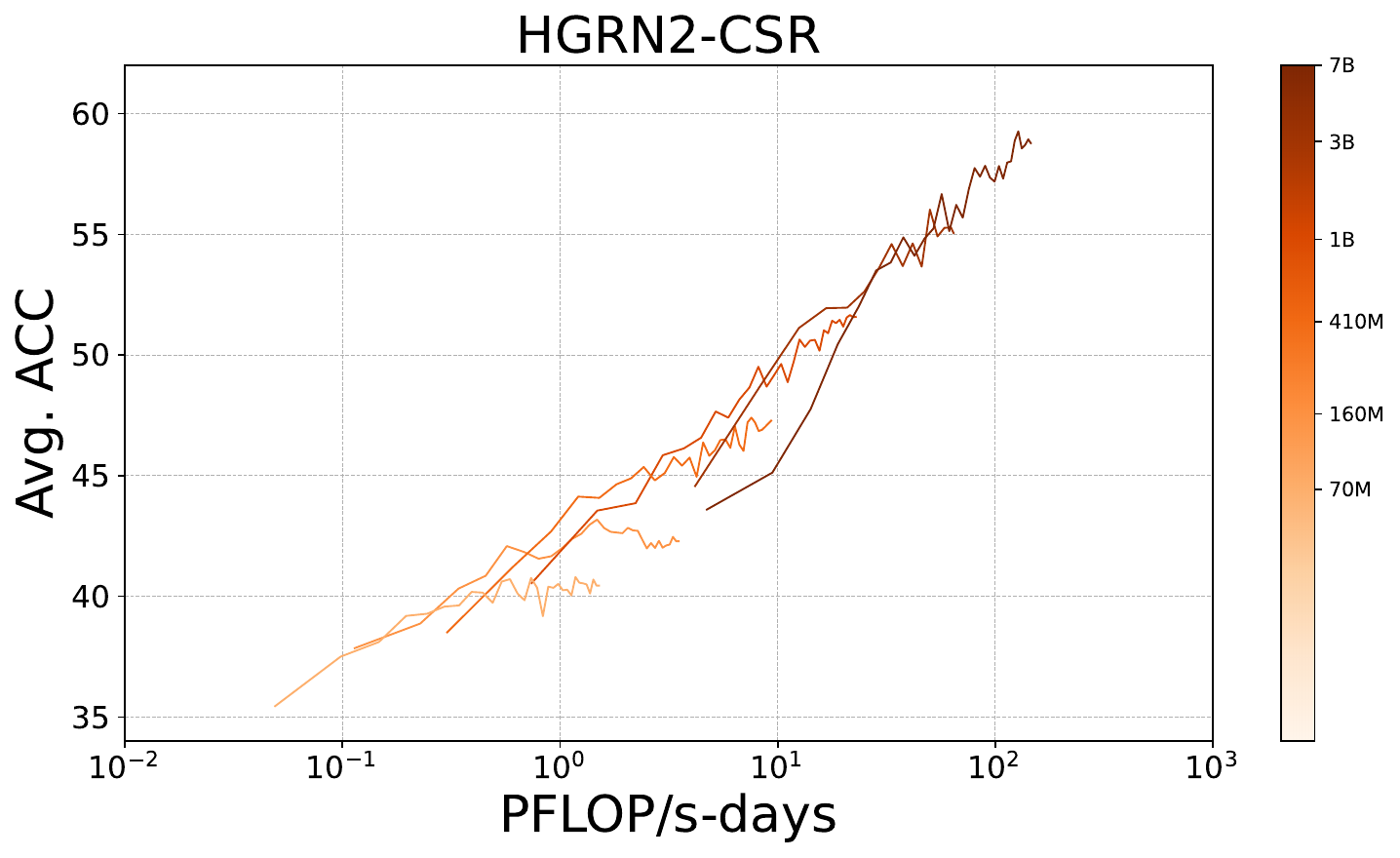} 
        \includegraphics[width=0.30\linewidth]{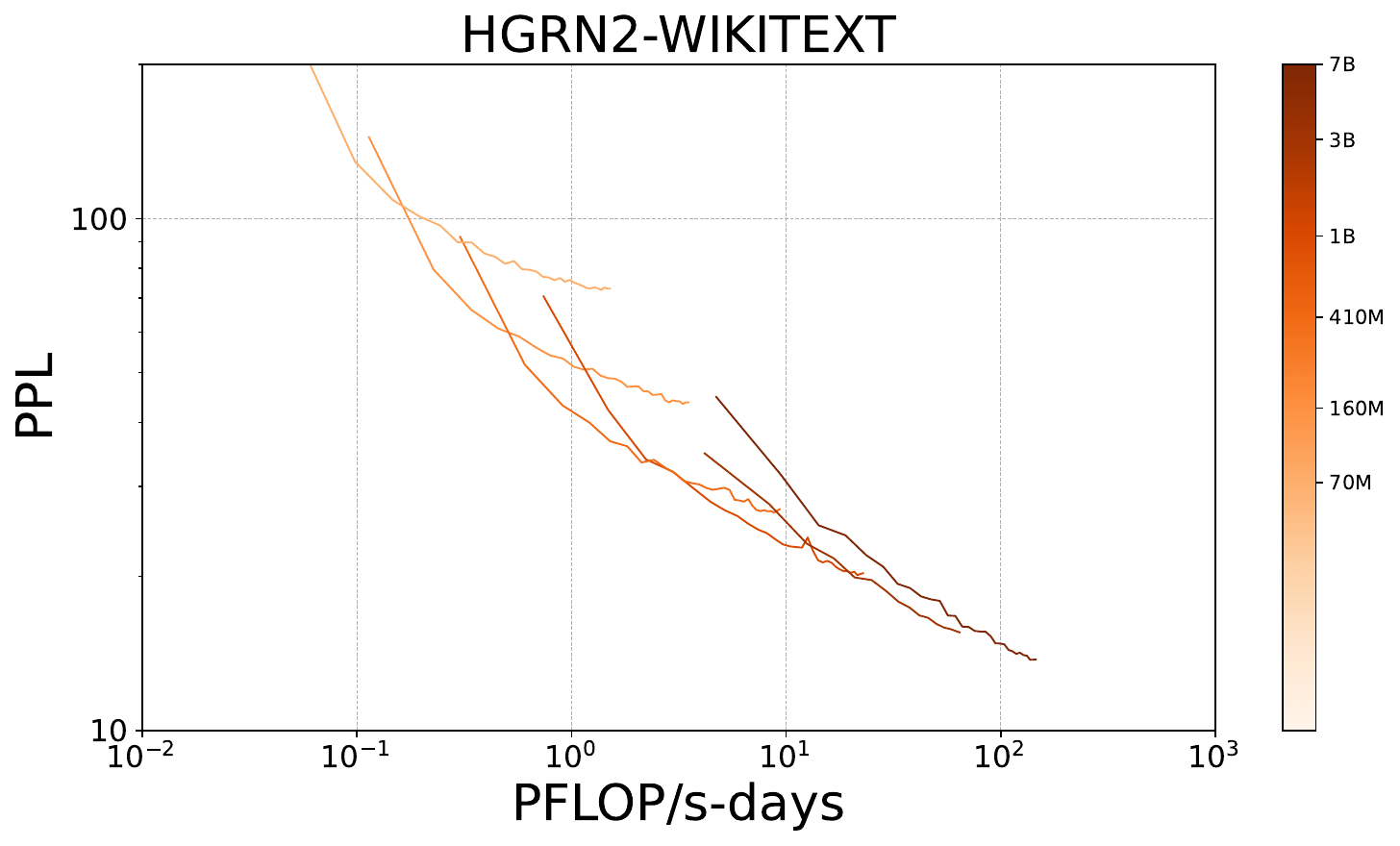} 
        \includegraphics[width=0.30\linewidth]{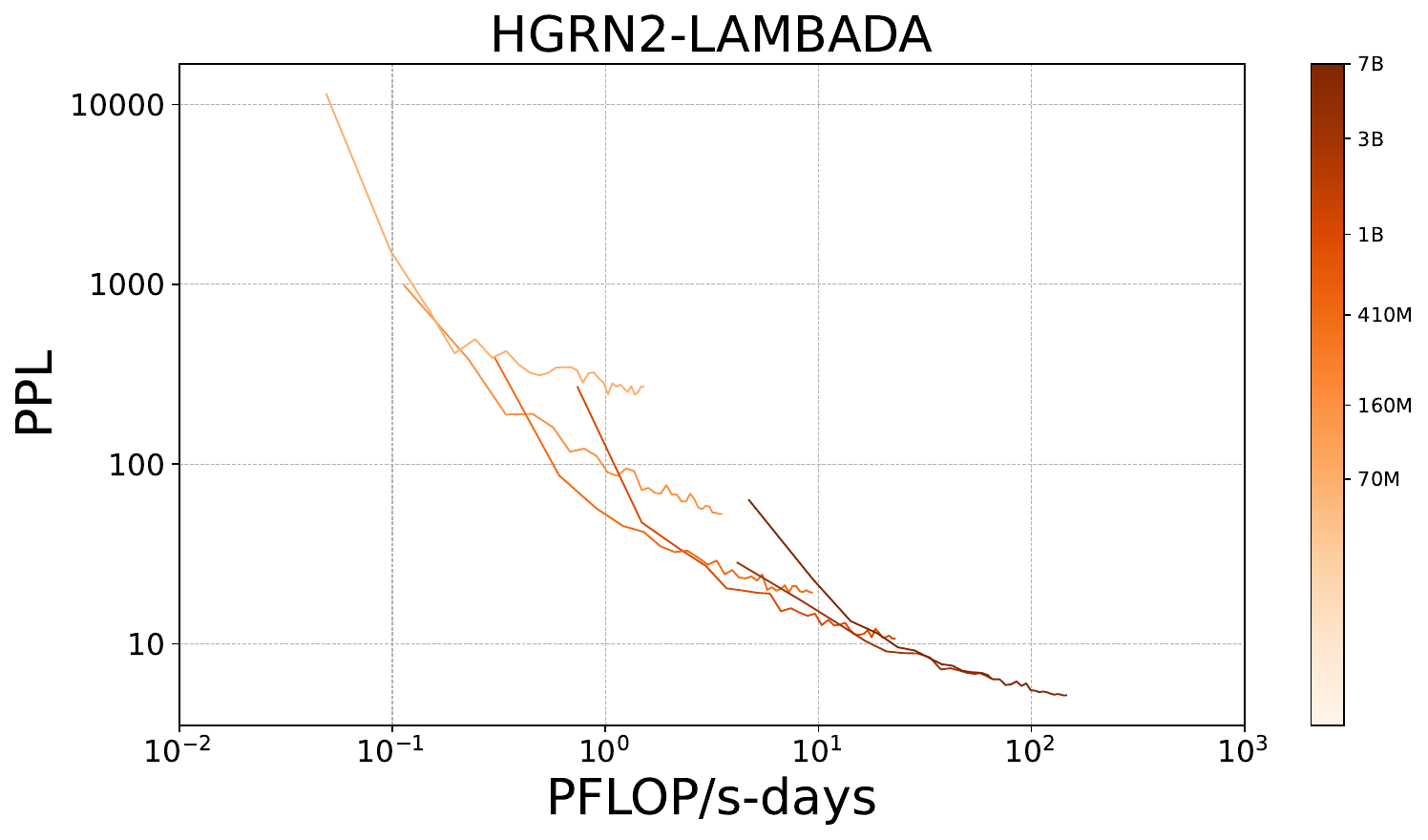} 
        \includegraphics[width=0.30\linewidth]{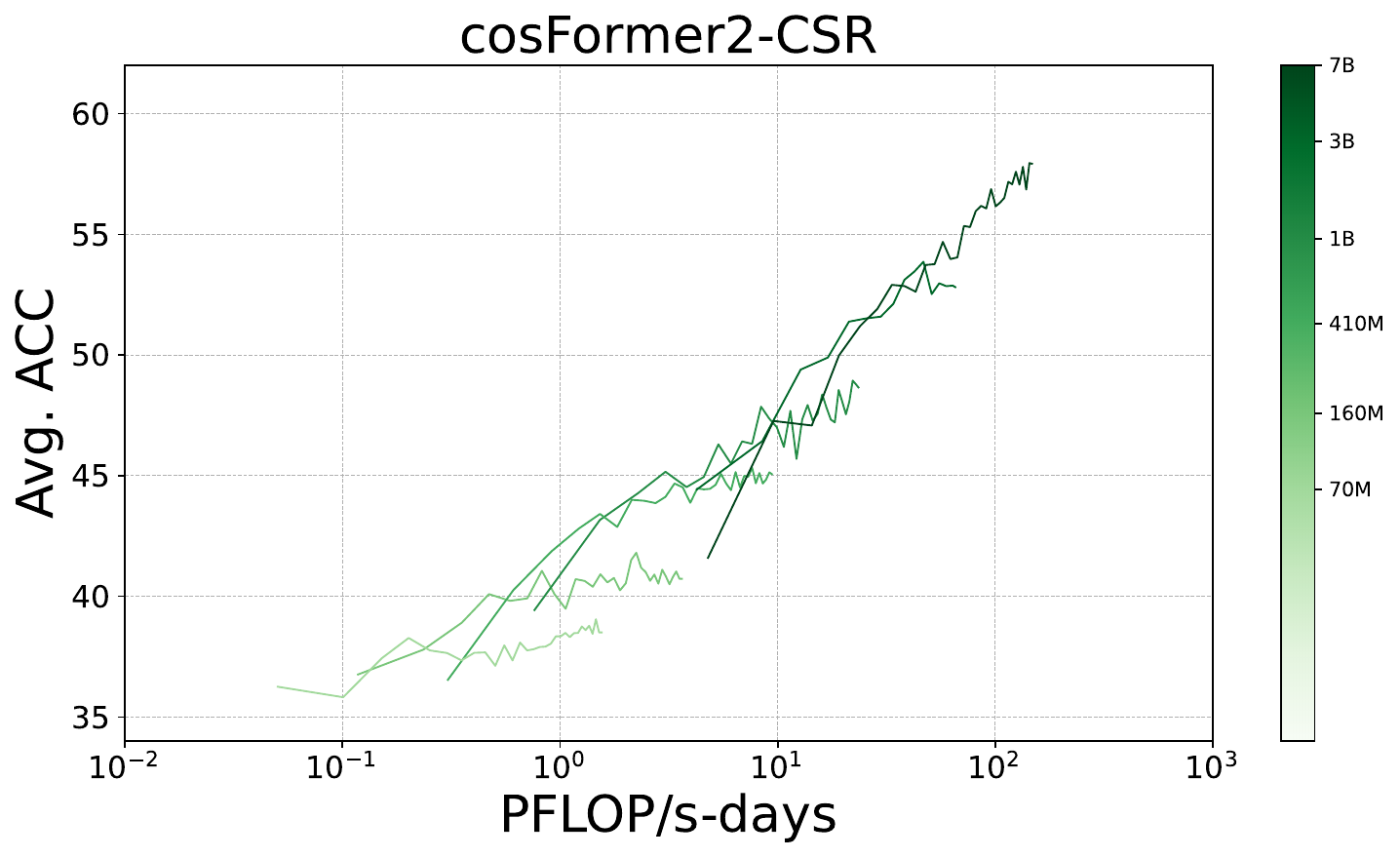} 
        \includegraphics[width=0.30\linewidth]{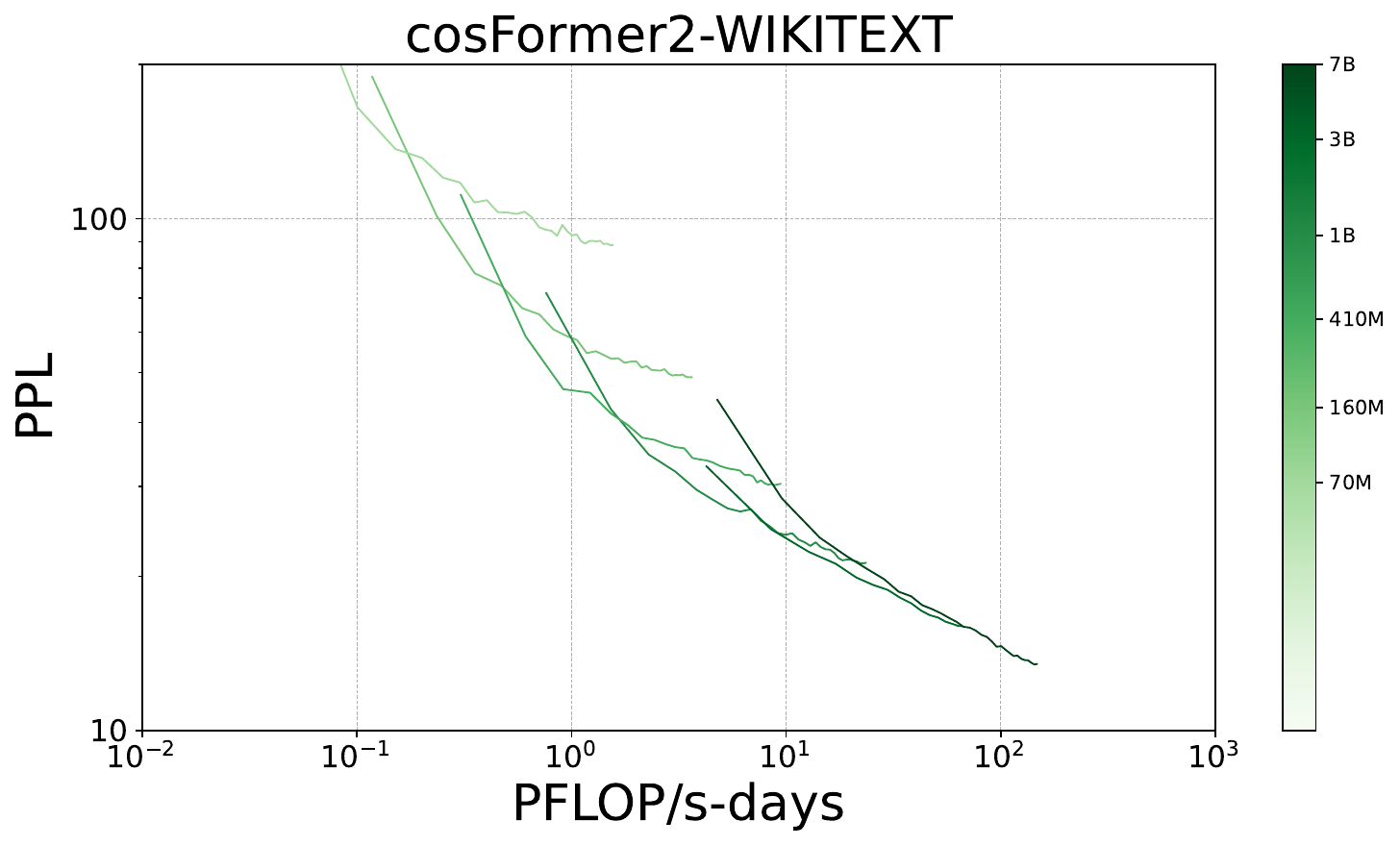} 
        \includegraphics[width=0.30\linewidth]{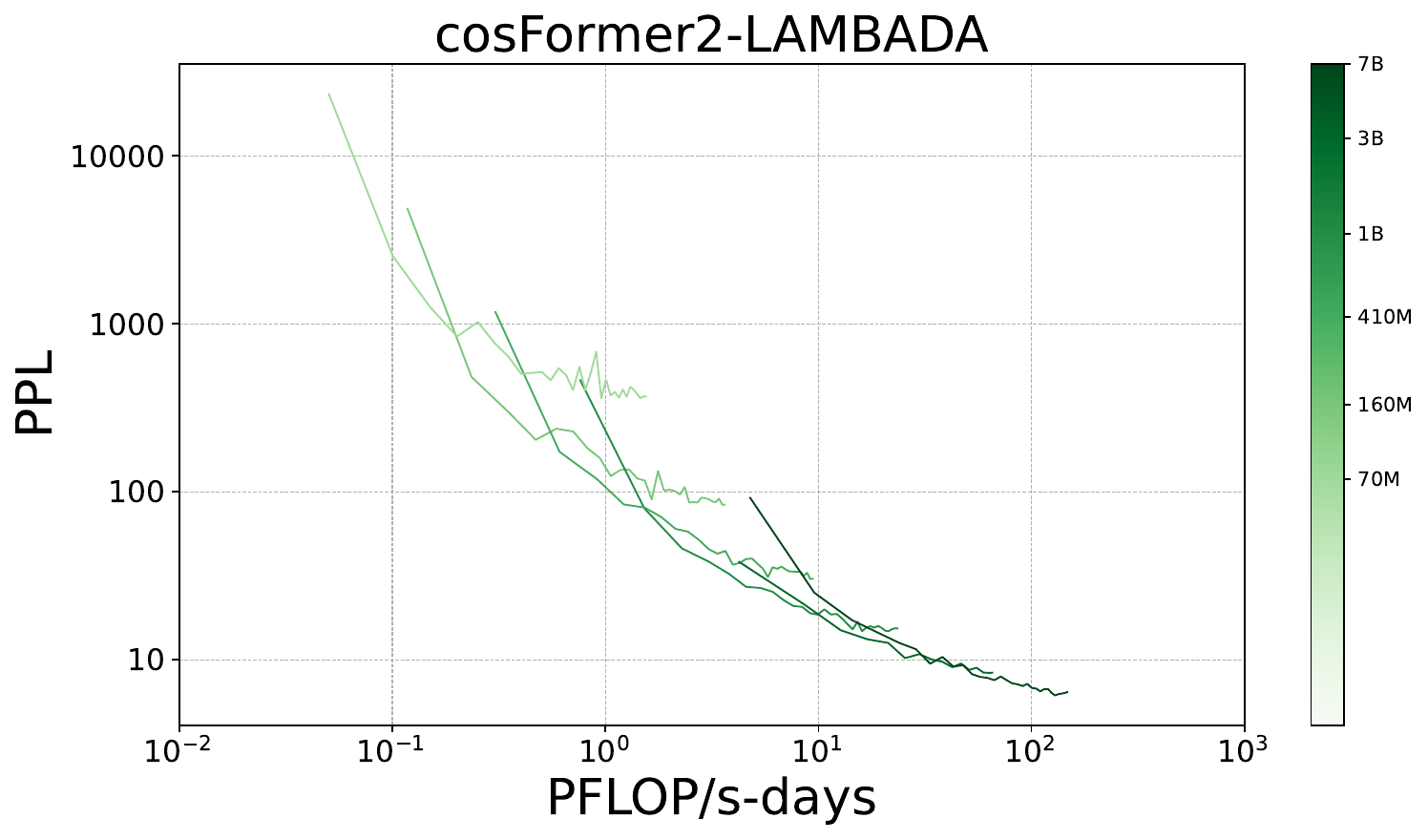} 
    \vspace{-4mm}
    \caption{\textbf{Comparative performance across distinct benchmarks} illustrating the scaling trends observed in evaluation metrics. The figure highlights the progressive improvement in model performance as the complexity and size of the models increase, underscoring the significance of scaling in enhancing benchmark outcomes.}
    \vspace{-3mm}
    \label{fig:scaling_eval}
    
\end{figure*}

\subsection{Downstream tasks}

Traditional scaling laws primarily focus on the relationship between computation power and training loss, typically measured by cross-entropy. However, this measure alone does not fully capture the capabilities of large language models~\cite{}. To address this, we expand our investigation to include scaling laws that correlate computation power with validation perplexity and common sense reasoning (CSR) scores in Fig.~\ref{fig:scaling_eval}. Additionally, we evaluate the performance of retrieval and generation capabilities using benchmarks such as Needle in A Haystack (NIAH) in Table~\ref{tab:scaling_law_benchmark} and SCROLLS in Table~\ref{tab:scaling_law_benchmark}. Each metric in these tasks provides a unique perspective on the strengths and limitations of LLMs.

\begin{table*}
\small

\centering
\caption{\textbf{Benchmark of Downstream Task: Common Sense Reasoning (CSR), Validation Perplexity, Needle in A Haystack (easy mode) and SCROLLS.} For CSR, Needle in A Haystack, and SCROLLS, higher scores indicate better performance. For Validation Perplexity, lower scores are preferable. PS: parameter size (billion). HS: HellaSwag. WG: WinoGrande. OBQA: OpenBookQA. WIKI: WIKITEXT-2. $\lambda$: LAMBADA. acc\_n.: acc\_norm. We provide the average score for CSR, the weighted average accuracy for NIAH, and the average score for SCROLLS. Detailed score breakdowns can be found in the "Experiments" section of the Appendix \ref{sec:exp}. }
\vspace{-3mm}
\setlength{\tabcolsep}{1.4mm}
\begin{tabular}{p{1.3cm}c|cccccccc|cc|cc}
\toprule
    Arch          & P.S.        & BoolQ & PIQA  & HS    & WG    & ARC-E & ARC-C & OBQA  & CSR & WIKI & $\lambda$  & NIAH &SCORLLS \\ \hline
                  & B           &acc &acc &acc\_n. &acc &acc &acc\_n.  &acc\_n.  & avg $\uparrow$ &ppl.$\downarrow$ &ppl. $\downarrow$ &w.a. $\uparrow$  &avg $\uparrow$\\ \midrule
    LLaMA         & 0.07 & 46.48 & 58.87 & 27.82 & 48.46 & 39.98 & 21.42 & 25.60 & 38.38            & 82.7              & 291.3           & 0.4           &\textbf{7.43} \\
    TNL           & 0.07 & 43.18 & 58.87 & 27.77 & 50.12 & 39.77 & 21.76 & 23.80 & 37.90            & 77.0              & 369.1           & \textbf{0.6}  &6.13\\
    HGRN2         & 0.07 & 56.57 & 59.19 & 28.05 & 52.01 & 38.64 & 22.61 & 26.00 & \textbf{40.44}   & \textbf{73.0}     & \textbf{270.1}  & 0.2           &7.32 \\
    cos2          & 0.07 & 47.61 & 60.94 & 28.12 & 49.72 & 37.33 & 22.18 & 23.60 & 38.50            & 88.8              & 369.9           & 0.1           &6.67\\ 
    \hline
    LLaMA        & 0.16 & 52.94 & 63.66 & 30.67 & 51.78 & 44.32 & 23.29 & 26.60 & 41.89             & 51.1              & 69.9            &6.0             &8.37 \\
    TNL          & 0.16 & 53.82 & 63.82 & 31.22 & 50.20 & 45.92 & 23.21 & 28.80 & \textbf{42.43}    & 44.9              & 71.1            &\textbf{7.5}    &7.77\\
    HGRN2        & 0.16 & 54.01 & 63.06 & 31.04 & 52.41 & 45.08 & 23.38 & 27.00 & 42.28             & \textbf{43.8}     & \textbf{52.8}   & 0.2            &8.29 \\
    cos2         & 0.16 & 45.47 & 63.28 & 29.72 & 52.41 & 44.49 & 22.53 & 27.20 & 40.73             & 49.0              & 83.8            &1.3             &\textbf{8.71}\\ 
    \hline
    LLaMA        & 0.41 & 54.04 & 67.19 & 38.75 & 52.17 & 49.24 & 23.72 & 30.00 & 45.02             & 29.8              & 25.1            &\textbf{52.3}             &10.51\\
    TNL          & 0.41 & 60.31 & 66.65 & 38.98 & 51.70 & 52.61 & 25.17 & 30.00 & 46.49             & 28.0              & 23.3            & 14.2            &7.55\\
    HGRN2        & 0.41 & 60.86 & 67.74 & 40.32 & 51.78 & 54.21 & 24.83 & 31.20 & \textbf{47.28}    & \textbf{27.0}     & \textbf{19.3}   & 4.8             &\textbf{10.93} \\
    cos2         & 0.41 & 57.40 & 66.27 & 36.65 & 50.59 & 51.81 & 23.72 & 29.00 & 45.06             & 30.3              & 30.3            & 9.6             &9.06\\
    \hline
    LLaMA        & 1 & 56.42 & 69.97 & 47.04 & 52.72 & 57.07 & 28.16 & 32.60 & 49.14                & 26.5              & 12.8            &\textbf{44.1}     &11.01 \\
    TNL          & 1 & 59.85 & 71.49 & 48.70 & 52.57 & 57.07 & 27.73 & 33.20 & 50.09                & 21.7              & 12.2            &28.0              &9.65\\
    HGRN2        & 1 & 59.17 & 71.65 & 49.52 & 54.38 & 60.27 & 28.07 & 33.40 & \textbf{50.92}       & \textbf{21.0}     & \textbf{10.9}   &10.0             &\textbf{11.08}\\
    cos2         & 1 & 44.28 & 70.73 & 45.55 & 50.51 & 55.22 & 27.30 & 31.00 & 46.37                & 21.2              & 15.5            & 10.9            &10.88\\ 
    \hline
    LLaMA        & 3 & 61.31 & 73.18 & 57.88 & 59.59 & 63.93 & 31.40 & 34.00 & 54.47                & 23.0              & 7.4             &\textbf{45.1}   &13.88 \\
    TNL          & 3 & 56.76 & 75.03 & 60.87 & 61.33 & 65.49 & 33.02 & 36.40 & \textbf{55.56}       & 16.4              & 6.6             &11.1             &12.26 \\
    HGRN2        & 3 & 55.47 & 74.10 & 61.48 & 58.64 & 65.61 & 34.47 & 35.60 & 55.06                & \textbf{15.6}     & \textbf{6.5}    &17.9             &\textbf{15.43}  \\
    cos2         & 3 & 50.92 & 74.27 & 57.38 & 57.30 & 63.22 & 31.40 & 35.20 & 52.81                & 16.0              & 8.4             &25.8             &12.75 \\
    \hline
    LLaMA        & 7 & 57.46 & 75.19 & 64.39 & 61.88 & 67.55 & 35.41 & 35.00 & 56.70                & 15.2              & 5.9             & \textbf{59.7}   &14.57\\
    TNL          & 7 & 62.63 & 76.22 & 66.29 & 61.48 & 67.76 & 38.23 & 37.80 & 58.63                & 14.1              & 5.5             & 20.5            &10.74 \\
    HGRN2        & 7 & 62.69 & 76.50 & 66.96 & 61.40 & 69.02 & 36.86 & 38.00 & \textbf{58.78}       & 13.8              & \textbf{5.2}    &30.8             &13.46\\
    cos2         & 7 & 65.02 & 76.33 & 63.93 & 59.19 & 66.96 & 36.43 & 37.60 & 57.92                & \textbf{13.5}     & 6.4             &23.6             &\textbf{15.15} \\
\bottomrule
\end{tabular}
\label{tab:scaling_law_benchmark}
\end{table*}

\noindent
\textbf{Validation perplexity}
Validation perplexity consistently decreases across all architectures as the number of model parameters increases, observed in both the WIKITEXT-2 and LAMBADA datasets. This trend underscores the scalability of linear complexity sequence models compared to vanila transformer models. When analyzing specific parameter sizes, HGRN2 architecture shows the best performance, closely followed by TNL. In the WIKITEXT-2 dataset, cosFormer2 surpasses LLaMA, while LLaMA performs better than cosFormer2 on the LAMBADA dataset.

\noindent
\textbf{CSR score}
The CSR scores for all linear complexity sequence models demonstrate scaling capabilities comparable to the transformer models. Specifically, HGRN2 is the only model that surpasses LLaMA at the 70M parameter level. In terms of 7B parameters, however, all linear complexity sequence models outscore LLaMA, suggesting that linear complexity sequence models exhibit enhanced scaling capabilities as the number of parameters increases. Notably, HGRN2 and TNL consistently outshine the other models. In contrast, cosFormer2 shows fluctuating performance compared to LLaMA.

Both CSR scores and validation perplexity highlight the strong scaling potential of HGRN2, TNL, and cosFormer2  in addressing linguistic and knowledge-based tasks in downstream tasks.

\noindent
\textbf{NIAH}
In evaluating the easy mode of the Needle in a Haystack (NIAH) task in 16K contexts, different architectures perform differently. Models with parameter sizes below 160M struggle to perform the tasks effectively. The vanilla transformer LLaMA maintains a success rate of about 50\%. TNL begins to show results in NIAH only after reaching 1B parameters, achieving a maximum success rate of about 10\%. Both HGRN2 and cosFormer2 start to display scaling capabilities in NIAH with over 410M parameters. Specifically, cosFormer2 achieves a maximum retrieval success rate of 25\% in 4K context, while HGRN2 performs slightly better with 30\% in 4.8K context. Linear complexity sequence models like cosFormer2 and HGRN2 tend to retrieve information from contexts shorter than their pre-training length of 8K. In terms of performance, the order is LLaMA $>$ cosFormer2 $=$ HGRN2 $>>$ TNL. Additionally, these linear complexity sequence models require large parameter sizes to effectively handle NIAH tasks.

\noindent
\textbf{SCROLLS}
Similar to NIAH, all architectures begin to effectively address the SCROLLS task starting at minimal 410 million parameters. All models with linear complexity sequences display a consistent scaling power comparable to LLaMA. TNL also requires a large parameter size (1B) for SCROLLS. The overall performance ranking is LLaMA and cosFormer2 at the top, followed by HGRN2, and then TNL.

\begin{table*}
\small
\centering
\caption{\textbf{Benchmark of Aspect Ratio and Model Capacity:} it covers models with \textit{1B} parameters each, featuring an \textit{8K} pre-training context. Key metrics include CSR, Validation PPL, NIAH (easy mode), and SCROLLS. HS: HellaSwag. WG: WinoGrande. OBQA: OpenBookQA. WIKI: WIKITEXT-2. $\lambda$: LAMBADA. $\text{acc\_n.}$: $\text{acc\_norm}$.}
\vspace{-3mm}
\setlength{\tabcolsep}{1.2mm}
\begin{tabular}{p{1.0cm}cc|cccccccc|cc|cc}
\toprule
Arch          & Dim & L. & BoolQ & PIQA  & HS    & WG    & ARC-E & ARC-C & OBQA  & CSR & WIKI & $\lambda$  & NIAH &SCORLLS \\ \hline
        &           &  &acc &acc &acc\_n. &acc &acc &acc\_n.  &acc\_n.  & avg $\uparrow$ &ppl.$\downarrow$ &ppl. $\downarrow$ &w.a. $\uparrow$  &avg $\uparrow$\\ \midrule

LLaMA & 1536 &32 & 60.98 & 69.91 & 46.74 & 54.85 & 56.94 & 28.50 & 30.00 & \textbf{49.70} & 26.2 & 13.4 & 44.1 &11.01 \\
LLaMA & 1792 &24  & 49.85 & 70.51 & 47.29 & 53.20 & 57.37 & 28.16 & 32.00 & 48.34 & \textbf{19.8} & 13.4 & \textbf{54.2} &\textbf{12.45} \\
LLaMA & 2048 &18  & 56.42 & 69.97 & 47.04 & 52.72 & 57.07 & 28.16 & 32.60 & 49.14 & 26.5 & \textbf{12.8} & 45.0 &12.20 \\
LLaMA & 3072 &8  & 55.44 & 69.53 & 44.25 & 51.62 & 52.78 & 26.54 & 30.20 & 47.19 & 24.8 & 16.7 & 46.0 &10.88 \\ \hline
cos2 & 1536  &32   & 56.42 & 70.57 & 45.99 & 52.01 & 57.49 & 26.11 & 32.00 & 48.66 & 21.3 & 15.3 & 10.9 &\textbf{10.88}\\
cos2 & 1792 &24  & 61.83 & 70.67 & 46.04 & 51.70 & 56.69 & 27.39 & 32.40 & \textbf{49.53} & \textbf{21.0} & \textbf{14.0} & 8.8  &10.31\\
cos2 & 2048 &18  & 44.28 & 70.73 & 45.55 & 50.51 & 55.22 & 27.30 & 31.00 & 46.37 & 21.2 & 15.5 & \textbf{12.3} &10.75 \\
cos2 & 3072 &8  & 43.52 & 69.86 & 43.38 & 50.83 & 53.79 & 26.62 & 32.60 & 45.80 & 23.4 & 18.3 & 6.5  &9.85\\
\bottomrule
\end{tabular}
\label{tab:aspect_ratio_benchmark}
\end{table*}

\begin{table*}
\small

\centering

\caption{\textbf{Benchmark of Pre-training Context Length:} it involves CSR, Validation PPL, NIAH (easy mode), and SCROLLS. All models tested have a parameter size of \textit{1B} and a hidden dimension of \textit{1536}. HS: HellaSwag. WG: WinoGrande. OBQA: OpenBookQA. WIKI: WIKITEXT-2. $\lambda$: LAMBADA. acc\_n.: acc\_norm.}
\vspace{-3mm}
\setlength{\tabcolsep}{1.4mm}
\begin{tabular}{p{1.2cm}c|cccccccc|cc|cc}
\toprule

Arch          & Len  & BoolQ & PIQA  & HS    & WG    & ARC-E & ARC-C & OBQA  & CSR & WIKI & $\lambda$  & NIAH &SCORLLS \\ \hline
        &   &acc &acc &acc\_n. &acc &acc &acc\_n.  &acc\_n.  & avg $\uparrow$ &ppl.$\downarrow$ &ppl. $\downarrow$ &w.a. $\uparrow$  &avg $\uparrow$\\ \midrule

TNL   & 2K  & 61.96 & 72.03 & 49.94 & 54.38 & 57.58 & 28.33 & 31.20 & 50.77 & 26.4 & \textbf{10.5} & 8.0  & 9.02  \\
TNL   & 4K  & 61.80 & 72.31 & 49.88 & 55.33 & 57.91 & 29.10 & 32.20 & \textbf{51.22} & 23.9 & 10.7 & 12.1 & \textbf{11.79} \\
TNL   & 8K  & 54.53 & 71.60 & 49.94 & 55.41 & 58.50 & 29.01 & 34.20 & 50.45 & \textbf{20.6} & 11.7 & \textbf{28.0} & 9.65  \\
TNL   & 16K & 49.05 & 71.06 & 49.51 & 51.46 & 57.53 & 28.16 & 31.60 & 48.34 & \textbf{20.6} & 11.3 & 14.2 & 8.74  \\
\hline
HGRN2 & 2K  & 62.23 & 72.25 & 50.68 & 54.70 & 60.02 & 30.03 & 33.40 & \textbf{51.90} & 22.2 & \textbf{10.0} & 2.3  & 11.60 \\
HGRN2 & 4K  & 61.77 & 70.95 & 51.21 & 53.59 & 60.19 & 30.89 & 31.20 & 51.40 & 20.9 & 10.6 & 2.1  & 11.46 \\
HGRN2 & 8K  & 59.54 & 71.82 & 50.65 & 54.85 & 60.40 & 29.61 & 34.20 & 51.58 & 20.3 & 10.7 & \textbf{10.0} & 11.08 \\
HGRN2 & 16K & 54.92 & 72.03 & 50.37 & 55.25 & 59.01 & 28.92 & 32.00 & 50.36 & \textbf{20.1} & 11.2 & 8.8  & \textbf{12.23} \\
\hline
cos2  & 2K  & 60.95 & 70.35 & 47.37 & 53.43 & 56.44 & 27.30 & 31.00 & \textbf{49.55} & 22.7 & \textbf{12.0} & 6.8  & 10.93 \\
cos2  & 4K  & 55.66 & 70.08 & 47.05 & 50.99 & 55.35 & 27.13 & 33.00 & 48.46 & \textbf{21.3} & 12.5 & 6.5  & 11.79 \\
cos2  & 8K  & 56.42 & 70.57 & 45.99 & 52.01 & 57.49 & 26.11 & 32.00 & 48.66 & \textbf{21.3} & 15.3 & \textbf{10.9} & 10.88 \\
cos2  & 16K & 62.51 & 69.86 & 44.61 & 52.72 & 54.25 & 26.02 & 32.20 & 48.88 & 22.1 & 16.9 & 9.6  & \textbf{13.04} \\
\bottomrule
\end{tabular}
\label{tab:seq_length}
\vspace{-3mm}
\end{table*}

\section{Discussion}\label{evaluation}

\subsection{Aspect ratio and model capacity}

Under the same model parameters, we can tweak the model architecture by adjusting the aspect ratio (hidden dimension and layers) and the dimension of attention heads. We analyze the aspect ratio for a 1B parameter model in Table~\ref{tab:aspect_ratio_benchmark}. Since we need to fix the parameters, a higher hidden dimension indicates fewer layers. For both LLaMA and cosFormer2, a hidden dimension below 2048 proves beneficial for CSR and validation perplexity.

In tasks involving retrieval and generation, LLaMA and cosFormer2 consistently show similar results for CSR and validation perplexity. However, a larger aspect ratio can lead to failures in these tasks. Specifically, cosFormer2 with a 3072 dimension results in a collapse in NIAH and SCROLLS evaluations. Models with linear complexity in their sequences are more sensitive to aspect ratio changes than the vanilla transformer models.

\subsection{Pre-training context length}

We further examined the impact of pre-training on the performance of downstream tasks. Table~\ref{tab:seq_length} indicates that CSR and validation perplexity for all linear complexity sequence models remains unaffected by pre-training context lengths of 2K, 4K, and 8K. However, extending the context length to 16K slightly degrades performance. 

When increasing the pre-training context length from 8K to 16K, all linear complexity sequence models fail to retrieve longer contexts in both NIAH and SCROLLS tasks. In contrast, LLaMA's retrieval capabilities double when the pre-training context length is increased to 16K from 8K. Moreover, shorter pre-training context lengths have a detrimental effect on retrieval tasks for linear complexity sequence models.

\subsection{Decay types for linear complexity models}
As outlined in our preliminary study, TNL, HGRN2, and cosFormer2 utilize three distinct decay strategies for linear complexity sequence models: data-independent decay, data-dependent decay, and no decay. Analyzing Tables~\ref{tab:scaling_law_benchmark} and~\ref{tab:seq_length}, we find that cosFormer2, which employs a no-decay linear attention mechanism, performs worse than TNL (which uses data-independent decay) in terms of CSR and validation perplexity. However, cosFormer2 shows superior retrieval capabilities in NIAH and SCROLLS tasks. Meanwhile, HGRN2, which uses data-dependent decay RNN, displays performance on par with no-decay linear attention in retrieval and generation tasks and matches the performance of data-independent decay in CSR and validation perplexity.

\subsection{Retrieval capacity of linear models}
Our experimental results suggest that linear complexity models have limited capacity in retrieval tasks. 
This is because linear models maintain a fixed size of hidden space, which makes it difficult to retain input information precisely. Then a natural question arises: "Why does softmax attention have this capability? Can we address this limitation for linear models?"

To answer this question, let us consider a softmax attention:
\begin{equation}
\mathbf{O} = \mathrm{Softmax}(\mathbf{Q} \mathbf{K}\top / \sqrt{d}) \mathbf{V}.
\end{equation}
It can be rewritten into a linear recurrent form as:
\begin{equation}
\begin{gathered}
s_t^0=0, 
s_t^{j}=s_{t}^{j-1}+\exp(\mathbf q_t \mathbf k_j^T/\sqrt d),\\
\mathbf o_t^j =(s_t^{j-1}/s_t^j)\mathbf o_{t}^{j-1} +(1-s_t^{j-1}/s_t^j) \mathbf v_j,  \\
\mathbf o_t=\mathbf o_t^t , 
j=1,\ldots, t.
\end{gathered}
\end{equation}
Note that the linear recurrence form of Linear Attention is as follows:
\begin{equation}
\begin{gathered}
\mathbf {kv}_t=0, 
\mathbf {kv}_t=\mathrm{diag}\{\lambda_t\}\mathbf {kv}_{t-1}+ \mathbf k_t  \mathbf v_t^\top\\
\mathbf o_t= \mathbf {kv}_t^\top \mathbf q_t , 
j=1,\ldots, t.
\end{gathered}
\end{equation}
% This means that softmax attention, at each time step $t$, needs to perform an $O(t)$ linear recurrence based on the current $\mathbf q_t$ (calculating $s_t^j$ and then performing the recursion), which we refer to as GTB. In contrast, Linear Attention does not perform this operation.

The softmax attention can be interpreted as an additive linear RNN ~\cite{qin2024you}. At each time step $t$, the hidden space is recomputed starting from the initial time $t_0 = 1$, a process referred to as "Going Through a Book (GTB)". This approach allows the model to accurately retain input information by revisiting previous information. For linear models, there is no recomputation. Therefore, linear models struggle to accurately retain input data without a GTB process.

\section{Related work}\label{related_work}
Scaling laws in large language models aim for an ideal balance between increasing the number of parameters and enlarging the training corpus, given limited computation resources \cite{jared_scaling_law_openai_2020, Henighan_scaling_2020, danny_scaling_transfer_openai_2021, jordan_chinchilla_2022, clark_scaling_law_moe_icml_2022}.
The initial scaling laws \cite{jared_scaling_law_openai_2020} use the test-time cross-entropy loss as a regression target to investigate its power-law correlations with model size, dataset size and training computation budget. \citet{jordan_chinchilla_2022} use three approaches to find the optimal model size and dataset size given a fixed computation budget. By 1) freezing model size and varying number of training tokens, 2) fixing FLOPs and changing model sizes and dataset sizes and 3) directly solving a constrained optimization equation, they conclude that models and the training corpus should be scaled equally when enlarging computing resources. They use the revised scaling law to train a compute-optimal model, \textit{Chinchilla}, that stands out across various benchmarks.
Other works extend scaling laws to multiple modalities \cite{Henighan_scaling_2020}, mixture of expert models \cite{clark_scaling_law_moe_icml_2022} and reinforcement learning \cite{jacob_scaling_law_rl_openai_2023}. 
Recently, \citet{hui_unraveling_2024,Bi2024deepseek} studied the influence of additional factors such as learning rate, context length, and batch size on the scaling-law coefficients. \cite{Isik2024downstream} studies scaling laws of downstream task performance in a transfer learning setting for the machine translation task.

\section{Conclusion}
Our comprehensive study has demonstrated that linear complexity language models, including TNL, HGRN2, and cosFormer2, exhibit competitive scaling capabilities akin to transformer-based models while also showcasing enhanced linguistic proficiency and knowledge retention. With rigorous training across a vast parameter range and extensive evaluation of diverse tasks, our findings validate these models as promising contenders for future large-scale language model development. 

\section*{Limitations}
\begin{itemize}
    \item We train all models on a fixed dataset, thus overlooking the influence of data distribution on scaling laws.
    \item For each model architecture, we only experiment with six different model sizes, resulting in fewer data points than previous works in terms of fitting the loss-computation curve.
    \item We use fixed learning rate scheduler and batch size across experiments.
\end{itemize}

\bibliography{reference}

\clearpage
\appendix

\section{Appendix}

\subsection{Model parameters and FLOPs}\label{model_params_flops}
Here we provided detailed FLOPs and the number of model parameters calculation for each model architecture. Some operations are omitted for simplicity, e.g. the FLOPs and parameters related to positional encoding, normalization, activation functions, and softmax of the final head, if applicable. We parameterize models with the following notations: 
\begin{itemize}
  \item $d$: attention hidden dimension.
  \item $h$: number of heads in attention. 
  \item $g$: GLU hidden dimension. (In all scenarios, we use $g=8/3 d$.)
  \item $l$: number of layers.
  \item $n$: input sequence length.
  \item $v$: vocabulary size.
  \item $b$: batch size.
  \item $t$: output gate bottleneck dimension.
  \item $B$ lightning attention/flash linear attention block size. (In all scenarios, we use $B=d/h$.)
  \item $e$: Tpe hidden dimension.
\end{itemize}

Similar to \citep{jared_scaling_law_openai_2020, jordan_chinchilla_2022}, we use a factor of 2 to represent the multiplication-accumulation in matrice products, and a factor of 3 to include both the forward and the backward pass. In the following discussion, we assume $f$ is the swish function.

\begin{table*}[t]
\small
    \centering
     \caption{\textbf{Model Parameters and FLOPs.} The listed numbers of parameters do not include the embedding parameters.}
     \setlength{\tabcolsep}{3.6 mm}
    \begin{tabular}{c|c|c|c|c|c|c|c|c|c}
    \toprule
        {Configuration} &{b}      &{n}       &{l}   &{d}      &{h}  &{g}      & {v}     &  {\makecell[c]{Parameters\\(Million)}}            & {\makecell[c]{PFLOPs /\\Step}} \\ \midrule
        LLaMA-70M     &480    &8192    &6   &512    &4  &1536   &100280 &20.5                       &1.6\\
        LLaMA-160M    &480    &8192    &12  &768    &6  &2048   &100280 &85.0                     &5.6\\
        LLaMA-410M    &480    &8192    &26  &1024   &8  &2816   &100280 &334.1                      &18.0\\
        LLaMA-1B      &480    &8192    &32  &1536   &16 &5632   &100280 &906.2                     &40.4\\
        LLaMA-3B      &480    &8192    &35  &2560   &20 &6912   &100280 &2775.8                     &99.6\\
        LLaMA-7B      &480    &8192    &32  &4096   &32 &11008  &100280 &6476.5                     &202.7\\ \midrule
        TNL-70M       &480    &8192    &6   &512    &4  &1536   &100280 &21.2                       &0.5\\
        TNL-160M      &480    &8192    &12  &768    &6  &2048   &100280 &87.3                      &2.2\\
        TNL-410M      &480    &8192    &25  &1024   &8  &2816   &100280 &327.7                      &7.9\\
        TNL-1B        &480    &8192    &32  &1536   &16 &5632   &100280 &918.6                     &22.3\\
        TNL-3B        &480    &8192    &35  &2560   &20 &6912   &100280 &2798.4                     &66.6\\
        TNL-7B        &480    &8192    &32  &4096   &32 &11008  &100280 &6509.6                     &154.4\\ \midrule
        HGRN2-70M     &480    &8192    &6   &512    &4  &1536   &100280 &20.5                       &0.5\\
        HGRN2-160M    &480    &8192    &12  &768    &6  &2048   &100280 &84.9                      &2.1\\
        HGRN2-410M    &480    &8192    &26  &1024   &8  &2816   &100280 &334.0                      &8.0\\
        HGRN2-1B      &480    &8192    &32  &1536   &16  &5632   &100280 &906.0                     &22.0\\
        HGRN2-3B      &480    &8192    &35  &2560   &20  &6912   &100280 &2775.5                     &66.0\\
        HGRN2-7B      &480    &8192    &32  &4096   &32  &11008  &100280 &6476.1                     &153.6\\ \midrule
        cos2-70M     &480    &8192    &6   &512    &4  &1536   &100280 &    21.3                    &0.5\\
        cos2-160M    &480    &8192    &12  &768    &6  &2048   &100280 &    87.5                   &2.3\\
        cos2-410M    &480    &8192    &25  &1024   &8  &2816   &100280 &     328.0                   &8.1\\
        cos2-1B      &480    &8192    &32  &1536   &16  &5632   &100280 &  919.1                     &22.7\\
        cos2-3B      &480    &8192    &35  &2560   &20  &6912   &100280 & 2799.0                     &67.4\\
        cos2-7B      &480    &8192    &32  &4096   &32  &11008  &100280 & 6511.0                     &155.6\\
    \bottomrule
    \end{tabular}
    \label{table:cmos}
    \vspace{-4mm}
\end{table*}

\subsubsection{Transformer-LLaMA}
\textbf{Equation}
Input embedding:
\begin{equation}
\small
\begin{aligned}
&\mathbf X^{(1)}= \mathrm{Lookup}(\mathbf X^{(0)}, \mathbf W_{in}), \\
&\mathbf X^{(0)}\in \mathbb R^n, \mathbf W_{in} \in \mathbb R^{d \times v}.
\end{aligned}
\end{equation}
Token mixer($s$-th layer):
\begin{equation}
\small
\begin{aligned}
&\mathbf {\bar X}^{(s)}= \mathrm{Norm}(\mathbf X^{(s)}), \\
&\mathbf Q^{(s)}_i,\mathbf K^{(s)}_i,\mathbf V^{(s)}_i=
\mathbf  {\bar X}^{(s)} \mathbf W_{q_i}^{(s)}, \mathbf  {\bar X}^{(s)} \mathbf W_{k_i}^{(s)}, \mathbf  {\bar X}^{(s)} \mathbf W_{v_i}^{(s)}, \\
&\mathbf { O}^{(s)}_i=\mathrm{Softmax}\left(\mathbf Q^{(s)}_i {\mathbf K^{(s)}_i}^\top /\sqrt{d/h} \right)\mathbf V^{(s)}_i, \\
&\mathbf O^{(s)}=\mathrm{Concat}[\mathbf { O}^{(s)}_1,\ldots, \mathbf { O}^{(s)}_h]\mathbf W_o^{(s)} + \mathbf X^{(s)},\\
&\mathbf X^{(s)}\in \mathbb R^{n\times d}, \mathbf W_{o}^{(s)}\in \mathbb R^{d\times d},  \\
&\mathbf W_{q_i}^{(s)},\mathbf W_{k_i}^{(s)},\mathbf W_{v_i}^{(s)} \in \mathbb R^{d\times d / h}, i=1,\ldots, h.
\end{aligned}
\end{equation}
Channel mixer($s$-th layer):
\begin{equation}
\small
\begin{aligned}
&\mathbf {\bar O}^{(s)}= \mathrm{Norm}(\mathbf O^{(s)}), \\
&\mathbf { U}^{(s)}, \mathbf {V}^{(s)}=\mathbf {\bar O}^{(s)}\mathbf W_u^{(s)}, \mathbf {\bar O}^{(s)}\mathbf W_v^{(s)}, \\
&\mathbf {X}^{(s+1)}=[\mathbf { U}^{(s)}\odot f(\mathbf {V}^{(s)})] \mathbf W_{down}^{(s)}+\mathbf { O}^{(s)},\\
&\mathbf O^{(s)} \in \mathbb R^{n\times d}, \\
&\mathbf W_u^{(s)}, \mathbf W_v^{(s)} \in \mathbb R^{d\times g}, 
 \mathbf W_{down}^{(s)} \in \mathbb R^{g\times d}.
\end{aligned}
\end{equation}
Output embedding:
\begin{equation}
\small
\begin{aligned}
&\mathbf {O}= \mathbf X^{(l+1)} \mathbf W_{out}, \\
&\mathbf X^{(l+1)} \in \mathbb R^{n\times d},
\mathbf W_{out}\in \mathbb R^{d\times v}.
\end{aligned}
\end{equation}

\textbf{FLOPs}
\begin{itemize}
 
   \item  Input embedding:  $[n] \times [d, v] \implies 2ndv $.

   \item Token mixer:  $8nd^2+4n^2d+3n^2h+4nd.$ %softmax: exp, sum, divide
   \begin{itemize}
       \item qkv projection: $$ [n, d] \times [d, 3d] \implies 3\times2nd^2.$$
       \item qk multiplication:$$ [h, n, d/h] \times [h, d/h, n]  \implies 2n^2d. $$
       \item RoPE: $4nd.$
       \item Softmax: exp, sum, divide $\implies 3n^2h.$
       \item (qk)v multiplication: $$ [h, n, n] \times [h, d/h, n] \implies 2n^2d.$$ 
       \item output projection: $$[n, d] \times [d, d] \implies 2nd^2.$$
   \end{itemize}

   \item  Channel mixer:  $6ndg + ng.$
\begin{itemize}
    \item u,v projection:  $$[n, d] \times [d, g] \implies 4\times ndg.$$  
    \item gating: $$[n, g] \odot [n, g] \implies  ng.$$  
    \item down projection: $$[n, g] \times [g, d] \implies 2\times ndg.$$
\end{itemize}
   \item  Output embedding:  $$[n, d] \times [d, v] \implies 2ndv .$$
   \item  Forward FLOPs: $bl\times(8nd^2+4n^2d+3n^2h + 4nd+ 6ndg + ng) + 4ndv$
   \item Total training FLOPS:
   $bl\times(24nd^2+12n^2d+9n^2h + 18ndg + 3ng) + 12ndv$
   \item Substituting $ g = 8/3d $ yields:
   \begin{equation*}
    \begin{gathered}
bl(72nd^2 + 12n^2d+9n^2h + 20nd)+12ndv \\
   =\underbrace{72bnld^2\left( 
    1 + \frac{n}{6d}+ \frac{5}{18d}+\frac{h}{8d^2}
   \right)}_{\text{Non-embedding term}}  \\+ \underbrace{12ndv}_{\text{Embedding term}}.
   \end{gathered}
   \end{equation*}

\end{itemize}
\textbf{Parameters}
\begin{itemize}
    \item Input \& output embedding (shared weights): $dv$.
    \item Token mixer: $4d^2$.
    \item Channel mixer: $3dg$.
    \item Total parameters: $4ld^2+3ldg+dv$.
    \item  Substituting $ g = 8/3d $ yields: $12ld^2 + dv$.
\end{itemize}

\subsubsection{Linear Attention - TNL (data independent decay)}
For TNL, we use $0<\lambda_i<1$ as the decay of head $i$ and use LA as the abbreviation for Lightning Attention.

\textbf{Equation}
Input embedding:
\begin{equation}
\small
\begin{aligned}
&\mathbf X^{(1)}= \mathrm{Lookup}(\mathbf X^{(0)}, \mathbf W_{in}), \\
&\mathbf X^{(0)}\in \mathbb R^n, \mathbf W_{in} \in \mathbb R^{d \times v}.
\end{aligned}
\end{equation}
Token mixer($s$-th layer):
\begin{equation}
\small
\begin{aligned}
&\mathbf {\bar X}^{(s)}= \mathrm{Norm}(\mathbf X^{(s)}), \\
&\mathbf Q^{(s)}_i,\mathbf K^{(s)}_i,\mathbf V^{(s)}_i=
f(\mathbf  {\bar X}^{(s)} \mathbf W_{q_i}^{(s)}), f(\mathbf  {\bar X}^{(s)} \mathbf W_{k_i}^{(s)}), \mathbf  {\bar X}^{(s)} \mathbf W_{v_i}^{(s)}, \\
&\mathbf G^{(s)}=\mathrm{Sigmoid}\left(\mathbf X \mathbf W_{gdown}^{(s)}\mathbf 
 W_{gup}^{(s)}\right), \\
&\mathbf { O}^{(s)}_i=\mathrm{LA}\left(\mathbf Q^{(s)}_i, {\mathbf K^{(s)}_i}, \mathbf V^{(s)}_i,\lambda_i \right), \\
&\mathbf O^{(s)}=\mathrm{Norm}\left( \mathrm{Concat}[\mathbf { O}^{(s)}_1,\ldots, \mathbf { O}^{(s)}_h] \right) \odot \mathbf G^{(s)} + \mathbf X^{(s)},\\
&\mathbf X^{(s)}\in \mathbb R^{n\times d}, \\ 
&\mathbf W_{gdown}^{(s)}\in \mathbb R^{d\times t},\mathbf 
 W_{gup}^{(s)} \in \mathbb R^{t\times d}, 
\mathbf W_{o}^{(s)}\in \mathbb R^{d\times d},  \\
&\mathbf W_{q_i}^{(s)},\mathbf W_{k_i}^{(s)},\mathbf W_{v_i}^{(s)} \in \mathbb R^{d\times d / h}, i=1,\ldots, h.
\end{aligned}
\end{equation}
Channel mixer($s$-th layer):
\begin{equation}
\small
\begin{aligned}
&\mathbf {\bar O}^{(s)}= \mathrm{Norm}(\mathbf O^{(s)}), \\
&\mathbf { U}^{(s)}, \mathbf {V}^{(s)}=\mathbf {\bar O}^{(s)}\mathbf W_u^{(s)}, \mathbf {\bar O}^{(s)}\mathbf W_v^{(s)}, \\
&\mathbf {X}^{(s+1)}=[\mathbf { U}^{(s)}\odot \mathbf {V}^{(s)}] \mathbf W_{down}^{(s)}+\mathbf { O}^{(s)},\\
&\mathbf O^{(s)} \in \mathbb R^{n\times d}, \\
&\mathbf W_u^{(s)}, \mathbf W_v^{(s)} \in \mathbb R^{d\times g}, 
 \mathbf W_{down}^{(s)} \in \mathbb R^{g\times d}.
\end{aligned}
\end{equation}
Output embedding:
\begin{equation}
\small
\begin{aligned}
&\mathbf {O}= \mathbf X^{(l+1)} \mathbf W_{out}, \\
&\mathbf X^{(l+1)} \in \mathbb R^{n\times d},
\mathbf W_{out}\in \mathbb R^{d\times v}.
\end{aligned}
\end{equation}

\textbf{FLOPs}
\begin{itemize}
   \item Input embedding:  $[n] \times [d, v] \implies 2ndv $.

   \item Token Mixer:   $C = 8nd^2 + \frac{4nd^2}{h} + \frac{nd^2}{Bh} + 4nBd + nBh + 4ntd + 2nd$. 
\begin{itemize}
    \item\text{qkv projection: }           $$[n, d] \times [d, 3d] \implies 3\times2nd^2.$$
    \item\text{lightning attention intra block: } $$\text{repeat } \frac{n}{B} \text{ times} \implies  4B^2d+B^2h.$$
    \item\text{lightning attention inter block: } $$\text{repeat } \frac{n}{B} \text{ times} \implies  \frac{2Bd^2}{h}.$$
    \item\text{kv update: } \text{repeat } $$\frac{n}{B} \text{ times} \implies  \frac{2Bd^2}{h}+\frac{d^2}{h}.$$
    \item\text{attention output update: } $$\text{repeat } \frac{n}{B} \text{ times} \implies  Bd.$$
    \item\text{output  gate: }                 $$[n, d] \times [d, t], \ [n, t] \times [t, d] \implies 4ntd.$$
    \item\text{gating: }                       $$[n, d] \odot [n, d] \implies nd.$$
    \item\text{output projection: }                 $$[n, d] \times [d, d] \implies 2nd^2.$$
\end{itemize}

   \item  Channel Mixer:  $6ndg + ng$. 
\begin{itemize}
    \item u,v projection:  $$[n, d] \times [d, g] \implies 4\times ndg.$$  
    \item gating: $$[n, g] \odot [n, g] \implies  ng.$$
    \item down projection: $$[n, g] \times [g, d] \implies 2\times ndg.$$
\end{itemize}
   \item  Output embedding:  $[n, d] \times [d, v] \implies 2ndv $.
   \item Forward FLOPs: $l \times (C + 6ndg + ng) + 4ndv$.
   \item  Total training FLOPs: $bl(3C + 18ndg + 3ng) + 12bndv$.
   \item Substituting $ g = 8/3d , B=t=d/h$ yields:
   \begin{equation*}
     \begin{aligned}
& C=8nd^2+12\frac{nd^2}{h}+4nd, \\
 &  \mathrm{FLOPs}= bl\left(72nd^2 + \frac{36nd^2}{h}+20nd \right)+12ndv\\
& =\underbrace{72bnld^2\left( 
    1 + \frac{1}{2h}+\frac{5}{18d}
   \right)}_{\text{Non-embedding term}}  + \underbrace{12ndv}_{\text{Embedding term}}.
   \end{aligned}
   \end{equation*}

\end{itemize}
\textbf{Parameters}
\begin{itemize}
    \item Input \& output embedding (shared weights): $dv$.
    \item Token Mixer: $4d^2+2dt$.
    \item Channel Mixer: $3dg$.
    \item Total parameters: $4ld^2+2ldt+3ldg+dv$.
      \item  Substituting $ g = 8/3d,t=d/h $ yields: $12ld^2 +2ld^2/h+ dv$.
\end{itemize}

\subsubsection{Linear RNN - HGRN2 (data dependent decay)}
We use FLA as the abbreviation for Flash Linear Attention. \\
\textbf{Equation}
Input embedding:
\begin{equation}
\small
\begin{aligned}
&\mathbf X^{(1)}= \mathrm{Lookup}(\mathbf X^{(0)}, \mathbf W_{in}), \\
&\mathbf X^{(0)}\in \mathbb R^n, \mathbf W_{in} \in \mathbb R^{d \times v}.
\end{aligned}
\end{equation}
Lower bound:
\begin{equation*}
\begin{aligned}
&\mathbf{\overline{LR}}_i= \mathrm{Softmax}(\mathbf{LR}_i,\mathrm{dim}=0), \\
&\mathbf{Lr}_i^{(s)}
=\mathrm{Cumsum}(\mathbf{\overline{LR}}_i,\mathrm{dim}=0)[s], \\
&\mathbf {LR}_i \in \mathbb R^{L\times d/h},i=1,\ldots, h.
\end{aligned}
\end{equation*}
Token mixer($s$-th layer):
\begin{equation}
\small
\begin{aligned}
&\mathbf {\bar X}^{(s)}= \mathrm{Norm}(\mathbf X^{(s)}), \\
&\mathbf {Og}^{(s)}_i,\mathbf {Fg}^{(s)}_i,\mathbf H^{(s)}_i=
\mathbf  {\bar X}^{(s)} \mathbf W_{og_i}^{(s)}, \mathbf  {\bar X}^{(s)} \mathbf W_{fg_i}^{(s)}, \mathbf  {\bar X}^{(s)} \mathbf W_{h_i}^{(s)}, \\
&\mathbf {Fg}^{(s)}_i=\mathbf{Lr}_i^{(s)}+(1-\mathbf{Lr}_i^{(s)})(\mathrm{Sigmoid}( \mathbf {Fg}^{(s)}_i)), \\
&\mathbf { O}^{(s)}_i=\mathrm{FLA}\left(\mathbf {Og}^{(s)}_i, {\mathbf {Fg}^{(s)}_i}, \mathbf H^{(s)}_i,1-{\mathbf {Fg}^{(s)}_i}\right), \\
&\mathbf O^{(s)}=\mathrm{Norm}\left( \mathrm{Concat}[\mathbf { O}^{(s)}_1,\ldots, \mathbf { O}^{(s)}_h] \right)  + \mathbf X^{(s)}, \\
&\mathbf X^{(s)}\in \mathbb R^{n\times d}, \\ 
&\mathbf W_{gdown}^{(s)}\in \mathbb R^{d\times t},\mathbf 
 W_{gup}^{(s)} \in \mathbb R^{t\times d}, 
\mathbf W_{o}^{(s)}\in \mathbb R^{d\times d},  \\
&\mathbf W_{og_i}^{(s)},\mathbf W_{fg_i}^{(s)},\mathbf W_{h_i}^{(s)} \in \mathbb R^{d\times d / h}, i=1,\ldots, h.
\end{aligned}
\end{equation}
Channel mixer($s$-th layer):
\begin{equation}
\small
\begin{aligned}
&\mathbf {\bar O}^{(s)}= \mathrm{Norm}(\mathbf O^{(s)}), \\
&\mathbf { U}^{(s)}, \mathbf {V}^{(s)}=\mathbf {\bar O}^{(s)}\mathbf W_u^{(s)}, \mathbf {\bar O}^{(s)}\mathbf W_v^{(s)}, \\
&\mathbf {X}^{(s+1)}=[\mathbf { U}^{(s)}\odot \mathbf {V}^{(s)}] \mathbf W_{down}^{(s)}+\mathbf { O}^{(s)},\\
&\mathbf O^{(s)} \in \mathbb R^{n\times d}, \\
&\mathbf W_u^{(s)}, \mathbf W_v^{(s)} \in \mathbb R^{d\times g}, 
 \mathbf W_{down}^{(s)} \in \mathbb R^{g\times d},
\end{aligned}
\end{equation}
Output embedding:
\begin{equation}
\small
\begin{aligned}
&\mathbf {O}= \mathbf X^{(l+1)} \mathbf W_{out}, \\
&\mathbf X^{(l+1)} \in \mathbb R^{n\times d},
\mathbf W_{out}\in \mathbb R^{d\times v}.
\end{aligned}
\end{equation}

\textbf{FLOPs}
\begin{itemize}
\item Input embedding:  $[n] \times [d, v] \implies 2ndv $.
 \item Lower bound: $4ld$.
   \item Token Mixer:   $C = 8nd^2 + \frac{4nd^2}{h} + \frac{nd^2}{Bh} + 4nBd + nBh+5nd$. 
\begin{itemize}
    \item\text{hidden state projection: }           $$[n, d] \times [d, 3d] \implies 3\times2nd^2.$$
      \item\text{forget gate compute: }           $4nd.$
    \item\text{fla intra block: } $$\text{repeat } \frac{n}{B} \text{ times} \implies  4B^2d+B^2h.$$
    \item\text{fla inter block: } $$\text{repeat } \frac{n}{B} \text{ times} \implies  \frac{2Bd^2}{h}.$$
    \item\text{state update: } \text{repeat } $$\frac{n}{B} \text{ times} \implies  \frac{2Bd^2}{h}+\frac{d^2}{h}.$$
    \item\text{attention output update: } $$\text{repeat } \frac{n}{B} \text{ times} \implies  Bd.$$
    \item\text{output projection: }                 $$[n, d] \times [d, d] \implies 2nd^2.$$
\end{itemize}

   \item  Channel Mixer:  $6ndg + ng$. 
\begin{itemize}
    \item u,v projection:  $$[n, d] \times [d, g] \implies 4\times ndg.$$  
    \item gating: $$[n, g] \odot [n, g] \implies  ng.$$  
    \item down projection: $$[n, g] \times [g, d] \implies 2\times ndg.$$
\end{itemize}
\item  Output embedding:  $[n, d] \times [d, v] \implies 2ndv $.
\item Forward FLOPs: $l \times (C + 6ndg + ng) + 4ndv+4dl$.
\item  Total training FLOPs: $bl(3C + 18ndg + 3ng) + 12bndv+12dl$.
 \item Substituting $ g = 8/3d , B=d/h$. yields:
   \begin{equation*}
     \begin{aligned}
&C=8nd^2+\frac{8nd^2}{h}+7nd, \\
&\mathrm{FLOPs}\\
 =& bl\left(72nd^2 + \frac{24nd^2}{h}+29nd \right)\\
 &+12ndv+12dl\\
=&\underbrace{72bnld^2\left( 
    1 + \frac{1}{3h}+\frac{29}{72d}
   \right)+12dl}_{\text{Non-embedding term}} \\
   & + \underbrace{12ndv}_{\text{Embedding term}}.
   \end{aligned}
   \end{equation*}
\end{itemize}

\textbf{Parameters}
\begin{itemize}
    \item Lower bound: $ld$.
    \item Input \& output embedding (shared weights): $dv$.
    \item Token Mixer: $4d^2$.
    \item Channel Mixer: $3dg$.
    \item Total parameters: $4ld^2+3ldg+dv+ld$.
       \item  Substituting $ g = 8/3d $ yields: $12ld^2 +dv+ld$.
\end{itemize}

\subsubsection{Linear Attention - cosFormer2}
For cosFormer2, we use $\theta_i \in \mathbb R^{d/h}$ as the Lrpe parameter of head $i$ and use LA as the abbreviation for Lightning Attention.

\textbf{Equation}
Input embedding:
\begin{equation}
\small
\begin{aligned}
&\mathbf X^{(1)}= \mathrm{Lookup}(\mathbf X^{(0)}, \mathbf W_{in}), \\
&\mathbf X^{(0)}\in \mathbb R^n, \mathbf W_{in} \in \mathbb R^{d \times v}.
\end{aligned}
\end{equation}
Tpe:
\begin{equation*}
    \mathbf X^{(1)}= \mathrm{Tpe}(\mathbf X^{(1)}).
\end{equation*}

Token mixer($s$-th layer):
\begin{equation}
\small
\begin{aligned}
&\mathbf {\bar X}^{(s)}= \mathrm{Norm}(\mathbf X^{(s)}), \\
&\mathbf Q^{(s)}_i,\mathbf K^{(s)}_i,\mathbf V^{(s)}_i=
f(\mathbf  {\bar X}^{(s)} \mathbf W_{q_i}^{(s)}), f(\mathbf  {\bar X}^{(s)} \mathbf W_{k_i}^{(s)}), \mathbf  {\bar X}^{(s)} \mathbf W_{v_i}^{(s)}, \\
&\mathbf Q^{(s)}_i=\mathrm{Concat}
[\cos(\theta_i^{(s)})\mathbf Q^{(s)}_i,
\sin(\theta_i^{(s)})\mathbf Q^{(s)}_i
], \\
&\mathbf K^{(s)}_i=\mathrm{Concat}
[\cos(\theta_i^{(s)})\mathbf K^{(s)}_i,
\sin(\theta_i^{(s)})\mathbf K^{(s)}_i
],   \\
&\mathbf G^{(s)}=\mathrm{Sigmoid}\left(\mathbf X \mathbf W_{gdown}^{(s)}\mathbf 
 W_{gup}^{(s)}\right), \\
&\mathbf { O}^{(s)}_i=\mathrm{LA}\left(\mathbf Q^{(s)}_i, {\mathbf K^{(s)}_i}, \mathbf V^{(s)}_i \right), \\
&\mathbf O^{(s)}=\mathrm{Norm}\left( \mathrm{Concat}[\mathbf { O}^{(s)}_1,\ldots, \mathbf { O}^{(s)}_h] \right) \odot \mathbf G^{(s)} + \mathbf X^{(s)}, \\
&\mathbf X^{(s)}\in \mathbb R^{n\times d}, \\ 
&\mathbf W_{gdown}^{(s)}\in \mathbb R^{d\times t},\mathbf 
 W_{gup}^{(s)} \in \mathbb R^{t\times d}, 
\mathbf W_{o}^{(s)}\in \mathbb R^{d\times d},  \\
&\mathbf W_{q_i}^{(s)},\mathbf W_{k_i}^{(s)},\mathbf W_{v_i}^{(s)} \in \mathbb R^{d\times d / h}, i=1,\ldots, h.
\end{aligned}
\end{equation}
Channel mixer($s$-th layer):
\begin{equation}
\small
\begin{aligned}
&\mathbf {\bar O}^{(s)}= \mathrm{Norm}(\mathbf O^{(s)}), \\
&\mathbf { U}^{(s)}, \mathbf {V}^{(s)}=\mathbf {\bar O}^{(s)}\mathbf W_u^{(s)}, \mathbf {\bar O}^{(s)}\mathbf W_v^{(s)}, \\
&\mathbf {X}^{(s+1)}=[\mathbf { U}^{(s)}\odot \mathbf {V}^{(s)}] \mathbf W_{down}^{(s)}+\mathbf { O}^{(s)},\\
&\mathbf O^{(s)} \in \mathbb R^{n\times d}, \\
&\mathbf W_u^{(s)}, \mathbf W_v^{(s)} \in \mathbb R^{d\times g}, 
 \mathbf W_{down}^{(s)} \in \mathbb R^{g\times d},
\end{aligned}
\end{equation}
Output embedding:
\begin{equation}
\small
\begin{aligned}
&\mathbf {O}= \mathbf X^{(l+1)} \mathbf W_{out}, \\
&\mathbf X^{(l+1)} \in \mathbb R^{n\times d},
\mathbf W_{out}\in \mathbb R^{d\times v}.
\end{aligned}
\end{equation}

\textbf{FLOPs}
\begin{itemize}
   \item Input embedding:  $[n] \times [d, v] \implies 2ndv $.
\item Tpe: $4nde$. 
\begin{itemize}
    \item Up projection: $[n, d] \times [d, e] \implies 2nde $.
    \item Recurrence: $nde$.
     \item Down projection: $[n, d, e] \implies nde $.
\end{itemize}
   \item Token Mixer:   $C = 8nd^2 + \frac{8nd^2}{h} + \frac{2nd^2}{Bh} + 6nBd + nBh + 4ntd + 2nd$.
\begin{itemize}
    \item\text{qkv projection: }           $$[n, d] \times [d, 3d] \implies 3\times2nd^2.$$
    \item Lrpe: $4nd$.
    \item\text{lightning attention intra block: } $$\text{repeat } \frac{n}{B} \text{ times} \implies  6B^2d+B^2h.$$
    \item\text{lightning attention inter block: } $$\text{repeat } \frac{n}{B} \text{ times} \implies  \frac{4Bd^2}{h}.$$
    \item\text{kv update: } \text{repeat } $$\frac{n}{B} \text{ times} \implies  \frac{4Bd^2}{h}+\frac{2d^2}{h}.$$
    \item\text{attention output update: } $$\text{repeat } \frac{n}{B} \text{ times} \implies  Bd.$$
    \item\text{output  gate: }                 $$[n, d] \times [d, t], \ [n, t] \times [t, d] \implies 4ntd.$$
    \item\text{gating: }                       $$[n, d] \odot [n, d] \implies nd.$$
    \item\text{output projection: }                 $$[n, d] \times [d, d] \implies 2nd^2.$$
\end{itemize}

   \item  Channel Mixer:  $6ndg + ng$. 
\begin{itemize}
    \item u,v projection:  $$[n, d] \times [d, g] \implies 4\times ndg.$$  
    \item gating: $$[n, g] \odot [n, g] \implies  ng.$$  
    \item down projection: $$[n, g] \times [g, d] \implies 2\times ndg.$$
\end{itemize}
   \item  Output embedding:  $[n, d] \times [d, v] \implies 2ndv $.
   \item Forward FLOPs: $l \times (C + 6ndg + ng) + 4ndv + 4nde$.
   \item  Total training FLOPs: $bl(3C + 18ndg + 3ng) + 12bndv +12bnde$.
   \item Substituting $ g = 8/3d , B=e=d/h$ yields:
   \begin{equation*}
     \begin{aligned}
& C=8nd^2+18\frac{nd^2}{h}+5nd, \\
 &  \mathrm{FLOPs}\\
 =& bl\left(72nd^2 + \frac{54nd^2}{h}+23nd \right)\\
 &+12ndv+\frac{12bnd^2}{h}\\
 =& \underbrace{72bnld^2\left( 
    1 + \frac{3}{4h}+\frac{23}{72d}
   \right)+\frac{12bnd^2}{h}}_{\text{Non-embedding term}}  \\
   &+ \underbrace{12ndv}_{\text{Embedding term}}.
   \end{aligned}
   \end{equation*}
\end{itemize}

\textbf{Parameters}
\begin{itemize}
    \item Input \& output embedding (shared weights): $dv$.
    \item Tpe: $de$.
    \item Token Mixer: $4d^2+2dt$.
    \item Channel Mixer: $3dg$.
    \item Total parameters: $4ld^2+2ldt+3ldg+dv+de$.
      \item  Substituting $ g = 8/3d,t=e=d/h $ yields: $12ld^2 +2ld^2/h+ dv+d^2/h$.
\end{itemize}

\section{Evaluations}

\subsection{Metrics of Needle in A Haystack }
\label{sec:niah_mertics}

\par We use four types metrics in NIAH evaluation: 
\par\textbf{Accuracy at a context length.} This averages the retrieval accuracy at a chosen context length across all depth steps (acc$@$seq\_len in Table \ref{niah_easy_full}).
\par\textbf{Accuracy less or equal to a context length.} This metric calculates the mean accuracy over a range of context lengths, across all depth steps (acc$\leq$seq\_len in Table \ref{niah_easy_full}). The total averaged accuracy is the accuracy less or equal to the maximum context length, which is 16k in all our experiments.
\par\textbf{Weighted average accuracy.} To further represent the levels of complexity for retrieving a needle in different depths and context lengths, we assign weights to each depth and context length. We assume larger weights for deeper and longer texts. We use geometric progression as a weight function for both aspects. Specifically, the weights are calculated as: $w_{d_i}=w_{d_0}{\alpha_d}^{i-1}$, $w_{c_i}=w_{c_0}{\alpha_c}^{i-1}$, where $w_{d_i}$ ($w_{c_i}$) is the weight for $i$-th depth step (context length), $\alpha_d$ ($\alpha_c$) is a constant greater than 1. Using the outer product, we obtain a weight map for all depth-length combinations. The weight map is applied when calcualting the average accuracy (weighted avg acc in Table \ref{niah_easy_full}). 
\par\textbf{NIAH score.} Through our experiments, we observe cases when two models achieve the same average accuracy but display different patterns in the NIAH heatmap. Using weighted average can assist in this situation. To better evaluate the model ability in such cases, we develop a penalty mechanism. We first binarize the NIAH score array for success and failure, which is originally ranged from 1 to 10 using a threshold. Then for each context length, we penalize the situations when models do not consistently succeed or fail in retrieving the needle across different depths. For each column of the score array, we find the longest continuous sequence of 1s (success). If the sequence does not exist, the largest penalty is assigned ($p=0$); If the sequence length equals the number of depth steps, no penalty is assigned ($p=1$); Otherwise, we count the number $n$ of continuous segments of either 1s (success) or 0s (failure), and assign penalty as $p=2^{(1-n)/2}$. Combining weighted average and penalty, we have the NIAH score (niah score in Table \ref{niah_easy_full}).

\section{Experiments}
\label{sec:exp}

\subsection{SCROLLS}
We assess models such as LLaMA, TNL, HGRN2, and cosFormer2 using the SCROLLS benchmark, focusing on different parameter sizes (refer to Table~\ref{tab:scrolls}), aspect ratios (see Table~\ref{tab:scrolls_aspect_ratio}), and context lengths (consult Table~\ref{tab:scrolls_len}).

The table~\ref{tab:scrolls} provides a detailed comparison of various models such as LLaMA, TNL, HGRN2, and cosFormer2 across multiple metrics on the SCROLLS benchmark. It outlines the performance of these models based on parameter size (ranging from 0.07 to 7 billion), as detailed in each row. Table~\ref{tab:scrolls} highlights that across all linear complexity sequence models and LLaMA, there is a general improvement in performance with increasing parameter sizes. Models exhibit varying sensitivity to parameter size across tasks; for example, LLaMA's NarrativeQA F1 score jumps from 4.70 to 22.31 as parameters increase. At higher sizes, HGRN2 tends to outperform TNL consistently, highlighting its superior scaling capability. Additionally, models show task-specific strengths, with cos2 excelling in ContractNLI at 7 billion parameters, showcasing its effectiveness with legal texts.

\begin{table*}
\small

\centering
\caption{\textbf{SCROLLS Benchmark Overview.} P.S.: Parameter Size. R.1/R.2/R.L: rouge-1/rouge-2/rouge-l. The term '-' indicates a failure in the specified task.}
\label{tab:scrolls}
\setlength{\tabcolsep}{0.5mm}
\begin{tabular}{p{1.0cm}c|ccccccc|c}
\toprule
Arch & \makecell[c]{P.S. \\(Billion)} & \makecell[c]{GovReport \\(R.1/R.2/R.L)}   & \makecell[c]{SummScreenfd\\(R.1/R.2/R.L)} & \makecell[c]{QmSum\\(R.1/R.2/R.L)}       & \makecell[c]{Qasper\\(F1)} & \makecell[c]{NarrativeQA\\(F1)} & \makecell[c]{Quality\\(EM)} &  \makecell[c]{ContractNLI\\(EM)} & \makecell[c]{SCROLLS\\(Avg)} \\ \midrule
LLaMA & 0.07 & 6.49/1.46/5.07   & 8.52/0.96/7.05   & 5.13/0.79/4.58   & 10.45 & 4.70  & 26.46 & 14.95 & 7.43  \\
TNL   & 0.07 & 2.64/0.8/2.27    & 6.08/0.49/4.95   & 2.51/0.59/2.22   & 8.00  & 4.06  & 27.18 & 17.94 & 6.13  \\
HGRN2 & 0.07 & 10.88/2.09/8.19  & 7.13/0.58/5.98   & 7.14/1.02/6.38   & 7.08  & 3.34  & 26.70 & 9.45  & 7.32  \\
cos2  & 0.07 & 6.21/1.36/5.08   & 6.43/0.66/5.59   & 4.98/0.58/4.36   & 8.27  & 3.04  & 27.47 & 12.63 & 6.67  \\
\midrule
LLaMA & 0.16 & 5.5/2.15/4.44    & 10.91/1.24/8.45  & 7.98/1.46/7.37   & 9.10  & 8.93  & 26.75 & 14.56 & 8.37  \\
TNL   & 0.16 & -                & 9.88/1.21/8.09   & 3.36/0.82/2.76   & 8.19  & 6.39  & 27.90 & 9.06  & 7.77  \\
HGRN2 & 0.16 & 13.69/2.71/10.1  & 6.61/0.5/6.02    & 7.33/0.99/6.61   & 8.24  & 7.18  & 25.55 & 10.90 & 8.29  \\
cos2  & 0.16 & 7.01/1.89/5.44   & 8.07/0.9/6.85    & 9.28/1.52/8.14   & 8.33  & 5.66  & 26.41 & 10.70 & 7.71  \\
\midrule
LLaMA & 0.41 & 8.21/3.54/6.21   & 11.31/1.56/8.65  & 10.66/2.07/9.42  & 17.82 & 15.39 & 27.95 & 13.89 & 10.51 \\
TNL   & 0.41 & 2.96/1.12/2.54   & 10.54/1.15/7.95  & 6.34/1.33/5.08   & 11.41 & 9.87  & 27.61 & 10.32 & 7.55  \\
HGRN2 & 0.41 & 15.33/3.54/10.91 & 7.35/0.76/6.17   & 8.32/1.22/7.4    & 12.36 & 10.87 & 26.37 & 31.53 & 10.93 \\
cos2  & 0.41 & 6.11/2.51/4.87   & 12.02/1.83/9.41  & 10.25/2.19/8.62  & 14.04 & 9.60  & 27.23 & 9.06  & 9.06  \\
\midrule
LLaMA & 1    & 12.91/3.06/9.38  & 9.47/0.84/7.72   & 10.93/2.24/9.43  & 22.77 & 16.03 & 28.43 & 9.93  & 11.01 \\
TNL   & 1    & 5.86/2.02/4.74   & 9.39/1.34/7.32   & 5.81/1.43/4.8    & 14.23 & 13.83 & 28.19 & 26.52 & 9.65  \\
HGRN2 & 1    & 14.86/4.21/10.45 & 11.4/1.44/9.16   & 10.9/2.28/9.68   & 16.21 & 15.09 & 27.76 & 10.61 & 11.08 \\
cos2  & 1    & 7.97/3.51/6.15   & 12.25/1.95/9.38  & 11.91/2.7/9.96   & 16.94 & 13.93 & 27.76 & 17.07 & 10.88 \\
\midrule
LLaMA & 3    & 11.16/4.88/8.14  & 11.89/1.9/9.3    & 16.08/4.25/12.87 & 28.57 & 20.77 & 30.44 & 20.15 & 13.88 \\
TNL   & 3    & -                & 9.65/1.56/7.17   & 11.37/2.97/9.14  & 21.20 & 17.70 & 28.95 & 12.92 & 12.26 \\
HGRN2 & 3    & 21.7/6.62/14.09  & 14.55/2.13/10.79 & 12.48/2.69/10.58 & 25.41 & 18.75 & 28.86 & 31.92 & 15.43 \\
cos2  & 3    & 14.69/5.37/9.98  & 11.33/1.74/8.77  & 15.38/3.53/12.68 & 25.10 & 18.05 & 29.72 & 9.35  & 12.75 \\
\midrule
LLaMA & 7    & 17.4/7.33/11.43  & 12.92/1.75/9.95  & 14.59/3.7/11.8   & 32.35 & 22.31 & 33.84 & 10.03 & 14.57 \\
TNL   & 7    & 5.36/2.29/4.41   & 11.17/1.72/8.46  & 12.02/3.1/9.46   & 24.12 & 19.24 & 29.15 & 9.16  & 10.74 \\
HGRN2 & 7    & 14.93/5.21/10.16 & 15.43/2.4/11.1   & 14.3/2.97/11.78  & 27.07 & 19.60 & 30.06 & 10.03 & 13.46 \\
cos2  & 7    & 19.97/7.36/12.92 & 14.31/2.4/10.57  & 13.72/3.27/11.34 & 23.94 & 18.70 & 30.73 & 27.68 & 15.15 \\
\bottomrule
\end{tabular}
\end{table*}

\begin{table*}
\small
\centering
\caption{\textbf{SCROLLS Benchmark by Aspect Ratio.} The term '-' indicates a failure in the specified task.}
\label{tab:scrolls_aspect_ratio}
\setlength{\tabcolsep}{0.8mm}
\begin{tabular}{p{1.0cm}c|ccccccc|c}
\toprule
Arch & \makecell[c]{Dim} & \makecell[c]{GovReport\\(R.1/R.2/R.L)}   & \makecell[c]{SummScreenfd\\(R.1/R.2/R.L)} & \makecell[c]{QmSum\\(R.1/R.2/R.L)}       & \makecell[c]{Qasper\\(F1)} & \makecell[c]{NarrativeQA\\(F1)} & \makecell[c]{Quality\\(EM)} &  \makecell[c]{ContractNLI\\(EM)} & \makecell[c]{SCROLLS\\(Avg)} \\ \midrule
LLaMA & 1536 & 12.91/3.06/9.38  & 9.47/0.84/7.72  & 10.93/2.24/9.43  & 22.77 & 16.03 & 28.43 & 9.93  & 11.01 \\
LLaMA & 1792 & 14.68/6.0/10.03  & 11.5/1.63/9.03  & 9.72/2.18/8.45   & 23.02 & 16.68 & 27.85 & 21.02 & 12.45 \\
LLaMA & 2048 & 7.62/3.59/5.89   & 11.07/1.61/8.5  & 14.49/3.51/11.85 & 19.06 & 16.62 & 27.85 & 26.90 & 12.20 \\
LLaMA & 3072 & 17.04/5.78/11.41 & 9.16/1.29/7.35  & 5.79/0.98/4.96   & 17.89 & 12.12 & 28.00 & 19.67 & 10.88 \\
\midrule
cos2  & 1536 & 7.97/3.51/6.15   & 12.25/1.95/9.38 & 11.91/2.7/9.96   & 16.94 & 13.93 & 27.76 & 17.07 & 10.88 \\
cos2  & 1792 & 6.1/2.52/4.88    & 11.06/1.49/8.41 & 12.46/2.87/10.24 & 17.80 & 13.90 & 27.76 & 14.56 & 10.31 \\
cos2  & 2048 & 7.63/3.22/5.89   & 12.94/1.71/9.84 & 11.35/2.33/9.69  & 17.68 & 13.39 & 27.76 & 16.30 & 10.75 \\
cos2  & 3072 & 9.22/2.85/7.01   & 9.03/1.05/7.45  & 12.96/2.99/10.96 & 13.89 & 12.16 & 27.47 & 11.09 & 9.85 \\
\bottomrule
\end{tabular}
\end{table*}

\begin{table*}
\small
\centering
\caption{\textbf{SCROLLS Benchmark by Pre-training Context Length.} The term '-' indicates a failure in the specified task.}
\label{tab:scrolls_len}
\setlength{\tabcolsep}{0.8mm}
\begin{tabular}{p{1.0cm}c|ccccccc|c}
\toprule
Arch & \makecell[c]{Len} & \makecell[c]{GovReport\\(R.1/R.2/R.L)}   & \makecell[c]{SummScreenfd\\(R.1/R.2/R.L)} & \makecell[c]{QmSum\\(R.1/R.2/R.L)}       & \makecell[c]{Qasper\\(F1)} & \makecell[c]{NarrativeQA\\(F1)} & \makecell[c]{Quality\\(EM)} &  \makecell[c]{ContractNLI\\(EM)} & \makecell[c]{SCROLLS\\(Avg)} \\ \midrule
TNL   & 2k  & 5.72/2.23/4.6    & 7.37/1.0/5.42   & 5.88/1.24/4.7    & 13.34 & 14.42 & 28.24 & 23.14 & 9.02  \\
TNL   & 4k  & 7.61/2.8/5.96    & 13.01/1.98/9.76 & 10.87/2.97/9.07  & 19.45 & 14.89 & 27.18 & 27.77 & 11.79 \\
TNL   & 8k  & 5.86/2.02/4.74   & 9.39/1.34/7.32  & 5.81/1.43/4.8    & 14.23 & 13.83 & 28.19 & 26.52 & 9.65  \\
TNL   & 16k & 4.54/1.66/3.67   & 9.41/0.96/7.4   & 7.12/1.61/5.79   & 16.61 & 13.68 & 28.09 & 13.11 & 8.74  \\
\midrule
HGRN2 & 2k  & 15.25/4.18/10.5  & 10.58/1.23/8.77 & 11.19/2.0/9.54   & 18.46 & 13.60 & 27.71 & 17.74 & 11.60 \\
HGRN2 & 4k  & 14.97/4.69/10.27 & 10.08/1.26/8.38 & 11.39/2.26/9.65  & 17.43 & 15.05 & 26.27 & 17.26 & 11.46 \\
HGRN2 & 8k  & 14.86/4.21/10.45 & 11.4/1.44/9.16  & 10.9/2.28/9.68   & 16.21 & 15.09 & 27.76 & 10.61 & 11.08 \\
HGRN2 & 16k & 21.7/5.67/14.25  & 11.25/1.25/9.07 & 11.69/2.43/10.07 & 20.70 & 15.01 & 26.80 & 9.06  & 12.23 \\
\midrule
cos2  & 2k  & 9.08/2.95/6.79   & 10.43/1.15/8.3  & 13.06/2.88/10.67 & 15.18 & 13.70 & 27.90 & 19.96 & 10.93 \\
cos2  & 4k  & 10.56/3.83/7.64  & 12.38/1.96/9.51 & 12.6/2.75/10.86  & 17.32 & 13.59 & 28.38 & 21.89 & 11.79 \\
cos2  & 8k  & 7.97/3.51/6.15   & 12.25/1.95/9.38 & 11.91/2.7/9.96   & 16.94 & 13.93 & 27.76 & 17.07 & 10.88 \\
cos2  & 16k & 17.92/5.71/12.17 & -               & 11.43/2.31/9.91  & 21.00 & 12.66 & 26.51 & 10.80 & 13.04 \\
\bottomrule
\end{tabular}
\end{table*}

\subsection{NIAH analysis}

The Needle in A Haystack (NIAH) evaluates language models in two modes: Easy Mode, where both the question and answer are embedded in a text for straightforward retrieval, and Standard Mode, where only the answer is embedded, requiring the model to comprehend the question and locate the answer, thereby adding complexity. In table~\ref{niah_easy_full},~\ref{niah_aspect_ratio} and ~\ref{niah_pretraining}, the upper sub-table displays the NIAH benchmark results in easy mode, while the lower sub-table shows the results in standard mode. 

Overall, LLaMA consistently outperforms other linear complexity sequence models in a variety of conditions, excelling in both retrieval-only and comprehension-inclusive tasks. Additionally, HGRN2 and cosFormer2 also demonstrate strong scaling capabilities, particularly in easy mode. TNL shows a more varied performance, performing decently in some contexts but not as uniformly strong as other models.

Across all architectures, performances are generally higher in the easy mode compared to the standard mode, which includes both retrieval and comprehension components. This suggests that the addition of comprehension tasks adds significant complexity and challenge.

The NIAH score, indicating efficiency in managing sparse or relevant information, is consistently highest for LLaMA, especially at larger context scales in easy mode. Both weighted average accuracy and average accuracy tend to follow similar trends, suggesting these metrics might be adjusted based on task difficulty or importance across various context scales.

\begin{table*}
\small
\centering
\caption{\textbf{Benchmark of Needle In A Haystack:} it presents accuracy metrics at four context scales: 2K, 4K, 8K, and 16K. Accuracies below the 4K and 8K thresholds are presented in the middle columns. Both average accuracy and weighted average accuracy are detailed, along with the NIAH score, in the rightmost columns.}
\label{niah_easy_full}
\setlength{\tabcolsep}{3.5mm}
\begin{tabular}{p{1.3cm}c|cccc|cc|cccc}
\toprule
Arch &P.S. & Acc &	Acc &	Acc &	Acc &	Acc &	Acc &	Acc &	Weighted &	NIAH \\
&B & $@$2k &	$@$4k &	$@$8k &	$@$16k &	$\leq$4k &	$\leq$8k &avg &	avg acc &	score \\ \midrule
\multicolumn{11}{c}{Easy Mode (Retrieval Only)} \\ \midrule
LLaMA & 0.07 & 3.2   & 1.3   & 0.0  & 0.0  & 1.0   & 0.7  & 0.4  & 0.4  & 0.2  \\
TNL   & 0.07 & 0.0   & 1.3   & 0.6  & 0.6  & 0.4   & 0.7  & 0.5  & 0.6  & 0.0  \\
HGRN2 & 0.07 & 0.0   & 0.0   & 0.0  & 0.0  & 0.0   & 0.3  & 0.2  & 0.2  & 0.0  \\
cos2  & 0.07 & 0.0   & 0.0   & 0.0  & 0.0  & 0.1   & 0.1  & 0.1  & 0.1  & 0.1  \\
\hline
LLaMA & 0.16 & 7.3   & 0.6   & 0.0  & 0.0  & 13.7  & 11.7 & 6.0  & 6.0  & 1.8  \\
TNL   & 0.16 & 2.5   & 3.5   & 24.4 & 4.4  & 5.6   & 10.1 & 7.3  & 7.5  & 1.7  \\
HGRN2 & 0.16 & 0.6   & 0.0   & 0.0  & 0.0  & 1.1   & 0.7  & 0.3  & 0.3  & 0.0  \\
cos2  & 0.16 & 0.0   & 5.4   & 0.0  & 0.0  & 2.5   & 2.1  & 1.2  & 1.3  & 0.9  \\
\hline
LLaMA & 0.41 & 100.0 & 97.1  & 97.8 & 0.0  & 99.3  & 99.5 & 56.4 & 52.3 & 45.5 \\
TNL   & 0.41 & 27.9  & 5.7   & 3.8  & 10.2 & 18.1  & 15.4 & 13.8 & 14.2 & 2.8  \\
HGRN2 & 0.41 & 8.6   & 6.3   & 1.3  & 0.0  & 17.0  & 9.3  & 4.9  & 4.8  & 1.4  \\
cos2  & 0.41 & 37.1  & 11.4  & 2.9  & 0.0  & 25.5  & 18.5 & 9.7  & 9.6  & 2.1  \\
\hline
LLaMA & 1    & 100.0 & 71.4  & 73.3 & 0.0  & 92.5  & 90.9 & 47.8 & 44.1 & 28.1 \\
TNL   & 1    & 43.5  & 8.9   & 21.3 & 8.6  & 30.8  & 28.7 & 27.6 & 28.0 & 2.7  \\
HGRN2 & 1    & 17.1  & 5.7   & 2.9  & 3.5  & 18.3  & 13.4 & 9.7  & 10.0 & 3.6  \\
cos2  & 1    & 54.9  & 5.7   & 2.9  & 0.0  & 37.8  & 22.7 & 11.5 & 10.9 & 2.0  \\
\hline
LLaMA & 3    & 97.1  & 100.0 & 82.9 & 0.6  & 95.4  & 93.9 & 48.8 & 45.1 & 29.9 \\
TNL   & 3    & 0.0   & 26.0  & 3.2  & 9.5  & 9.7   & 12.9 & 10.4 & 11.1 & 2.3  \\
HGRN2 & 3    & 58.4  & 11.4  & 2.9  & 7.3  & 46.4  & 28.9 & 18.0 & 17.9 & 4.2  \\
cos2  & 3    & 97.1  & 34.3  & 8.6  & 0.0  & 86.8  & 54.7 & 27.8 & 25.8 & 14.3 \\
\hline
LLaMA & 7    & 100.0 & 100.0 & 87.0 & 0.0  & 100.0 & 98.4 & 62.9 & 59.7 & 44.7 \\
TNL   & 7    & 43.5  & 14.3  & 12.4 & 18.1 & 38.8  & 26.8 & 20.2 & 20.5 & 7.8  \\
HGRN2 & 7    & 100.0 & 28.6  & 14.6 & 11.4 & 83.8  & 50.1 & 31.3 & 30.8 & 17.1 \\
cos2  & 7    & 97.1  & 37.1  & 8.6  & 0.0  & 78.6  & 48.0 & 25.1 & 23.6 & 11.6 \\

\midrule
\multicolumn{11}{c}{Standard Mode (Retrieval and Comprehension)} \\
\midrule

LLaMA & 0.07 & 1.9   & 0.0   & 0.6  & 0.0  & 0.6   & 0.3  & 0.2  & 0.2  & 0.1  \\
TNL   & 0.07 & 0.0   & 0.6   & 0.0  & 0.0  & 0.3   & 0.4  & 0.5  & 0.5  & 0.1  \\
HGRN2 & 0.07 & 0.0   & 0.6   & 0.0  & 0.0  & 0.4   & 0.3  & 0.2  & 0.2  & 0.1  \\
cos2  & 0.07 & 0.0   & 0.0   & 0.0  & 0.0  & 0.2   & 0.1  & 0.1  & 0.1  & 0.0  \\
\hline
LLaMA & 0.16 & 1.9   & 0.6   & 11.4 & 0.0  & 11.8  & 7.9  & 4.0  & 3.6  & 0.7  \\
TNL   & 0.16 & 1.3   & 7.0   & 23.5 & 0.0  & 5.2   & 7.0  & 6.3  & 6.4  & 1.3  \\
HGRN2 & 0.16 & 0.0   & 0.0   & 0.0  & 0.0  & 0.8   & 0.4  & 0.2  & 0.2  & 0.0  \\
cos2  & 0.16 & 0.0   & 0.0   & 0.0  & 0.0  & 0.7   & 0.4  & 0.2  & 0.2  & 0.1  \\
\hline
LLaMA & 0.41 & 100.0 & 100.0 & 92.7 & 0.0  & 100.0 & 99.6 & 55.4 & 51.4 & 43.5 \\
TNL   & 0.41 & 12.4  & 5.1   & 5.7  & 8.9  & 12.8  & 12.0 & 10.3 & 10.4 & 1.3  \\
HGRN2 & 0.41 & 0.0   & 0.0   & 0.0  & 0.0  & 8.3   & 4.2  & 2.3  & 2.1  & 0.4  \\
cos2  & 0.41 & 0.0   & 8.6   & 3.5  & 0.0  & 3.7   & 3.3  & 1.8  & 1.9  & 1.1  \\
\hline
LLaMA & 1    & 100.0 & 82.9  & 47.6 & 0.0  & 97.9  & 83.7 & 43.2 & 40.1 & 24.4 \\
TNL   & 1    & 8.3   & 7.0   & 34.6 & 22.2 & 20.9  & 22.9 & 23.6 & 24.0 & 1.2  \\
HGRN2 & 1    & 12.7  & 5.7   & 14.6 & 0.0  & 15.9  & 11.6 & 7.5  & 7.4  & 1.8  \\
cos2  & 1    & 76.8  & 5.7   & 2.9  & 0.0  & 24.3  & 13.6 & 7.0  & 6.6  & 2.0  \\
\hline
LLaMA & 3    & 99.4  & 100.0 & 42.9 & 0.0  & 98.3  & 85.5 & 43.4 & 40.1 & 20.0 \\
TNL   & 3    & 2.9   & 17.1  & 2.9  & 44.8 & 8.7   & 11.6 & 10.3 & 10.8 & 3.1  \\
HGRN2 & 3    & 54.0  & 5.7   & 2.9  & 8.9  & 32.2  & 18.0 & 12.7 & 12.7 & 3.1  \\
cos2  & 3    & 54.9  & 9.2   & 8.6  & 0.0  & 38.7  & 24.1 & 12.8 & 12.5 & 4.7  \\
\hline
LLaMA & 7    & 100.0 & 100.0 & 85.4 & 0.0  & 100.0 & 96.2 & 58.6 & 55.4 & 45.5 \\
TNL   & 7    & 28.9  & 13.7  & 19.7 & 44.1 & 29.4  & 21.8 & 18.8 & 19.4 & 6.3  \\
HGRN2 & 7    & 65.4  & 5.7   & 5.1  & 4.1  & 48.5  & 28.6 & 18.4 & 18.3 & 7.5  \\
cos2  & 7    & 74.3  & 5.7   & 11.4 & 0.0  & 51.6  & 31.0 & 16.8 & 16.0 & 7.6  \\
 
\bottomrule
\end{tabular}%
% }
\end{table*}

\begin{table*}
\small
\centering
\caption{\textbf{Benchmark of Needle In A Haystack by Aspect Ratio}}
\label{niah_aspect_ratio}
% \resizebox{\columnwidth}{!}{%
\setlength{\tabcolsep}{3.0mm}
\begin{tabular}{p{1.3cm}c|cccc|cc|cccc}
\toprule
Arch &Dim & Acc &	Acc &	Acc &	Acc &	Acc &	Acc &	Acc &	Weighted &	NIAH \\
& & $@$2k &	$@$4k &	$@$8k &	$@$16k &	$\leq$4k &	$\leq$8k &avg &	avg acc &	score \\ \midrule
\multicolumn{11}{c}{Easy Mode (Retrieval Only)} \\ \midrule
LLaMA & 1536 & 100.0 & 71.4  & 73.3 & 0.0 & 92.5  & 90.9 & 47.8 & 44.1 & 28.1 \\
LLaMA & 1792 & 100.0 & 100.0 & 82.9 & 0.0 & 100.0 & 96.8 & 57.4 & 54.2 & 45.0 \\
LLaMA & 2048 & 100.0 & 100.0 & 62.9 & 0.0 & 100.0 & 92.9 & 48.7 & 45.0 & 39.2 \\
LLaMA & 3072 & 100.0 & 100.0 & 94.3 & 0.3 & 92.5  & 95.6 & 49.8 & 46.0 & 38.1 \\
\hline
cos2  & 1536 & 54.9  & 5.7   & 2.9  & 0.0 & 37.8  & 22.7 & 11.5 & 10.9 & 2.0  \\
cos2  & 1792 & 20.0  & 5.7   & 5.7  & 0.0 & 31.8  & 18.2 & 9.2  & 8.8  & 3.6  \\
cos2  & 2048 & 82.9  & 5.7   & 2.9  & 0.0 & 49.6  & 26.4 & 13.4 & 12.3 & 4.7  \\
cos2  & 3072 & 1.0   & 5.7   & 2.9  & 0.0 & 21.7  & 13.6 & 6.9  & 6.5  & 1.6  \\
% \bottomrule
\midrule
\multicolumn{11}{c}{Standard Mode (Retrieval and Comprehension)} \\
\midrule
LLaMA & 1536 & 100.0 & 82.9  & 47.6 & 0.0 & 97.9  & 83.7 & 43.2 & 40.1 & 24.4 \\
LLaMA & 1792 & 100.0 & 100.0 & 57.1 & 0.0 & 100.0 & 92.3 & 51.5 & 48.3 & 38.8 \\
LLaMA & 2048 & 100.0 & 100.0 & 57.1 & 0.6 & 100.0 & 90.8 & 48.3 & 44.9 & 38.3 \\
LLaMA & 3072 & 97.1  & 100.0 & 74.3 & 0.0 & 97.2  & 92.3 & 47.1 & 43.3 & 26.5 \\
\hline
cos2  & 1536 & 76.8  & 5.7   & 2.9  & 0.0 & 24.3  & 13.6 & 7.0  & 6.6  & 2.0  \\
cos2  & 1792 & 5.7   & 3.5   & 1.9  & 0.0 & 11.5  & 6.9  & 3.8  & 3.9  & 2.0  \\
cos2  & 2048 & 24.8  & 5.7   & 2.9  & 0.0 & 25.0  & 13.3 & 6.8  & 6.4  & 4.1  \\
cos2  & 3072 & 0.3   & 2.9   & 0.0  & 0.0 & 5.9   & 3.9  & 2.0  & 2.0  & 1.0  \\
\bottomrule
\end{tabular}%
% }
\end{table*}

\begin{table*}
\small
\centering
\caption{\textbf{Benchmark of Needle In A Haystack by Pre-training Context Length}}
\label{niah_pretraining}
% \resizebox{\columnwidth}{!}{%
\setlength{\tabcolsep}{3.0mm}
\begin{tabular}{p{1.3cm}c|cccc|cc|cccc}
\toprule
Arch &Len & Acc &	Acc &	Acc &	Acc &	Acc &	Acc &	Acc &	Weighted &	NIAH \\
& & $@$2k &	$@$4k &	$@$8k &	$@$16k &	$\leq$4k &	$\leq$8k &avg &	avg acc &	score \\ \midrule
\multicolumn{11}{c}{Easy Mode (Retrieval Only)} \\ \midrule
TNL   & 2k  & 58.1 & 7.3  & 2.9  & 4.8  & 31.0 & 15.5 & 8.7  & 8.0  & 1.6 \\
TNL   & 4k  & 25.7 & 17.1 & 2.9  & 11.1 & 23.9 & 15.4 & 11.6 & 12.1 & 4.2 \\
TNL   & 8k  & 43.5 & 8.9  & 21.3 & 8.6  & 30.8 & 28.7 & 27.6 & 28.0 & 2.7 \\
TNL   & 16k & 43.2 & 13.3 & 16.8 & 4.4  & 21.4 & 16.5 & 13.7 & 14.2 & 2.7 \\
\hline
HGRN2 & 2k  & 0.0  & 2.9  & 3.5  & 0.0  & 1.1  & 2.9  & 2.1  & 2.3  & 1.0 \\
HGRN2 & 4k  & 5.7  & 2.9  & 1.3  & 0.6  & 4.0  & 3.3  & 2.1  & 2.1  & 0.8 \\
HGRN2 & 8k  & 17.1 & 5.7  & 2.9  & 3.5  & 18.3 & 13.4 & 9.7  & 10.0 & 3.6 \\
HGRN2 & 16k & 20.0 & 8.3  & 11.1 & 3.8  & 23.0 & 13.0 & 8.7  & 8.8  & 3.2 \\
\hline
cos2  & 2k  & 62.9 & 0.0  & 0.0  & 0.0  & 34.2 & 16.1 & 7.8  & 6.8  & 2.1 \\
cos2  & 4k  & 65.4 & 10.5 & 0.0  & 0.0  & 30.3 & 14.8 & 7.2  & 6.5  & 1.7 \\
cos2  & 8k  & 54.9 & 5.7  & 2.9  & 0.0  & 37.8 & 22.7 & 11.5 & 10.9 & 2.0 \\
cos2  & 16k & 17.1 & 2.9  & 2.9  & 0.0  & 22.9 & 14.2 & 9.4  & 9.6  & 3.9 \\
\midrule
\multicolumn{11}{c}{Standard Mode (Retrieval and Comprehension)} \\
\midrule
TNL   & 2k  & 14.0 & 3.5  & 3.8  & 17.5 & 16.3 & 9.0  & 6.4  & 6.5  & 2.0 \\
TNL   & 4k  & 5.7  & 10.2 & 3.5  & 12.1 & 11.4 & 8.5  & 7.9  & 8.4  & 2.7 \\
TNL   & 8k  & 8.3  & 7.0  & 34.6 & 22.2 & 20.9 & 22.9 & 23.6 & 24.0 & 1.2 \\
TNL   & 16k & 10.2 & 12.1 & 16.5 & 9.2  & 17.7 & 14.5 & 11.3 & 11.3 & 1.4 \\
\hline
HGRN2 & 2k  & 0.6  & 0.0  & 3.5  & 0.0  & 0.7  & 1.6  & 1.3  & 1.4  & 0.6 \\
HGRN2 & 4k  & 3.2  & 1.3  & 1.3  & 0.0  & 4.6  & 3.6  & 2.0  & 1.8  & 0.2 \\
HGRN2 & 8k  & 12.7 & 5.7  & 14.6 & 0.0  & 15.9 & 11.6 & 7.5  & 7.4  & 1.8 \\
HGRN2 & 16k & 5.4  & 9.2  & 0.6  & 17.5 & 11.3 & 6.5  & 4.7  & 4.7  & 1.1 \\
\hline
cos2  & 2k  & 37.5 & 0.0  & 0.0  & 0.0  & 21.9 & 10.3 & 5.0  & 4.4  & 0.4 \\
cos2  & 4k  & 34.3 & 5.7  & 0.0  & 0.0  & 21.4 & 10.2 & 5.0  & 4.5  & 1.5 \\
cos2  & 8k  & 76.8 & 5.7  & 2.9  & 0.0  & 24.3 & 13.6 & 7.0  & 6.6  & 2.0 \\
cos2  & 16k & 5.7  & 2.9  & 2.9  & 0.0  & 4.6  & 3.5  & 2.8  & 3.0  & 1.6 \\

\bottomrule
\end{tabular}%
% }
\end{table*}

\subsection{NIAH heatmaps by easy mode}
The figures below provide a heatmap visualization of NIAH in easy mode.

\begin{figure*}
    \centering
    \includegraphics[width=1\linewidth]{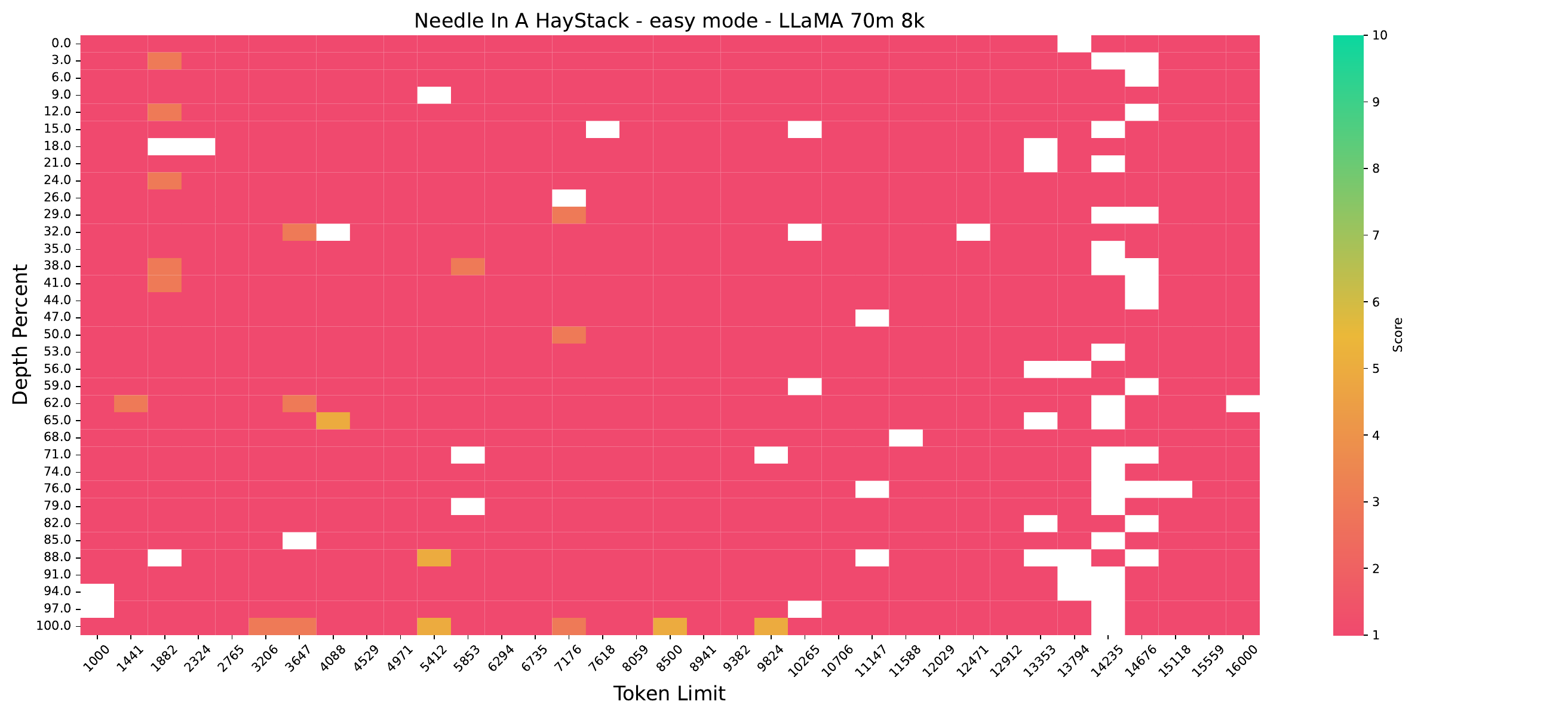}
    \includegraphics[width=1\linewidth]{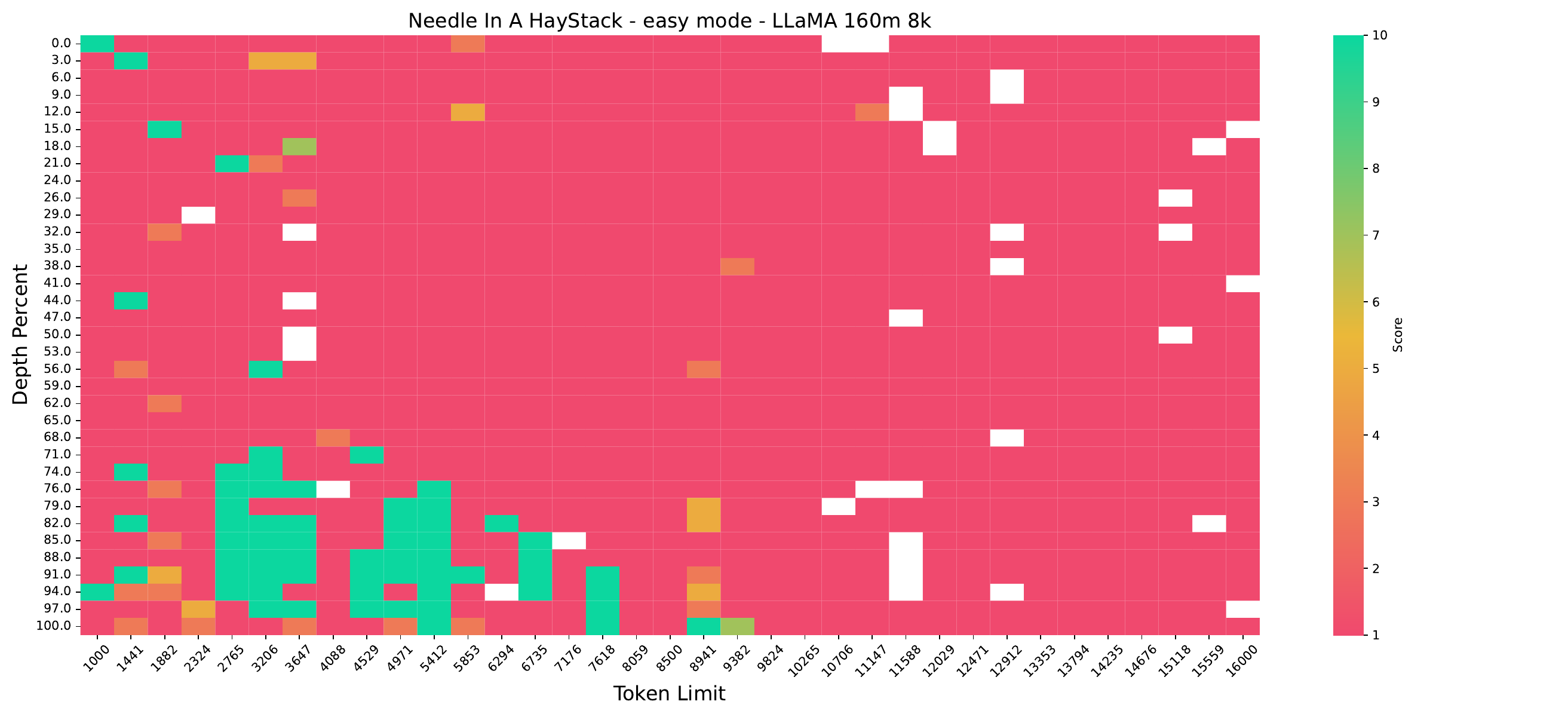}
    \includegraphics[width=1\linewidth]{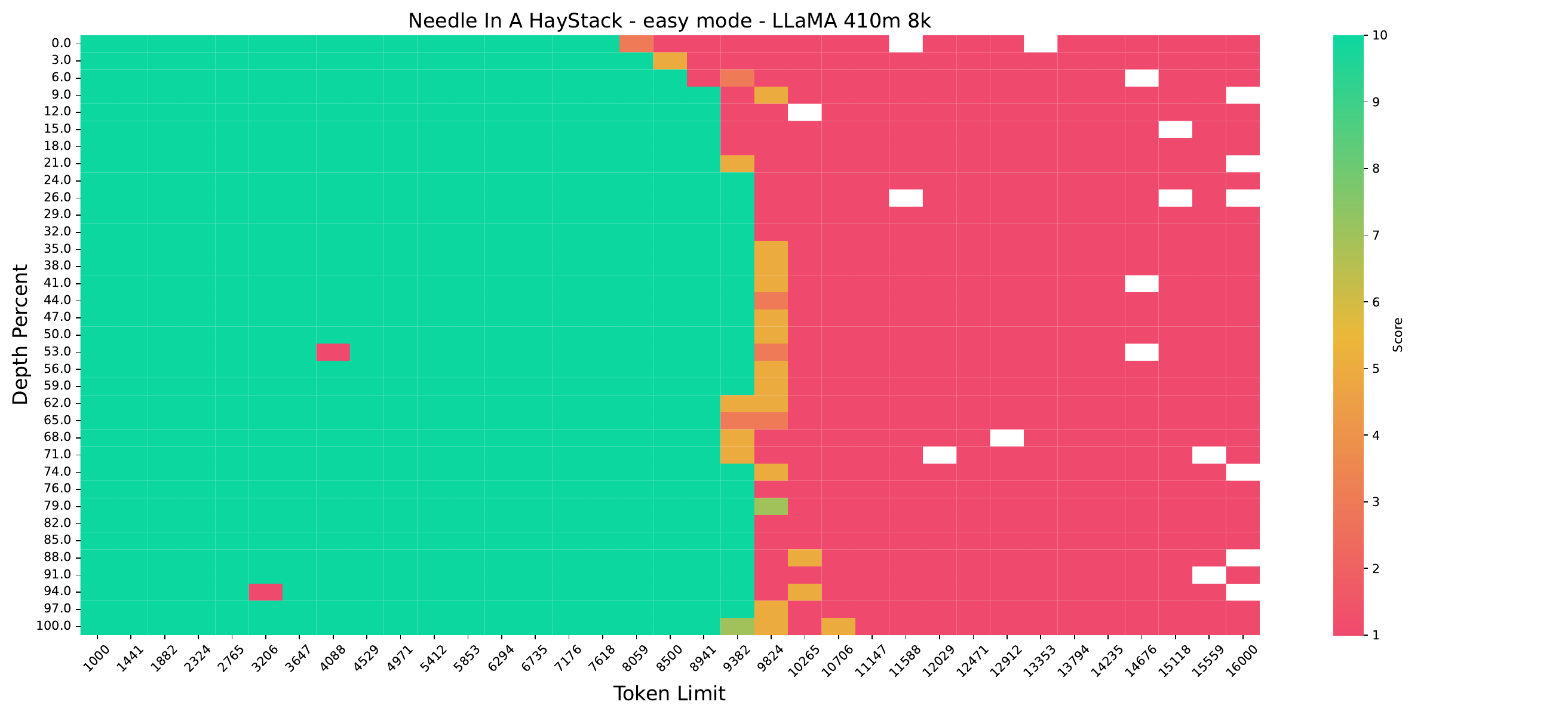}
\end{figure*}

\begin{figure*}
    \centering
    \includegraphics[width=1\linewidth]{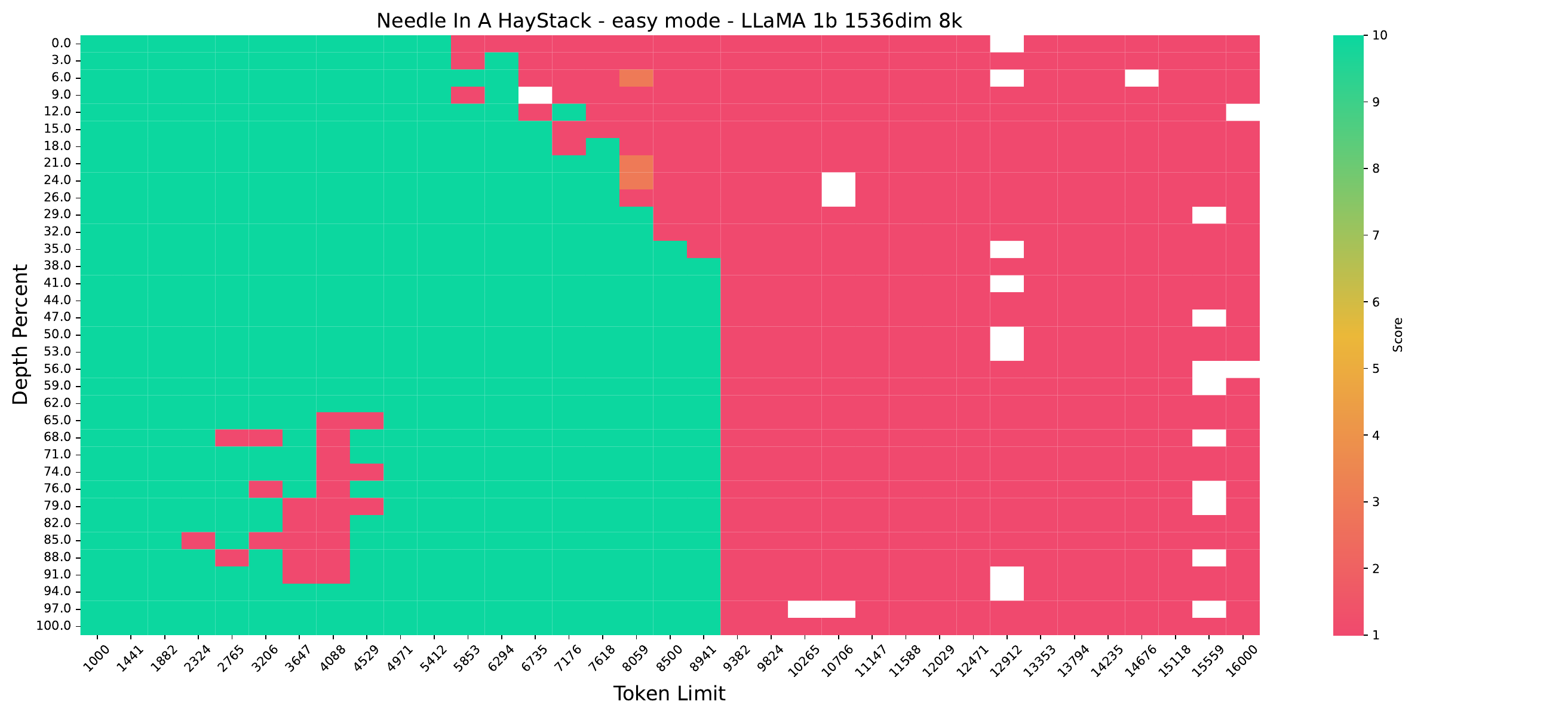}
    \includegraphics[width=1\linewidth]{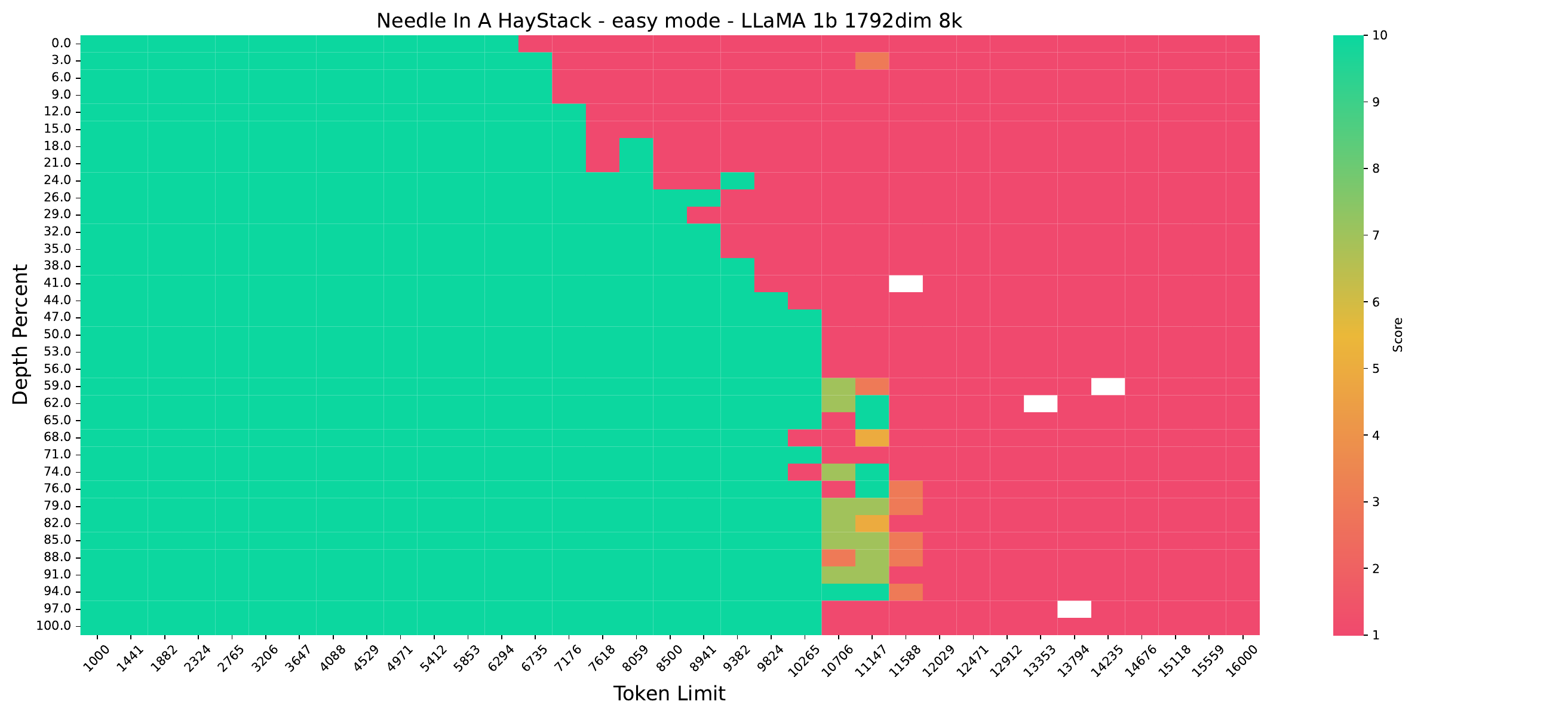}
    \includegraphics[width=1\linewidth]{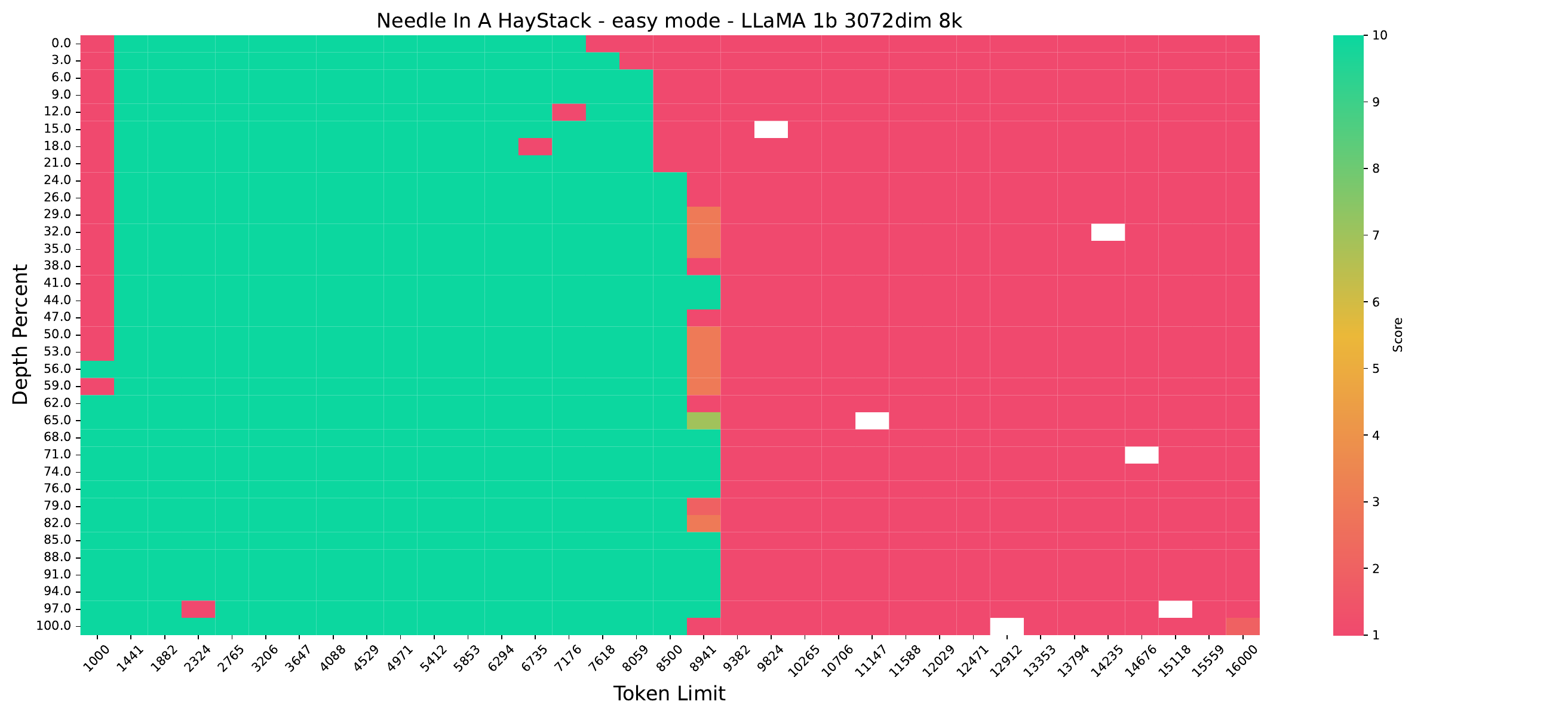}
\end{figure*}

\begin{figure*}
\centering
\includegraphics[width=1\linewidth]{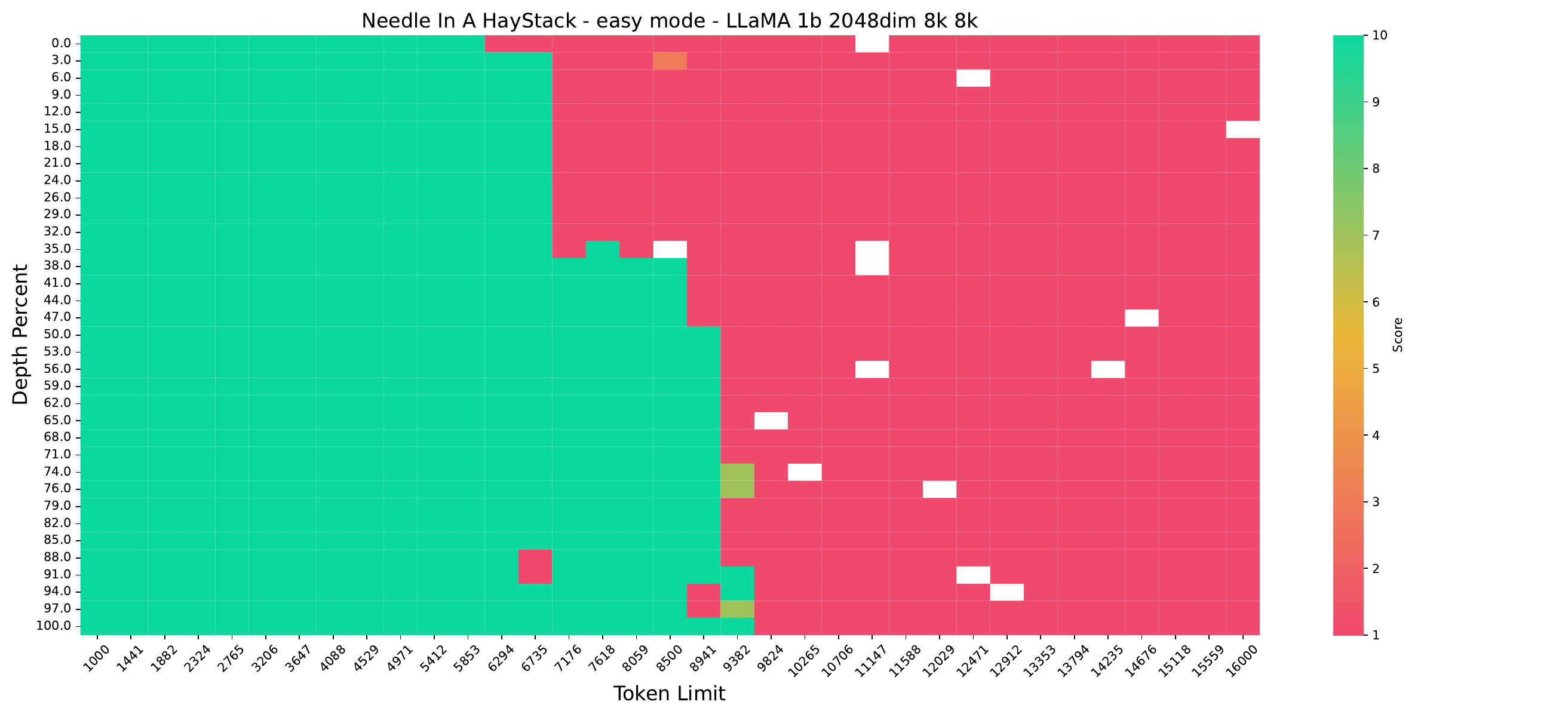}
\includegraphics[width=1\linewidth]{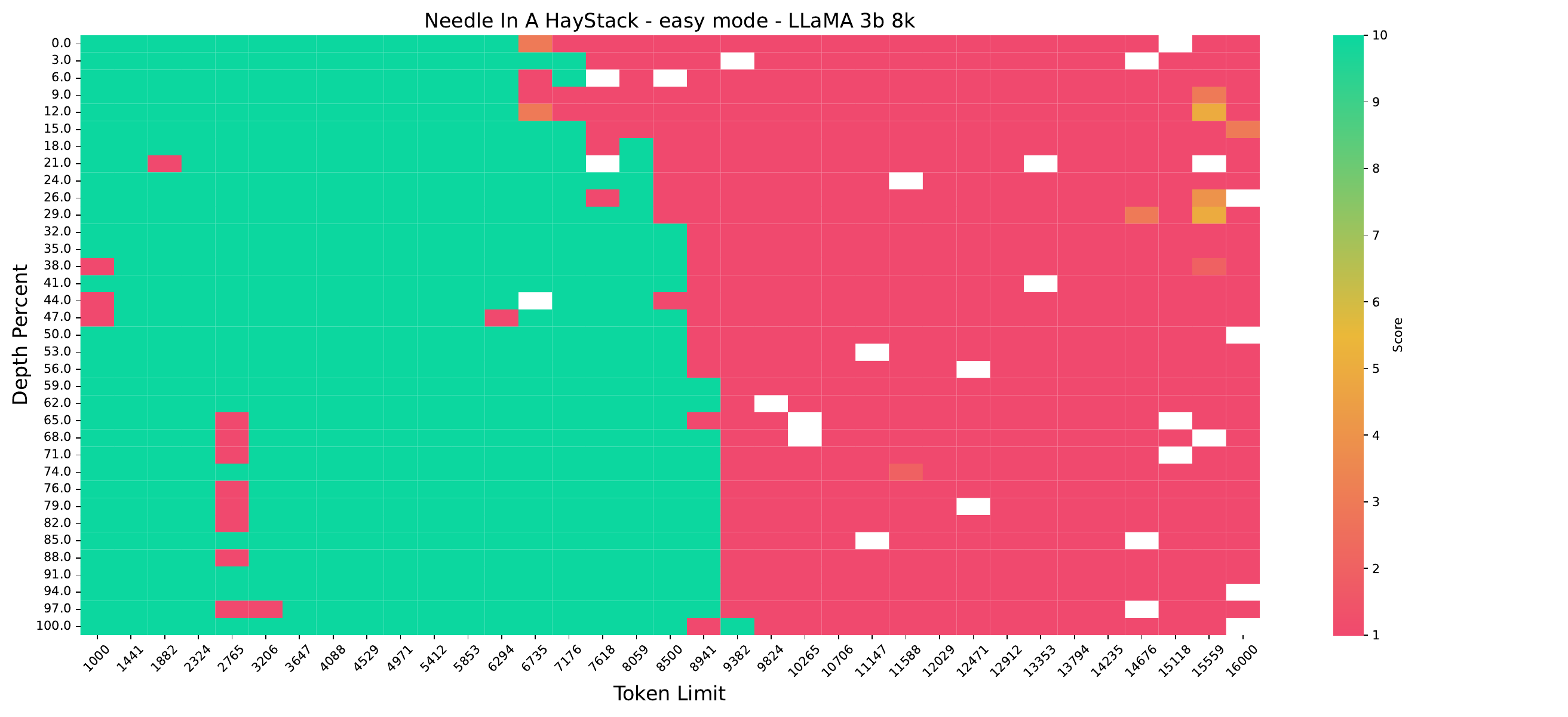}
\includegraphics[width=1\linewidth]{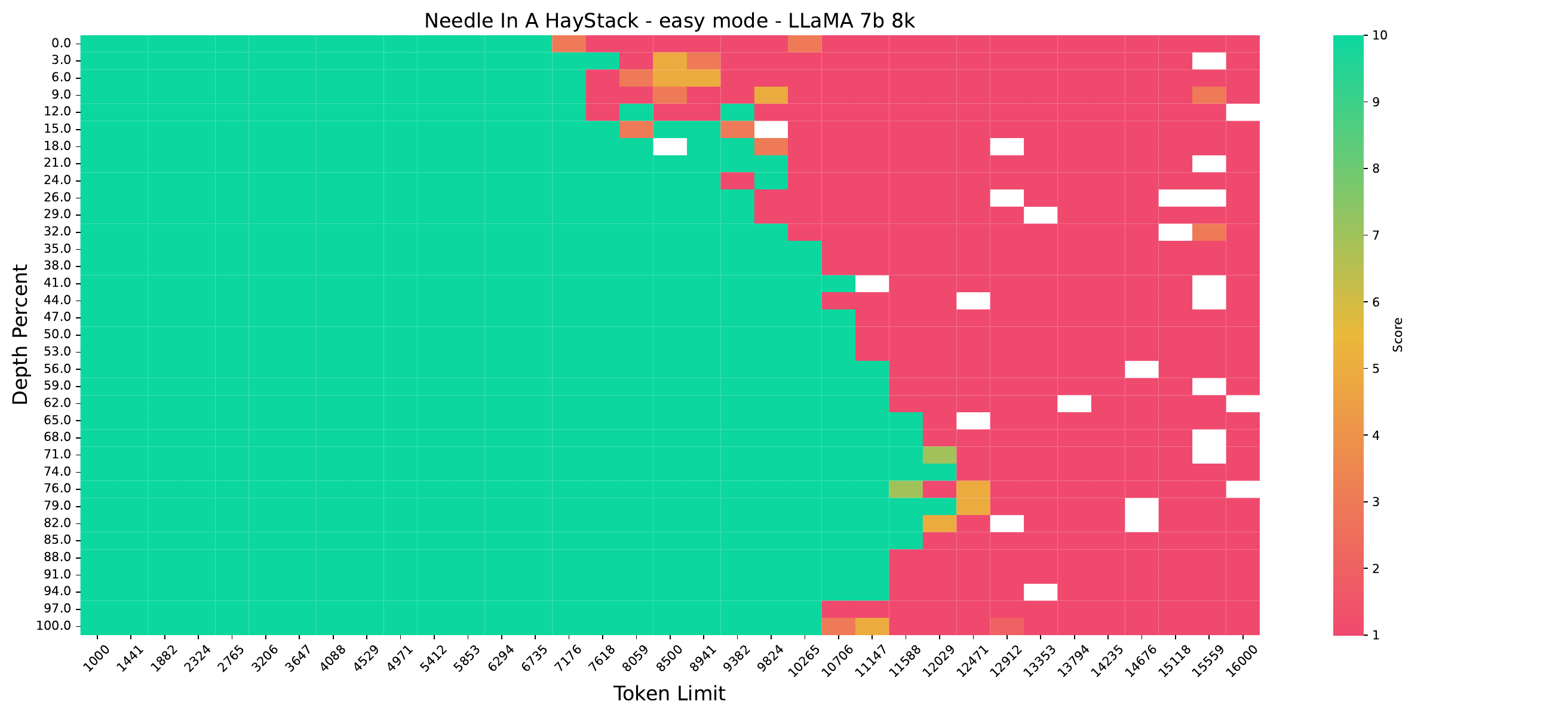}
\end{figure*}

\begin{figure*}
\centering
\includegraphics[width=1\linewidth]{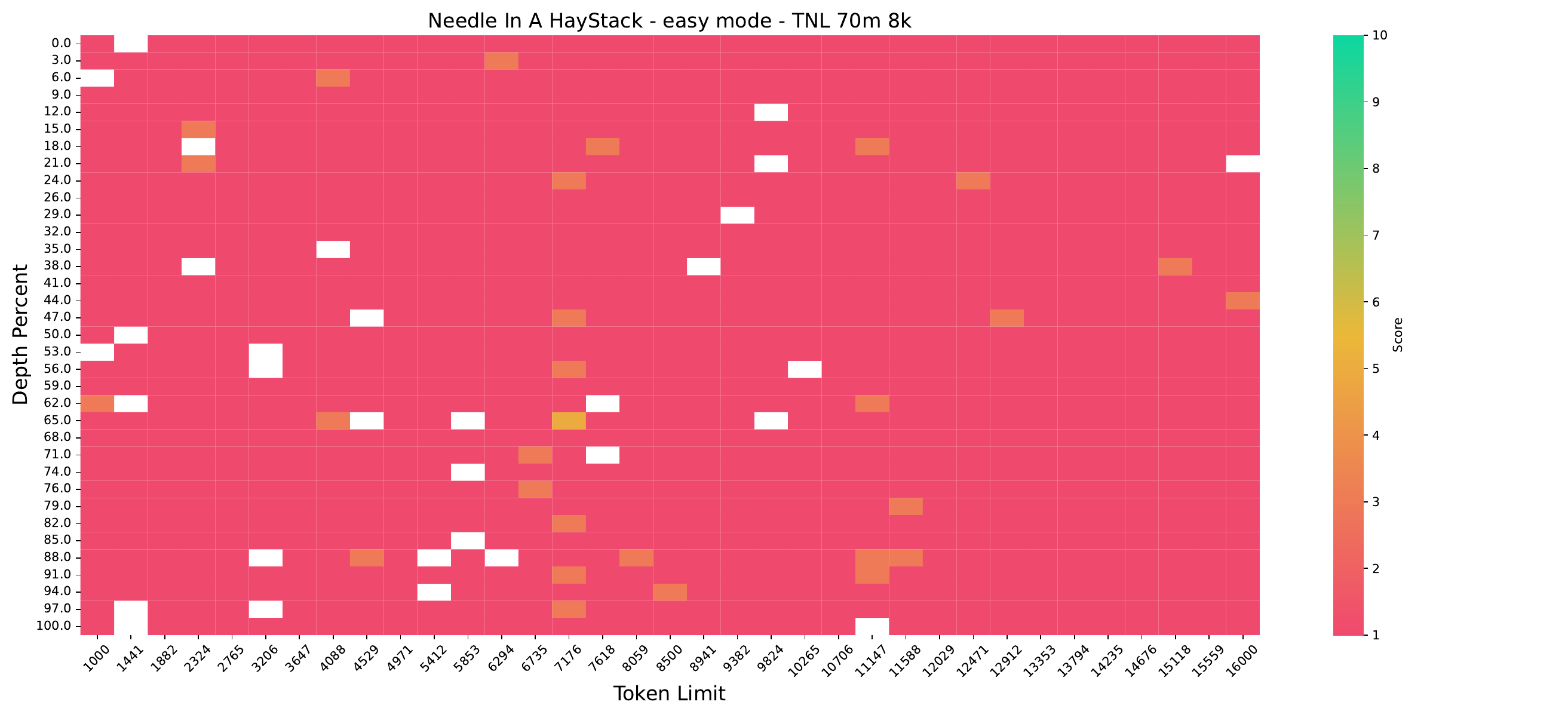}
\includegraphics[width=1\linewidth]{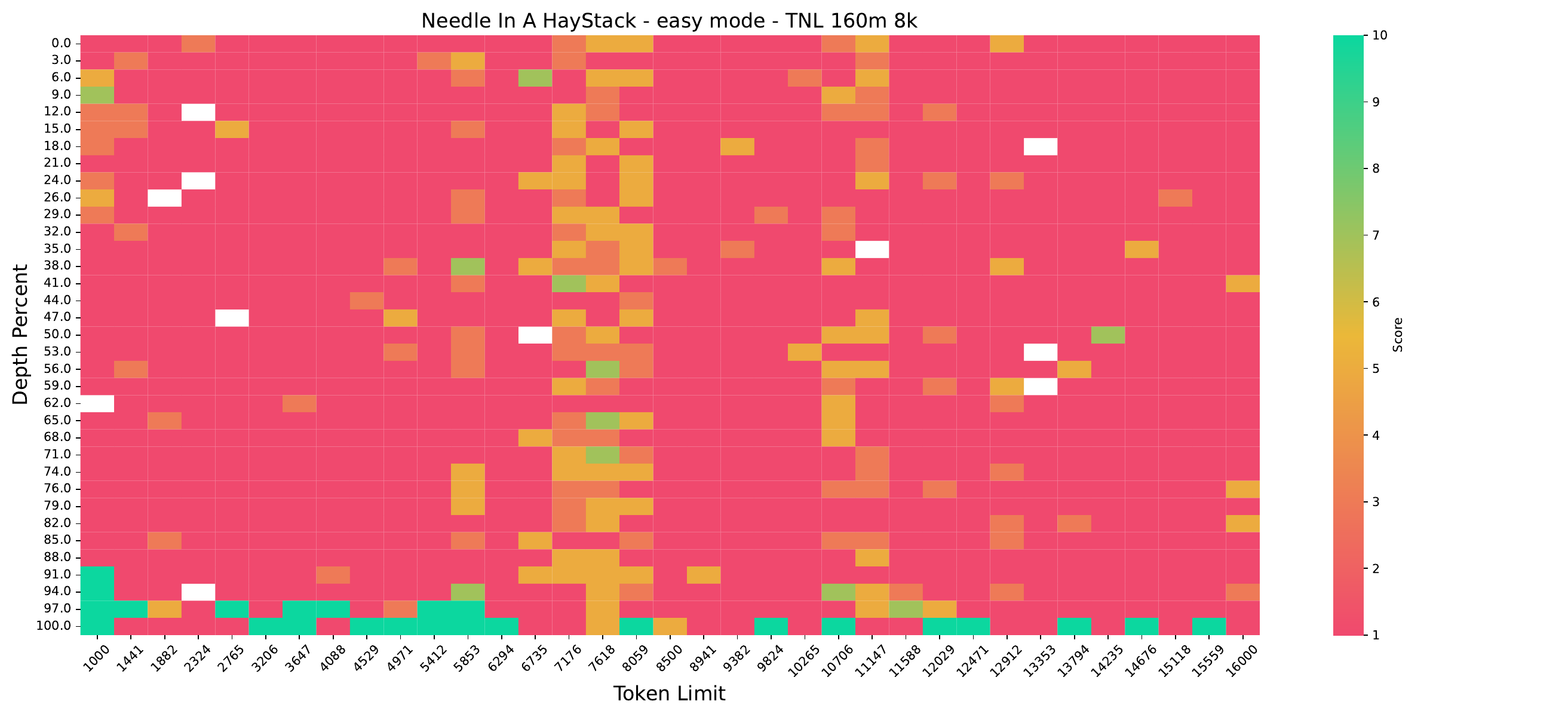}
\includegraphics[width=1\linewidth]{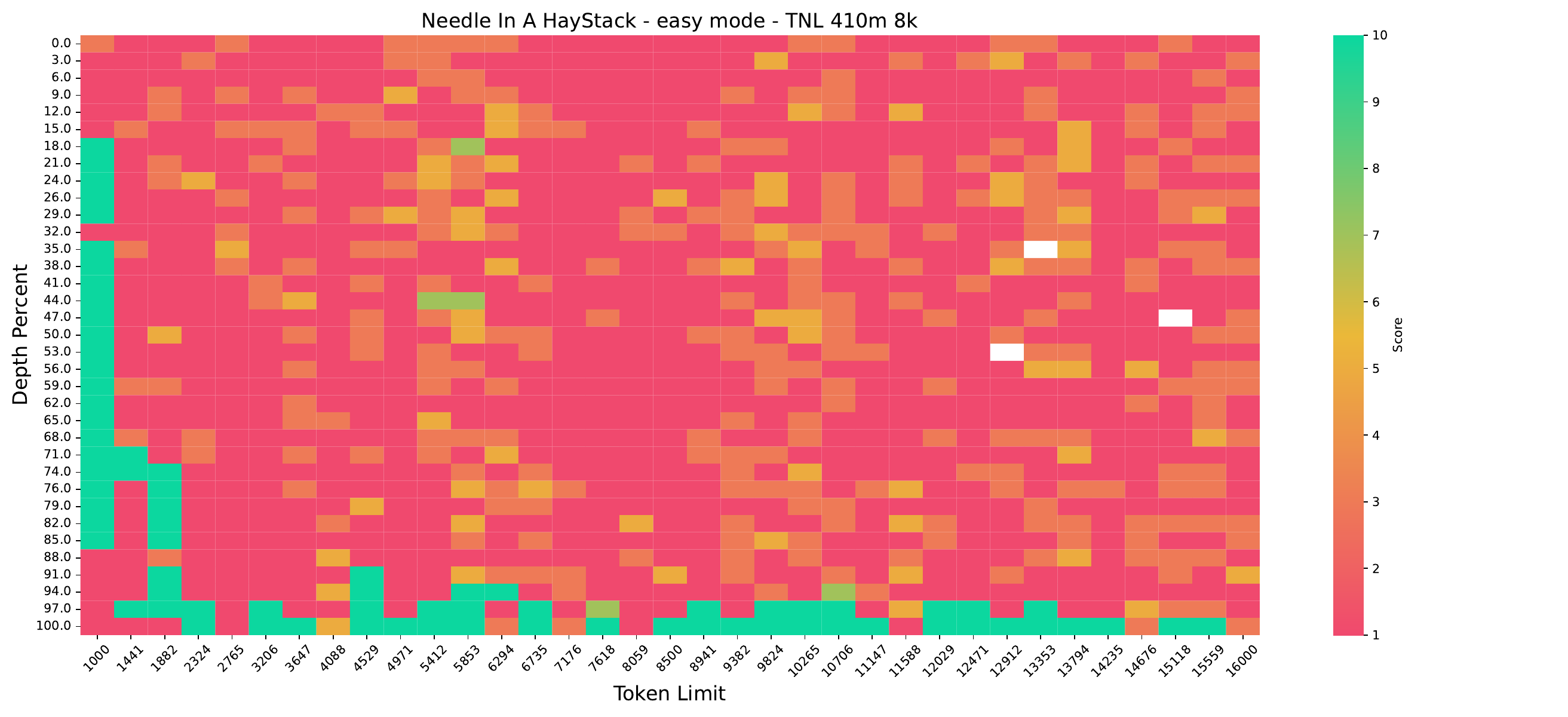}
\end{figure*}

\begin{figure*}
\centering
\includegraphics[width=1\linewidth]{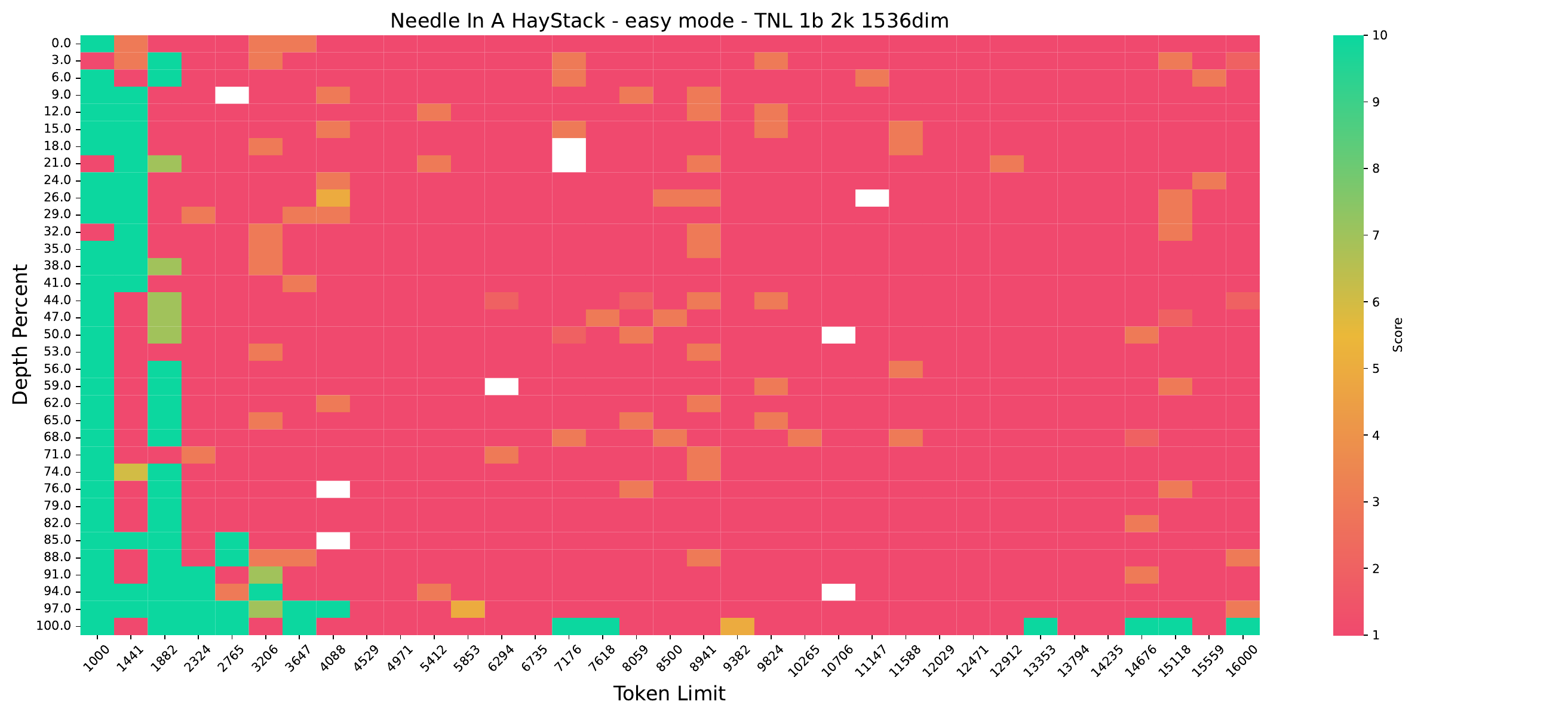}
\includegraphics[width=1\linewidth]{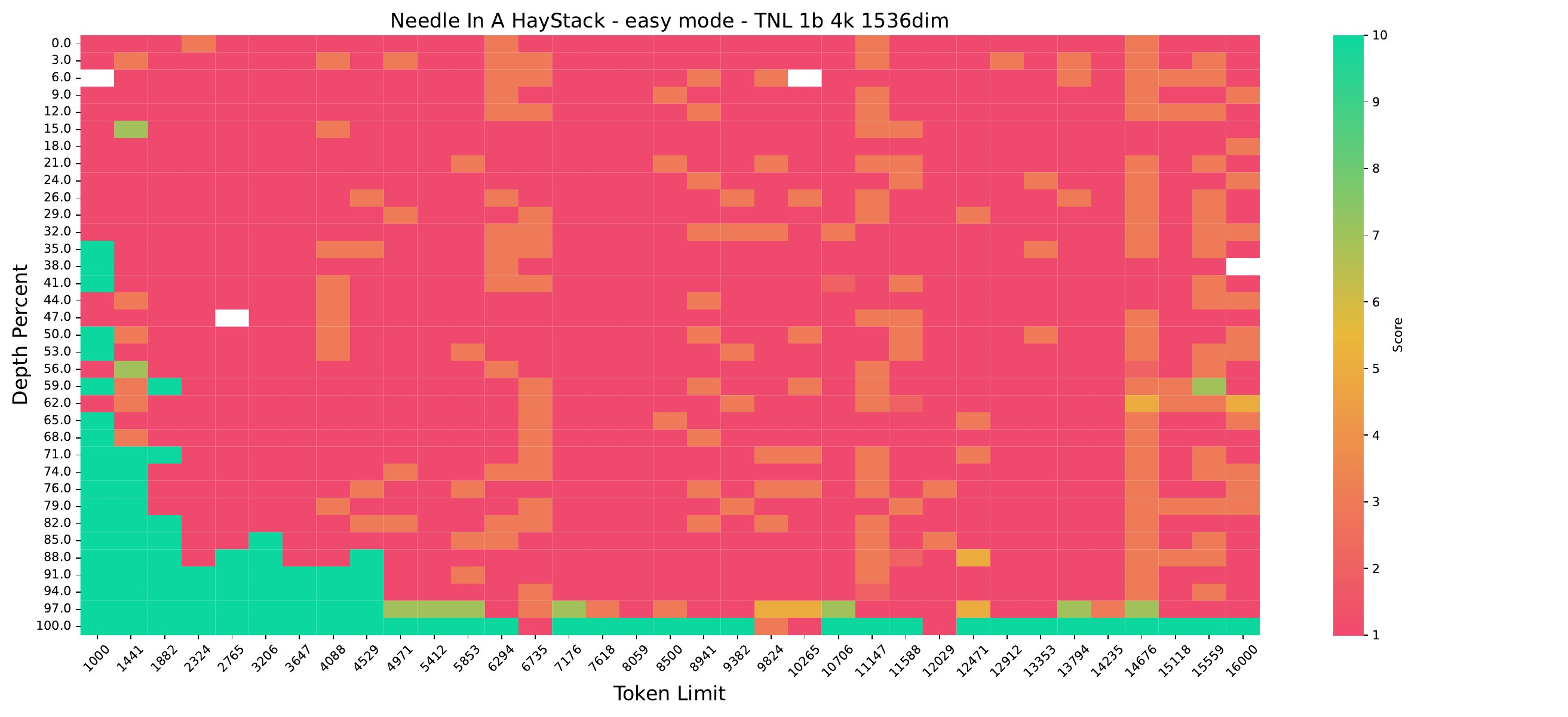}
\includegraphics[width=1\linewidth]{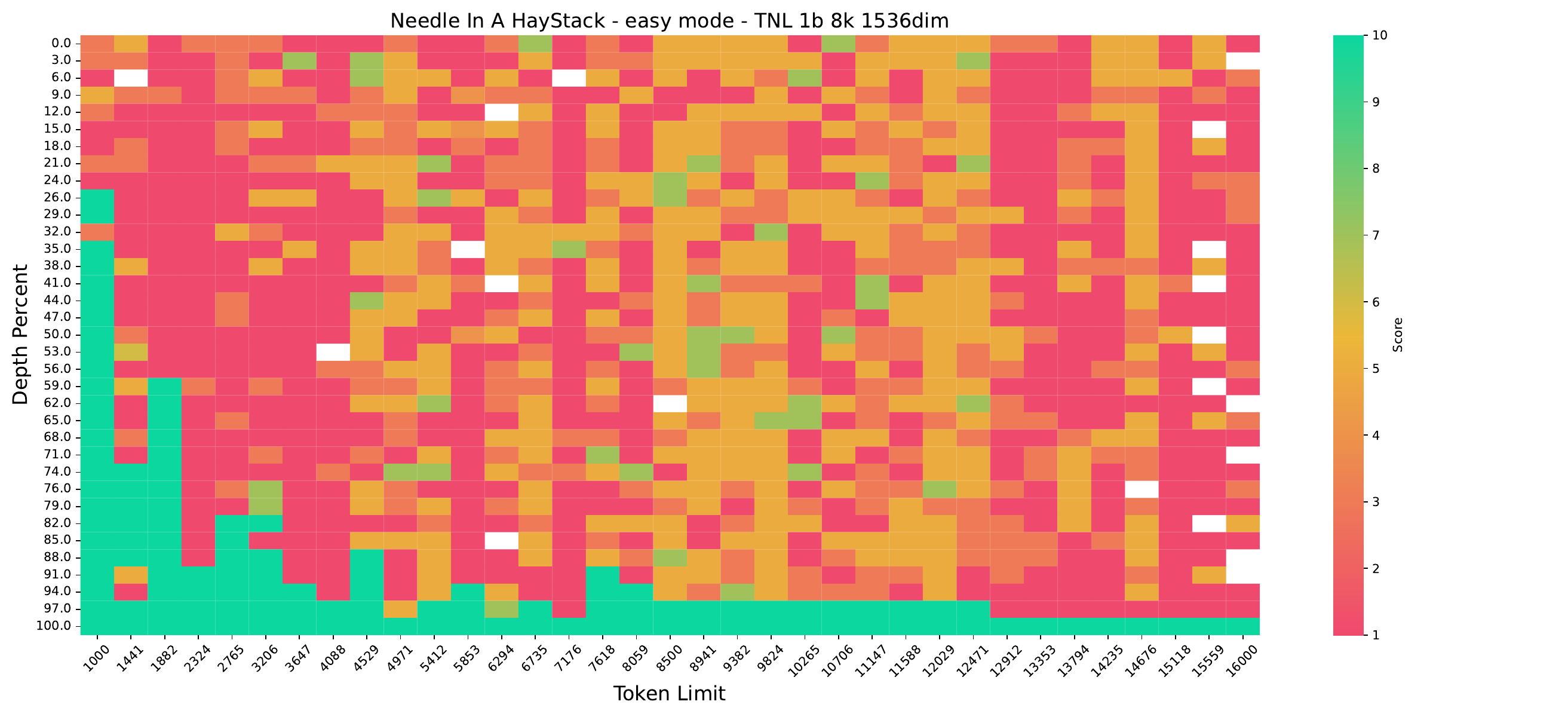}
\end{figure*}

\begin{figure*}
\centering
\includegraphics[width=1\linewidth]{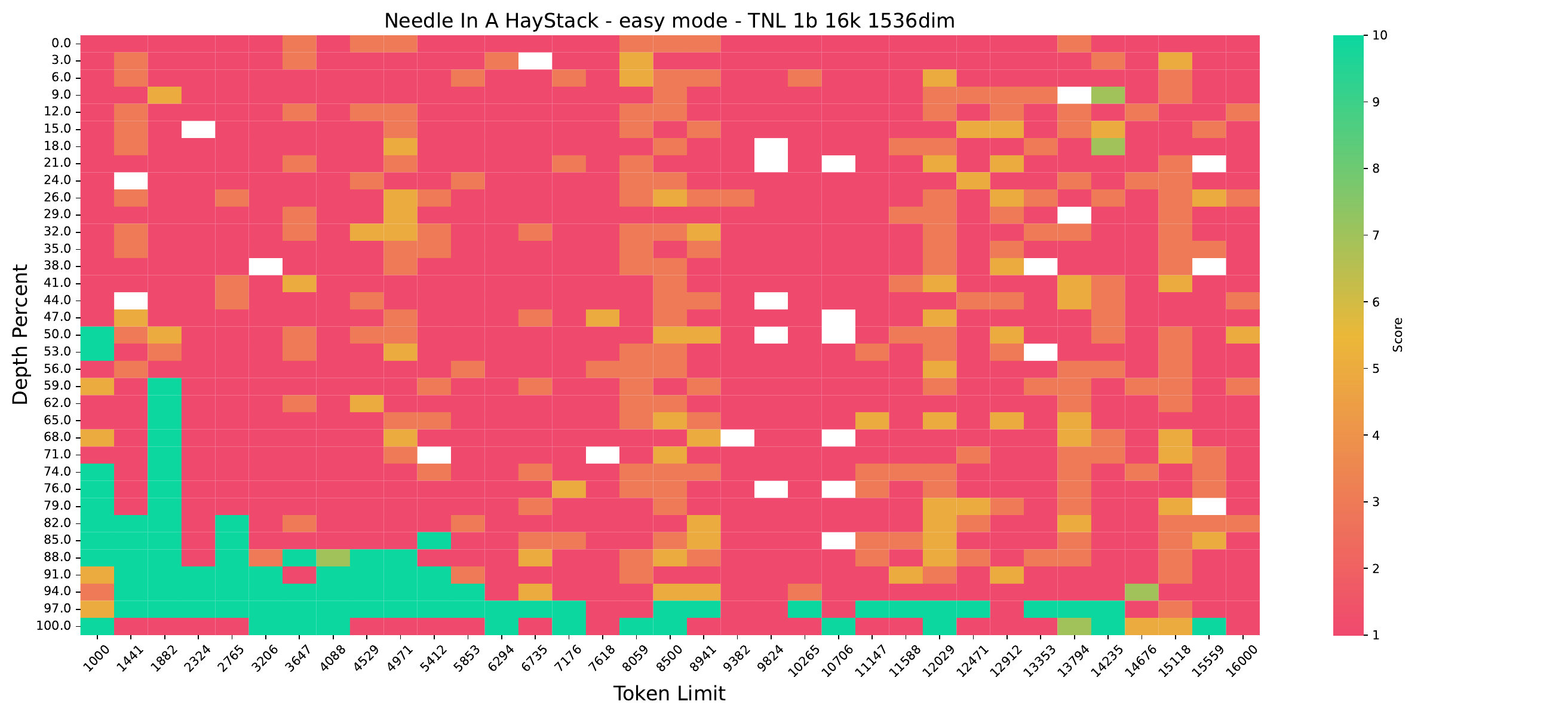}
\includegraphics[width=1\linewidth]{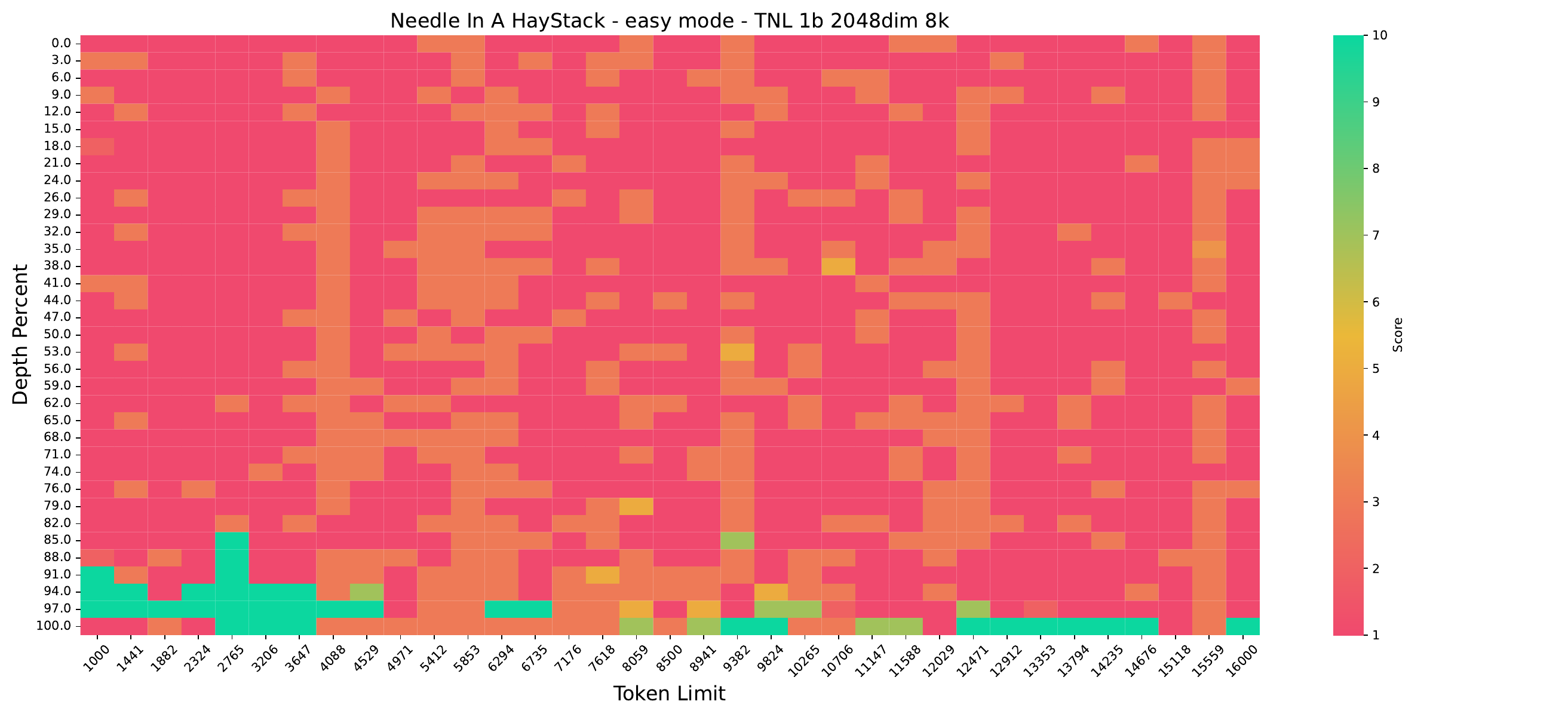}
\includegraphics[width=1\linewidth]{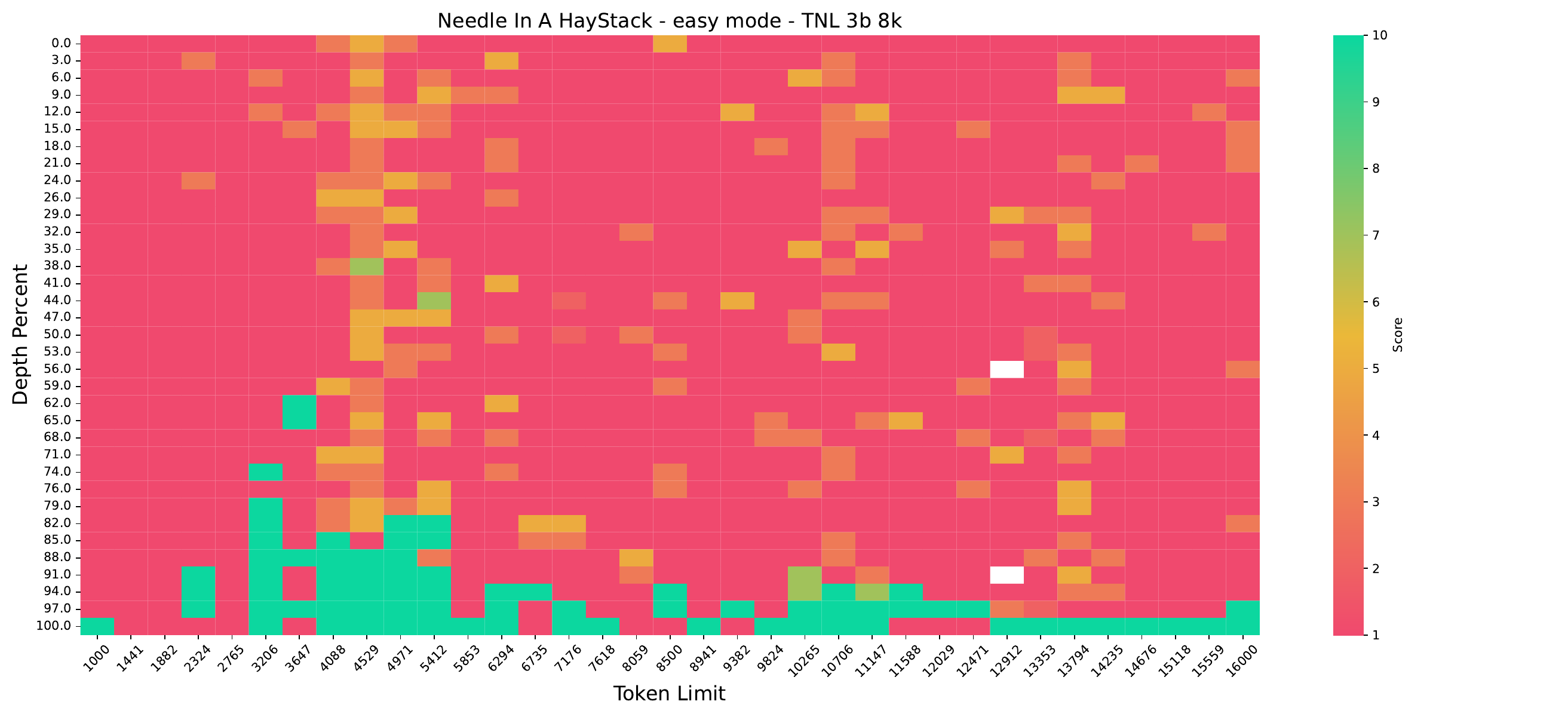}
\end{figure*}

\begin{figure*}
\centering
\includegraphics[width=1\linewidth]{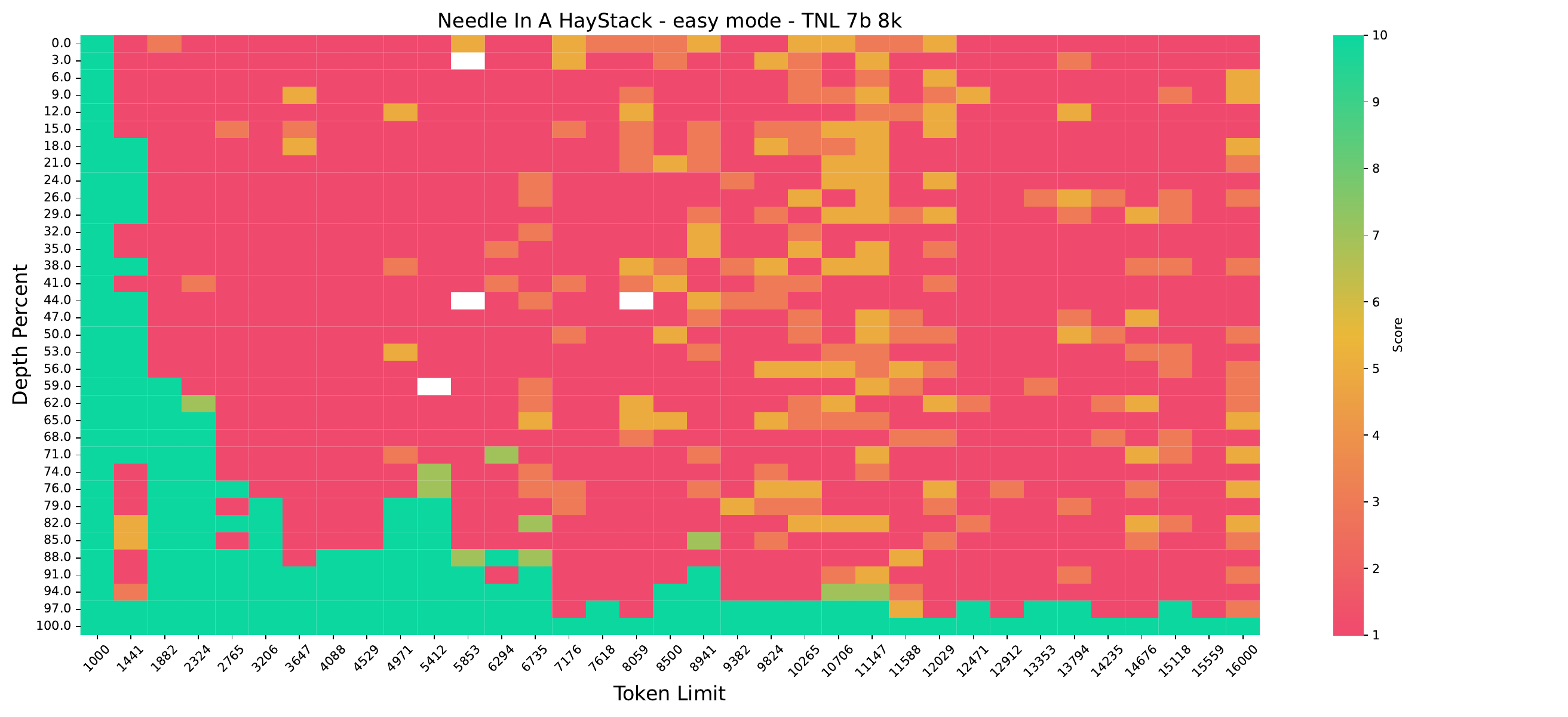}
\includegraphics[width=1\linewidth]{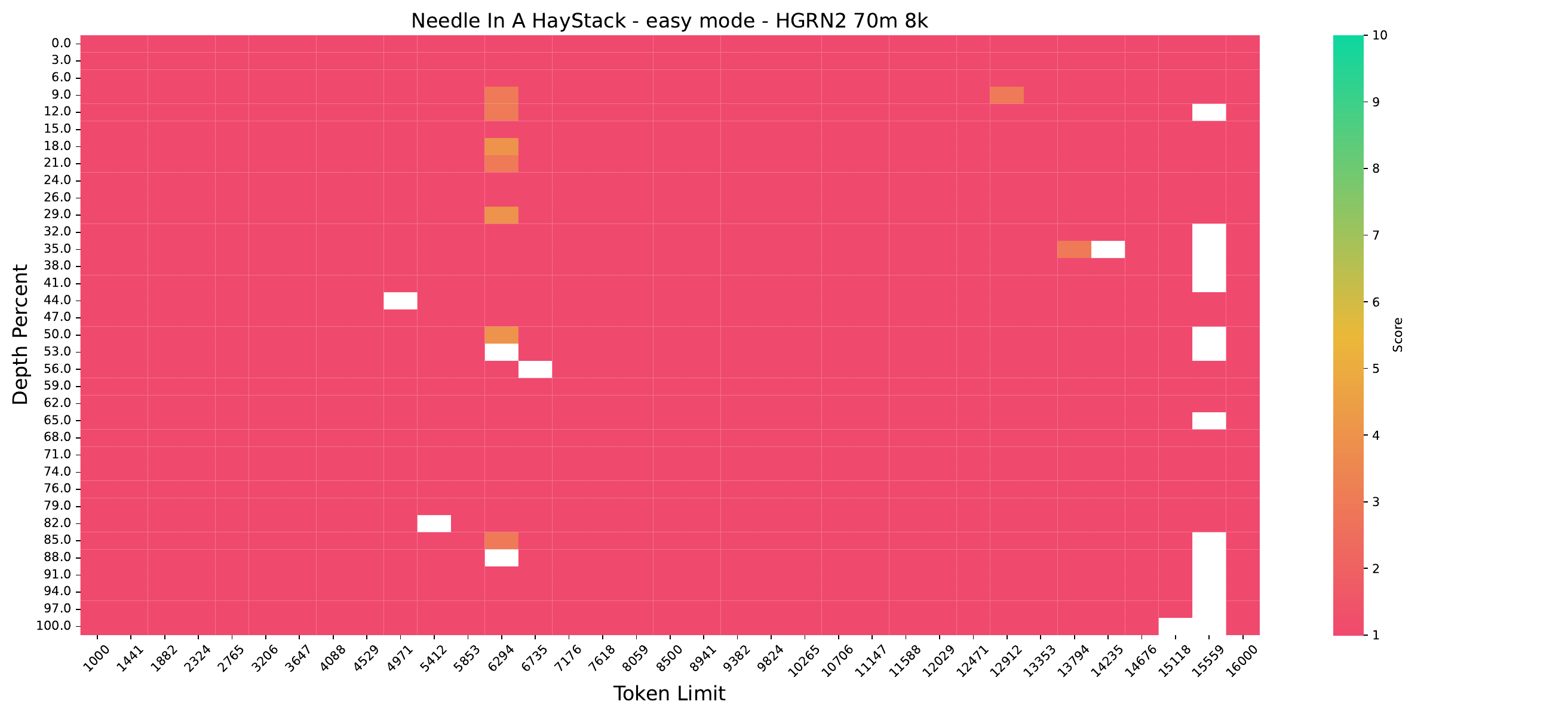}
\includegraphics[width=1\linewidth]{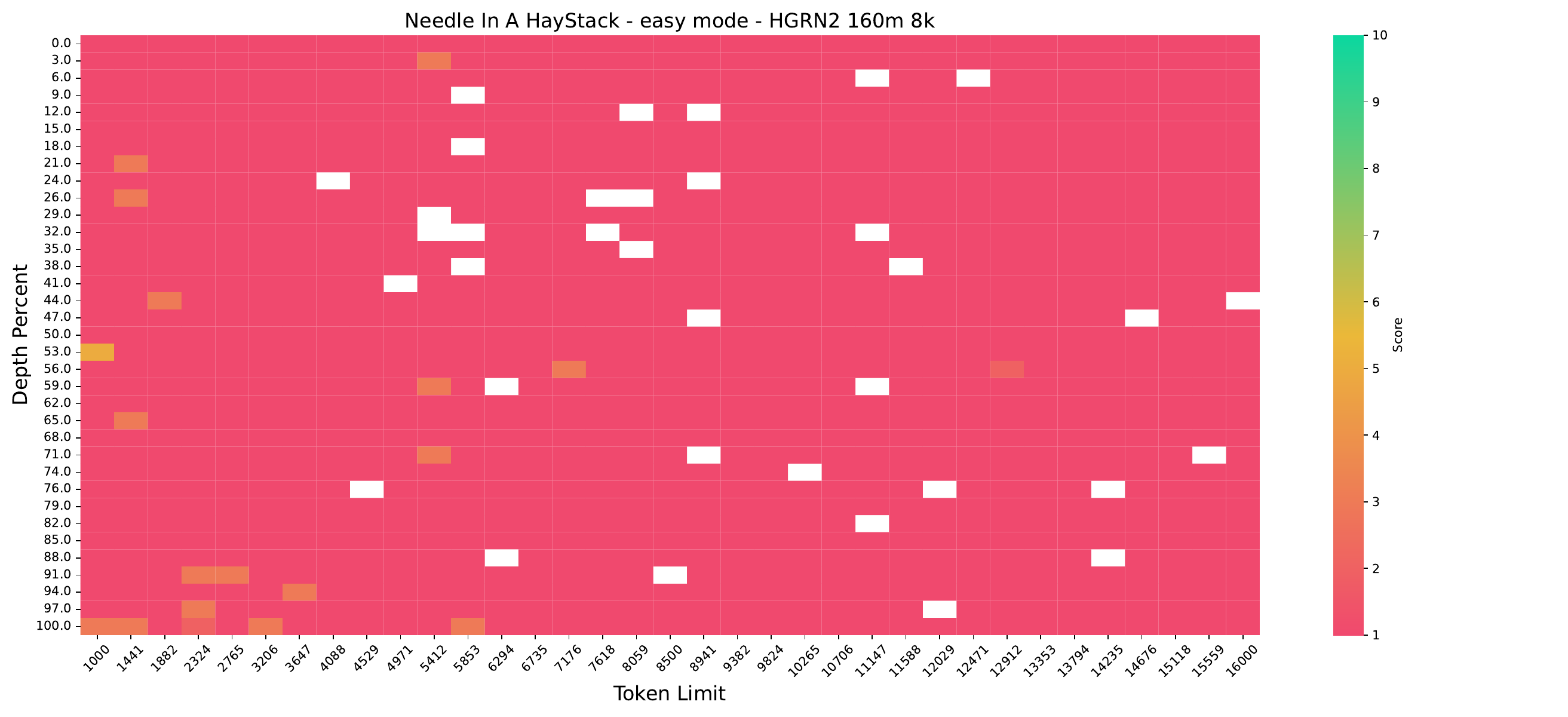}
\end{figure*}

\begin{figure*}
\centering
\includegraphics[width=1\linewidth]{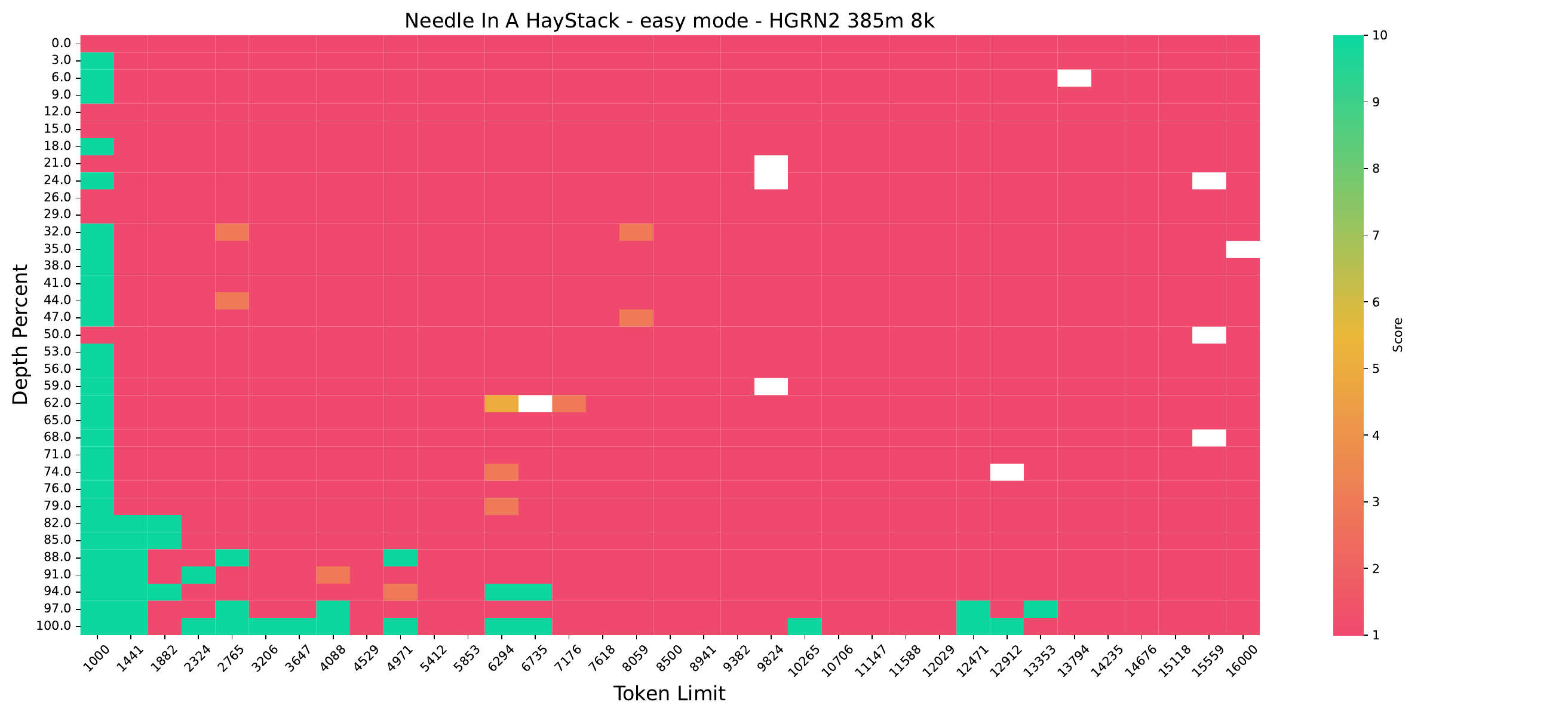}
\includegraphics[width=1\linewidth]{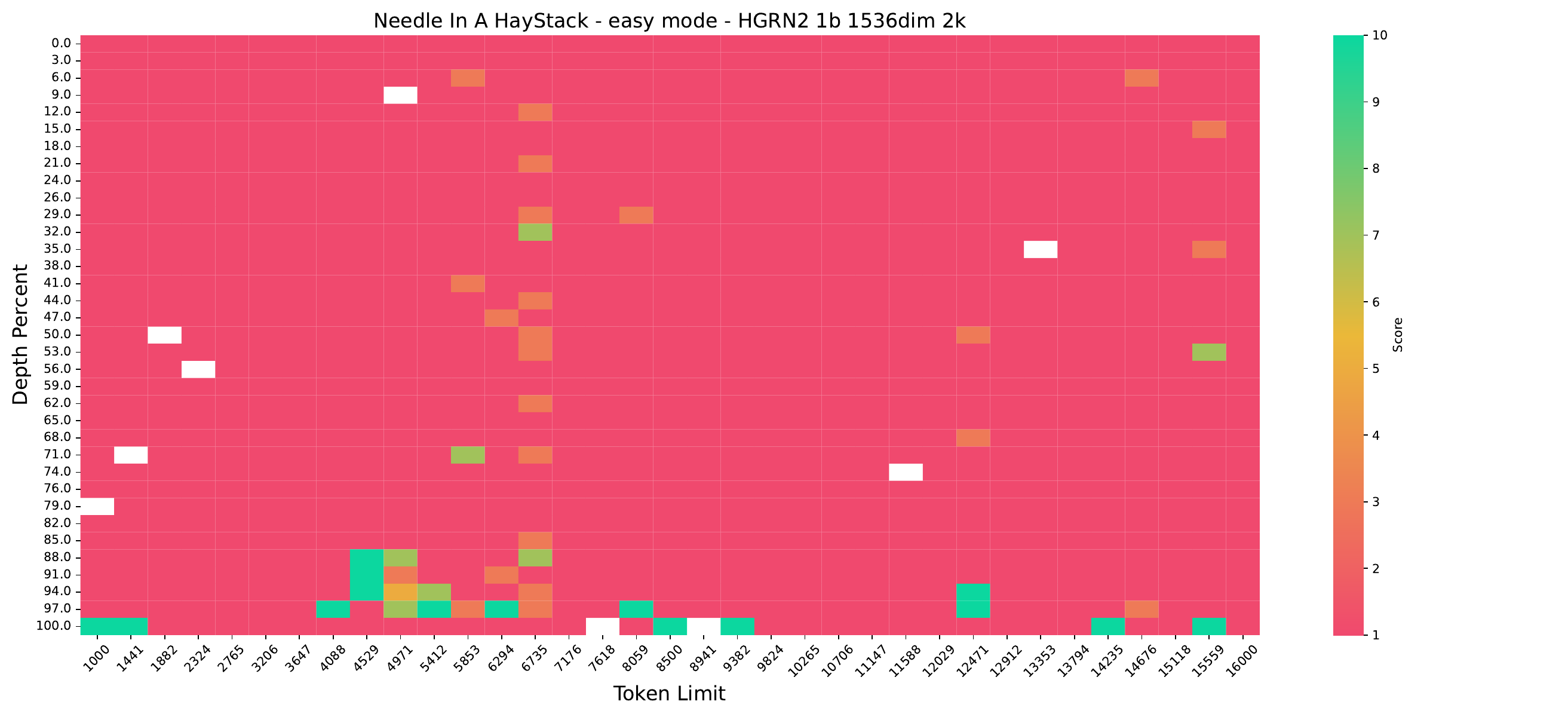}
\includegraphics[width=1\linewidth]{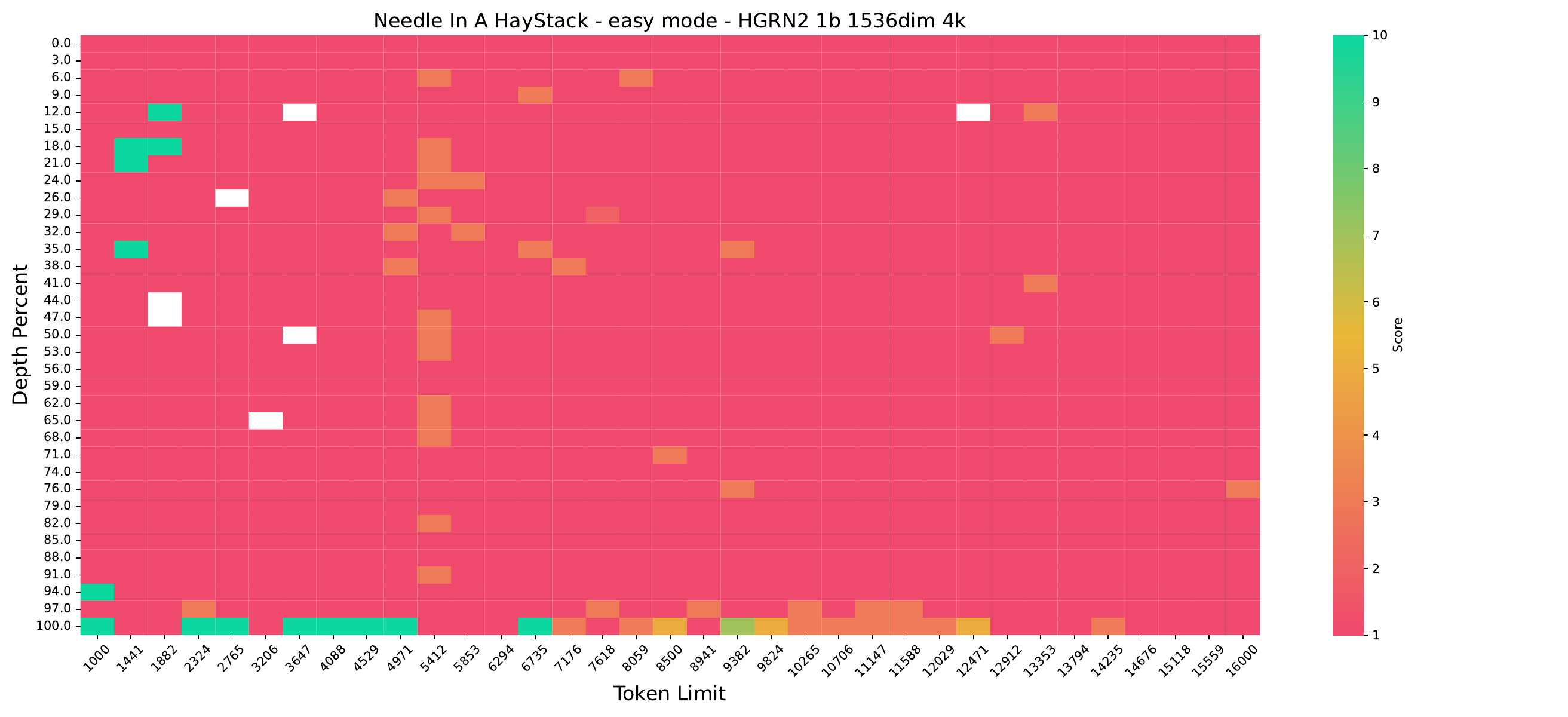}
\end{figure*}

\begin{figure*}
\centering
\includegraphics[width=1\linewidth]{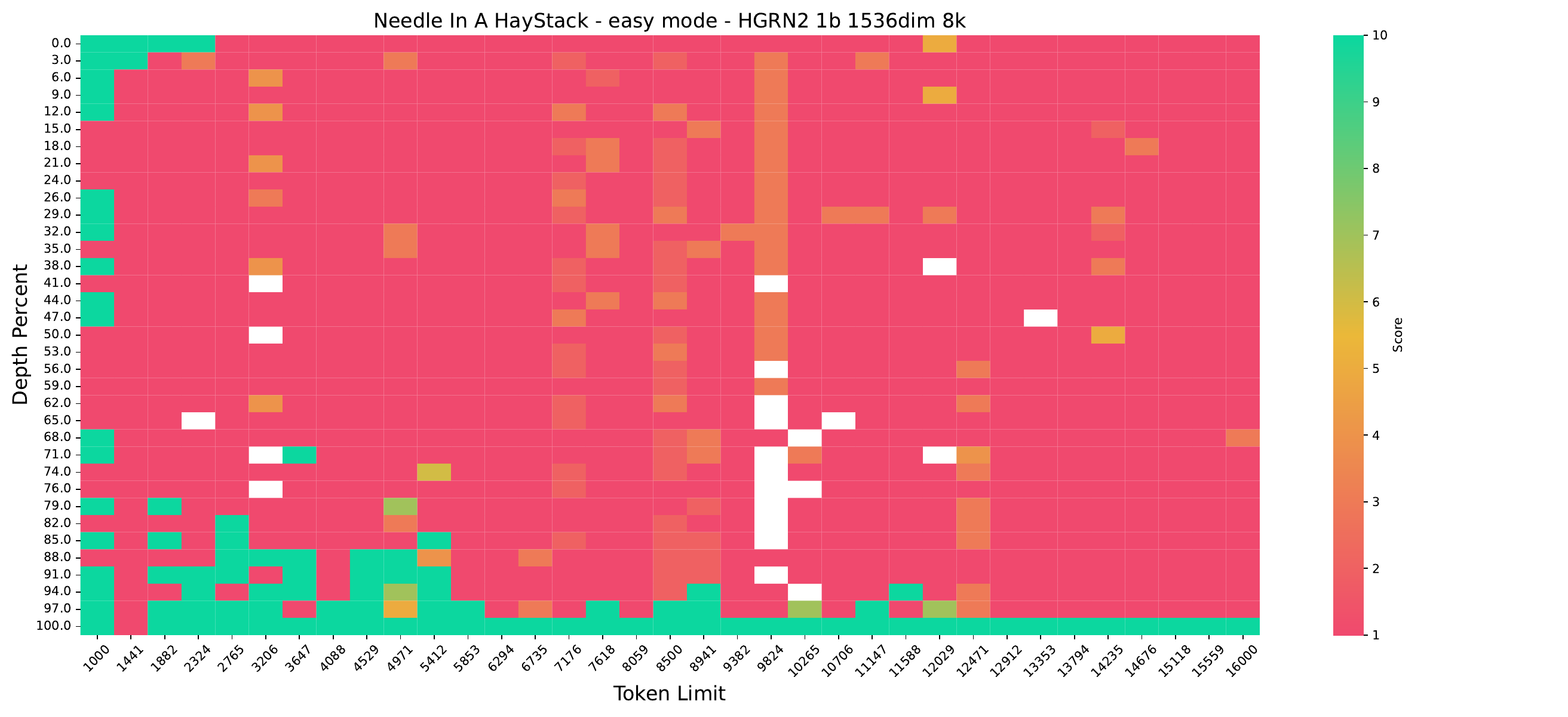}
\includegraphics[width=1\linewidth]{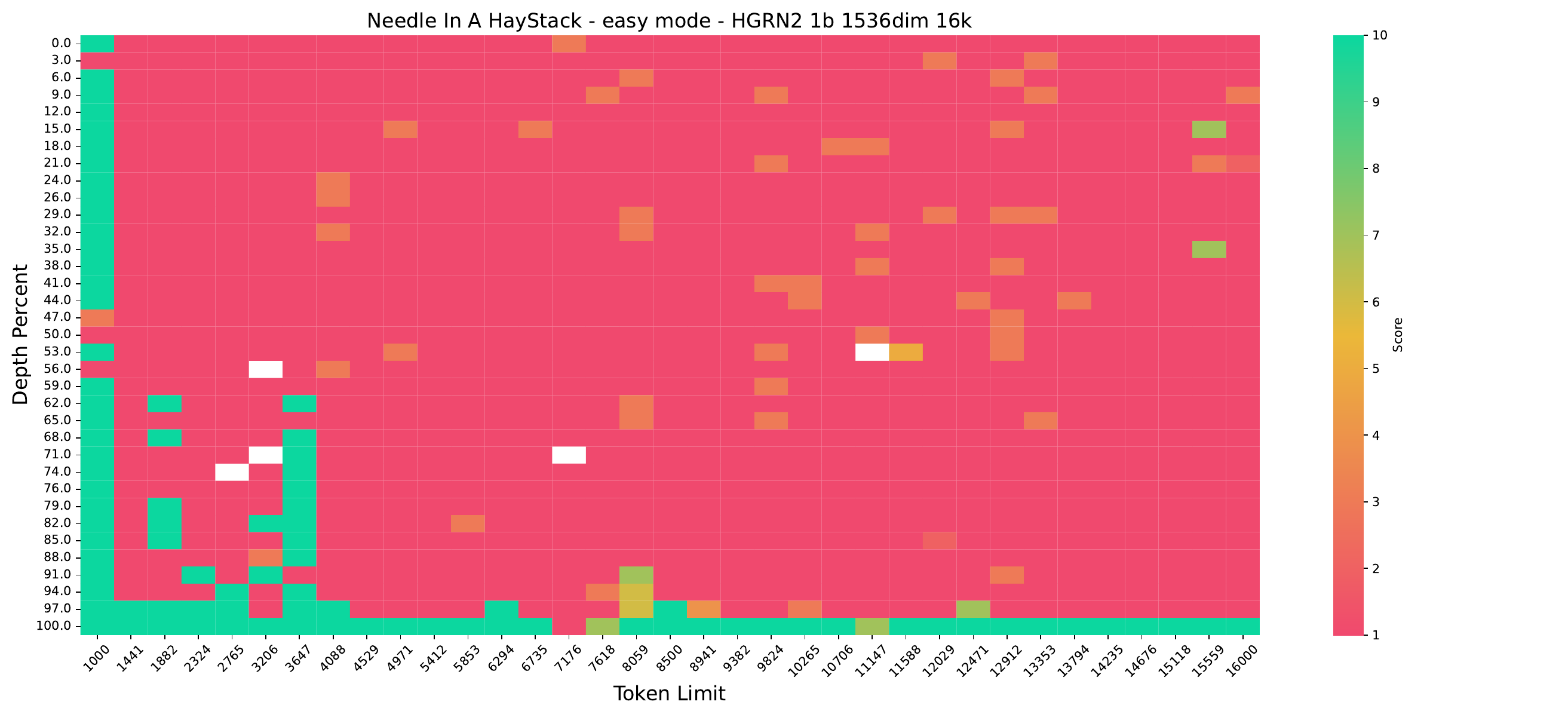}
\includegraphics[width=1\linewidth]{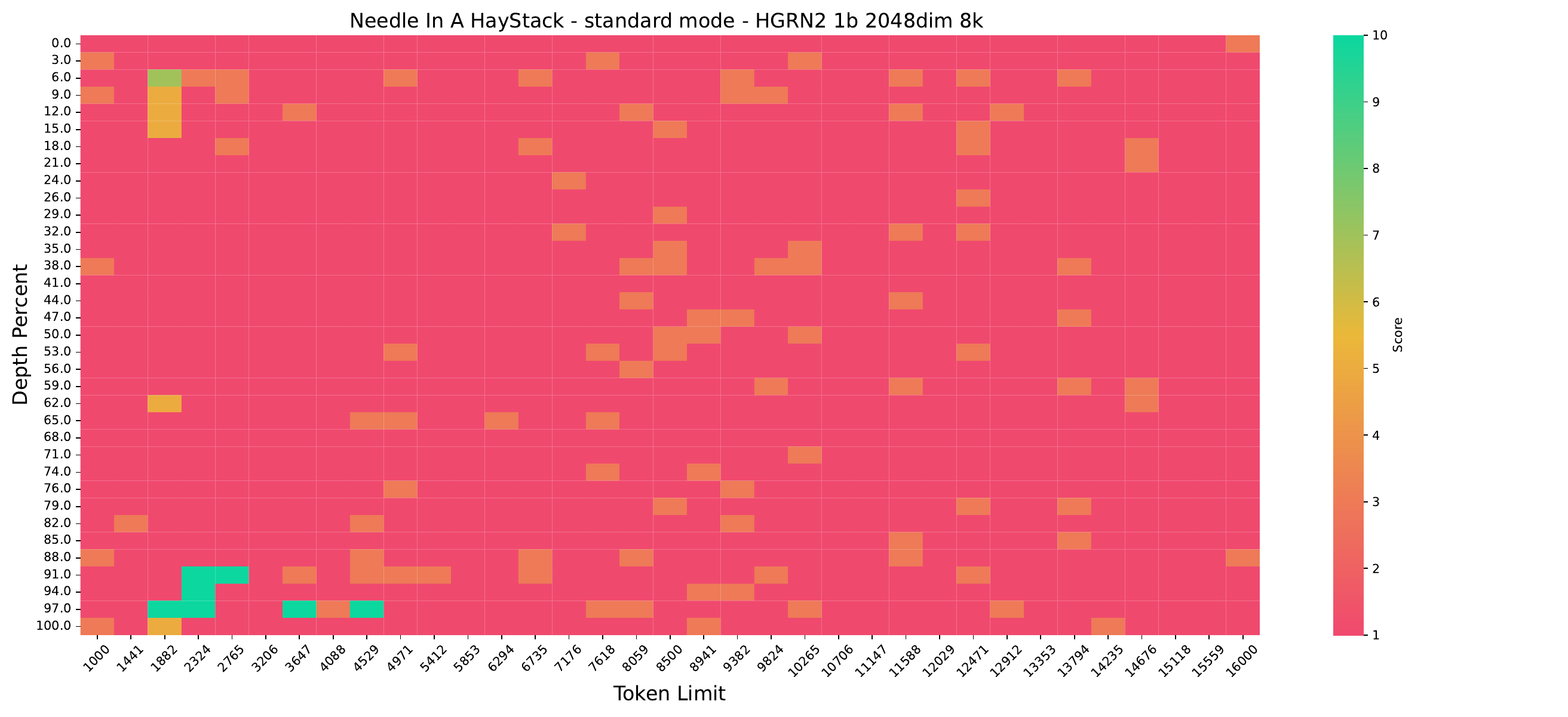}
\end{figure*}

\begin{figure*}
\centering
\includegraphics[width=1\linewidth]{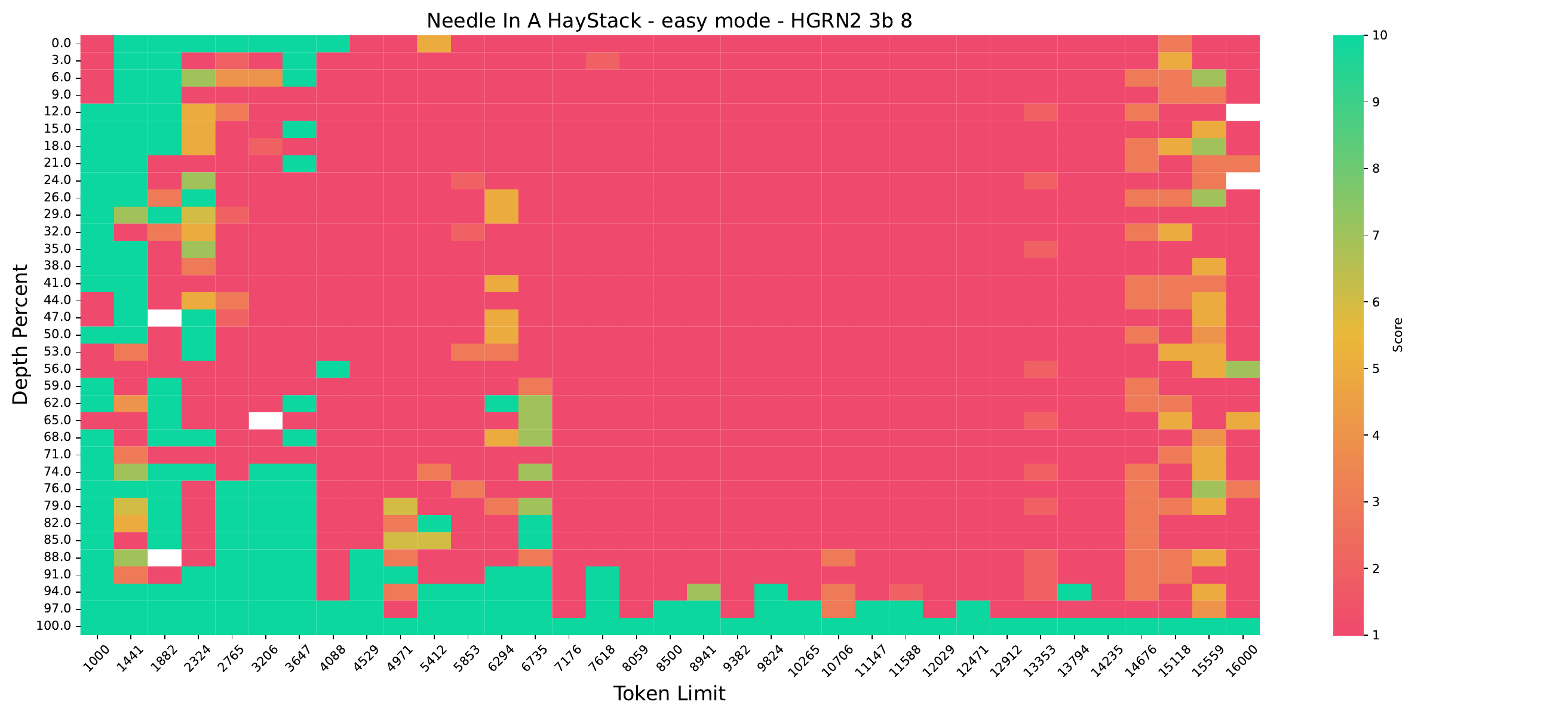}
\includegraphics[width=1\linewidth]{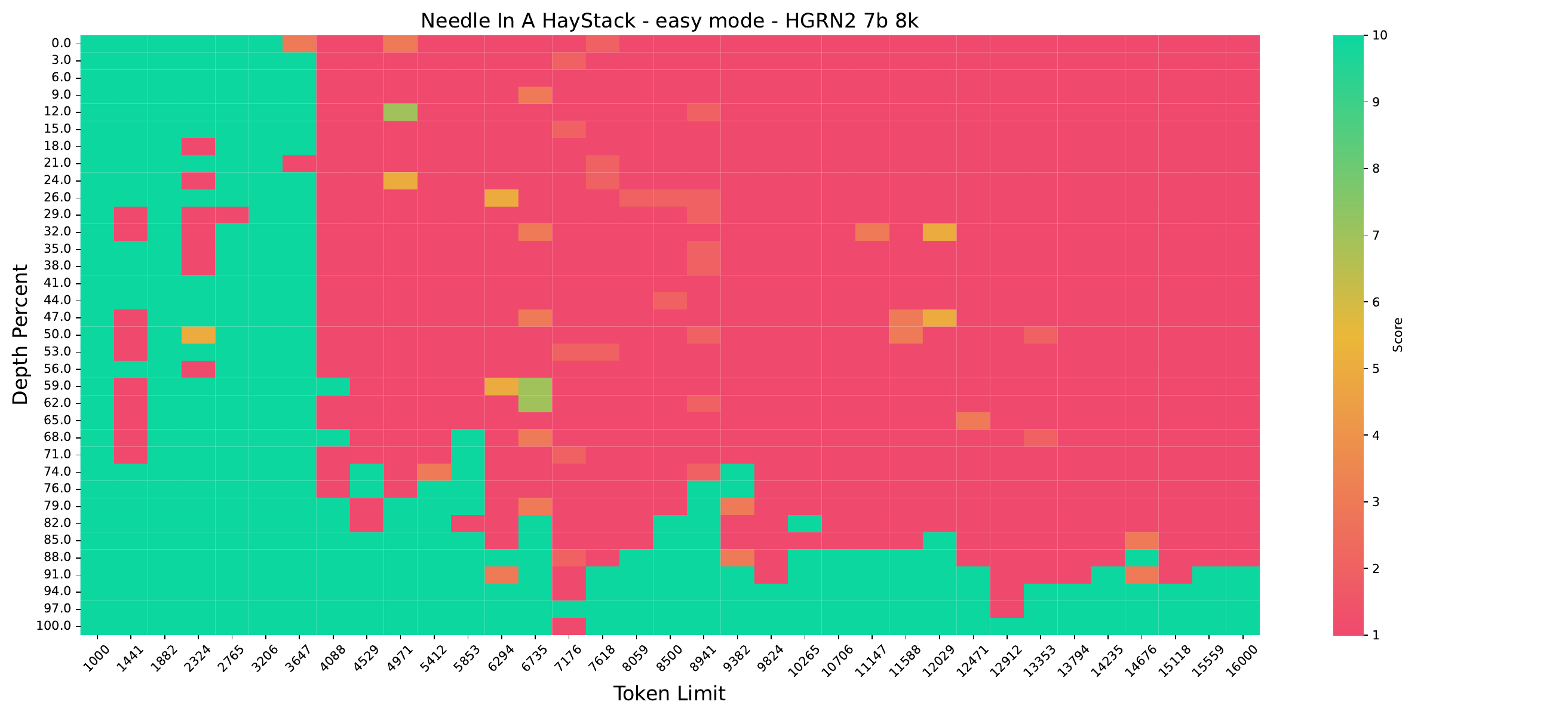}
\includegraphics[width=1\linewidth]{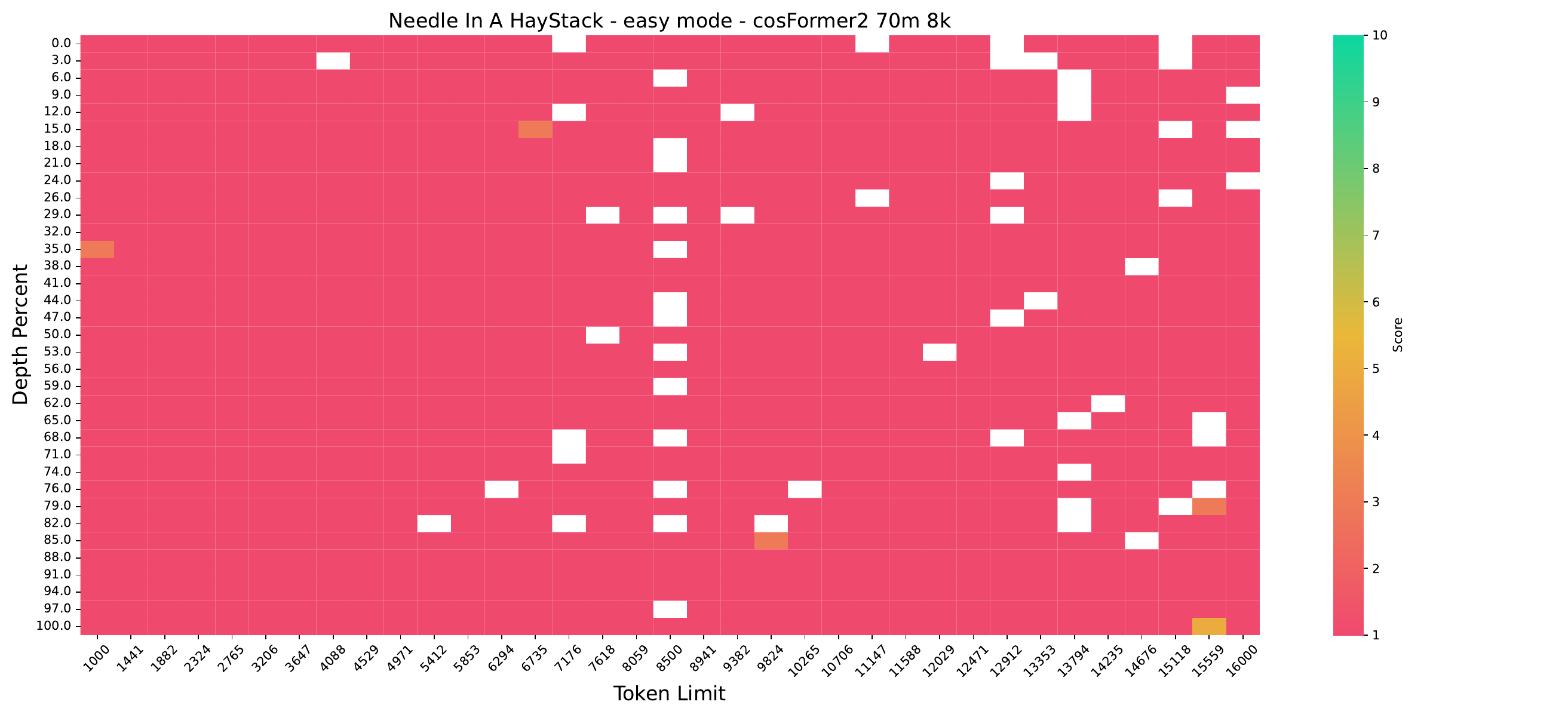}
\end{figure*}

\begin{figure*}
\centering
\includegraphics[width=1\linewidth]{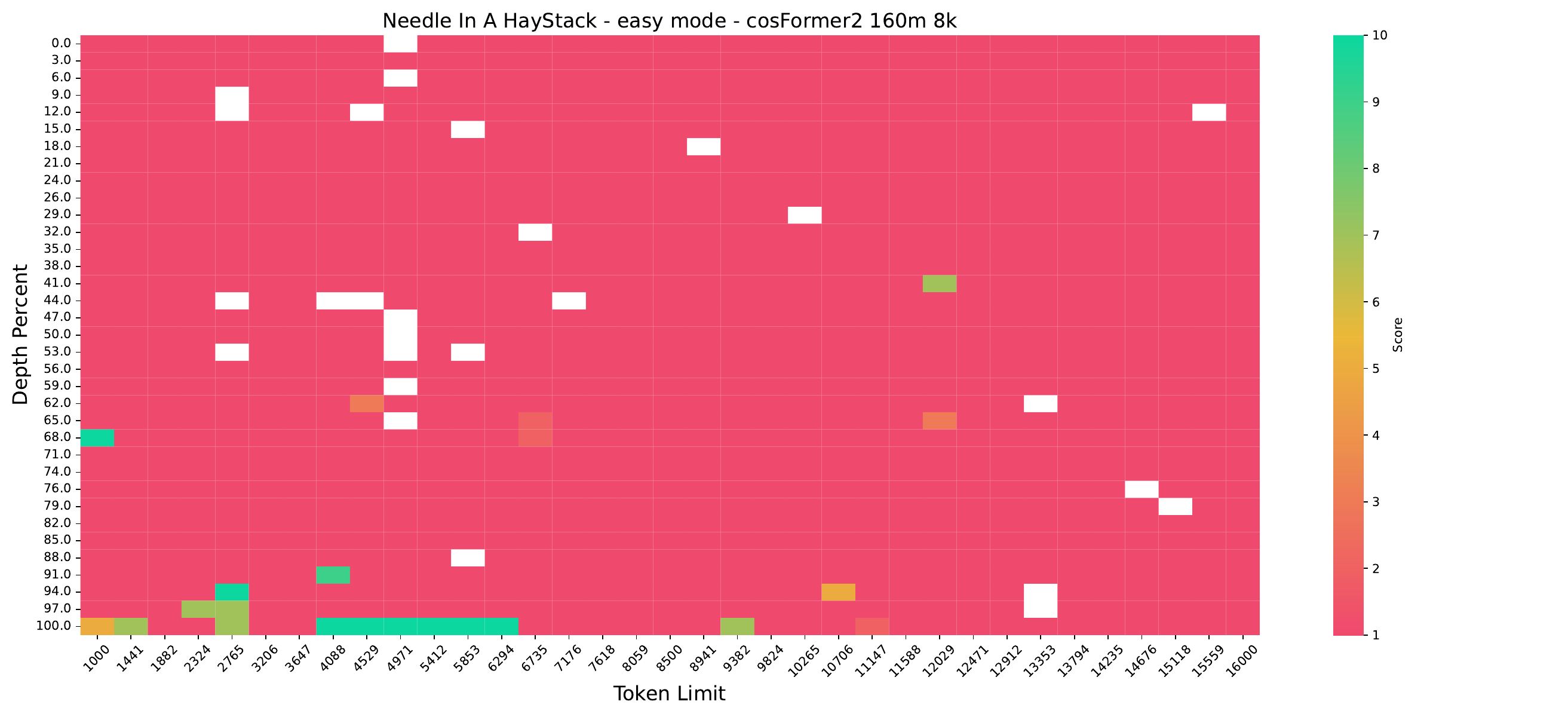}
\includegraphics[width=1\linewidth]{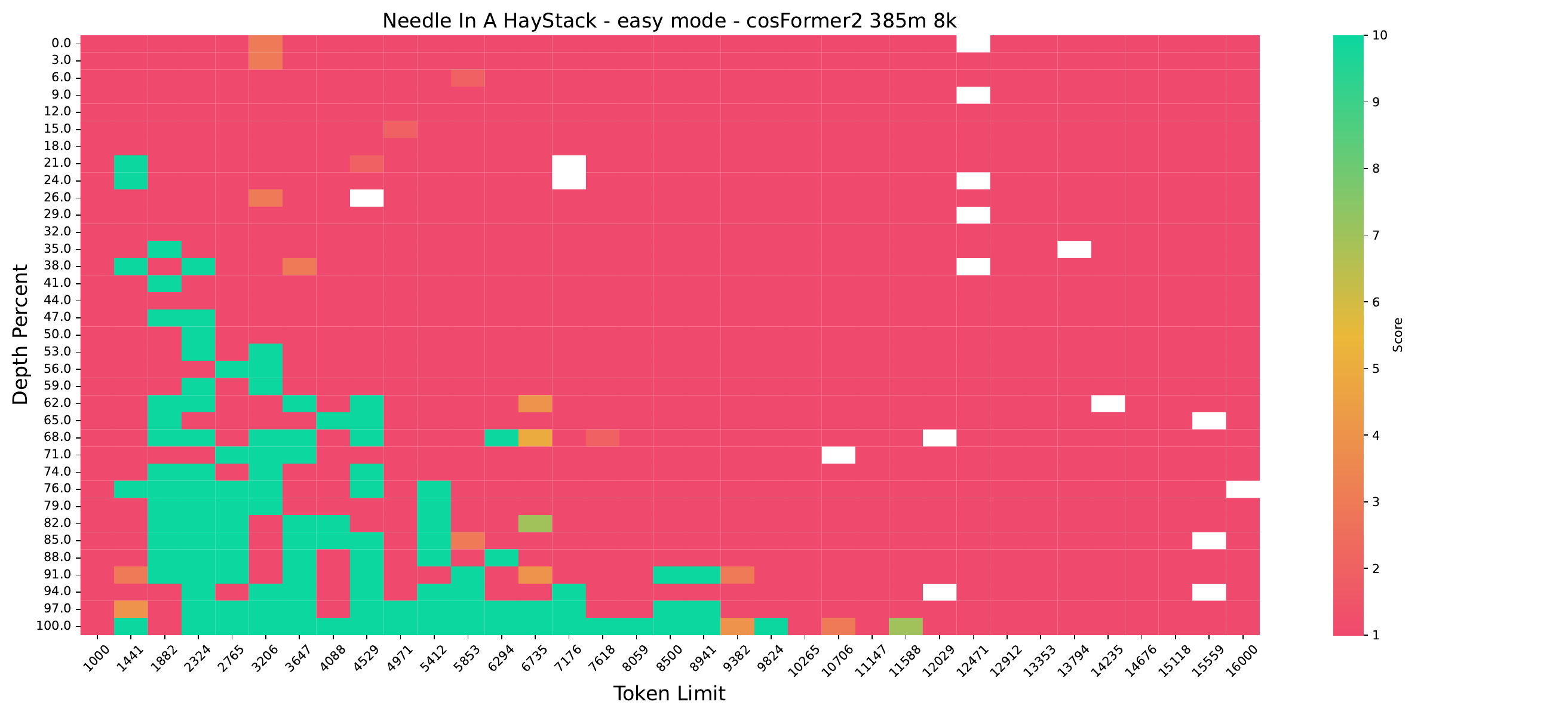}
\includegraphics[width=1\linewidth]{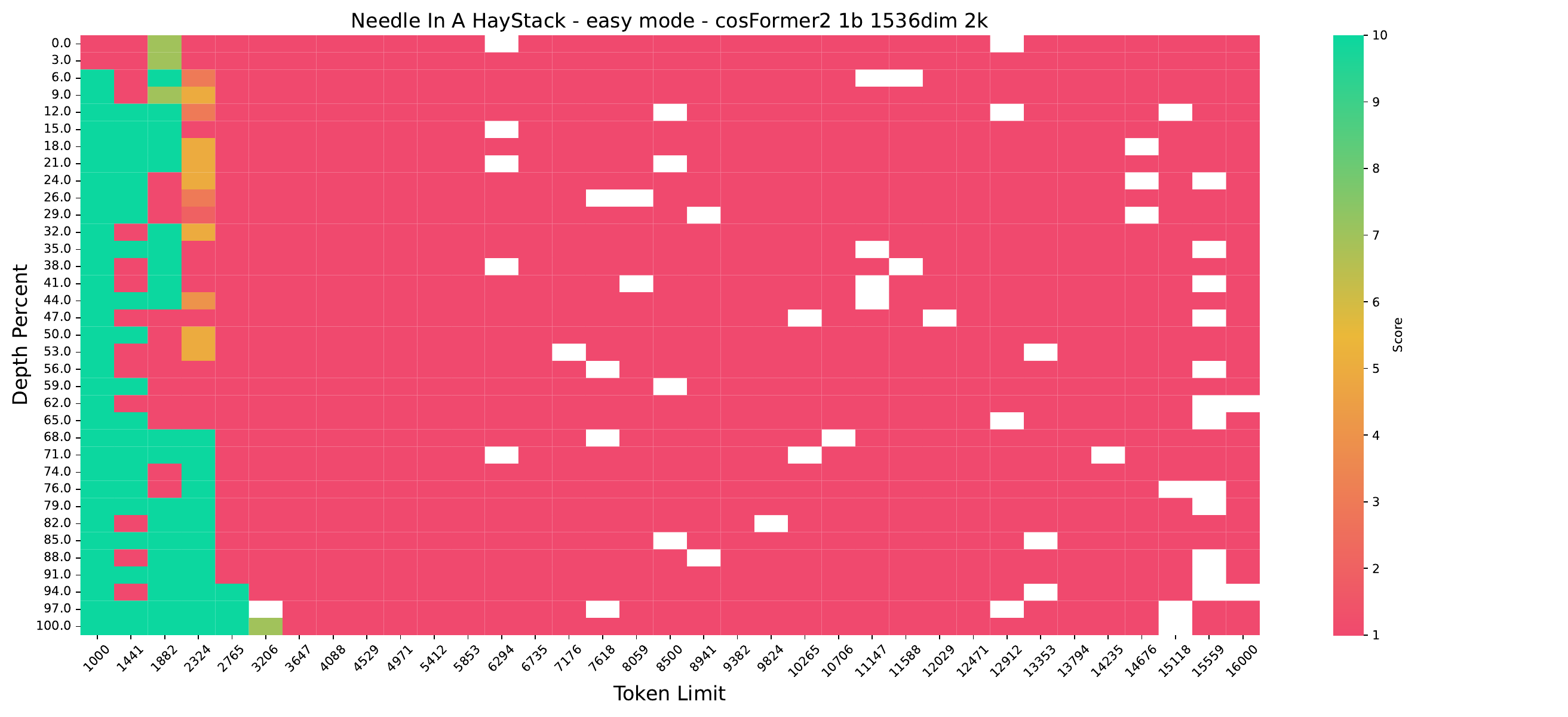}
\end{figure*}

\begin{figure*}
\centering
\includegraphics[width=1\linewidth]{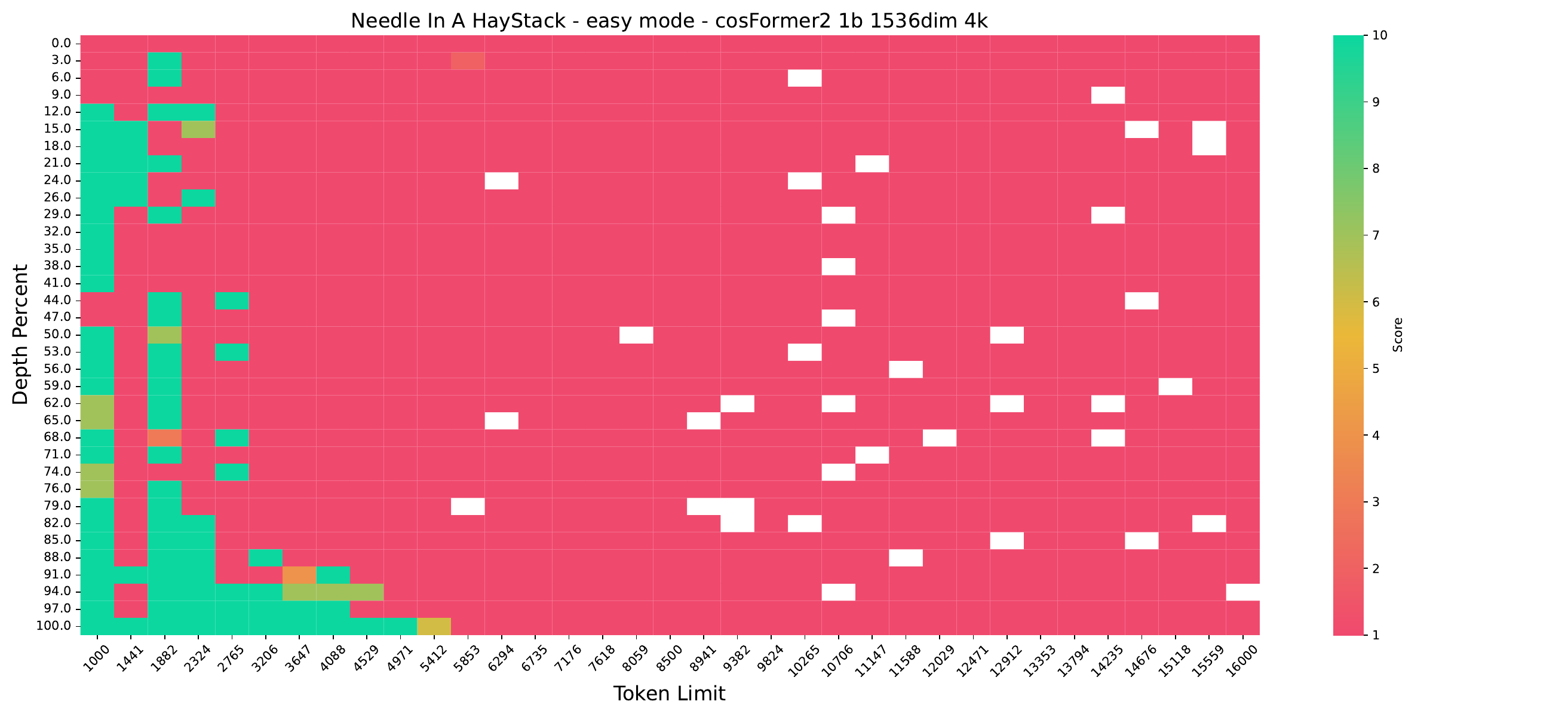}
\includegraphics[width=1\linewidth]{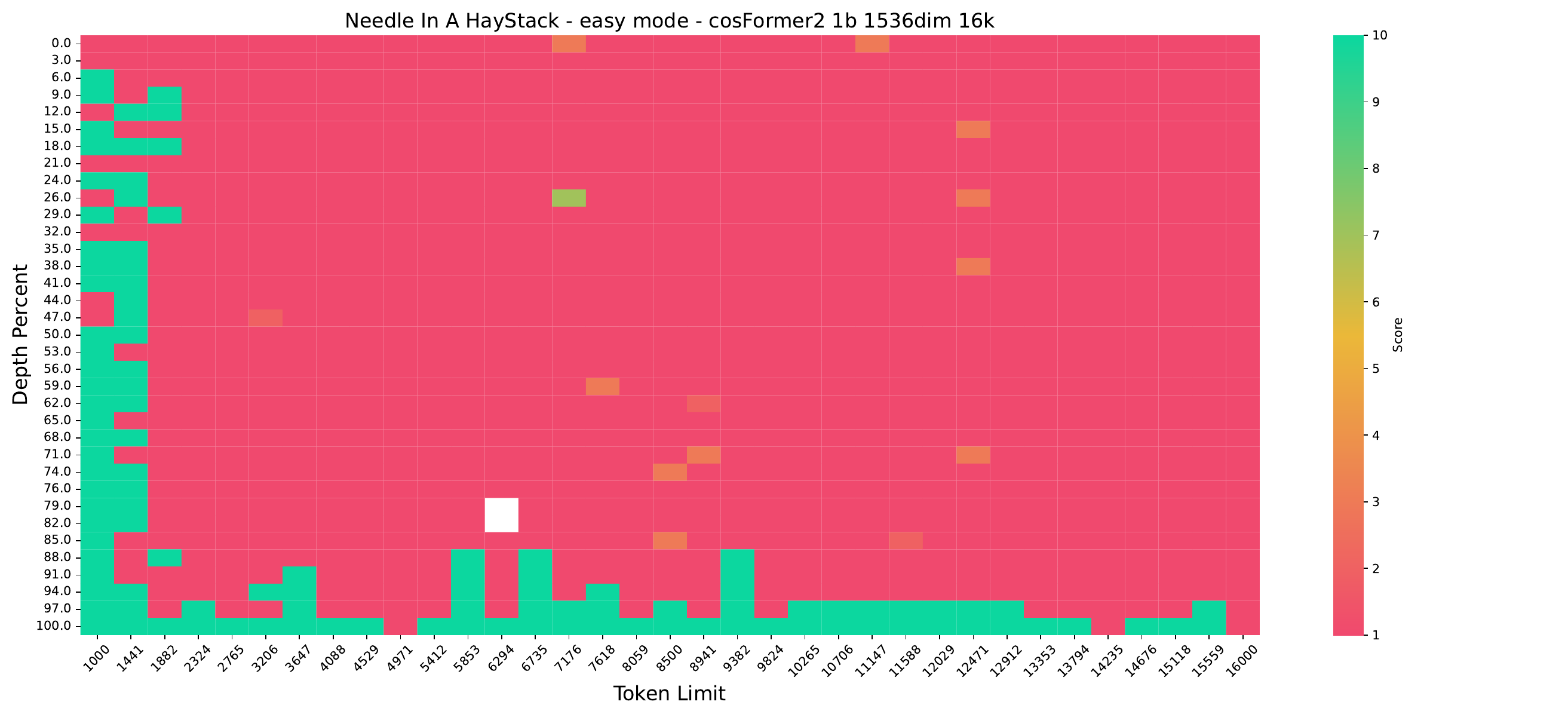}
\includegraphics[width=1\linewidth]{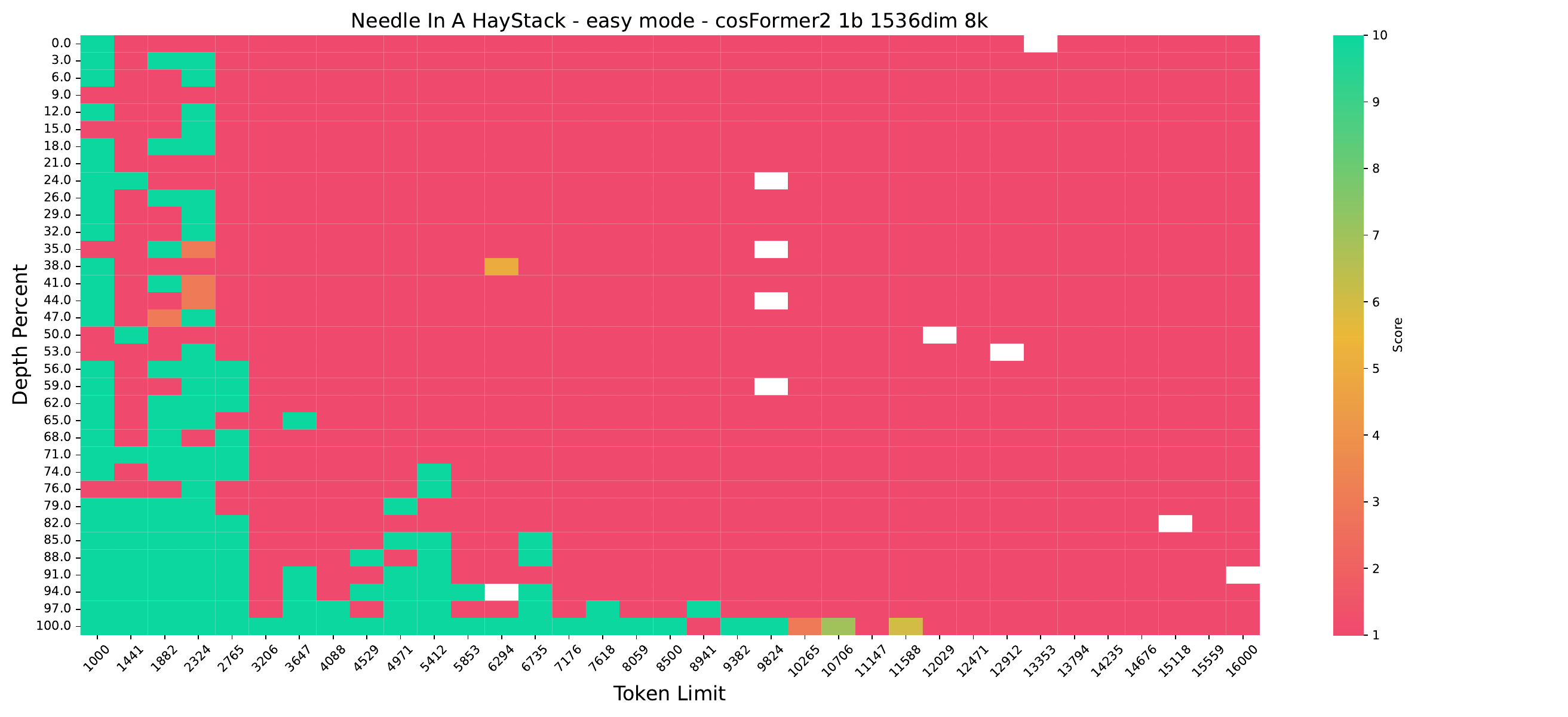}
\end{figure*}

\begin{figure*}
\centering
\includegraphics[width=1\linewidth]{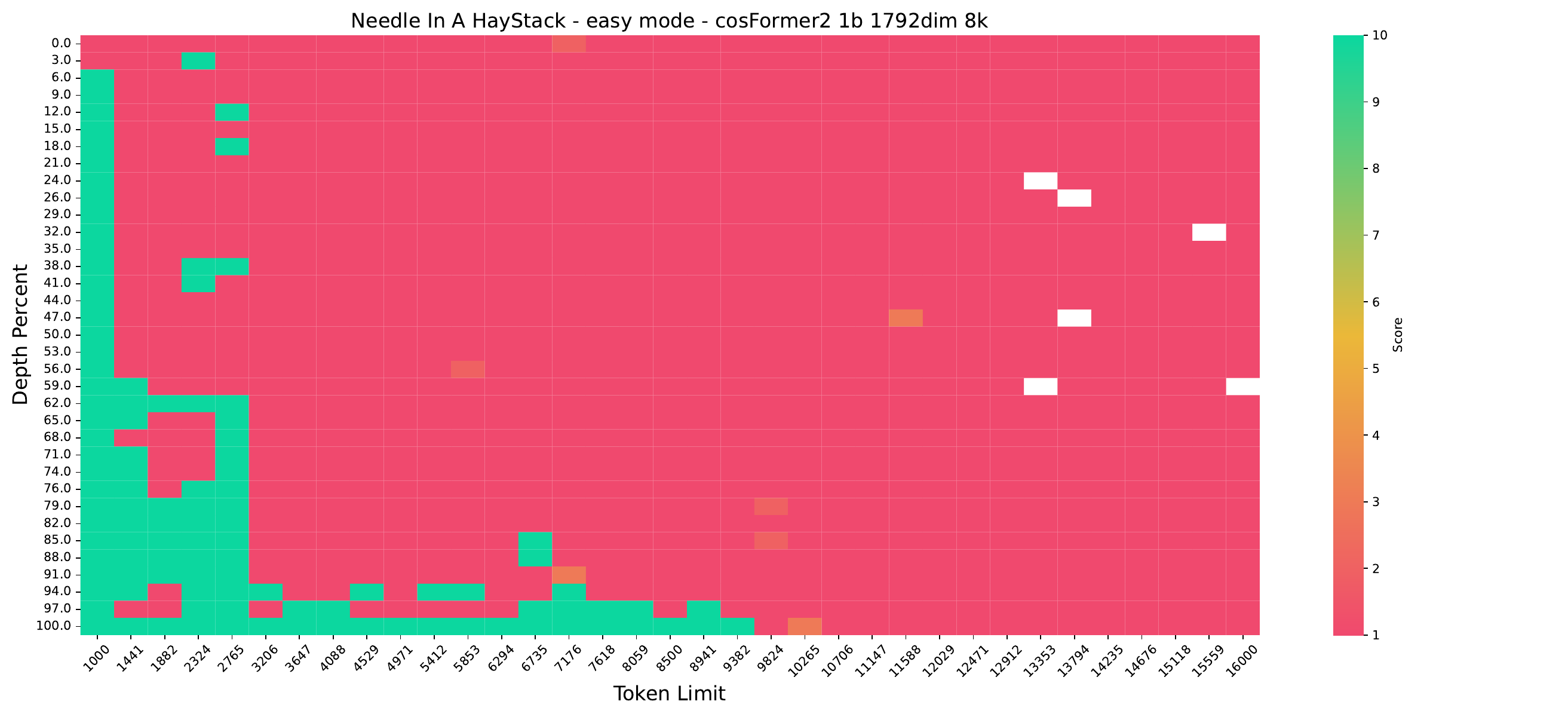}
\includegraphics[width=1\linewidth]{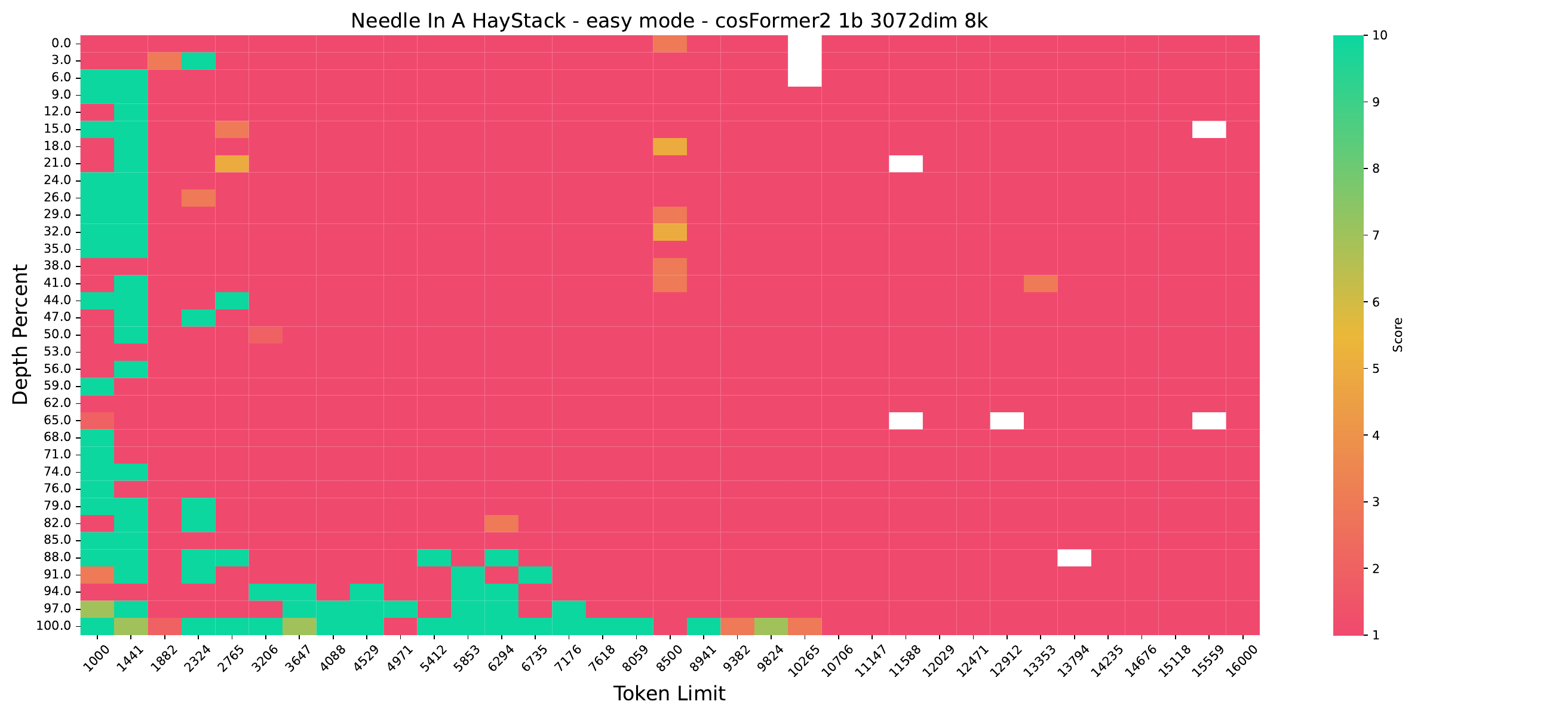}
\includegraphics[width=1\linewidth]{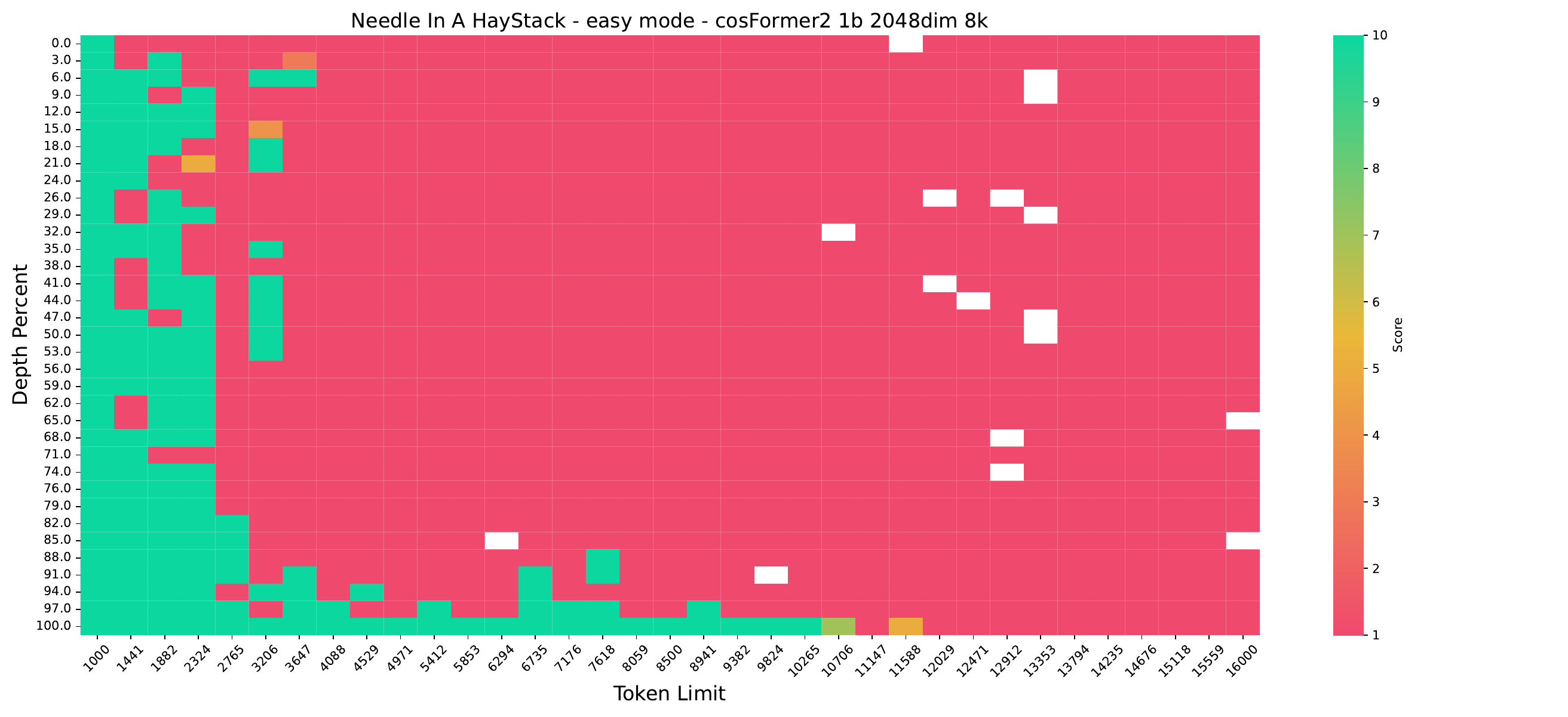}
\end{figure*}

\begin{figure*}
\centering
\includegraphics[width=1\linewidth]{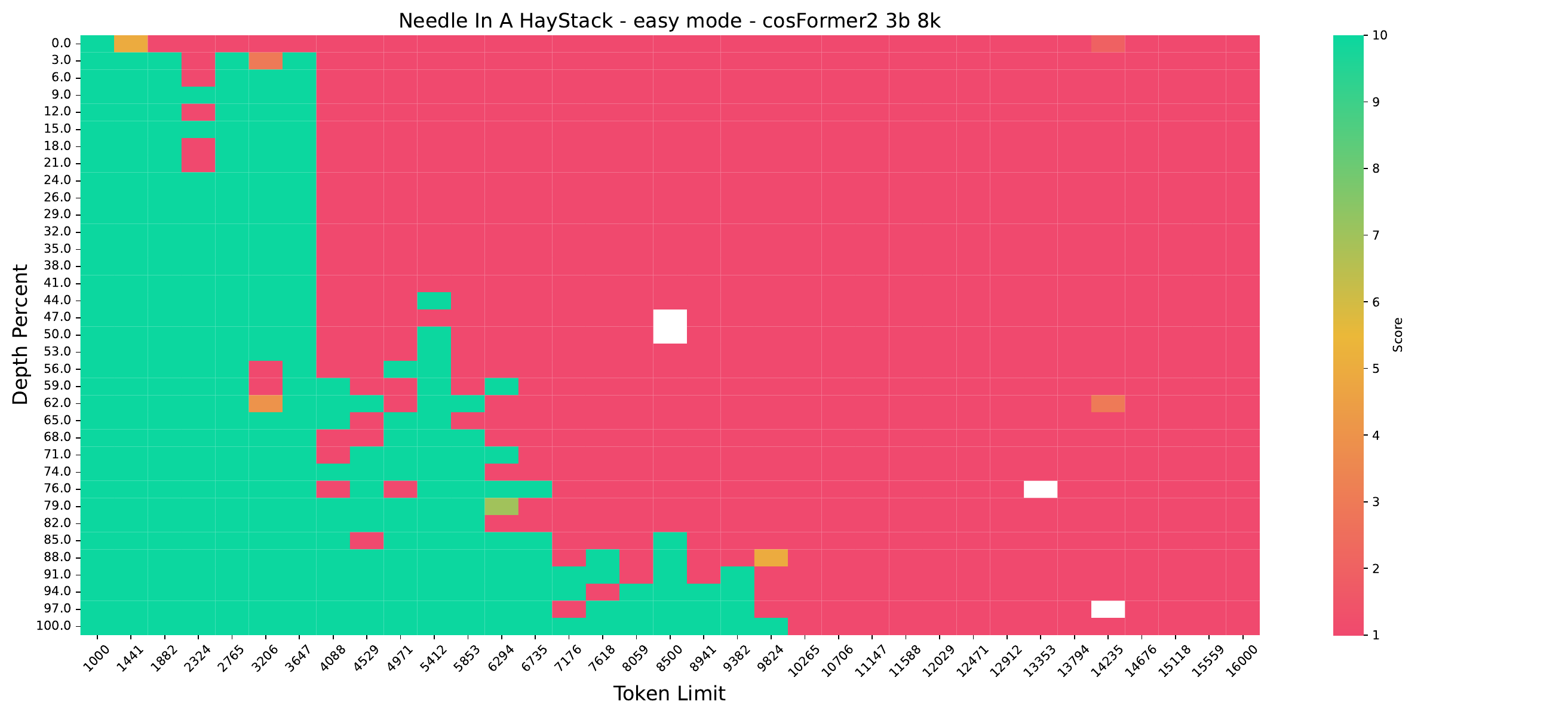}
\includegraphics[width=1\linewidth]{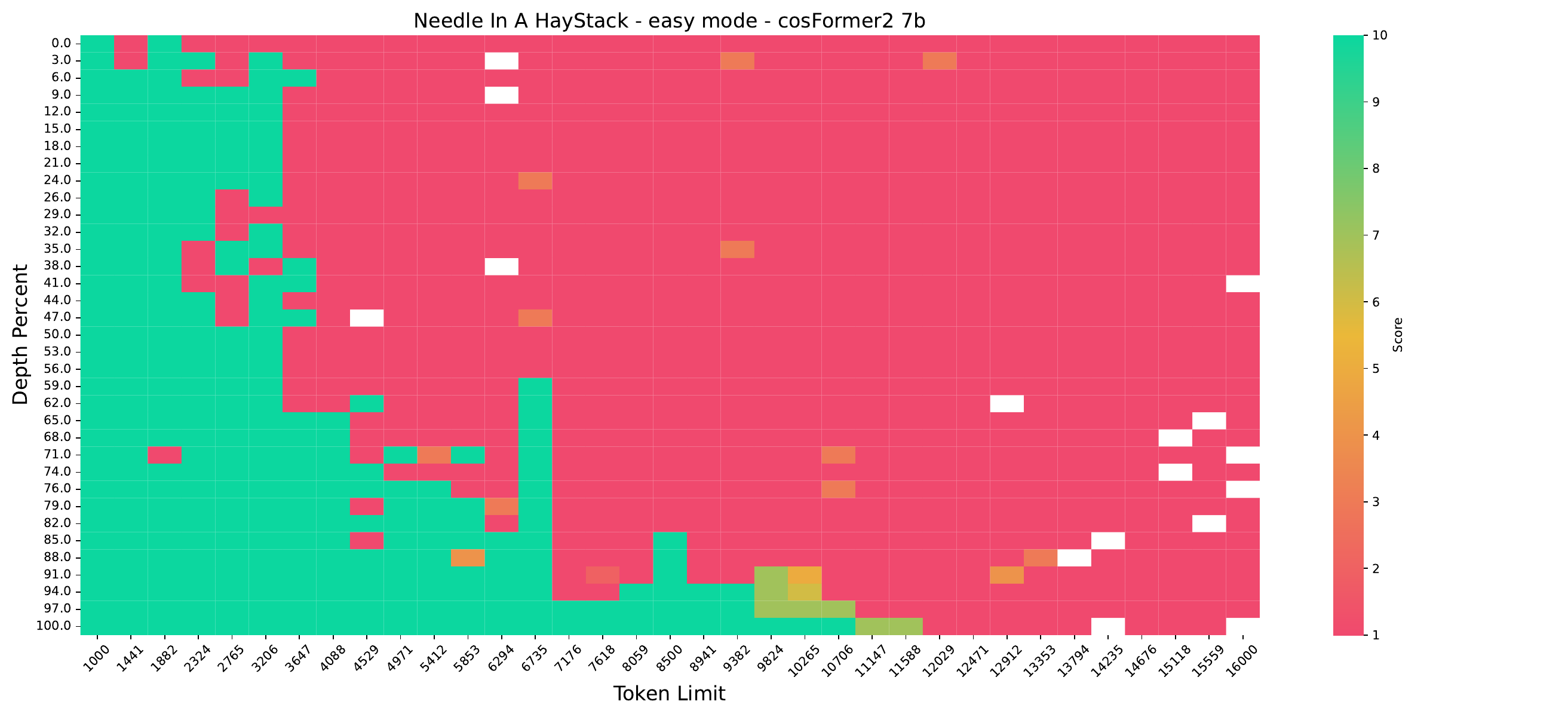}
\end{figure*}

\subsection{NIAH heatmap by standard mode}
The figures below provide a heatmap visualization of NIAH in standard mode.

\begin{figure*}
    \centering
    \includegraphics[width=1\linewidth]{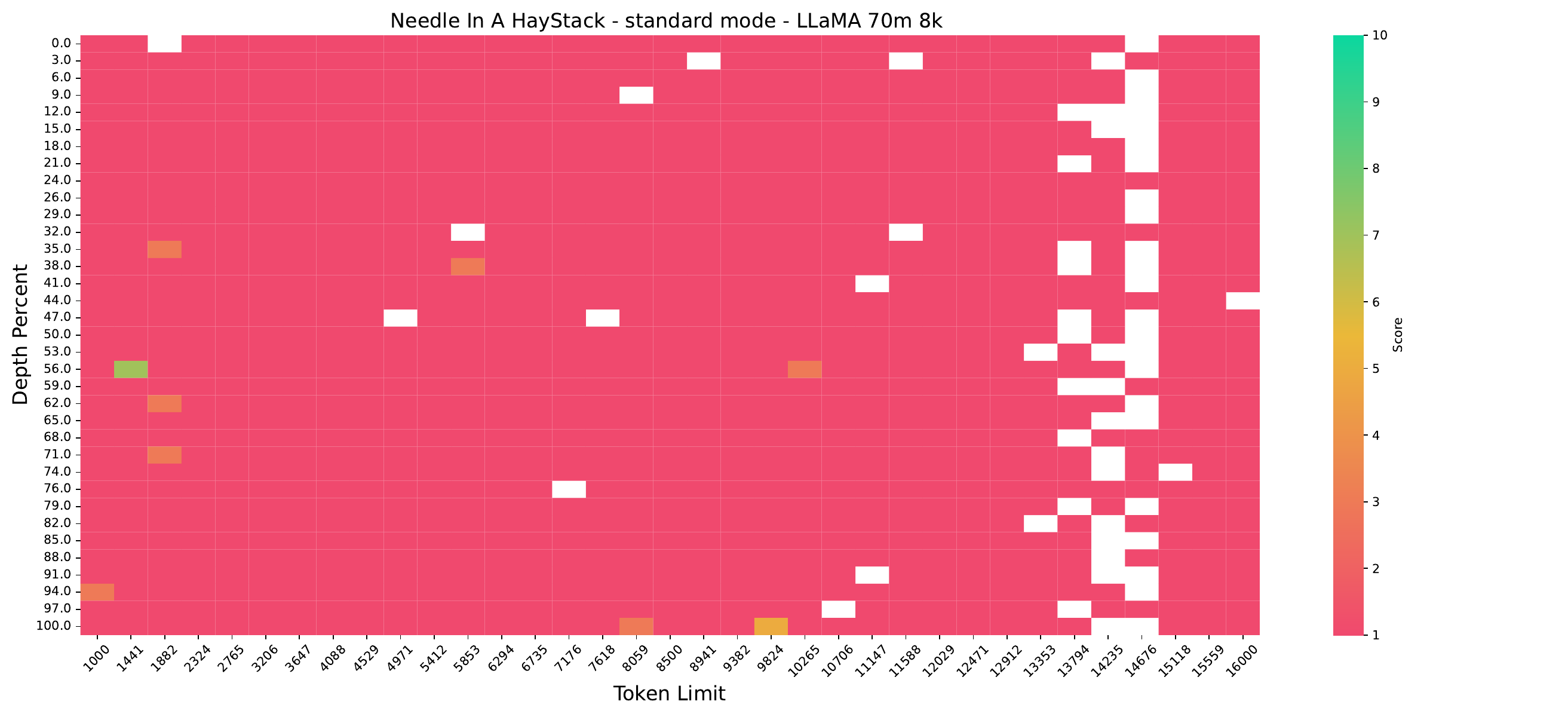}
    \includegraphics[width=1\linewidth]{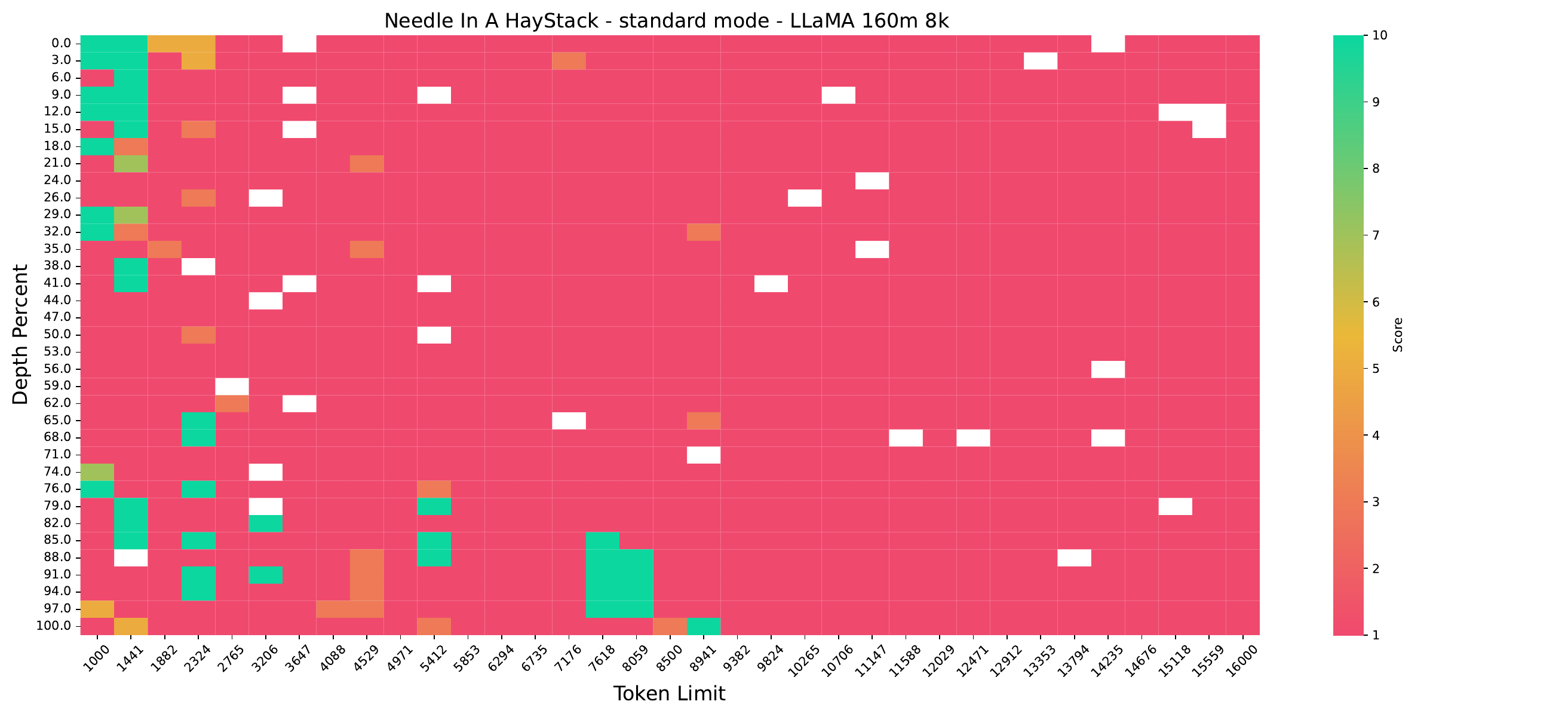}
    \includegraphics[width=1\linewidth]{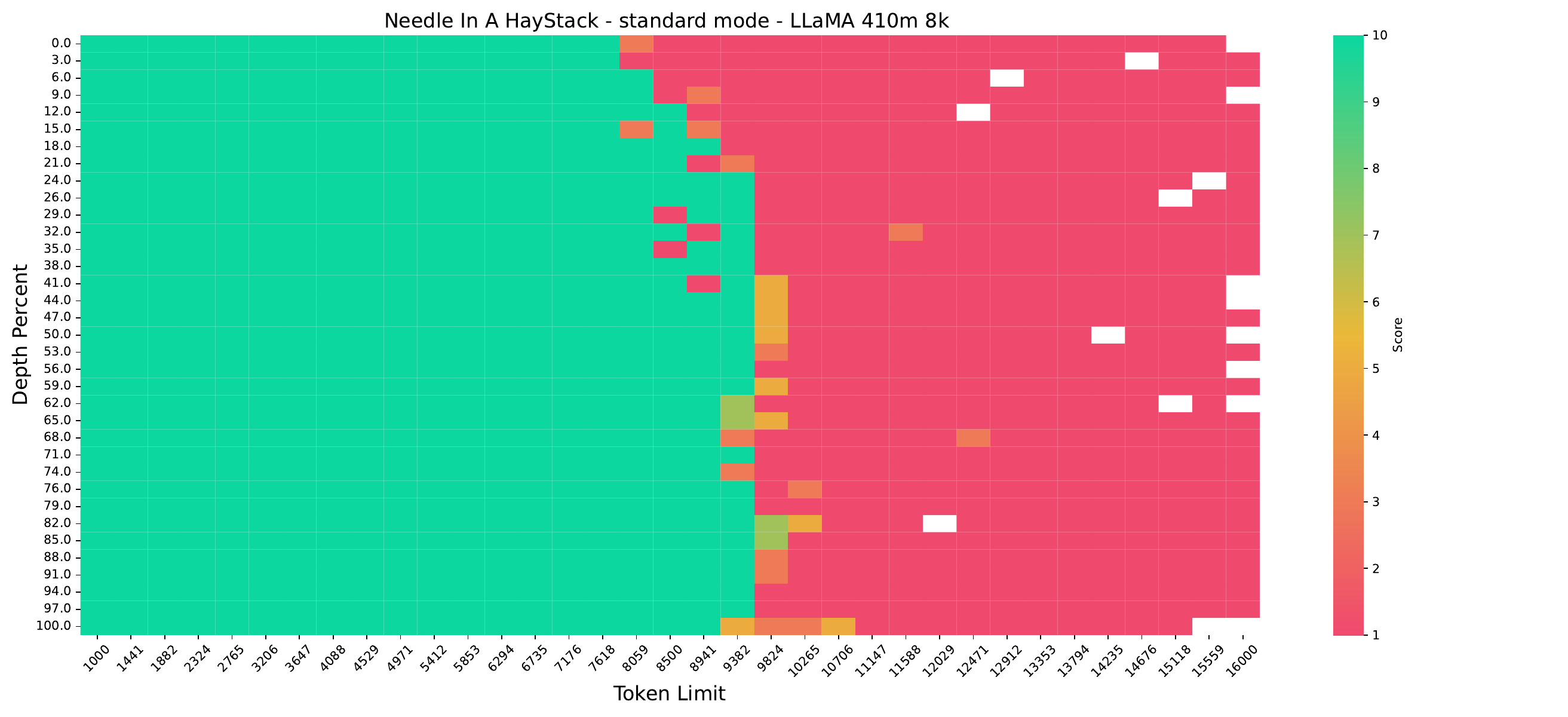}
\end{figure*}

\begin{figure*}
    \centering
    \includegraphics[width=1\linewidth]{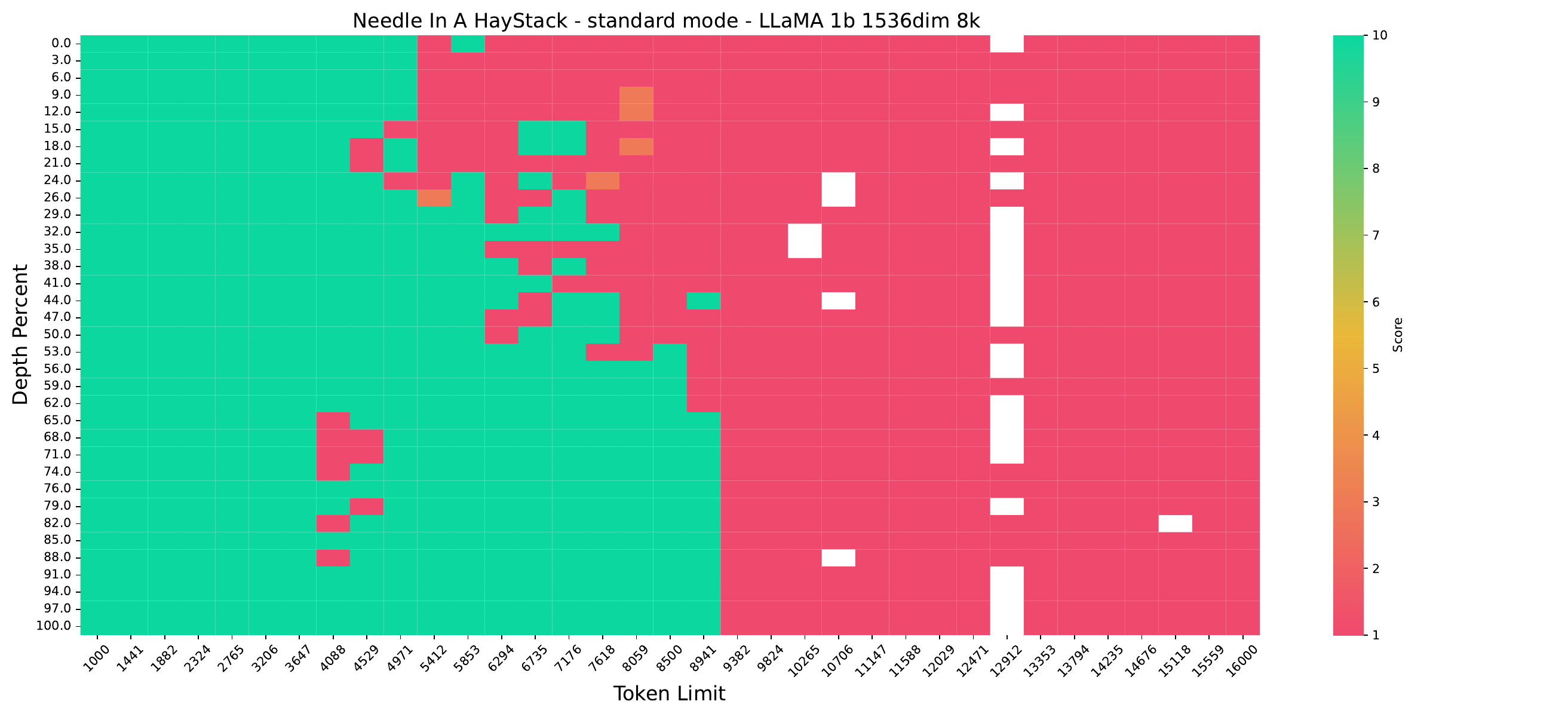}
    \includegraphics[width=1\linewidth]{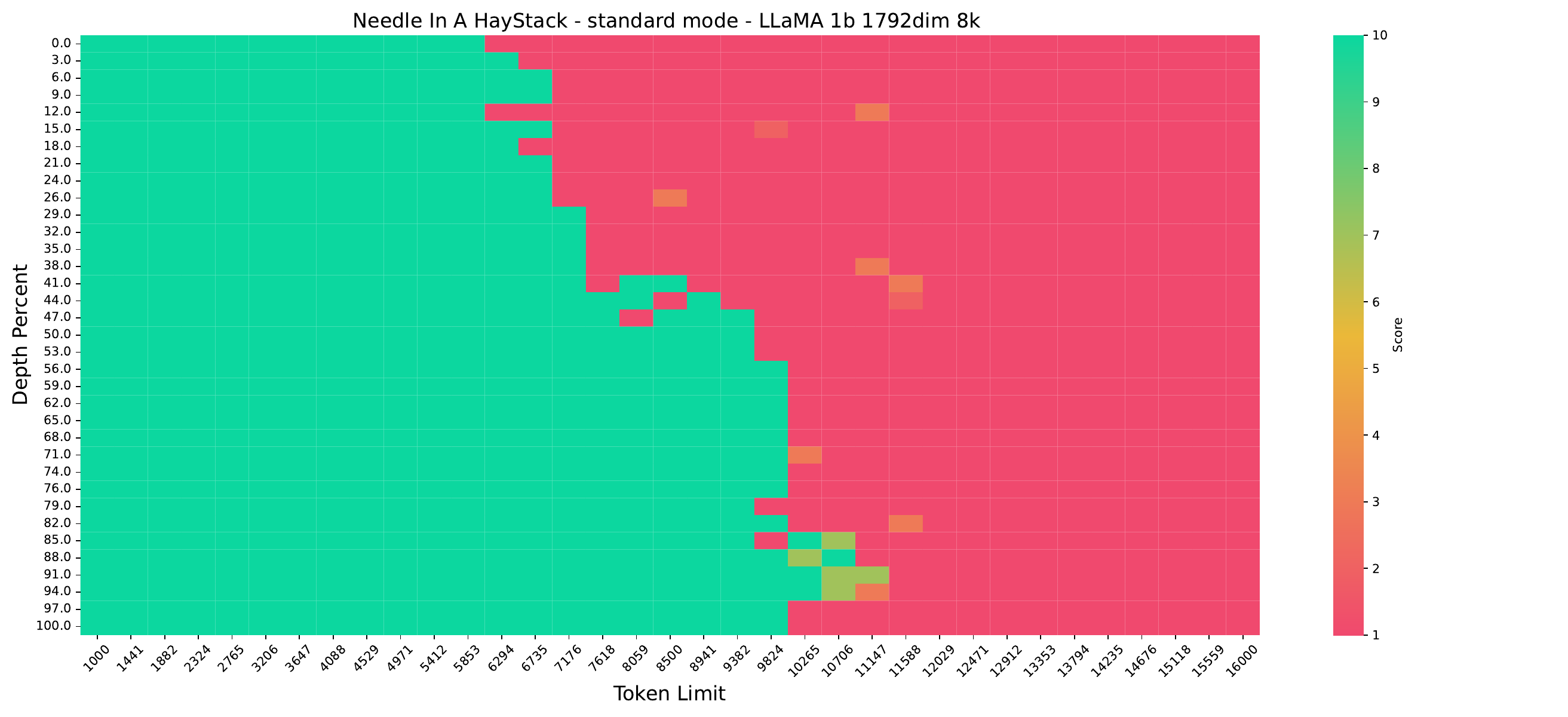}
    \includegraphics[width=1\linewidth]{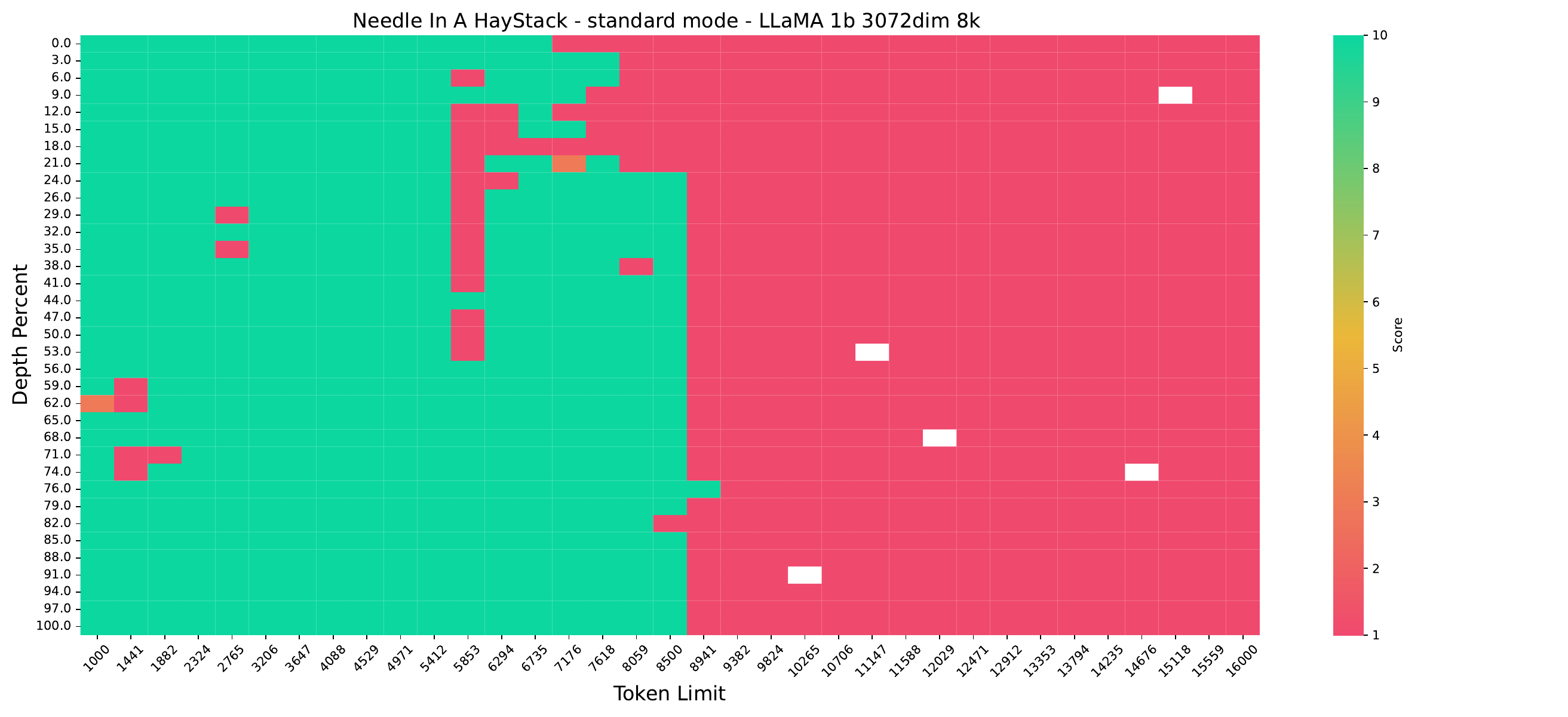}
\end{figure*}

\begin{figure*}
\centering
\includegraphics[width=1\linewidth]{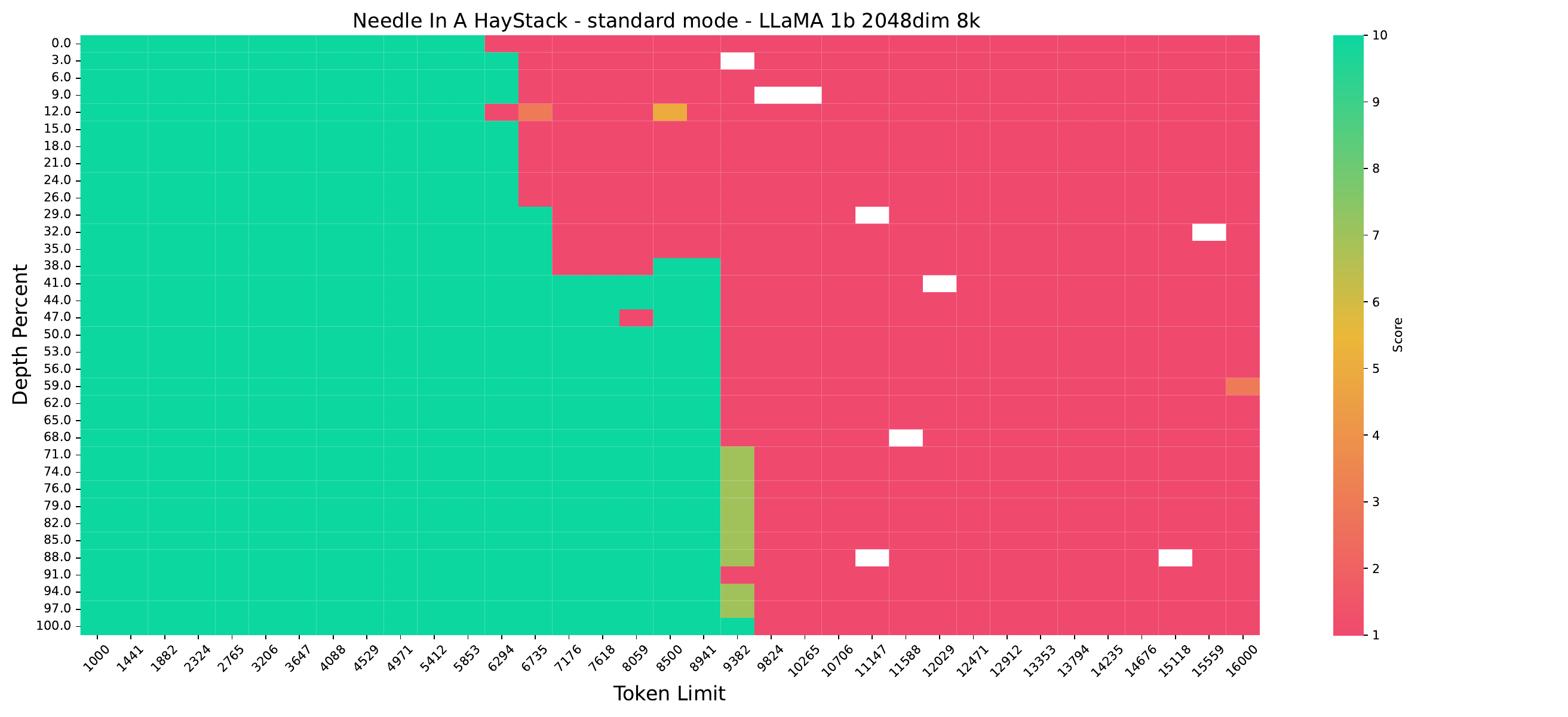}
\includegraphics[width=1\linewidth]{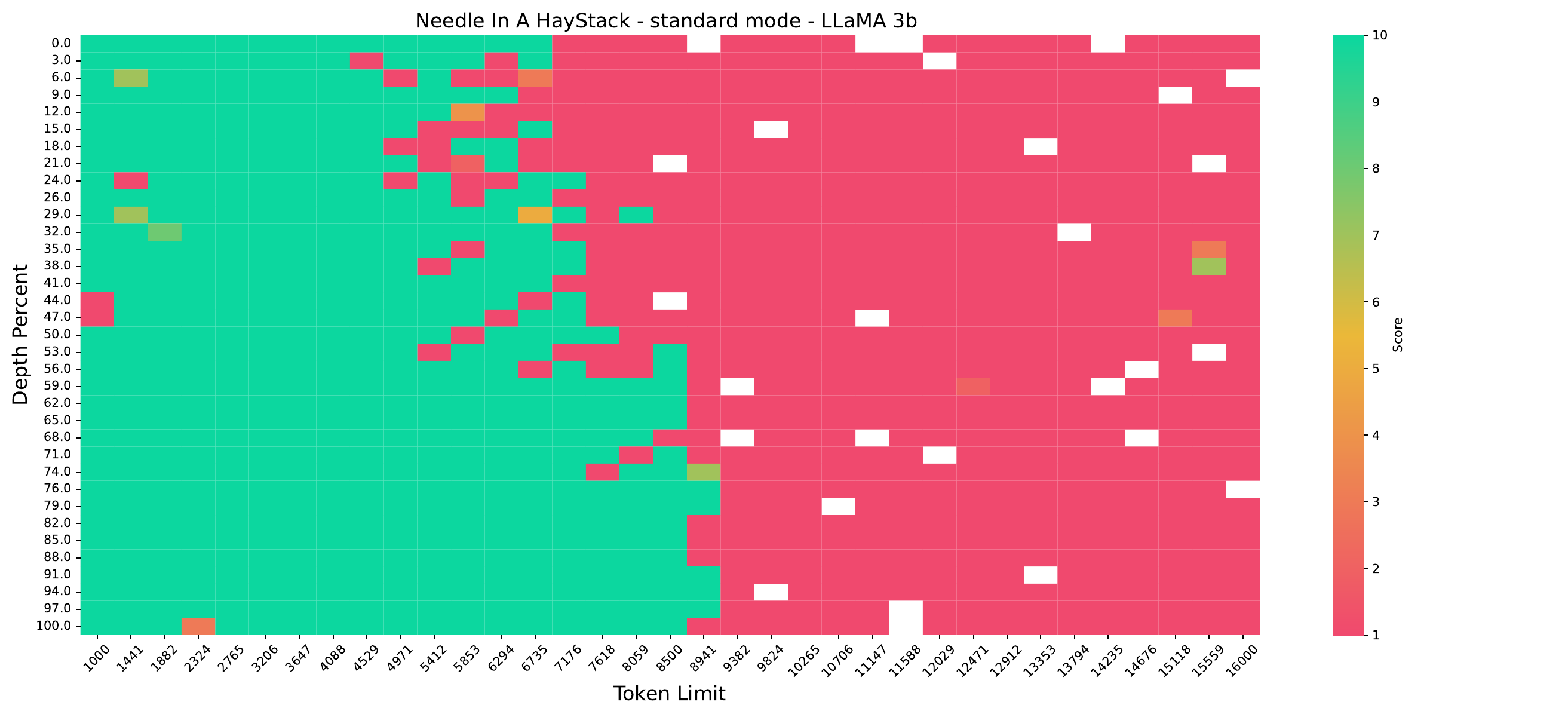}
\includegraphics[width=1\linewidth]{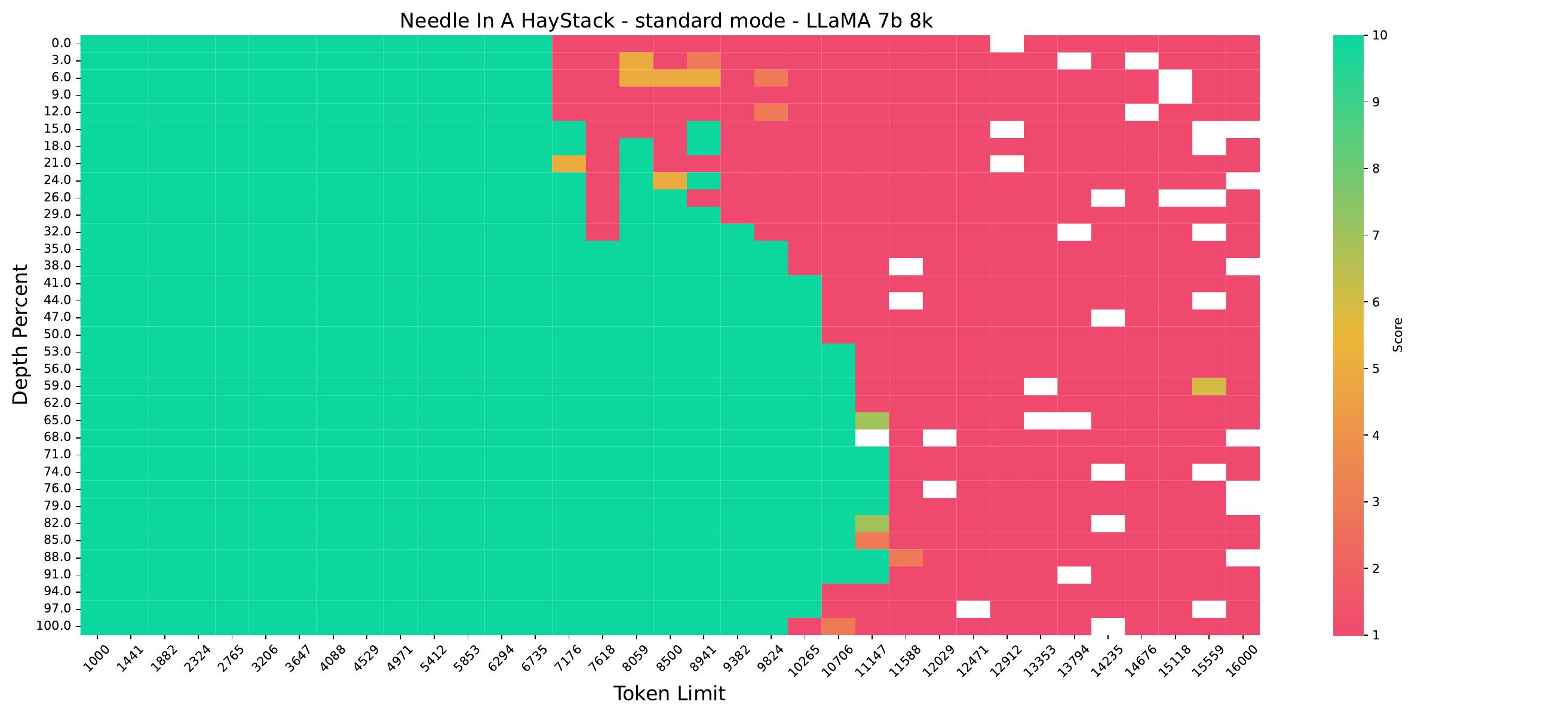}
\end{figure*}

\begin{figure*}
\centering
\includegraphics[width=1\linewidth]{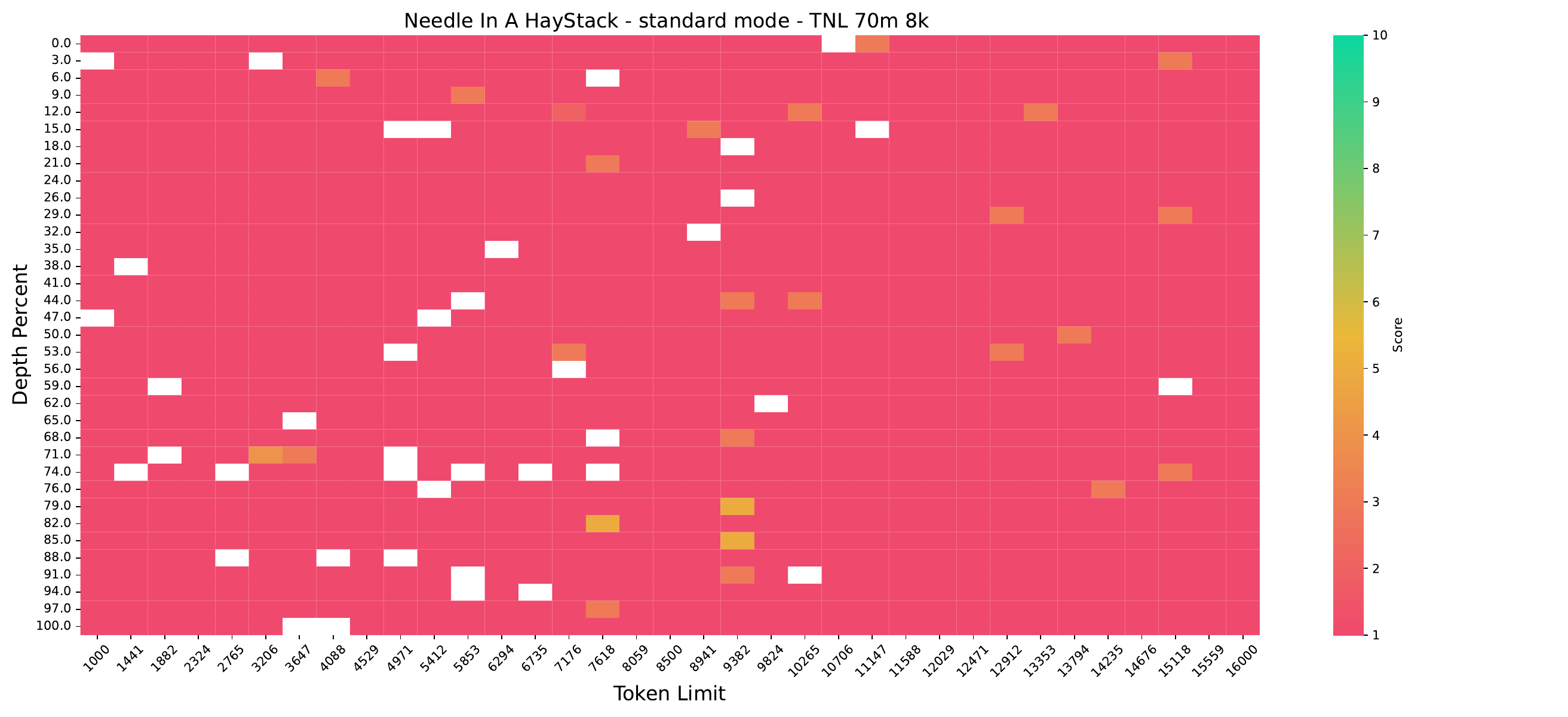}
\includegraphics[width=1\linewidth]{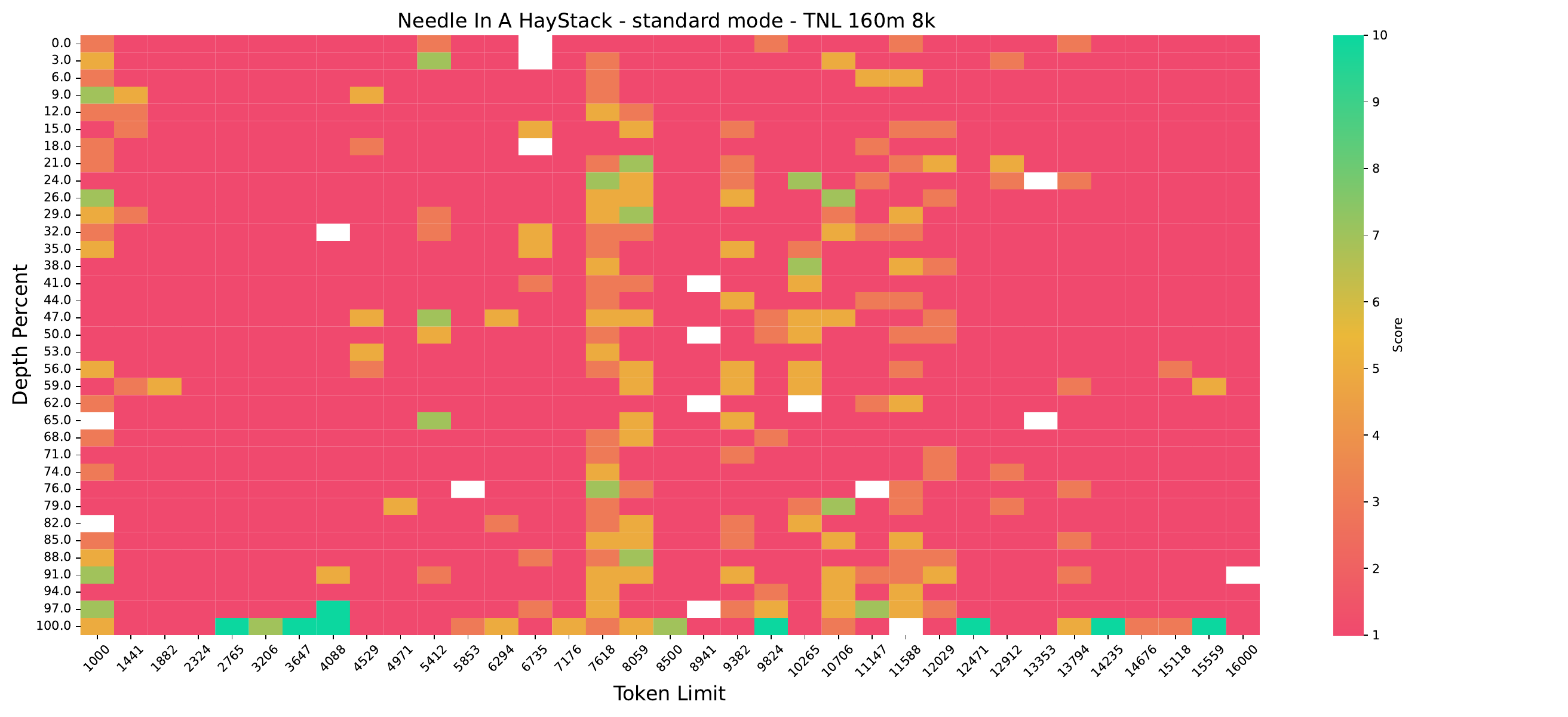}
\includegraphics[width=1\linewidth]{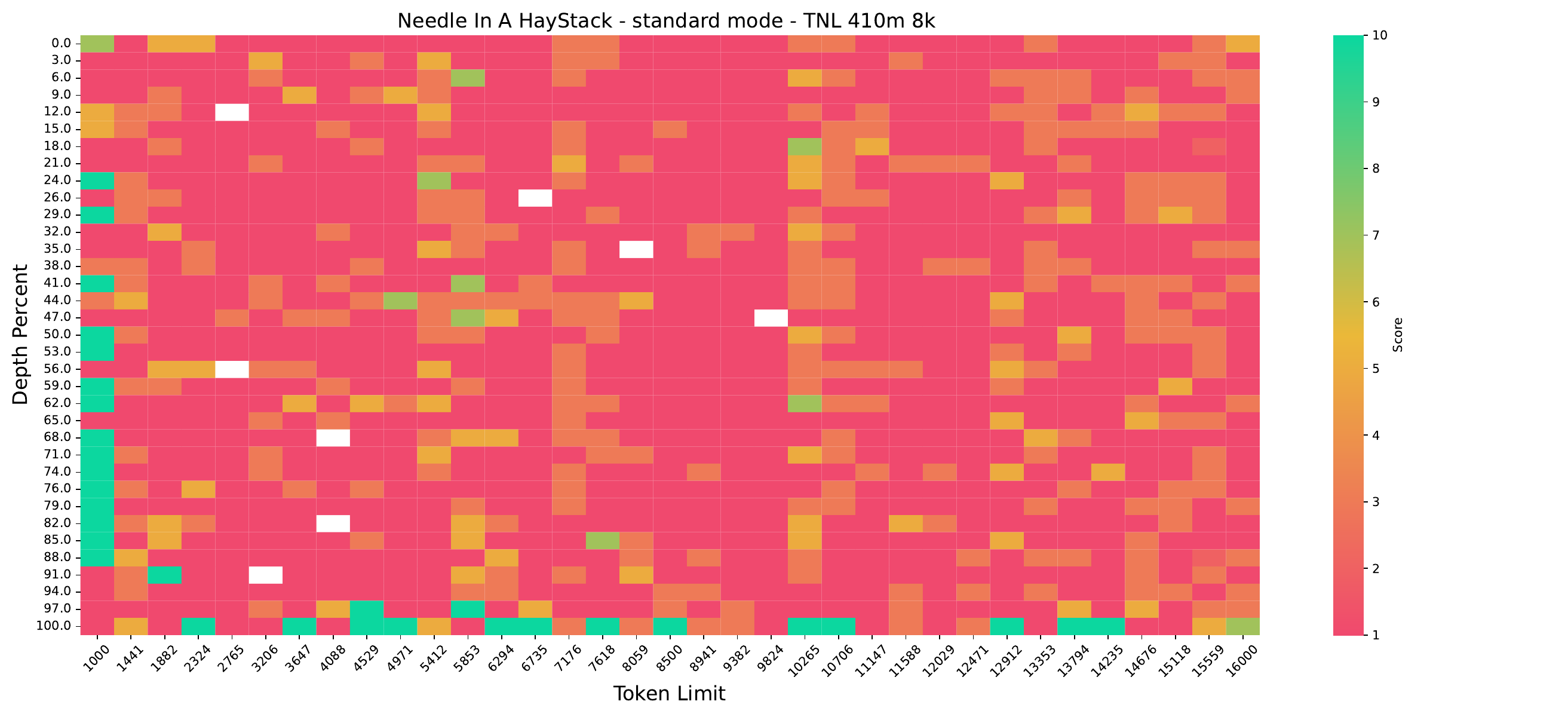}
\end{figure*}

\begin{figure*}
\centering
\includegraphics[width=1\linewidth]{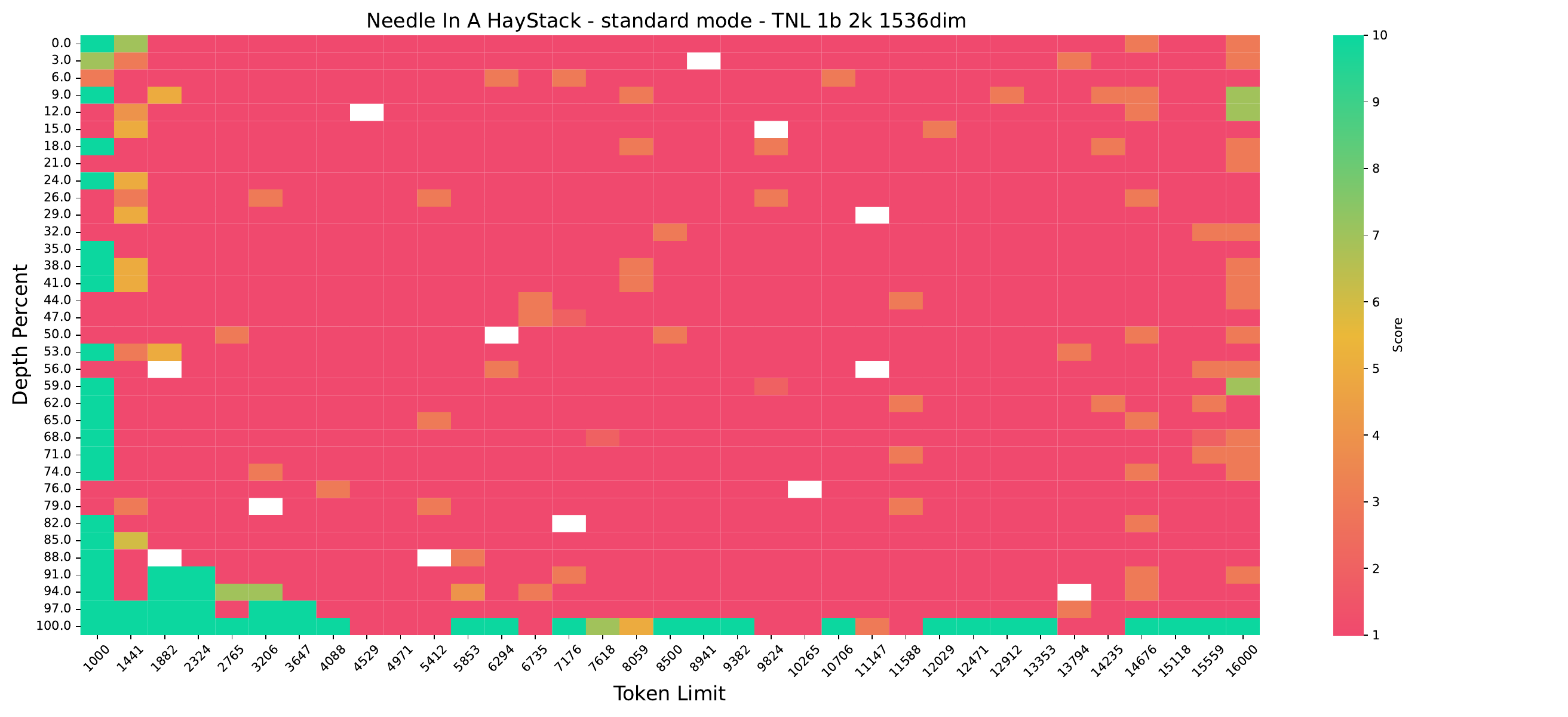}
\includegraphics[width=1\linewidth]{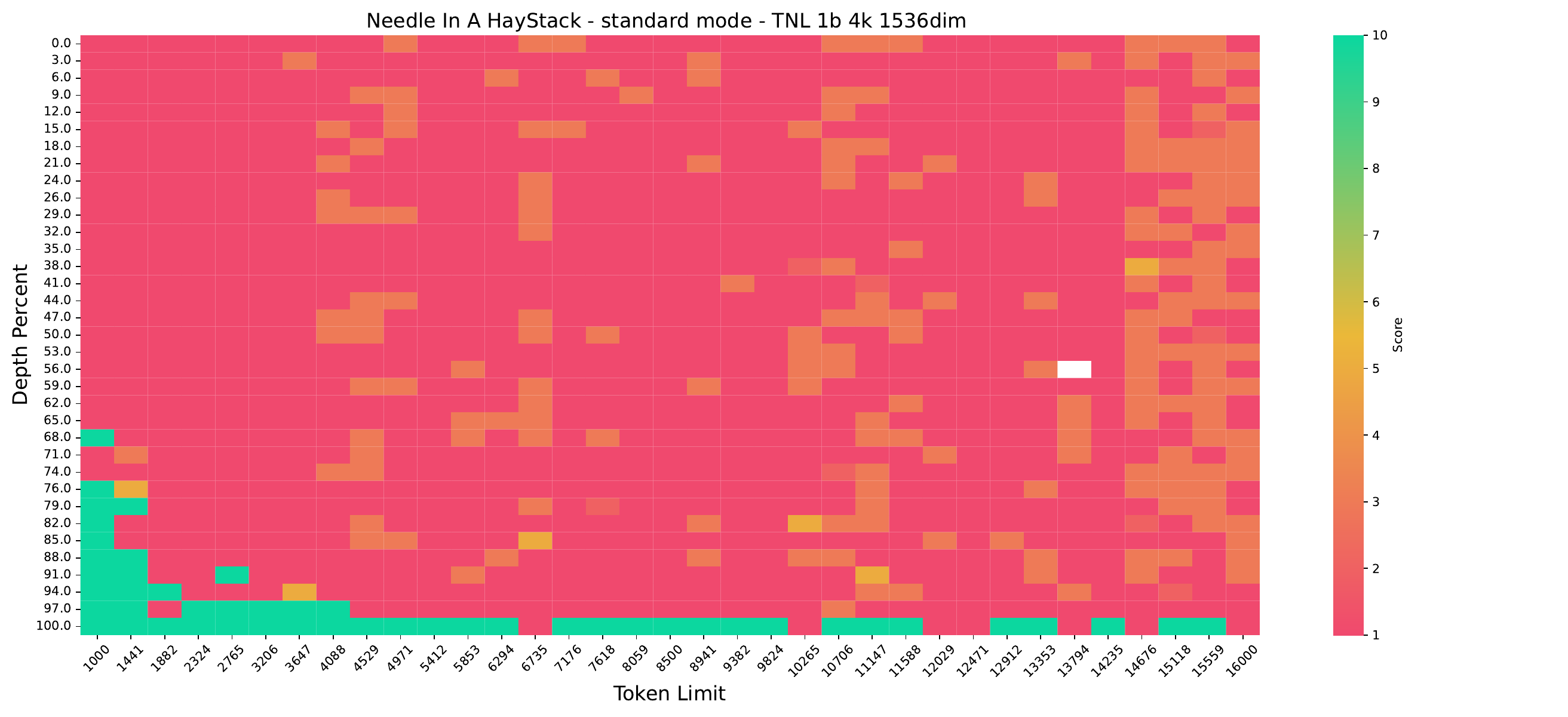}
\includegraphics[width=1\linewidth]{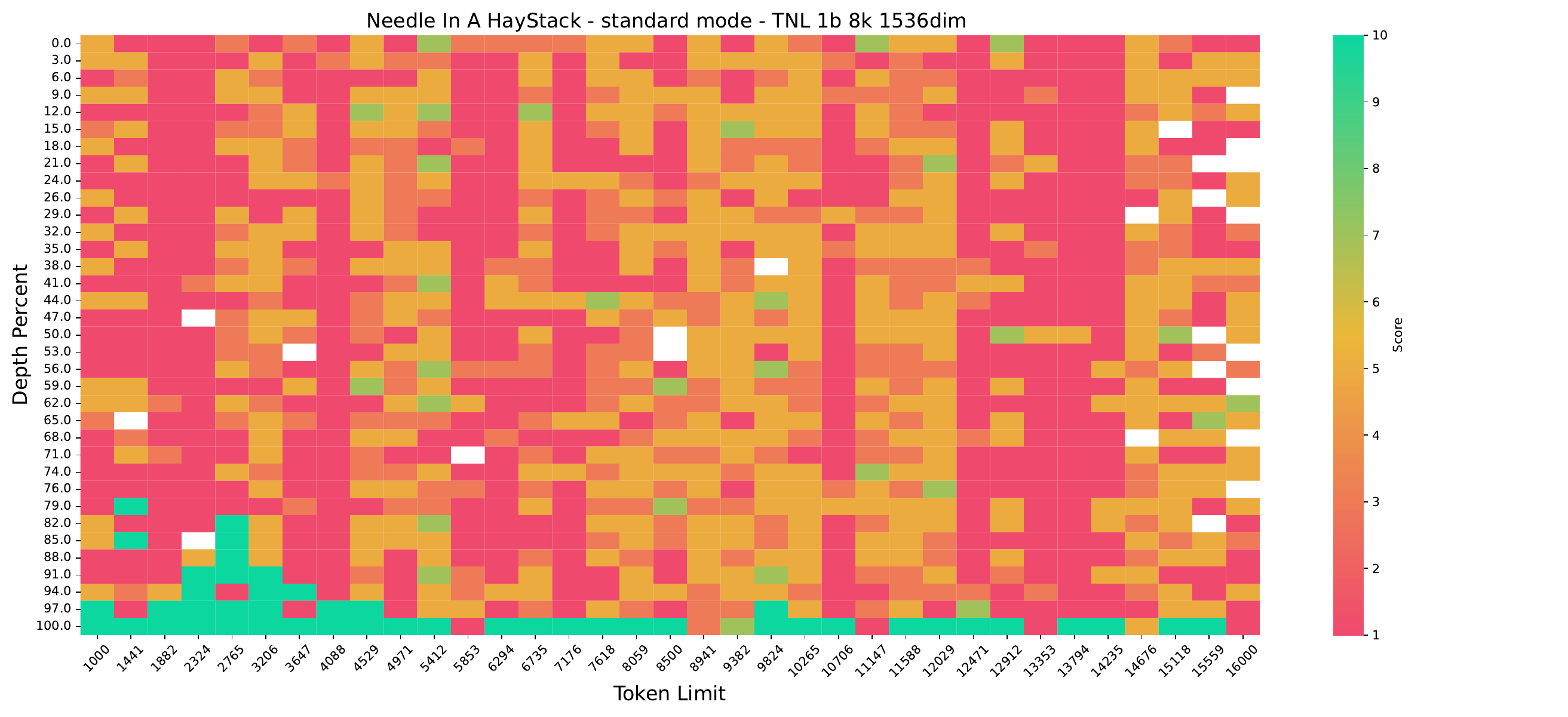}
\end{figure*}

\begin{figure*}
\centering
\includegraphics[width=1\linewidth]{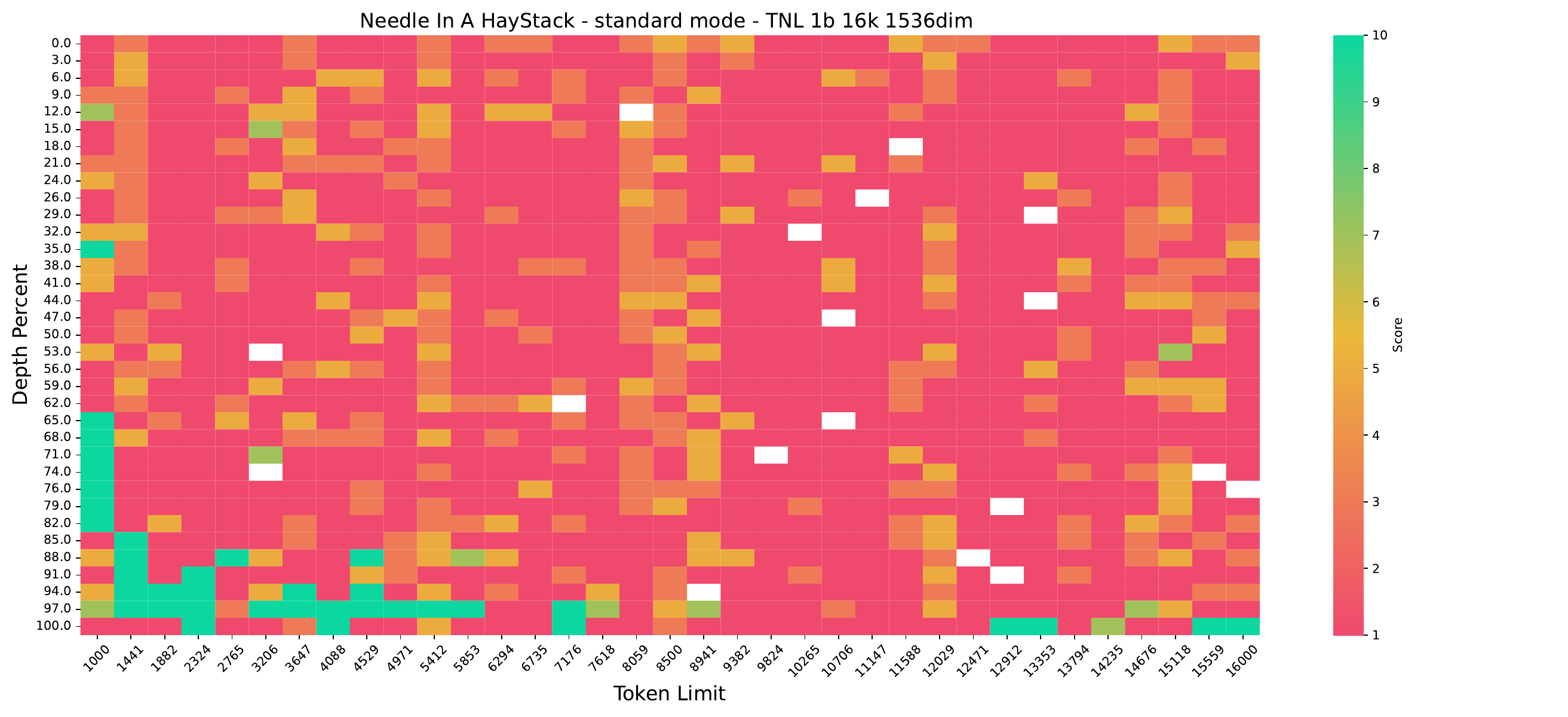}
\includegraphics[width=1\linewidth]{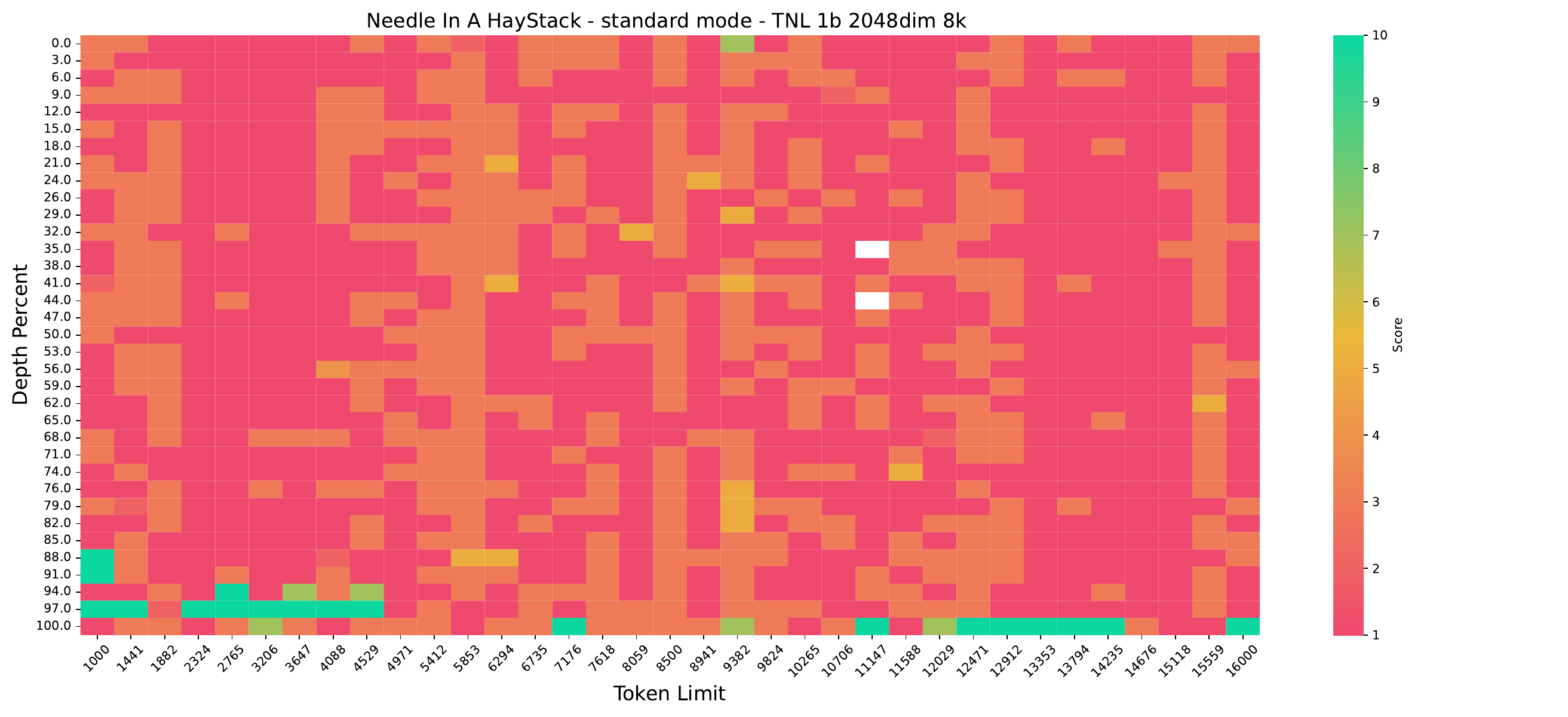}
\includegraphics[width=1\linewidth]{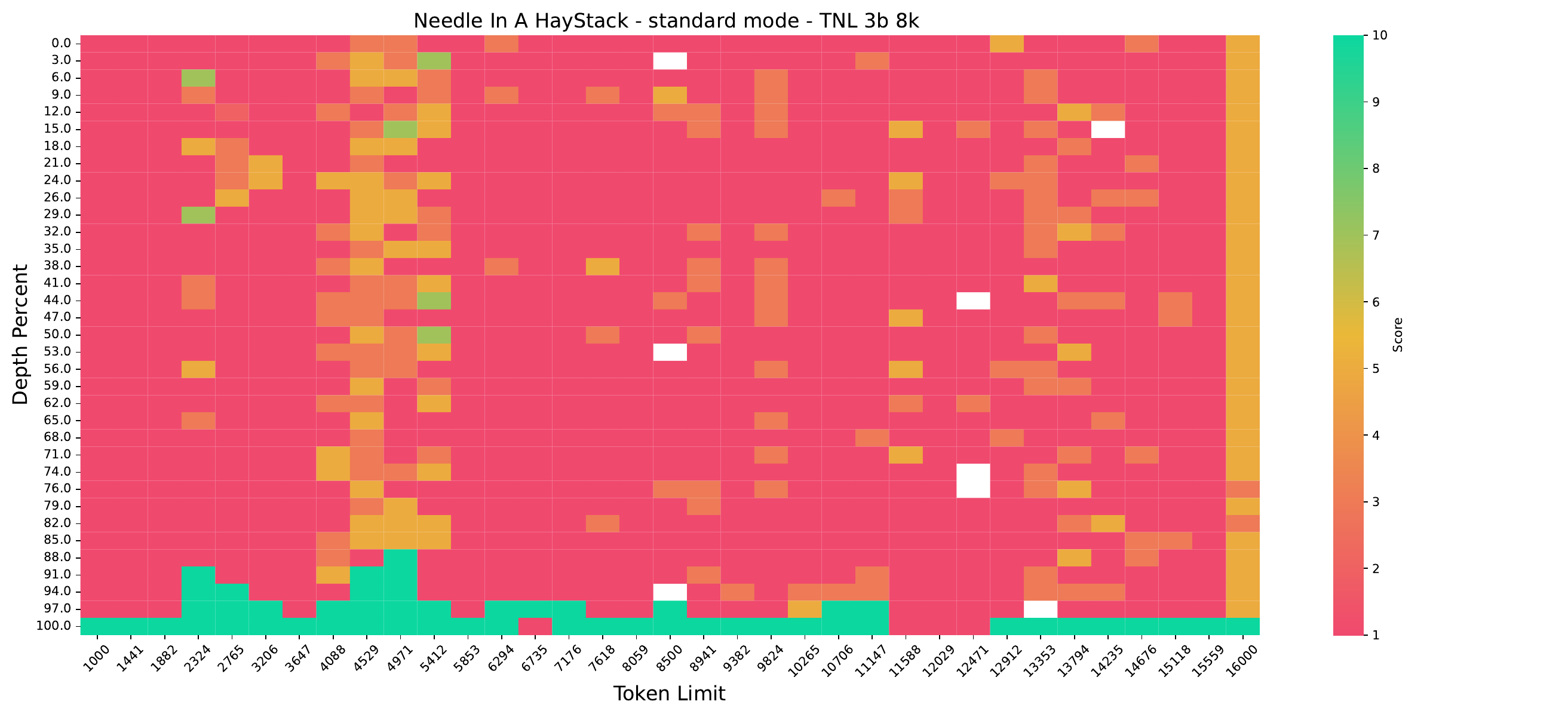}
\end{figure*}

\begin{figure*}
\centering
\includegraphics[width=1\linewidth]{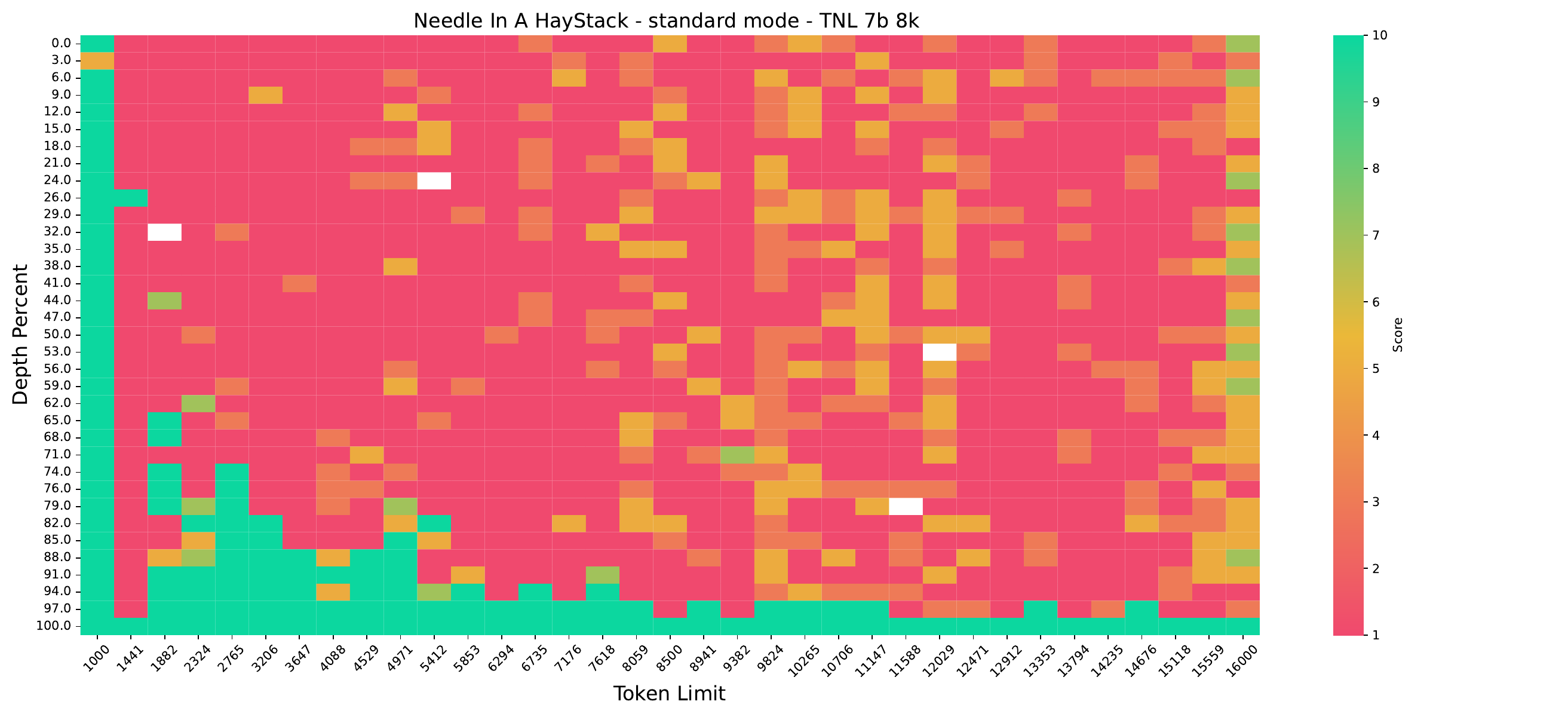}
\includegraphics[width=1\linewidth]{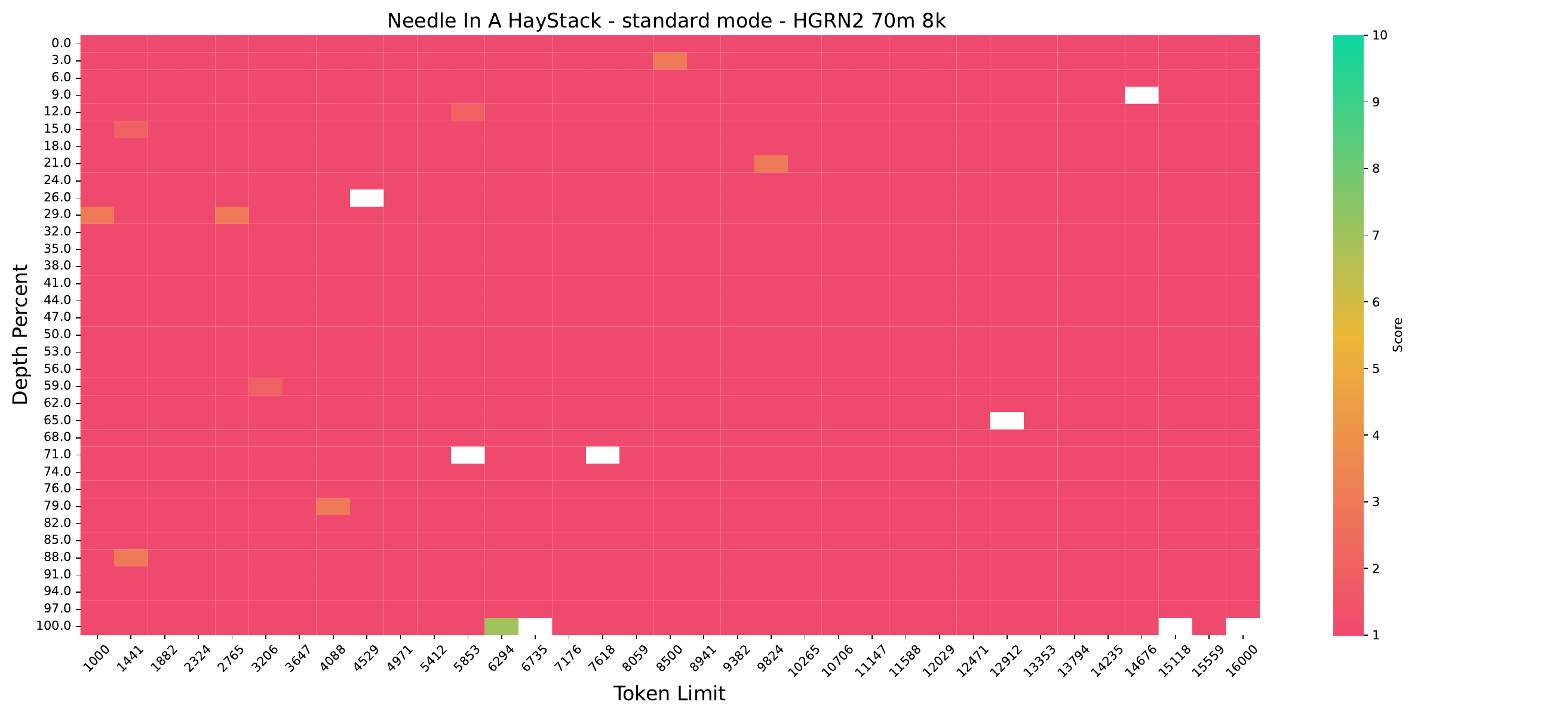}
\includegraphics[width=1\linewidth]{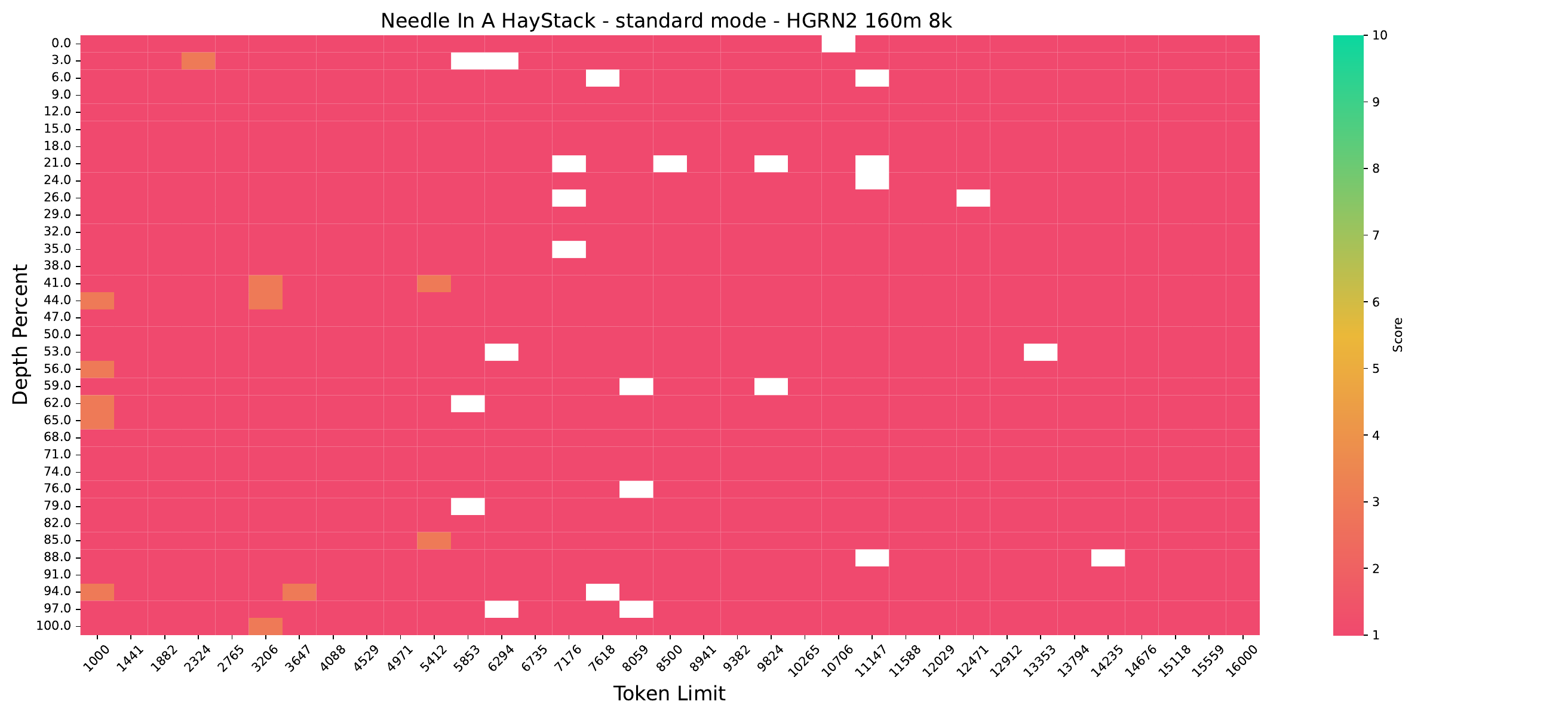}
\end{figure*}

\begin{figure*}
\centering
\includegraphics[width=1\linewidth]{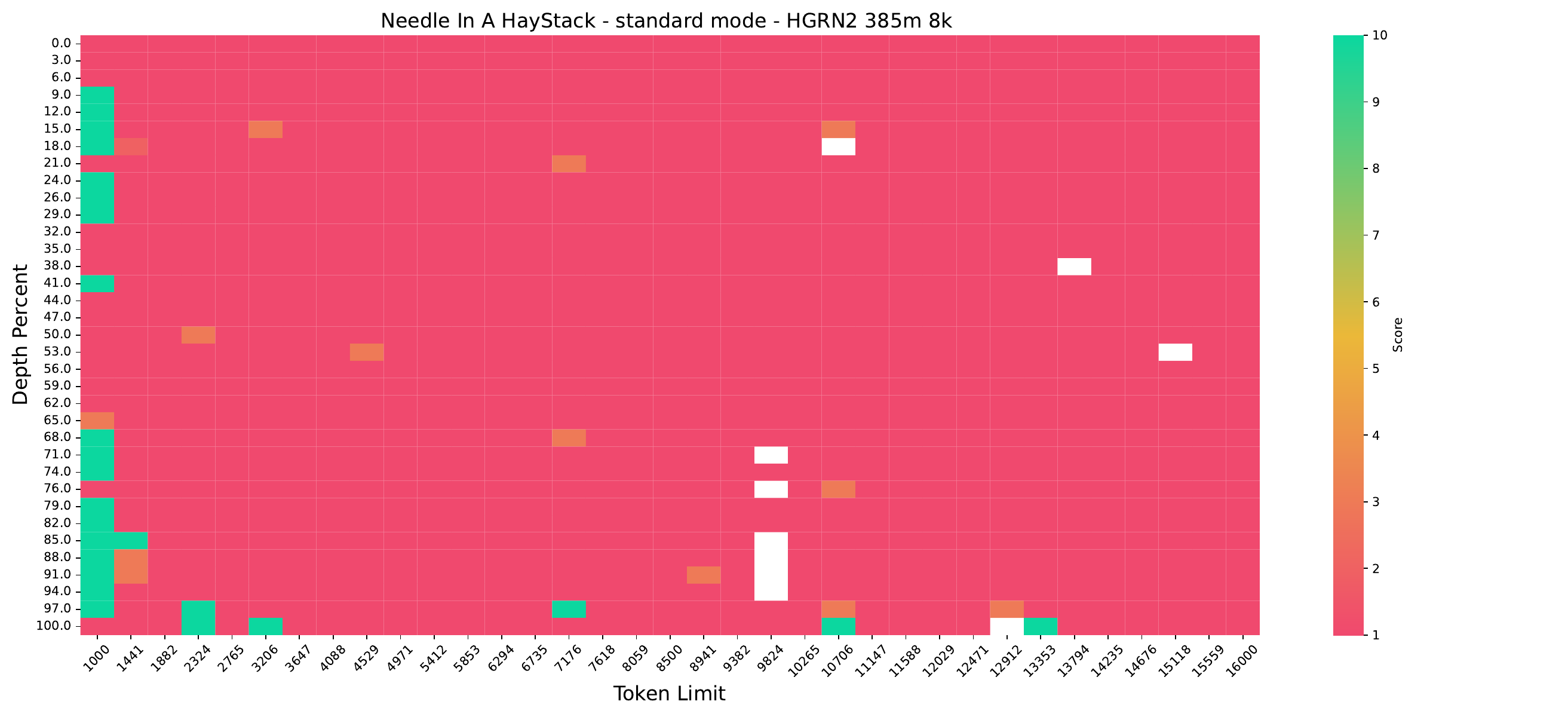}
\includegraphics[width=1\linewidth]{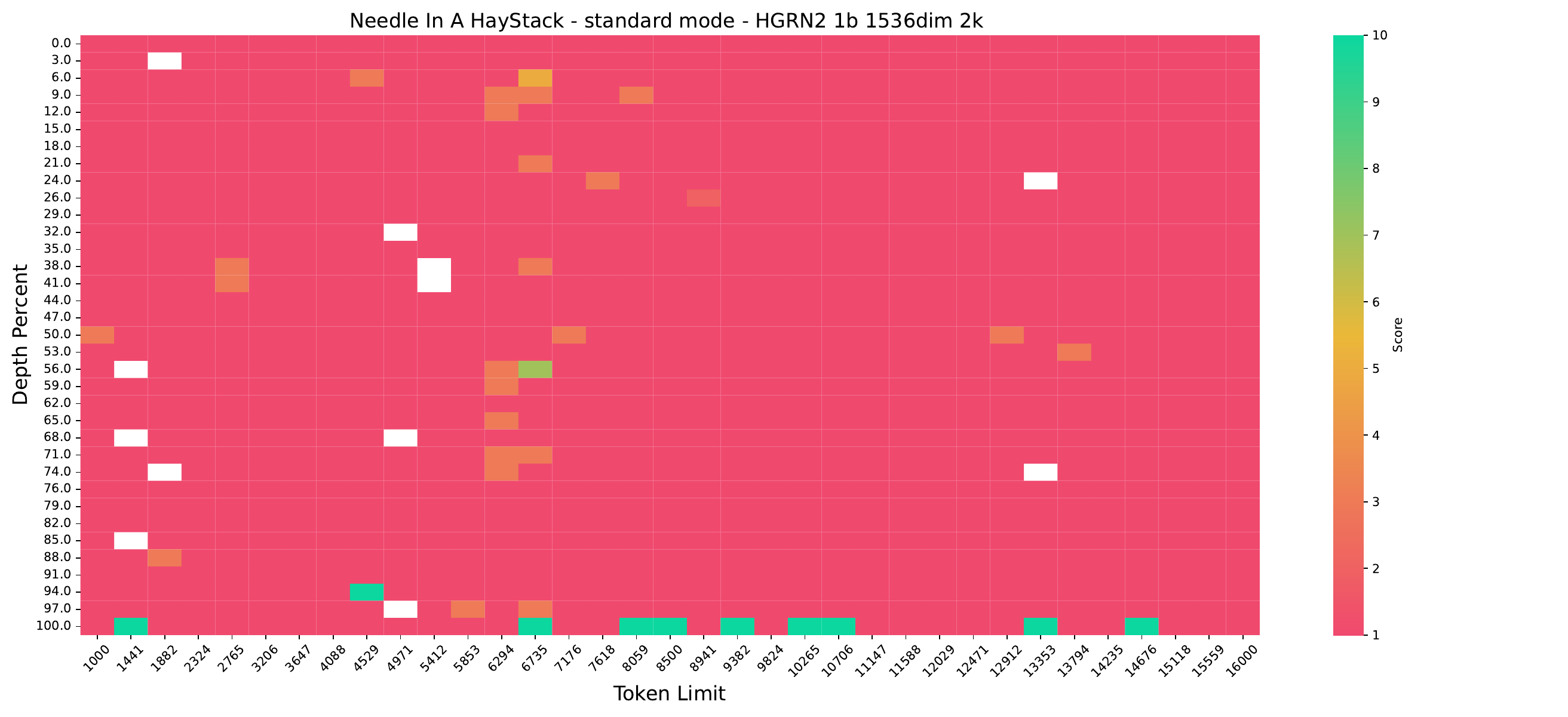}
\includegraphics[width=1\linewidth]{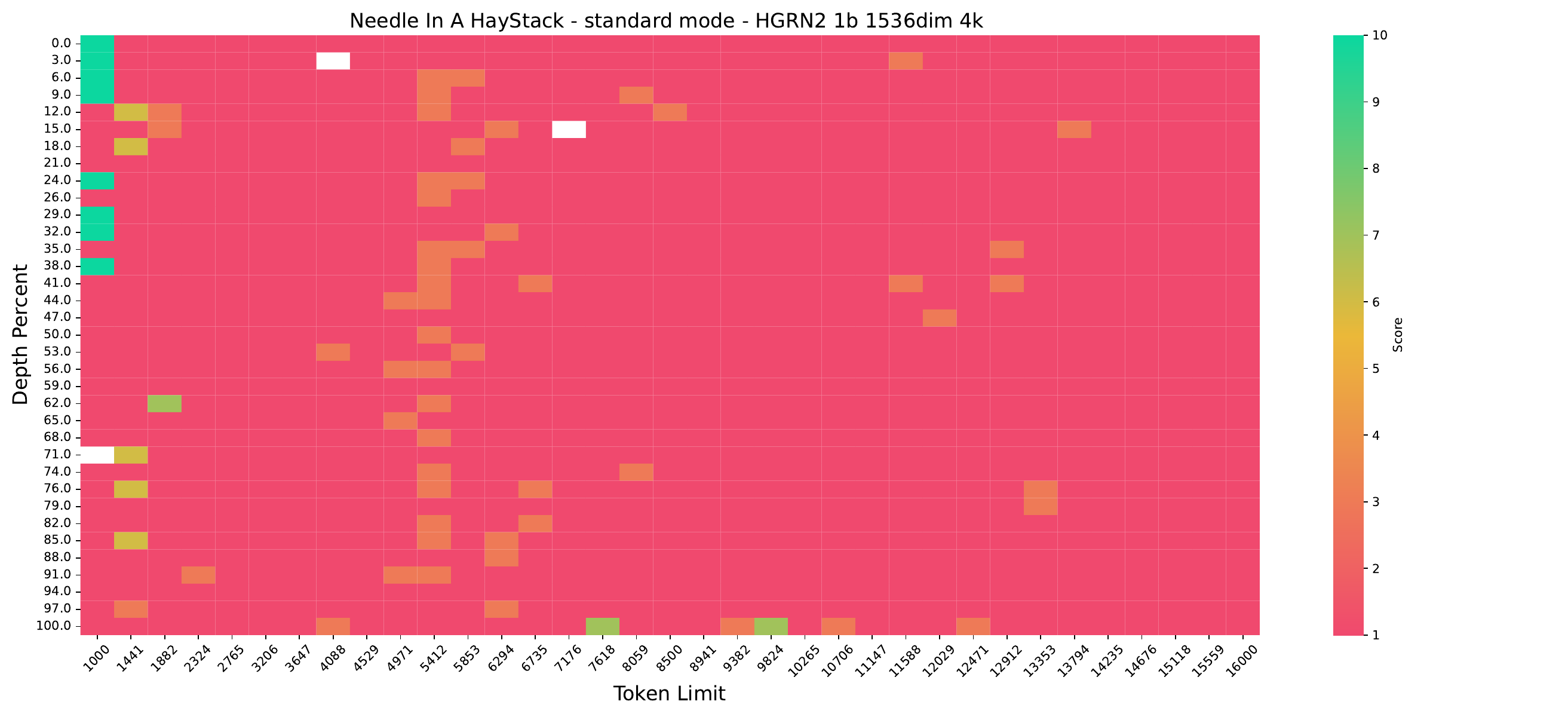}
\end{figure*}

\begin{figure*}
\centering
\includegraphics[width=1\linewidth]{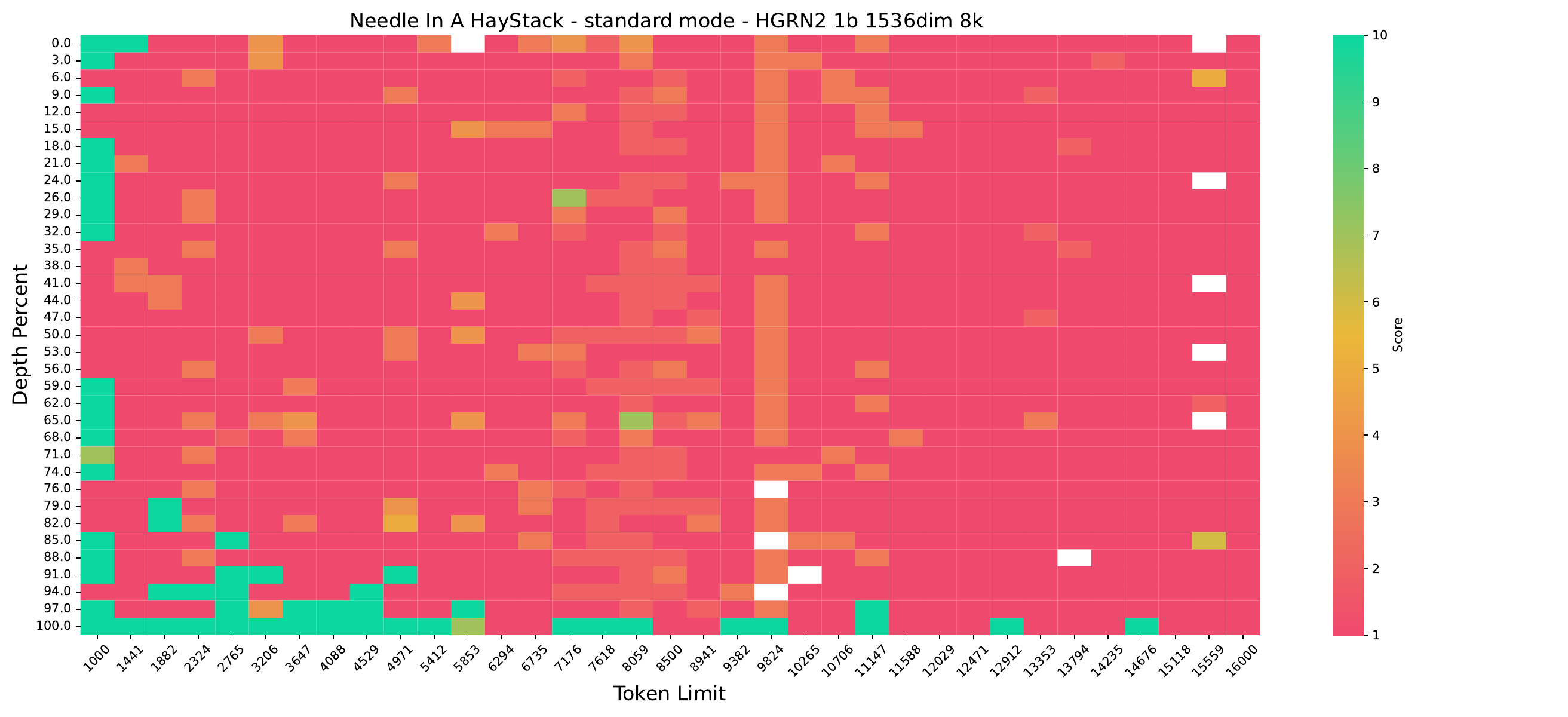}
\includegraphics[width=1\linewidth]{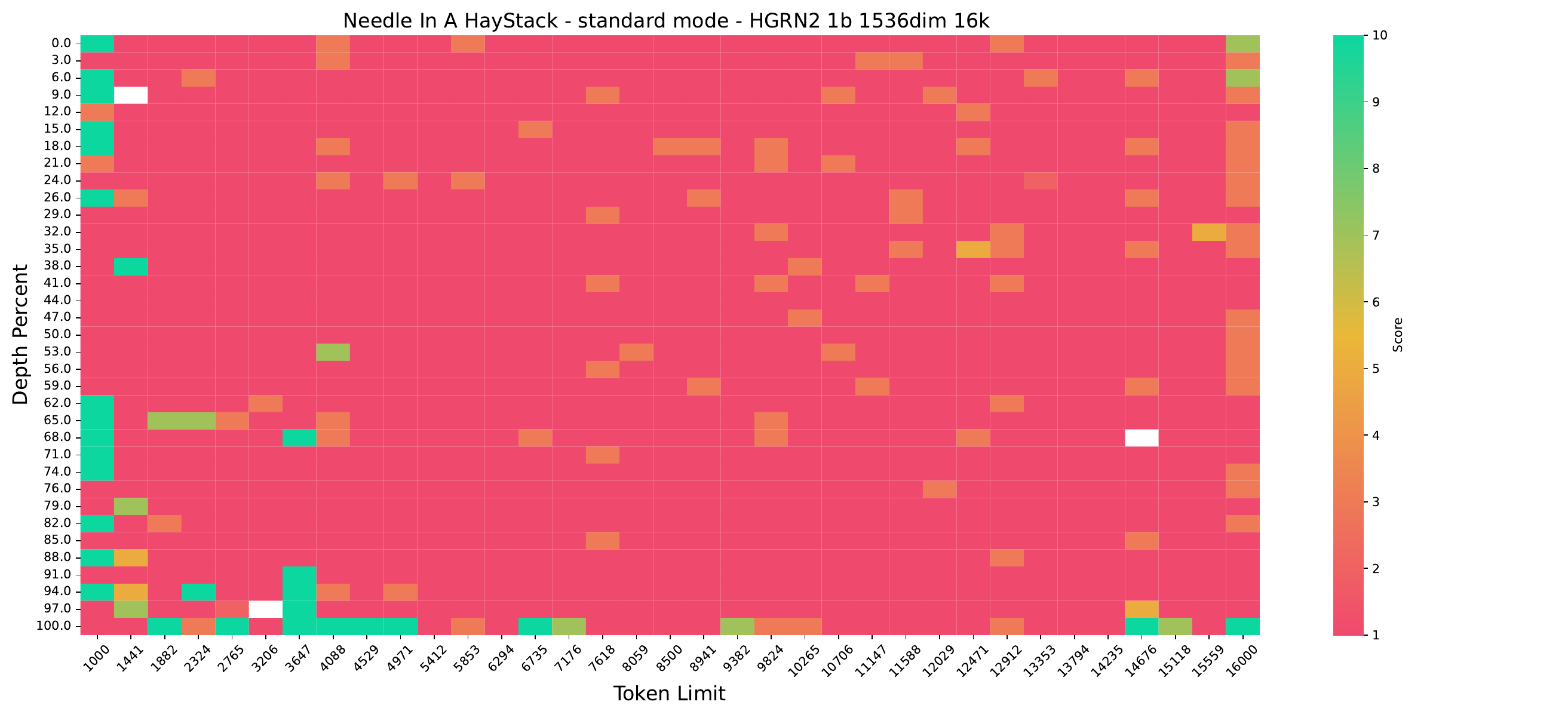}
\includegraphics[width=1\linewidth]{figures/niah/hgrn2_1b-2048dim-8k-ck1-comprehension.pdf}
\end{figure*}

\begin{figure*}
\centering
\includegraphics[width=1\linewidth]{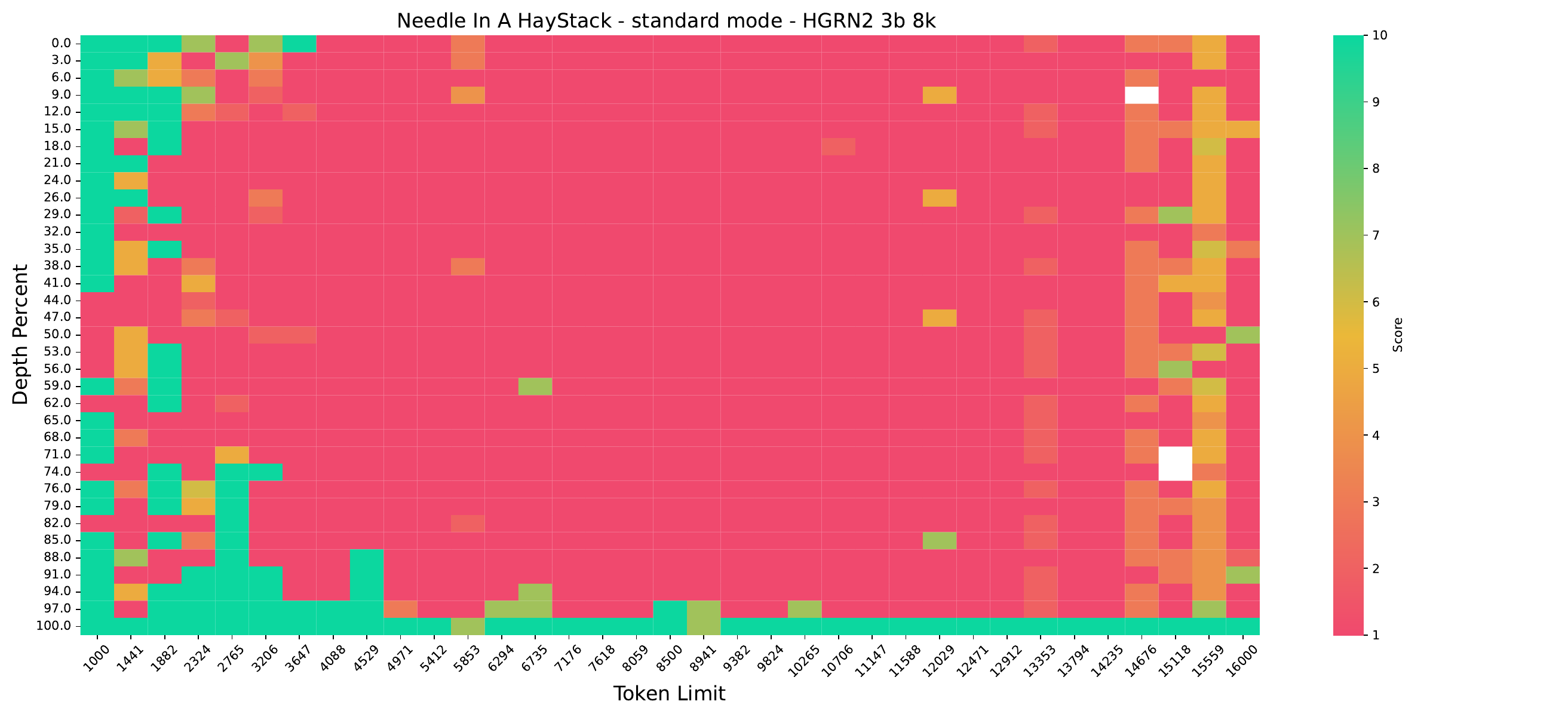}
\includegraphics[width=1\linewidth]{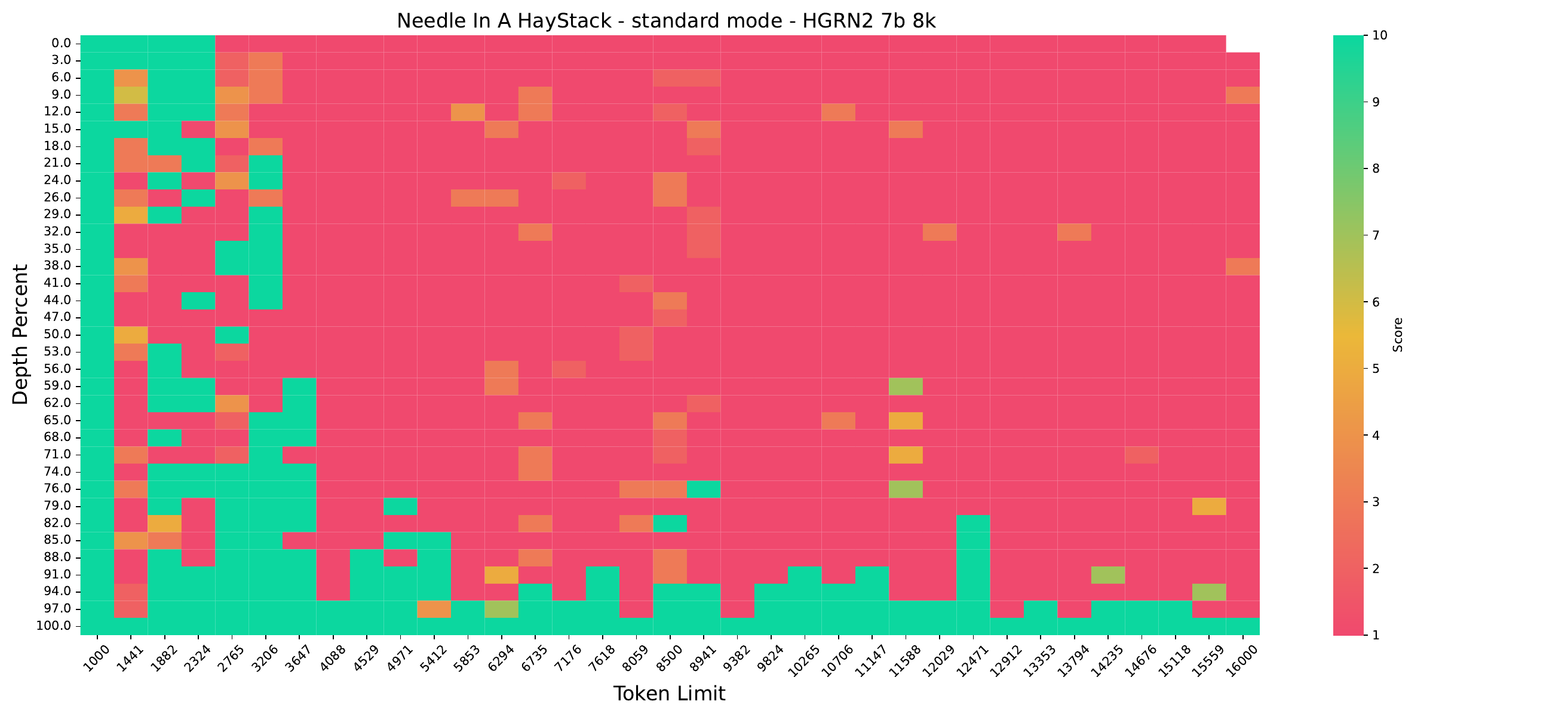}
\includegraphics[width=1\linewidth]{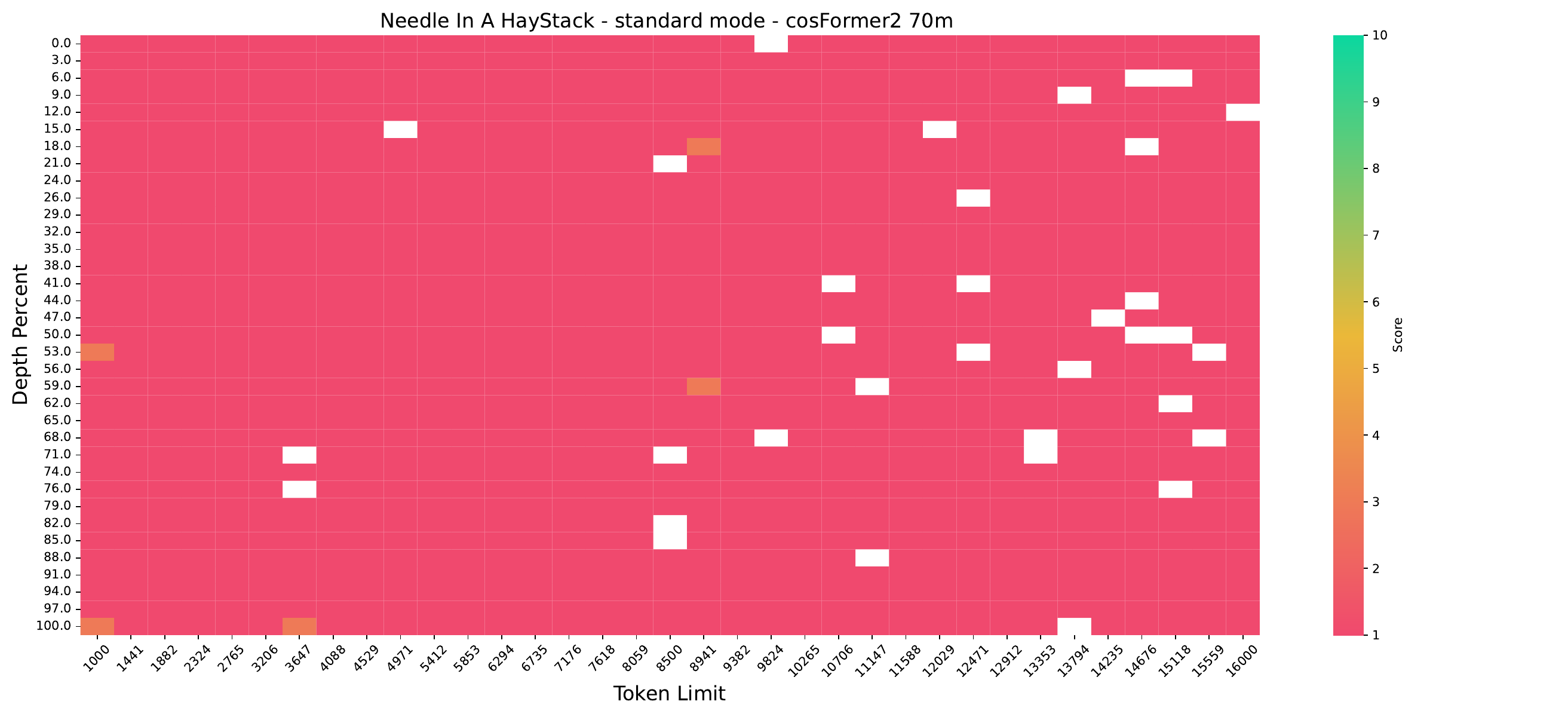}
\end{figure*}

\begin{figure*}
\centering
\includegraphics[width=1\linewidth]{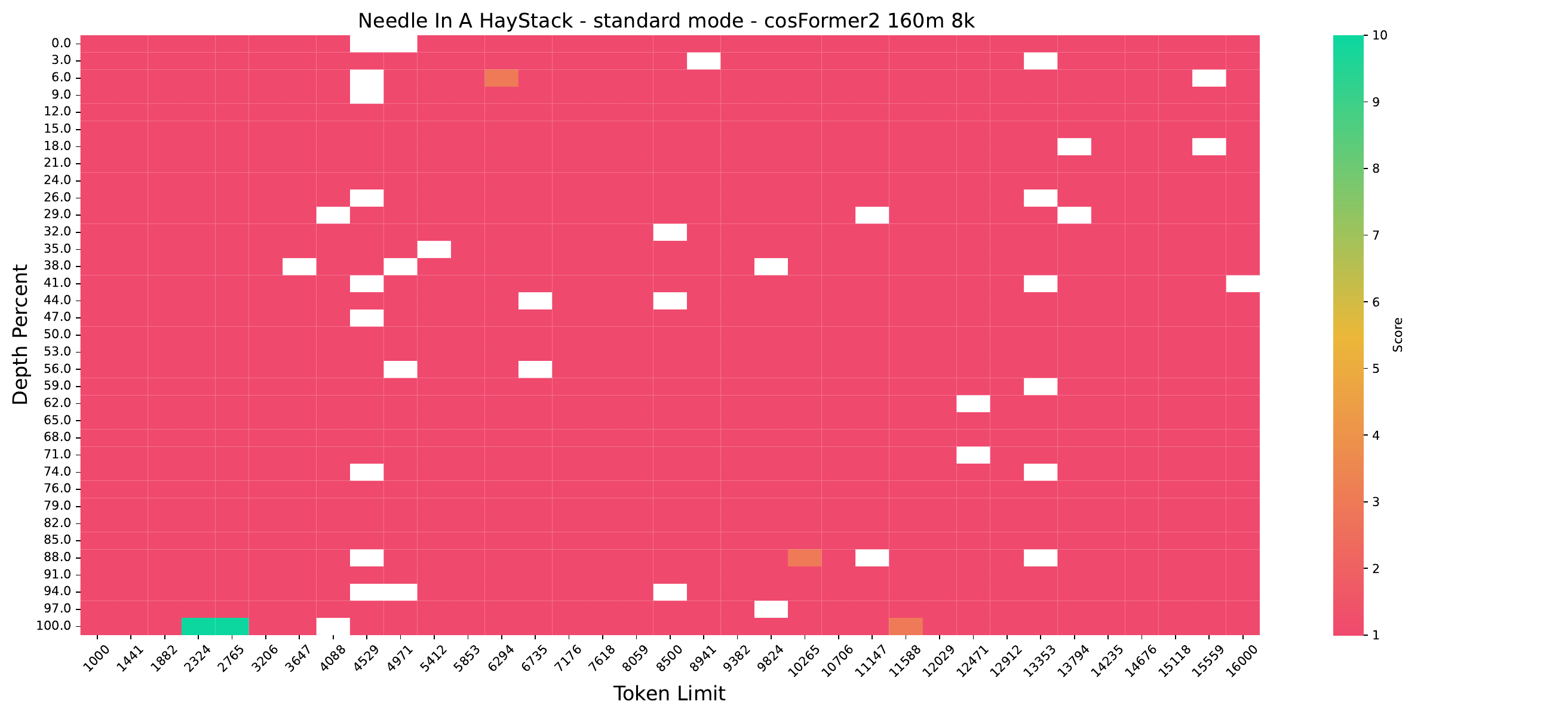}
\includegraphics[width=1\linewidth]{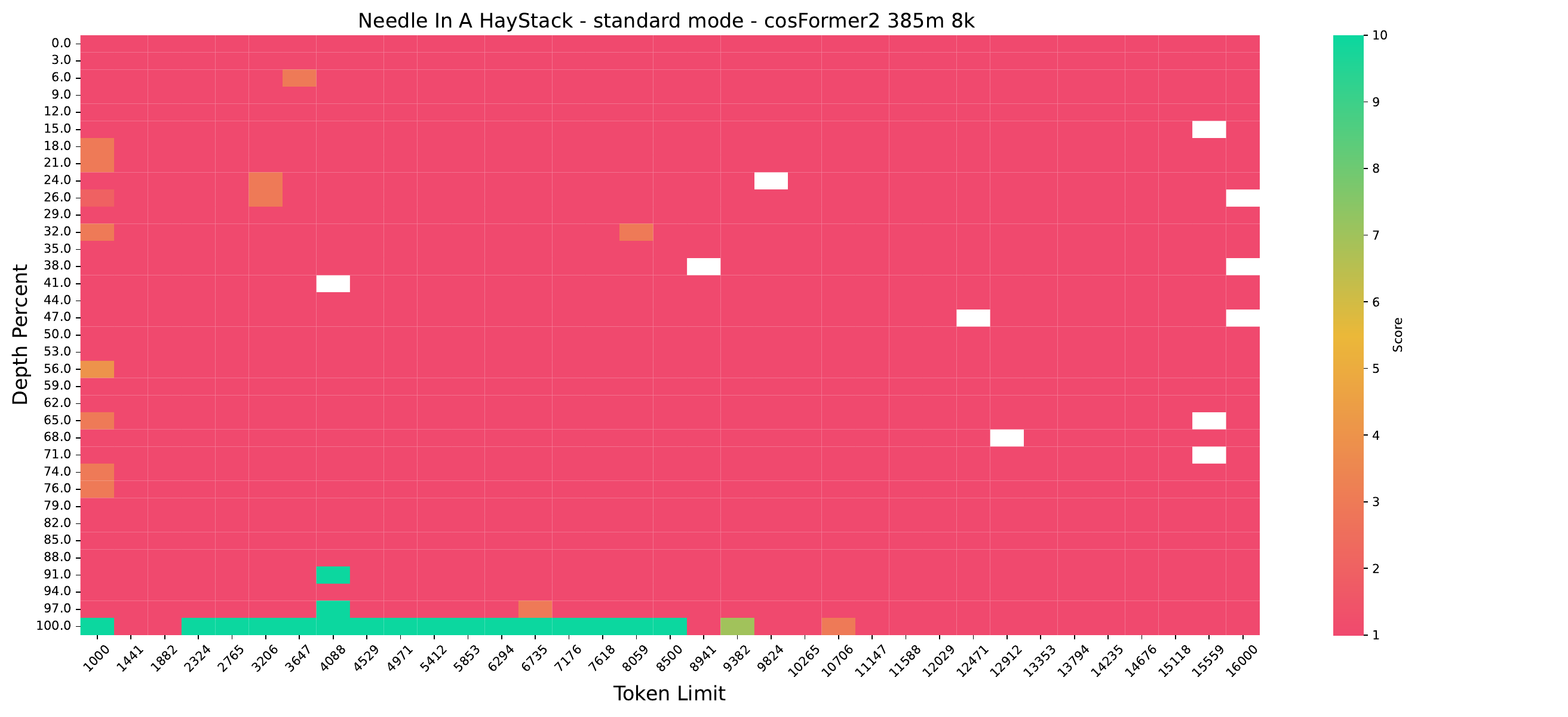}
\includegraphics[width=1\linewidth]{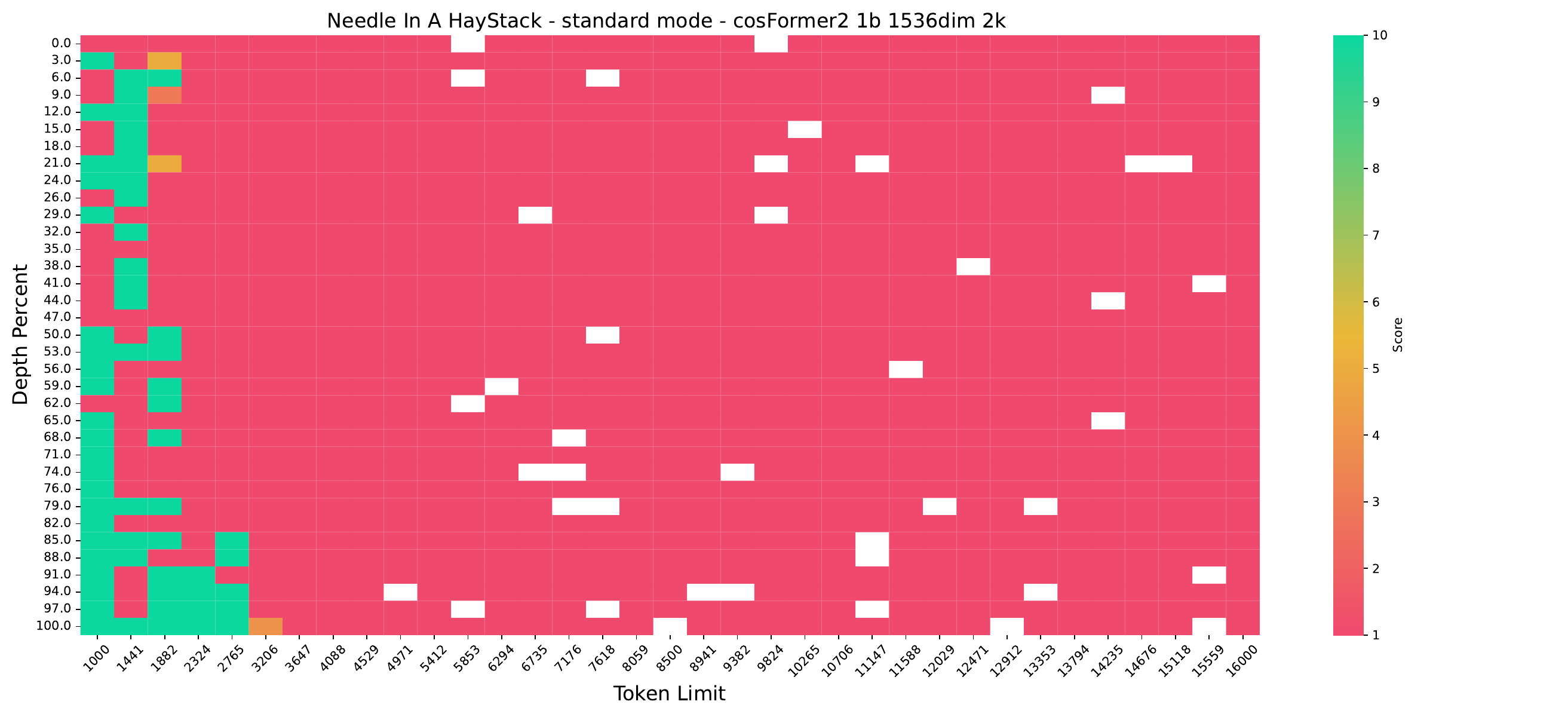}
\end{figure*}

\begin{figure*}
\centering
\includegraphics[width=1\linewidth]{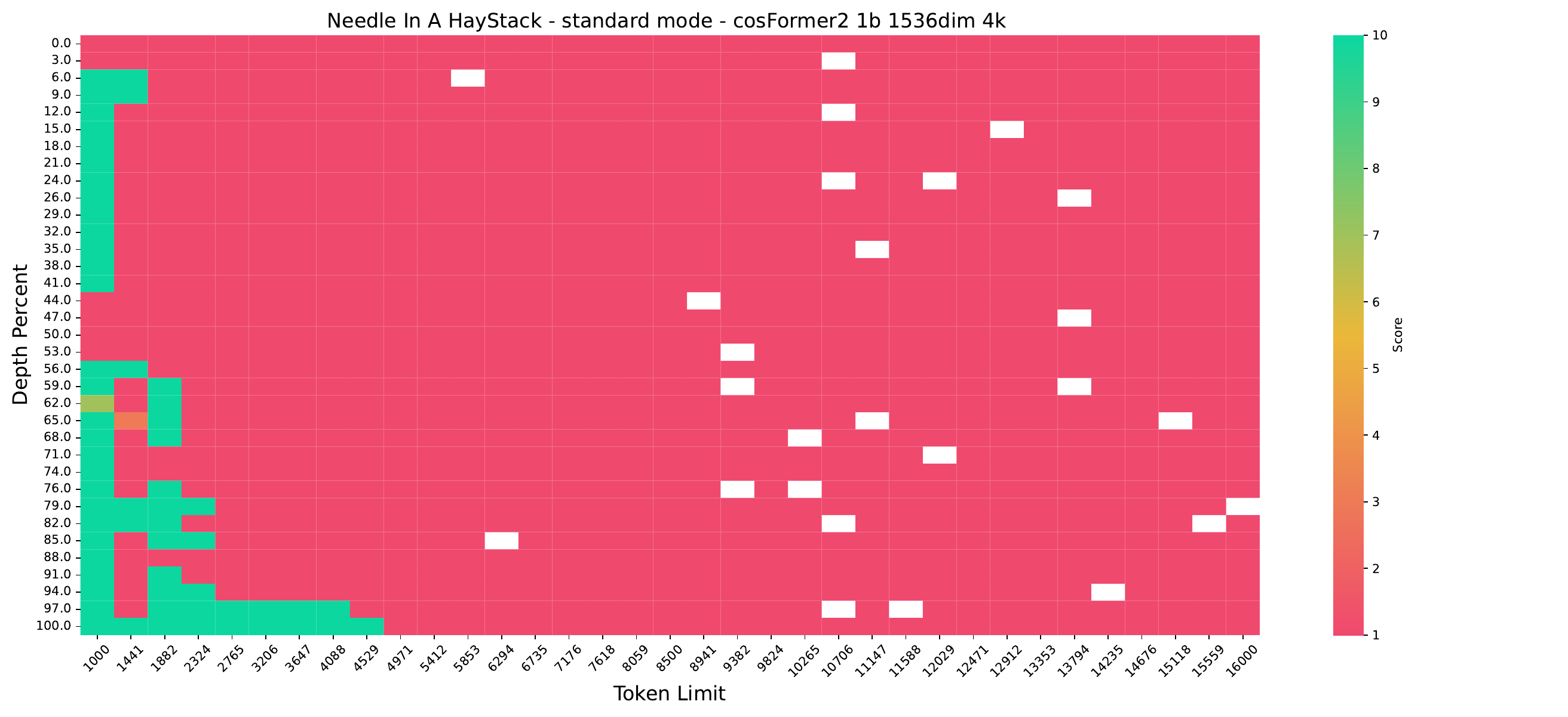}
\includegraphics[width=1\linewidth]{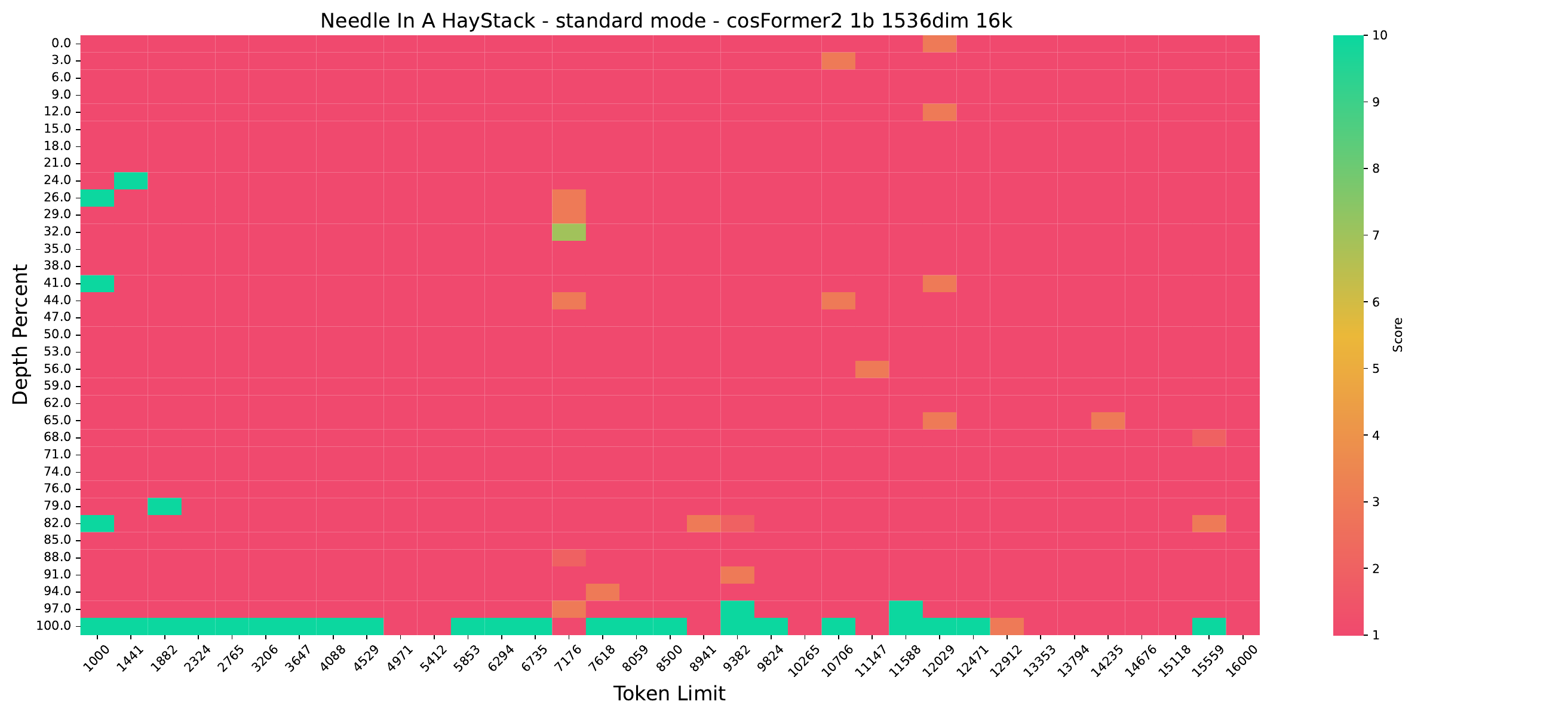}
\includegraphics[width=1\linewidth]{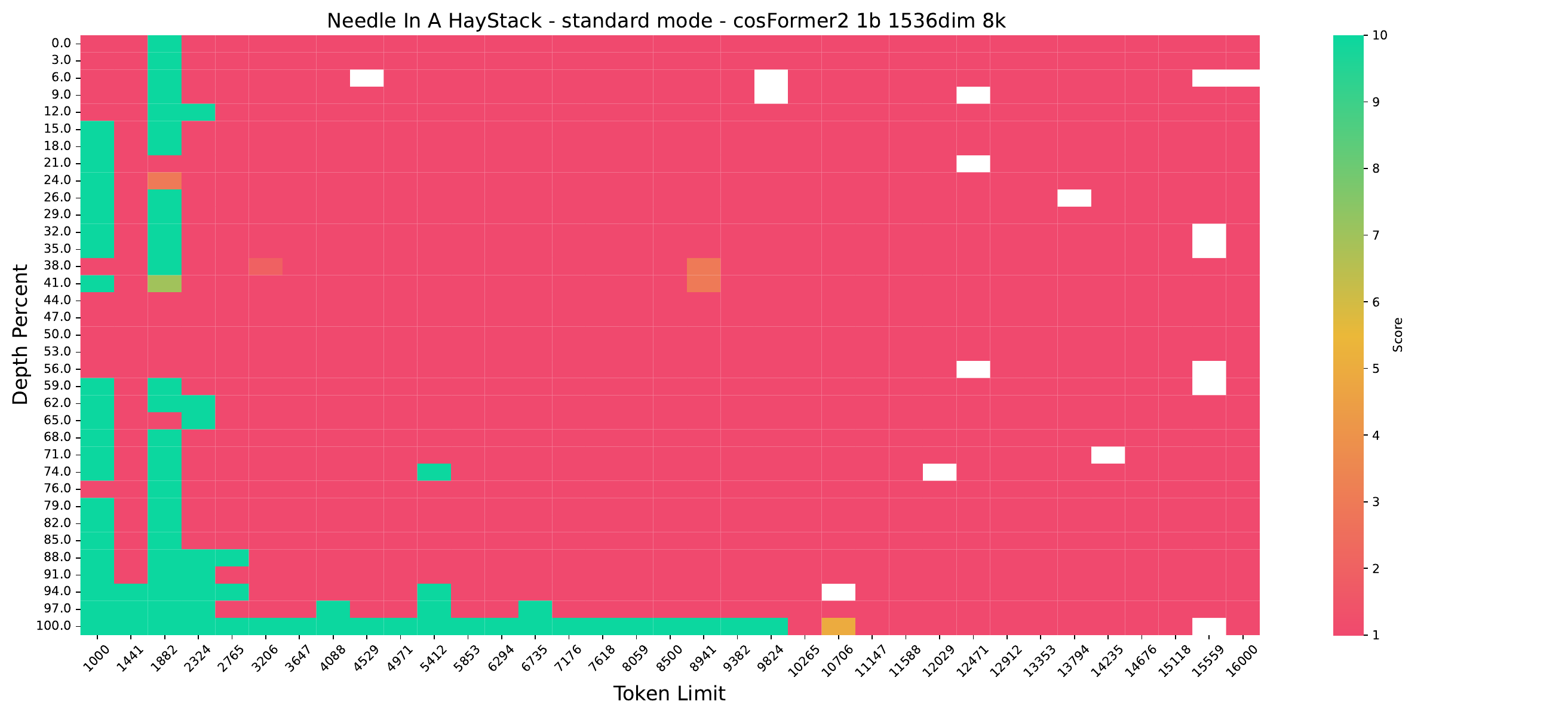}
\end{figure*}

\begin{figure*}
\centering
\includegraphics[width=1\linewidth]{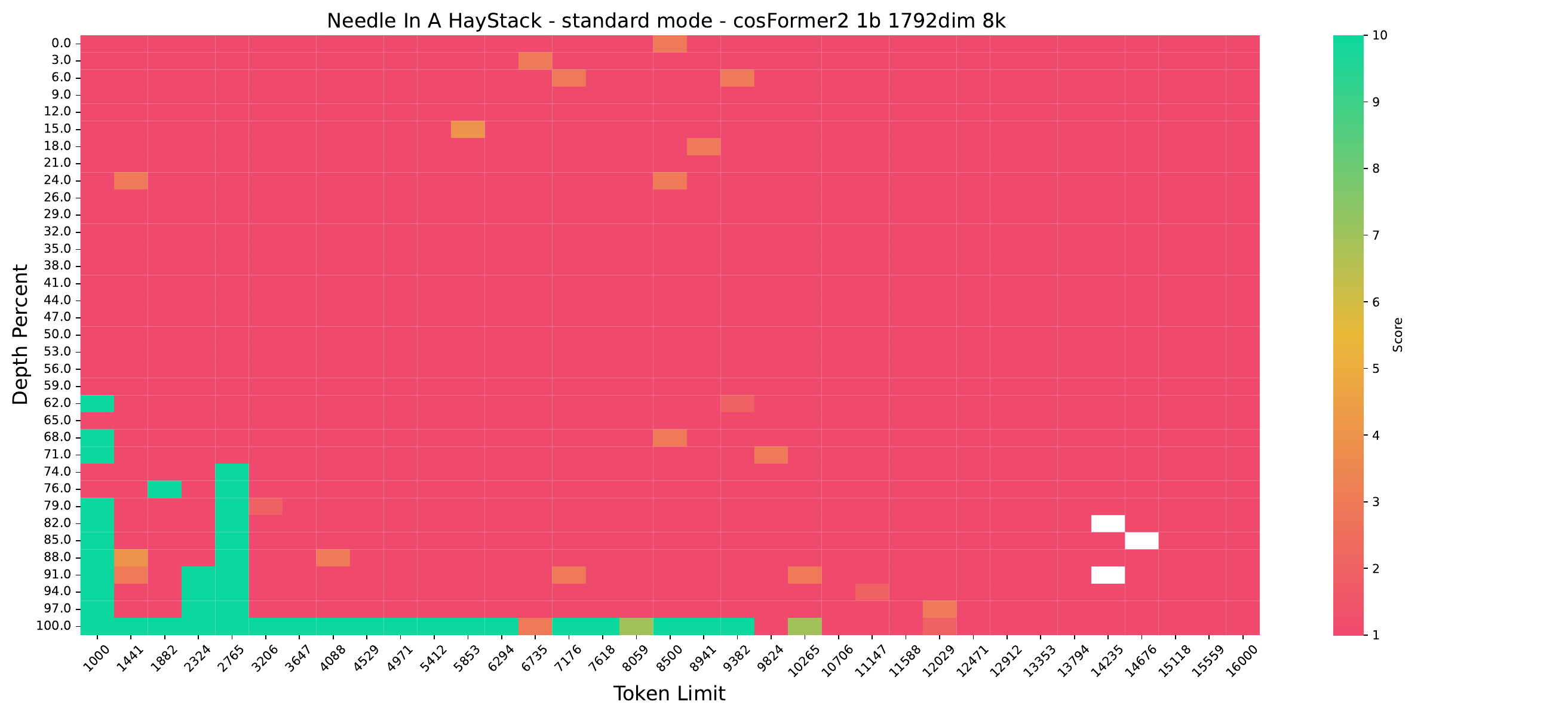}
\includegraphics[width=1\linewidth]{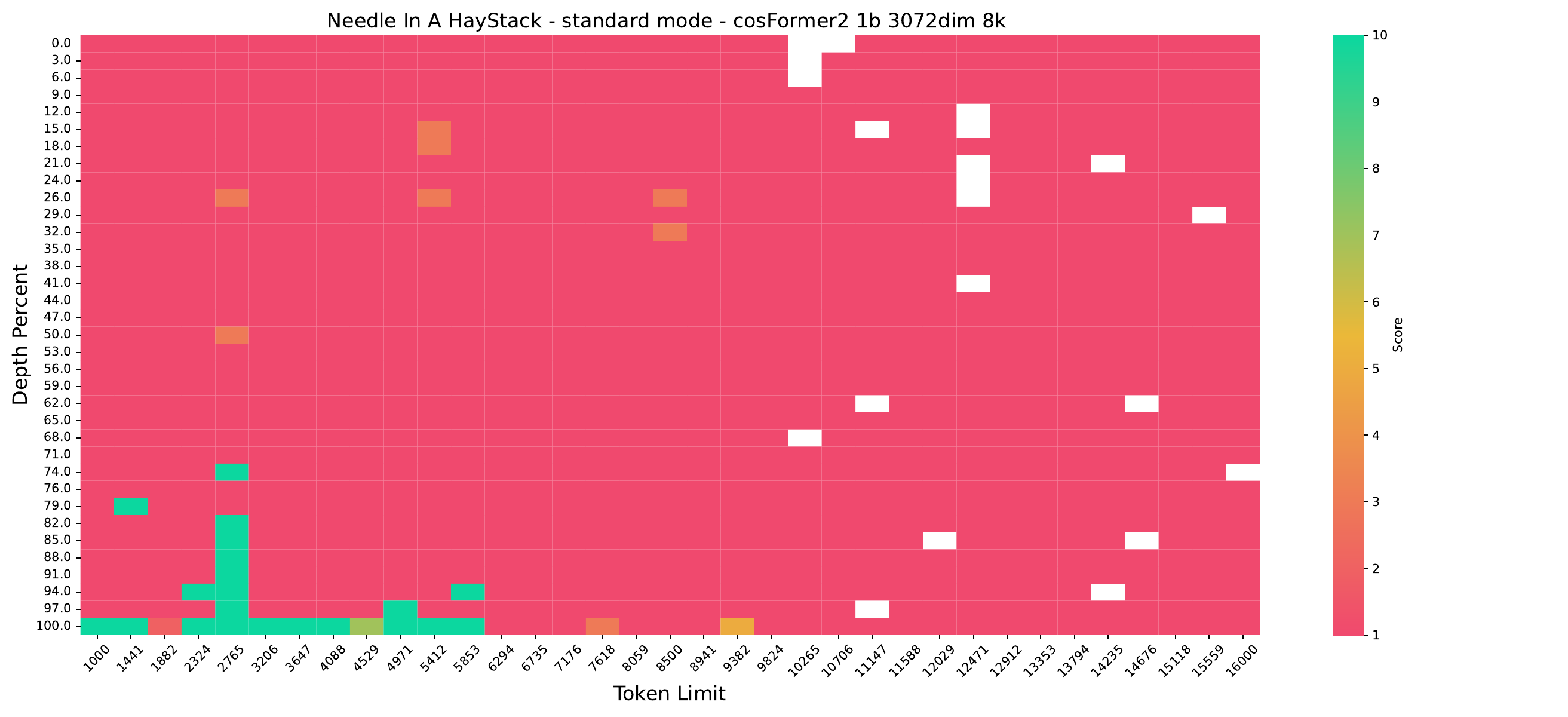}
\includegraphics[width=1\linewidth]{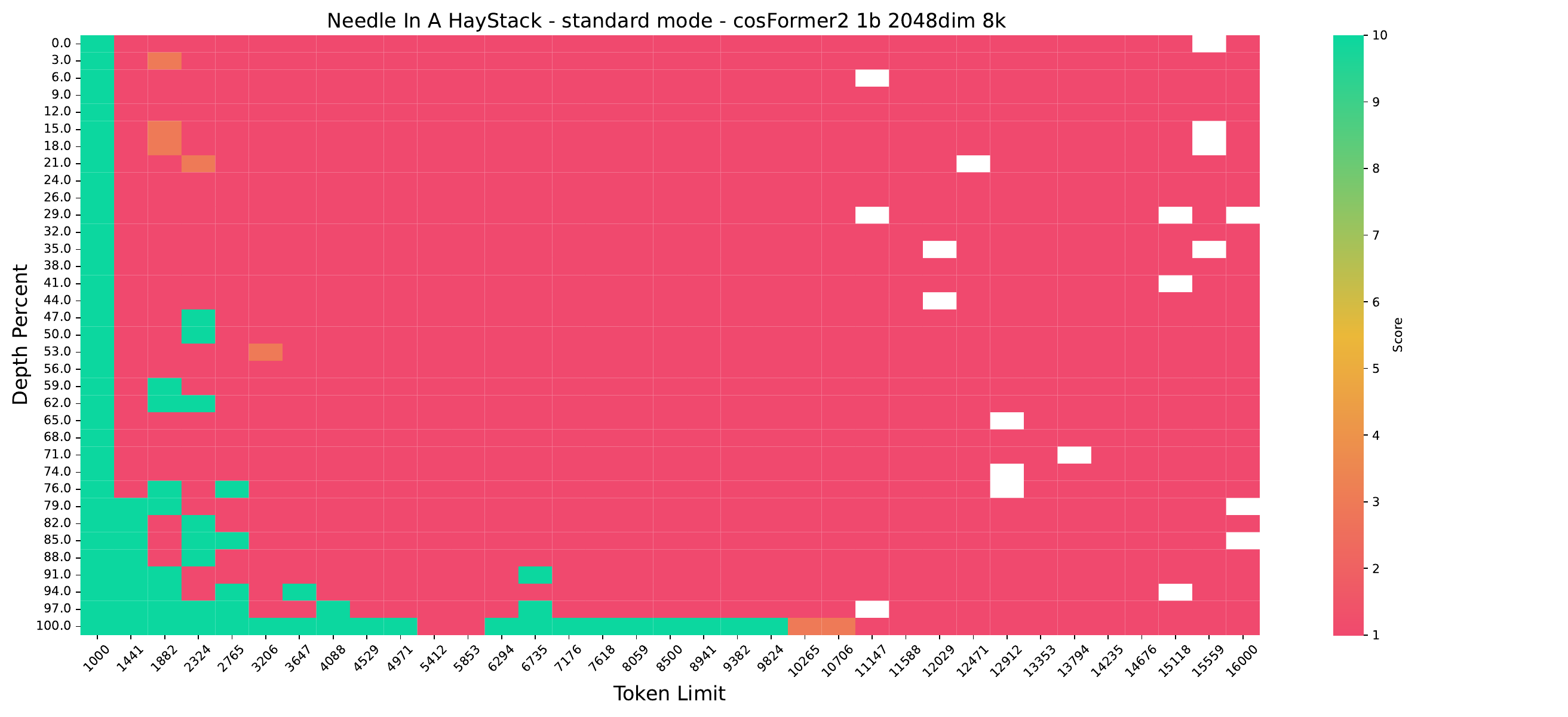}
\end{figure*}

\begin{figure*}
\centering
\includegraphics[width=1\linewidth]{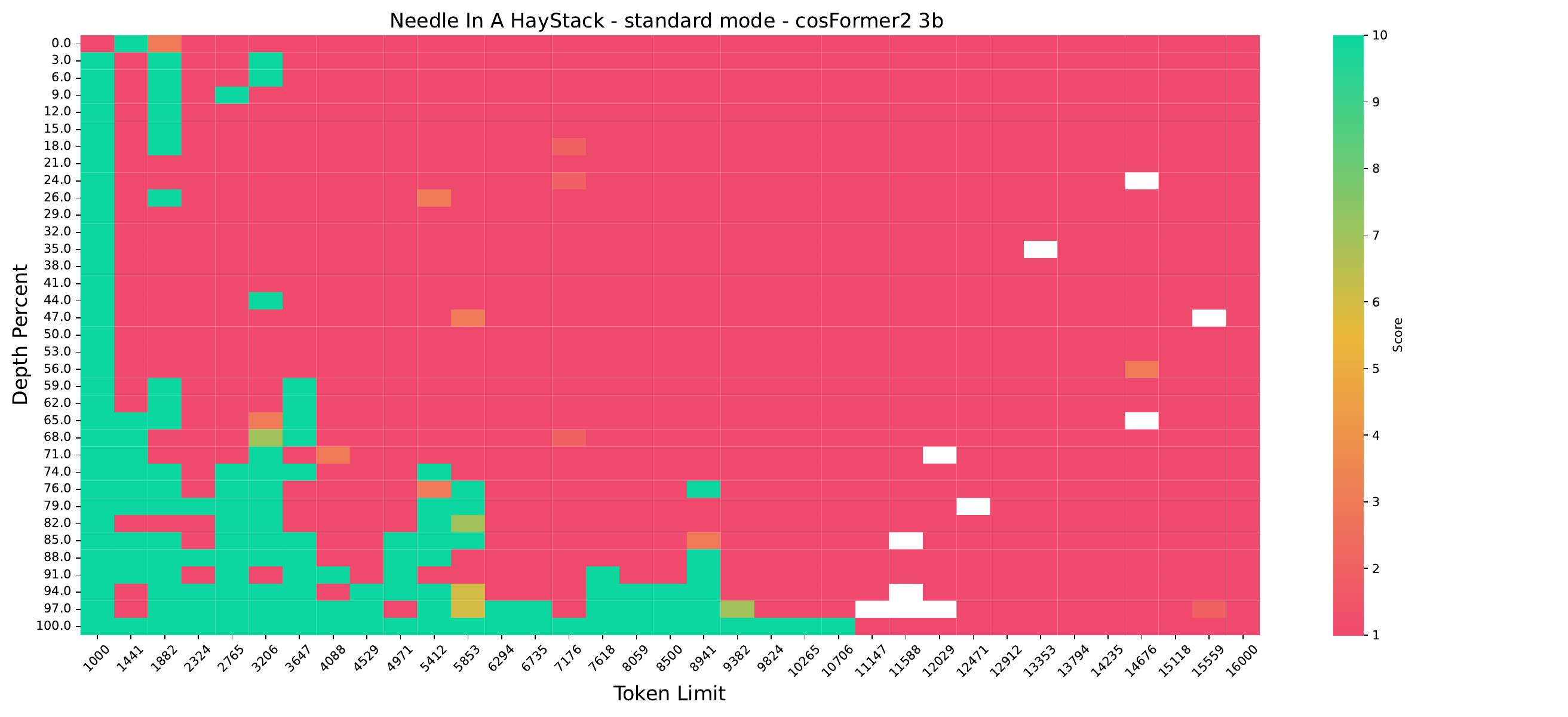}
\includegraphics[width=1\linewidth]{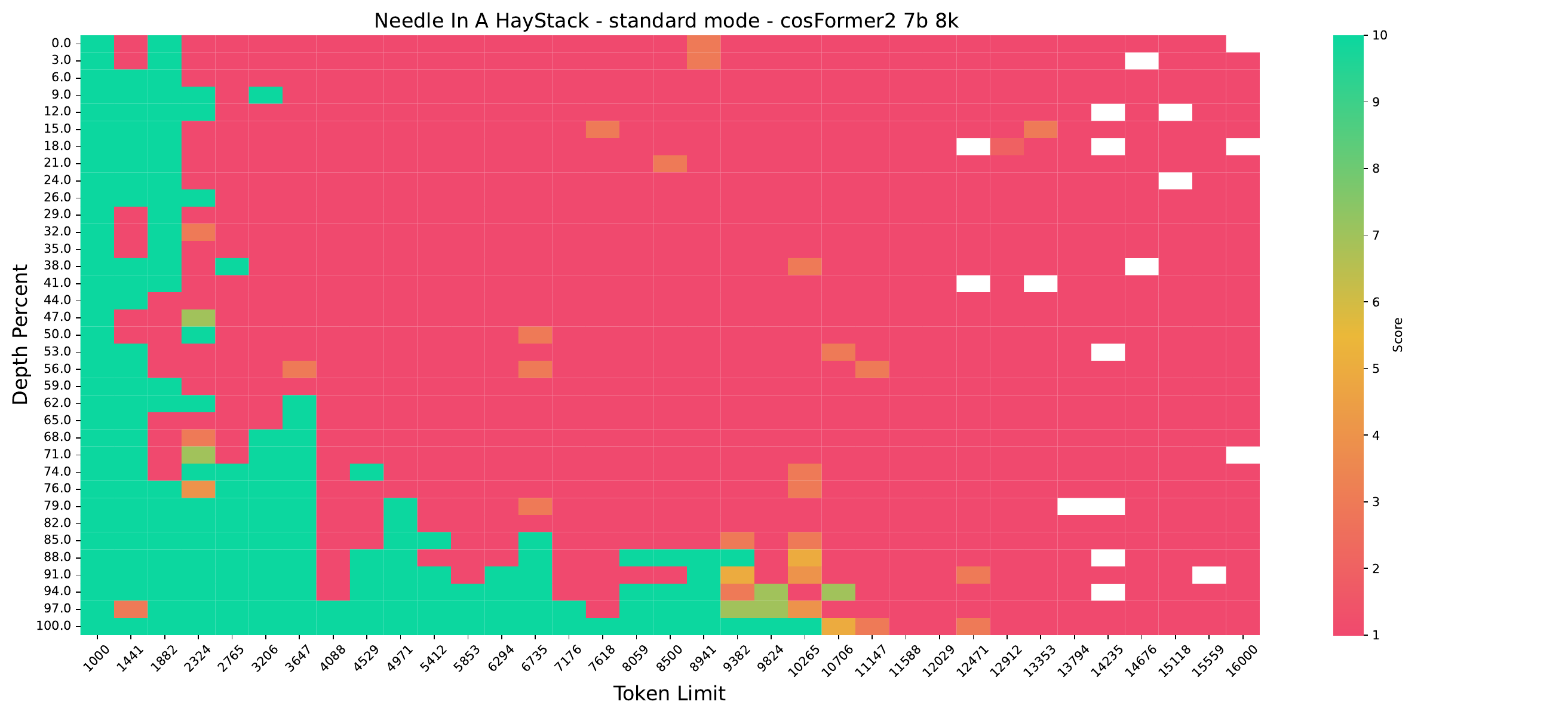}
\end{figure*}

\end{document}